\documentclass[11pt]{article}

\usepackage{amsmath}    
\usepackage{amssymb}    
\usepackage{amsthm}     
\usepackage{mathtools}  
\usepackage{bm}         
\usepackage{cancel}     

\usepackage{graphicx}   
\usepackage{float}      

\usepackage{booktabs}   
\usepackage{tabularx}   
\usepackage{array}      
\usepackage{multirow}   
\usepackage{makecell}   

\usepackage{xcolor}     
\usepackage{caption}    
\usepackage{subfig}     

\usepackage{natbib}     
\usepackage[bookmarks=false,hidelinks]{hyperref} 

\usepackage{appendix}  
\usepackage{verbatim}  
\usepackage{enumitem}  

\usepackage[margin=1in]{geometry} 

\theoremstyle{plain}
\newtheorem{theorem}{Theorem}[section]
\newtheorem{proposition}[theorem]{Proposition}
\newtheorem{lemma}[theorem]{Lemma}
\newtheorem{corollary}[theorem]{Corollary}

\theoremstyle{definition}

\newtheorem{assumption}{Assumption}[section]
\newtheorem{example}[theorem]{Example}

\theoremstyle{remark}
\newtheorem{remark}{Remark}[section]

\allowdisplaybreaks
\newcommand{\Var}{\operatorname{Var}}
\newcommand{\Cov}{\operatorname{Cov}}
\newcommand{\norm}[1]{\left\lVert #1\right\rVert}
\newcommand{\inner}[2]{\left\langle #1,#2\right\rangle}

\date{}
\begin{document}

\author{Xiaoran Jiang \\ 
  Department of Statistics \\ 
    George Mason University \\
    Fairfax, VA, 22030
 \and 
 Anand N. Vidyashankar \\
    Department of Statistics \\ 
     George Mason University \\ 
     Fairfax, VA, 22030 } 
\title{Federated Nonparametric Regression with Heterogeneous Clients}
\maketitle

\begin{abstract}
\sloppy
Whether a federated predictor benefits an individual client depends on both population heterogeneity and estimation uncertainty. We study this question for Nadaraya--Watson regression in a triangular array with potentially different fitting and evaluation distributions. Each client transmits local kernel summaries, allowing the server to construct a covariate-density-weighted federated estimator without centralizing individual observations. A client combines its fitted local predictor with the federated predictor through a prescribed weight, and gain is the paired reduction in squared prediction loss on an independent audit observation. Exact identities characterize both the gain of the realized fitted pair and its repeated-fitting mean. We also establish pointwise inference for the federated estimator. A two-stage analysis of client gains separates fitting variability from audit variability, retains the covariance created when a client's observations enter both its local estimator and the common server estimator, and yields fixed-block simultaneous inference across clients. Robustness is examined under two perturbation models. A Kullback--Leibler neighborhood provides a variance-sensitive lower certificate in a regular positive-gain-variance regime, while an $L_2$ neighborhood yields an exact least-favorable regression direction and robust value. Exact-model, vanishing-variance, and exact-zero-direction boundaries are kept separate because their scalar inputs are nonregular. The framework thus distinguishes decisions about a realized fitted pair from claims about the learning procedure under repeated fitting.
\fussy
\end{abstract}

\setlength{\emergencystretch}{2em}

\section{Introduction}
\label{sec:introduction}

Federated learning allows a central server to aggregate information across multiple clients while each client retains its raw observations \citep{McMahanEtAl2017,KairouzEtAl2021}. In the nonparametric setting considered here, each client fits a Nadaraya--Watson regression estimator \citep{Nadaraya1964,Watson1964}, and the information sent to the server consists of kernel numerator and denominator summaries. Their aggregate estimates a covariate-density-weighted combination of the client regression functions rather than a simple average.  This target is natural for prediction, but it leaves a separate participation question: when does access to the federated predictor improve a client's out-of-sample prediction relative to its local predictor?

Persistent population gain requires a persistent displacement between the local procedure target and the client's evaluation regression, together with a favorable correction from the federation.  Conditional-regression shift between fitting and evaluation provides a concrete instance under the shrinking-bandwidth assumptions used here.  Even when the local procedure is correctly targeted, federation may still yield finite-sample gain through reduced estimator variability.  We therefore formulate predictive gain directly as the paired reduction in squared prediction loss and infer it from an independent client audit sample rather than estimating several risks separately.

Statistical heterogeneity across clients is a central challenge in federated learning \citep{LiEtAl2020FedProx}, and recent work has also developed inferential theory for heterogeneous decentralized federated estimators \citep{GuChen2024}. The closest personalized-federated-learning literature studies interpolation between local and global models, adaptive mixture weights, and participation incentives \citep{MansourMohriRoSuresh2020, DengKamaniMahdavi2020, DonahueKleinberg2021}.  Agnostic federated learning optimizes over mixtures of client distributions \citep{MohriSivekSuresh2019}, while distributed nonparametric aggregation studies heterogeneous workers and aggregate risk \citep{ChenLiXu2026}.  Inference for predictive error and improvement has also been developed for cross-validation under algorithmic stability \citep{BayleEtAl2020}. Our focus is different: the estimand is client-specific paired predictive gain under moving heterogeneous fitting and evaluation laws; the local estimator is constructed from a client’s kernel numerator and denominator summaries, while weighted versions of those same summaries enter the server aggregate, inducing local–server dependence; and the resulting covariance is retained in realized-fit, repeated-fitting, simultaneous-client, and robustness calculations.

The paper has four contributions. First, we establish pointwise consistency and asymptotic normality for the federated Nadaraya--Watson estimator with a growing heterogeneous client
population, and derive exact population and repeated-fitting gain identities that separate persistent target correction from finite-sample variance rescue.  Second, an independent audit sample gives conditional Berry--Esseen inference \citep{Petrov1975} for the realized fitted pair, while scalar Nadaraya--Watson expansions give procedure-level inference in a nondegenerate first-order regime.  Third, the common server fit induces a non-diagonal fitting covariance across clients; fixed-block Gaussian limits and direct all-client summability bounds retain that dependence.  The countable-product formulation is used only to organize compatible finite-dimensional limits.  Fourth, we develop two gain-specific robustness layers: a Kullback--Leibler (KL) \citep{AhmadiJavid2012,Lam2016} lower certificate in a regular positive-gain-variance regime and an exact $L_2$ least-favorable regression analysis.  Their vanishing-variance and exact-zero-direction boundaries are separate nonsmooth problems.

All targets may move with the stage.  The primary fitting theorem is nondegenerate first order and does not determine the sign of gain; exact-model, vanishing-variance, exact-zero-direction, and local transition regimes are stated separately.

The remainder of the paper is organized as follows. Section~\ref{sec:setup} introduces the federated regression framework and prediction protocol. Section~\ref{sec:fnw-pointwise} establishes pointwise inference for the federated estimator. Sections~\ref{sec:gain} and \ref{sec:gain-inference} develop the predictive-gain characterization and its statistical inference, respectively. Sections~\ref{sec:kl-robust} and \ref{sec:l2-robust} investigate robustness under KL distributional perturbations and $L_2$ regression-function perturbations. Section~\ref{sec:supp-numerical} presents numerical experiments, followed by a concluding discussion. Detailed technical conditions and proofs are provided in the Appendix.

\section{Federated regression and prediction protocol}
\label{sec:setup}
We consider a sequence of federated regression systems indexed by a stage parameter $t$, allowing the number of participating clients, their underlying distributions, and their sample sizes to vary across stages. Within this general framework, we define a weighted federated Nadaraya–Watson target and construct its estimator through client-level summaries, without requiring the transmission of individual observations. We then introduce a two-stage prediction and evaluation protocol, in which local and federated estimators are constructed using fitting samples, while independent evaluation samples are used to assess the predictive gain of personalized estimators relative to their local counterparts. This separation between fitting and evaluation provides the foundation for the inferential analysis developed in subsequent sections.
\subsection{Triangular array of client systems}
\label{subsec:triangular-array}

Let $t\ge1$ index a sequence of federated systems.  At stage $t$, $K_t$ clients participate.  Client $j\le K_t$ is associated with a probability law $P_{j,t}$ on $(X,Y)\in\mathbb{R}^2$ and contributes a fitting sample
\begin{equation}
\mathcal{D}^{\mathrm{fit}}_{j,t}=\{Z_{i,j,t}:1\le i\le n_{j,t}\},
\qquad Z_{i,j,t}\stackrel{\mathrm{iid}}{\sim}P_{j,t}.
\label{eq:fit-sample}
\end{equation}
The fitting samples are independent across clients.  The client laws, sample sizes, estimation weights, kernels, and bandwidths may all depend on $t$.  We allow $K_t\to\infty$ and impose information-growth conditions only with the corresponding asymptotic results.

Let $p_{j,t}$ be the covariate density under $P_{j,t}$ with respect to a fixed $\sigma$-finite reference measure, and let
\begin{align}
m_{j,t}(x)&=\bm{E}_{P_{j,t}}[Y\mid X=x],\\
g_{j,t}(x)&=p_{j,t}(x)m_{j,t}(x).
\end{align}
The deterministic estimation weights satisfy
\begin{equation}
w_{j,t}\ge0,
\qquad
\sum_{j=1}^{K_t}w_{j,t}=1.
\end{equation}
Sample-size weighting, $w_{j,t}=n_{j,t}/N_t$ with $N_t=\sum_jn_{j,t}$, is a leading example.

\subsection{Federated target and estimator}
\label{subsec:federated-estimator}

The estimation weights introduced above determine how client-level information is combined. Using these weights, we first define the aggregated population numerator and denominator as
\begin{equation}
A_t(x)=\sum_{j=1}^{K_t}w_{j,t}g_{j,t}(x),
\qquad
B_t(x)=\sum_{j=1}^{K_t}w_{j,t}p_{j,t}(x).
\end{equation}
Whenever $B_t(x)>0$, the federated population target is
\begin{equation}
m_t^{\star}(x)
=\frac{A_t(x)}{B_t(x)}
=\frac{\sum_jw_{j,t}p_{j,t}(x)m_{j,t}(x)}{\sum_jw_{j,t}p_{j,t}(x)}.
\label{eq:federated-target}
\end{equation}
The target is allowed to move with $t$.

Client $j$ uses a kernel $\mathcal K_{j,t}$ and bandwidth $h_{j,t}>0$, with
\begin{equation}
\psi_{j,t}(x-X)
=\frac1{h_{j,t}}\mathcal K_{j,t}\left(\frac{x-X}{h_{j,t}}\right).
\end{equation}
It computes
\begin{align}
\widehat g_{j,t}(x)
&=\frac1{n_{j,t}}\sum_{i=1}^{n_{j,t}}
\psi_{j,t}(x-X_{i,j,t})Y_{i,j,t},\\
\widehat p_{j,t}(x)
&=\frac1{n_{j,t}}\sum_{i=1}^{n_{j,t}}
\psi_{j,t}(x-X_{i,j,t}),
\end{align}
and transmits the weighted summaries
\begin{equation}
a_{j,t}(x)=w_{j,t}\widehat g_{j,t}(x),
\qquad
b_{j,t}(x)=w_{j,t}\widehat p_{j,t}(x).
\end{equation}
The server forms
\begin{align}
\widehat A_t(x)&=\sum_ja_{j,t}(x),
&\widehat B_t(x)&=\sum_jb_{j,t}(x),\\
\widehat m_t(x)&=\widehat A_t(x)/\widehat B_t(x).
\label{eq:federated-nw}
\end{align}
This construction also specifies the communication protocol. Only $(a_{j,t}(x),b_{j,t}(x))$ are transmitted.  The implemented ratio uses a denominator floor and fixed clipping; under the inactive-stabilization and inactive-clipping conditions in the Appendix, it has the same pointwise limit as the raw ratio.  Exact prediction at a new $x$ uses one additional summary round; grid interpolation is a separate approximation.

\subsection{Fitting and evaluation samples}
\label{subsec:fit-eval-split}

Having specified the federated estimator and the corresponding communication protocol, we now introduce the prediction and evaluation framework. In particular, we distinguish the information used to construct the predictors from the independent data used to assess their predictive performance. Let
\begin{equation}
\mathcal{F}_t^{\mathrm{fit}}
=\sigma\{\mathcal{D}^{\mathrm{fit}}_{j,t}:1\le j\le K_t\}
\end{equation}
be the fitting sigma-field.  Client $j$ uses the local Nadaraya--Watson predictor
\begin{equation}
L_{j,t}(x)=\widehat m_{j,t}^{L}(x)=\frac{\widehat g_{j,t}(x)}{\widehat p_{j,t}(x)},
\end{equation}
from its fitting sample, while all clients share the server predictor $F_t(x)=\widehat m_t(x)$.  Because client $j$ contributes to the server fit, $L_{j,t}$ and $F_t$ are generally dependent.

For gain evaluation, client $j$ has an independent second-stage sample
\begin{equation}
\mathcal{D}^{\mathrm{eval}}_{j,t}
=\{Z^{\mathrm{eval}}_{r,j,t}:1\le r\le N_{j,t}^{\mathrm{eval}}\},
\qquad
Z^{\mathrm{eval}}_{r,j,t}\stackrel{\mathrm{iid}}{\sim}Q_{j,t},
\label{eq:eval-sample}
\end{equation}
where $Q_{j,t}$ is the client's stage-$t$ evaluation law.  The evaluation law may equal the current client law $P_{j,t}$ or represent a future client population.  The evaluation samples are independent across clients and independent of $\mathcal{F}_t^{\mathrm{fit}}$.  This independent audit design is used throughout the gain theory.  Leave-out or foldwise cross-fitting may be used for fitting-side nuisance estimates, but ordinary foldwise reuse of observations for both predictor fitting and gain evaluation has a different fold-averaged estimand and dependence structure and is not covered here.

Fix a prescribed local weight $\alpha\in[0,1)$ and put $\vartheta=1-\alpha$.  The personalized predictor is
\begin{equation}
M_{j,t}(x)=\alpha L_{j,t}(x)+\vartheta F_t(x).
\label{eq:personalized-predictor}
\end{equation}

The weight $\alpha$ controls the balance between local information and information borrowed from the federation. When $\alpha=0$, the personalized predictor reduces to the federated predictor, whereas values of $\alpha$ approaching one place increasing emphasis on the local fit. The restriction $\alpha<1$ ensures that the federated component receives a positive weight. Such borrowing may reduce estimation variability, but it may also introduce systematic bias when the client-specific regression function differs from the federated target. Consequently, an improvement in predictive performance is not guaranteed. This motivates the client-specific predictive gain analysis developed in the subsequent sections, where the personalized predictor is compared with its local counterpart under the evaluation law $Q_{j,t}$.

\section{Pointwise inference for the federated estimator}
\label{sec:fnw-pointwise}

We now turn to the inferential properties of the federated estimator. Since the personalized predictor incorporates the shared server fit $F_t=\widehat m_t$, understanding the estimation error of the federated component is a natural first step toward analyzing client-specific predictive gain. We therefore establish pointwise consistency and asymptotic normality under the triangular-array framework introduced above, allowing both the number of participating clients and their underlying distributions to vary with $t$.

For a fixed query point $x$, we characterize the fluctuations of the ratio estimator $\widehat m_t(x)$ around its population target $m_t^\star(x)$. Define the linearized estimation error
\begin{align}
\Delta_t^{\mathrm{NW}}(x)&=\frac{\widehat A_t(x)-m_t^\star(x)\widehat B_t(x)}{B_t(x)}.
\end{align}
Its variance and the aggregate kernel variance scale are given by
\begin{align}
V_t(x)&=\Var\{\Delta_t^{\mathrm{NW}}(x)\},\\
q_t&=\sum_{j=1}^{K_t}\frac{w_{j,t}^2}{n_{j,t}h_{j,t}}.
\end{align}
The quantity $q_t$ captures the combined effects of client weights, sample sizes, and bandwidths on the estimation variability, while $V_t(x)$ provides the variance normalization for the asymptotic distribution.

To establish the pointwise results, we impose the following regularity conditions at the fixed query point $x$. All suprema and infima over clients are taken over the active triangular array $1\le j\le K_t$, allowing the client laws $P_{j,t}$ to vary with $t$. The densities in these conditions are with respect to Lebesgue measure on $\mathbb{R}$.

\begin{assumption}[Uniform kernel regularity and information growth]
\label{fnw:ass:kernel-growth}
For every active pair $(j,t)$, $\mathcal K_{j,t}$ is measurable and satisfies
\begin{align}
\int_{\mathbb{R}}\mathcal K_{j,t}(u)\,d u=1, \qquad \int_{\mathbb{R}}u\mathcal K_{j,t}(u)\,d u=0.
\end{align}
There are finite constants $C_{K,1}$, $C_{K,2}$, $C_{K,R}$,
and $C_{K,2}^{(2)}$, and a constant $c_{K,R}>0$, such that
\begin{align}
&\sup_{t,j}\int_{\mathbb{R}}|\mathcal K_{j,t}(u)|\,d u\le C_{K,1},\\
&\sup_{t,j}\int_{\mathbb{R}}u^2|\mathcal K_{j,t}(u)|\,d u\le C_{K,2},\\
&c_{K,R}\le \inf_{t,j}\int_{\mathbb{R}}\mathcal K_{j,t}(u)^2\,d u
\le \sup_{t,j}\int_{\mathbb{R}}\mathcal K_{j,t}(u)^2\,d u\le C_{K,R},\\
&\sup_{t,j}\int_{\mathbb{R}}u^2\mathcal K_{j,t}(u)^2\,d u\le C_{K,2}^{(2)}.
\end{align}
The bandwidths and local sample sizes satisfy
\begin{align*}
\max_{1\le j\le K_t}h_{j,t}\to0,
\qquad
\min_{1\le j\le K_t}n_{j,t}h_{j,t}\to\infty.
\end{align*}
\end{assumption}

The kernel conditions are standard in nonparametric regression and are satisfied by commonly used symmetric kernels, including the Gaussian kernel and compactly supported kernels such as the Epanechnikov kernel. The bandwidth and sample-size conditions ensure that smoothing vanishes while the effective local sample size grows. Importantly, these requirements hold uniformly over all participating clients, accommodating a federated system in which the number of clients may increase with $t$.

\begin{assumption}[Uniform client-law smoothness and second moments]
\label{fnw:ass:client-regularity}
For every active pair $(j,t)$, $p_{j,t}$ and $g_{j,t}=p_{j,t}m_{j,t}$ are twice continuously differentiable.  There are finite constants $C_{p,0}$, $C_{p,2}$, and $C_{g,2}$ such that
\begin{align*}
\sup_{t,j,z}p_{j,t}(z)&\le C_{p,0}, &
\sup_{t,j,z}|p_{j,t}''(z)|&\le C_{p,2}, &
\sup_{t,j,z}|g_{j,t}''(z)|&\le C_{g,2}.
\end{align*}
Define
\begin{align}
s_{j,t}(z)=p_{j,t}(z)\bm{E}_{P_{j,t}}[Y^2\mid X=z].
\label{fnw:eq:s-second-moment}
\end{align}
There is a neighborhood $U_x$ of $x$, finite constants $L_p$, $L_g$, $L_s$, and $C_s$, such that for all $z_1,z_2\in U_x$,
\begin{align}
|p_{j,t}(z_1)-p_{j,t}(z_2)|\le L_p|z_1-z_2|, \qquad |g_{j,t}(z_1)-g_{j,t}(z_2)|\le L_g|z_1-z_2|.
\end{align}
\begin{align}
|s_{j,t}(z_1)-s_{j,t}(z_2)|\le L_s|z_1-z_2|, \qquad \sup_{t,j,z}s_{j,t}(z)\le C_s.
\end{align}
\end{assumption}

The smoothness conditions control the kernel estimation bias, while the local regularity of the density-weighted second moments supports the pointwise variance approximation. These requirements hold uniformly across clients to accommodate the growing federated system.

\begin{assumption}[Pointwise denominator and weighted variance nondegeneracy]
\label{fnw:ass:pointwise}
There are constants $c_B>0$ and $c_{V,0}(x)>0$ such that
\begin{equation}
\inf_t B_t(x)\ge c_B.
\label{fnw:eq:denominator-floor}
\end{equation}
Moreover, with $q_t=\sum_{j=1}^{K_t}w_{j,t}^2/(n_{j,t}h_{j,t})$, for all sufficiently large $t$,
\begin{equation}
\frac{1}{B_t(x)^2q_t}
\sum_{j=1}^{K_t}
\frac{w_{j,t}^2}{n_{j,t}h_{j,t}}
R(\mathcal K_{j,t})p_{j,t}(x)
\bm{E}_{P_{j,t}}\!\left([Y-m_t^\star(x)]^2\mid X=x\right)
\ge c_{V,0}(x).
\label{fnw:eq:weighted-variance-floor}
\end{equation}
\end{assumption}

Unlike a clientwise nondegeneracy condition, this assumption requires positive density and variance contributions only at the aggregated level. Individual clients are therefore allowed to contribute little or no information at the query point $x$, provided that the weighted federation retains sufficient information. This condition ensures that the pointwise variance remains comparable to the aggregate scale $q_t$, preventing degeneracy of the asymptotic normalization.

\begin{assumption}[Uniform higher-order moments]
\label{fnw:ass:higher-moment}
There is a constant $\delta>0$ and finite constants $C_{K,2+\delta}$ and $C_{Y,2+\delta}$ such that
\begin{align}
\sup_{t,j}\int_{\mathbb{R}}|\mathcal K_{j,t}(u)|^{2+\delta}\,d u\le C_{K,2+\delta}.
\end{align}
\begin{align}
\sup_{t,j,z}p_{j,t}(z)\bm{E}_{P_{j,t}}[|Y|^{2+\delta}\mid X=z]\le C_{Y,2+\delta}.
\end{align}
\end{assumption}
These higher-order moment conditions provide the tail control needed for asymptotic normality and consistency of the variance estimator.

\begin{assumption}[Growing-system undersmoothing]
\label{fnw:ass:undersmoothing}
With $q_t=\sum_{j=1}^{K_t}w_{j,t}^2/(n_{j,t}h_{j,t})$,
\begin{align}
\frac{\sum_{j=1}^{K_t}w_{j,t}h_{j,t}^2}{\sqrt{q_t}}\to0.
\end{align}
\end{assumption}

The undersmoothing condition ensures that the kernel smoothing bias is asymptotically negligible relative to the stochastic scale $\sqrt{q_t}$. Under the preceding smoothness conditions, the aggregate bias is controlled by the weighted bandwidth term $\sum_j w_{j,t}h_{j,t}^2$. Requiring this term to vanish faster than $\sqrt{q_t}$ allows the limiting distribution to be centered at the federated population target $m_t^\star(x)$ without an explicit bias correction.

The preceding assumptions distinguish the conditions needed for consistency from those required for asymptotic normality. The first three assumptions ensure pointwise consistency and a nondegenerate stochastic scale, while the additional higher-order moment and undersmoothing conditions support the centered normal approximation. Under Assumptions~\ref{fnw:ass:kernel-growth}--\ref{fnw:ass:pointwise}, the pointwise variance satisfies $V_t(x)\asymp q_t$.

To estimate the pointwise variance without transmitting individual observations, each client additionally provides the second-moment summaries
\begin{align}
\mathcal V_{j,t}^{(r)}(x)&=\frac{w_{j,t}^2}{n_{j,t}^2}
\sum_{i=1}^{n_{j,t}}\psi_{j,t}^2(x-X_{i,j,t})Y_{i,j,t}^r,
\qquad r=0,1,2.
\end{align}
Using these summaries, the server constructs the distributed variance estimator
\begin{align}
\widehat V_t(x)&=\frac{1}{\widehat B_t(x)^2}
\sum_{j=1}^{K_t}\left[
\mathcal V_{j,t}^{(2)}(x)
-2\widehat m_t(x)\mathcal V_{j,t}^{(1)}(x)
+\widehat m_t(x)^2\mathcal V_{j,t}^{(0)}(x)
\right].
\label{fnw:eq:distributed-variance}
\end{align}
Thus, the variance estimator requires only three additional scalar summaries per client at each query point. Its consistency, established in the following theorem, allows pointwise inference to be conducted without access to the underlying client-level observations.

\begin{theorem}[Pointwise inference]
\label{fnw:thm:pointwise-inference}
Under Assumptions~\ref{fnw:ass:kernel-growth}--\ref{fnw:ass:pointwise}, $\widehat m_t(x)-m_t^\star(x)\xrightarrow{\bm{P}}0$.  If Assumptions~\ref{fnw:ass:higher-moment} and \ref{fnw:ass:undersmoothing} also hold,
\begin{align}
\frac{\widehat m_t(x)-m_t^\star(x)}{\sqrt{\widehat V_t(x)}}&\Longrightarrow N(0,1),\\
\frac{\widehat V_t(x)}{V_t(x)}&\xrightarrow{\bm{P}}1.
\end{align}
\end{theorem}

The complete proof is provided in Appendix~\ref{app:fnw-pointwise}.

\section{Predictive gain and positivity}
\label{sec:gain}

The preceding section establishes the pointwise inferential properties of the federated estimator. However, these results do not, by themselves, guarantee that incorporating the federated component improves prediction for an individual client. We therefore shift our focus from estimation accuracy to predictive performance, comparing the personalized predictor $M_{j,t}$ with the local predictor $L_{j,t}$ under the client's evaluation law $Q_{j,t}$. In particular, we investigate when borrowing information from the federation leads to a positive predictive gain.

\subsection{Paired gain, fitted gain, and marginal gain}
\label{subsec:gain-objects}
We distinguish fitted and marginal notions of predictive gain. For $z=(x,y)$, define the paired gain
\begin{align}
W_{j,t}(z)
&=\{y-L_{j,t}(x)\}^2-\{y-M_{j,t}(x)\}^2\notag\\
&=2\vartheta\{y-L_{j,t}(x)\}
   \{F_t(x)-L_{j,t}(x)\}-\vartheta^2\{F_t(x)-L_{j,t}(x)\}^2.
\label{eq:paired-gain-main}
\end{align}
Conditional on $\mathcal{F}_t^{\mathrm{fit}}$, the gain of the realized fitted predictors is
\begin{equation}
G_{j,t}^{\mathrm{fit}}
=\int W_{j,t}(z)Q_{j,t}(\,d z)
=\bm{E}[W_{j,t}(Z)\mid\mathcal{F}_t^{\mathrm{fit}}].
\label{eq:fitted-gain-main}
\end{equation}
Its repeated-fitting mean is the deterministic stage-$t$ marginal gain
\begin{equation}
\Gamma_{j,t}=\bm{E}_t^{\mathrm{fit}}G_{j,t}^{\mathrm{fit}},
\label{eq:marginal-gain-main}
\end{equation}
where $\bm{E}_t^{\mathrm{fit}}$ denotes expectation under the joint law of all stage-$t$ fitting samples.  The evaluation estimator is
\begin{equation}
\widehat G_{j,t}
=\frac1{N_{j,t}^{\mathrm{eval}}}\sum_{r=1}^{N_{j,t}^{\mathrm{eval}}}
W_{j,t}(Z^{\mathrm{eval}}_{r,j,t}).
\label{eq:empirical-gain-main}
\end{equation}
The exact two-stage decomposition is
\begin{equation}
\widehat G_{j,t}-\Gamma_{j,t}
=\{G_{j,t}^{\mathrm{fit}}-\Gamma_{j,t}\}
+\{\widehat G_{j,t}-G_{j,t}^{\mathrm{fit}}\}.
\label{eq:two-stage-main}
\end{equation}
The decomposition corresponds to distinct scientific questions.  Conditional inference for $G_{j,t}^{\mathrm{fit}}$ concerns the realized fitted pair, whereas inference for $\Gamma_{j,t}$ concerns the learning procedure under repeated fitting. The same fitted-versus-marginal distinction is used for both robustness layers. 

Define
\begin{align}
A_{j,t}^{\mathrm{fit}}
&=\bm{E}\left[\{Y-L_{j,t}(X)\}\{F_t(X)-L_{j,t}(X)\}\mid\mathcal{F}_t^{\mathrm{fit}}\right],\\
D_{j,t}^{\mathrm{fit}}
&=\bm{E}\left[\{F_t(X)-L_{j,t}(X)\}^2\mid\mathcal{F}_t^{\mathrm{fit}}\right].
\end{align}
Then
\begin{equation}
G_{j,t}^{\mathrm{fit}}
=2\vartheta A_{j,t}^{\mathrm{fit}}-\vartheta^2D_{j,t}^{\mathrm{fit}}.
\label{eq:gain-A-D}
\end{equation}
Thus a prescribed borrowing amount $\vartheta>0$ gives positive fitted gain exactly when $2A_{j,t}^{\mathrm{fit}}>\vartheta D_{j,t}^{\mathrm{fit}}$.  If $D_{j,t}^{\mathrm{fit}}>0$, some positive amount of borrowing is beneficial exactly when $A_{j,t}^{\mathrm{fit}}>0$.

\subsection{Population procedure targets}
\label{subsec:population-target-gain}
To isolate the deterministic target-shift mechanism behind persistent gain,
we first replace the fitted predictors by their population procedure targets. Let $m_{j,t}^{L}$ and $m_t^{F}$ be deterministic local and federated procedure targets.  For finite bandwidths these are the corresponding smoothed ratio targets.  Let
\begin{align}
m_{j,t}^{Q}(x)&=\bm{E}_{Q_{j,t}}[Y\mid X=x],\\
\norm{f}_{j,t}^2&=\int f(x)^2Q_{j,t,X}(\,d x).
\end{align}
Define
\begin{equation}
b^{P}_{L,j,t}=m_{j,t}^{L}-m_{j,t}^{Q},
\qquad
b^{P}_{F,j,t}=m_t^{F}-m_{j,t}^{Q}.
\end{equation}

\begin{proposition}[Exact population gain]
\label{prop:population-gain-main}
For every $\alpha\in[0,1)$,
\begin{align}
G_{j,t}^{P}(\alpha)
=\vartheta\bigl\{
&\norm{b^{P}_{L,j,t}}_{j,t}^2-\norm{b^{P}_{F,j,t}}_{j,t}^2+\alpha\norm{b^{P}_{F,j,t}-b^{P}_{L,j,t}}_{j,t}^2
\bigr\}.
\label{eq:population-gain-main}
\end{align}
If $\norm{b^{P}_{F,j,t}-b^{P}_{L,j,t}}_{j,t}>0$, some positive amount of borrowing helps if and only if
\begin{equation}
\inner{b^{P}_{F,j,t}}{b^{P}_{L,j,t}}_{j,t}
<\norm{b^{P}_{L,j,t}}_{j,t}^2.
\label{eq:population-some-gain}
\end{equation}
Every nontrivial mixture has positive gain when
\begin{equation}
\norm{b^{P}_{F,j,t}}_{j,t}<\norm{b^{P}_{L,j,t}}_{j,t}.
\label{eq:population-all-gain}
\end{equation}
\end{proposition}

\begin{corollary}[Correctly targeted local procedure]
\label{cor:correctly-targeted-local}
If $m_{j,t}^{L}=m_{j,t}^{Q}$, then
\begin{equation}
G_{j,t}^{P}(\alpha)
=-\vartheta^2\norm{m_t^{F}-m_{j,t}^{Q}}_{j,t}^2\le0.
\label{eq:correct-local-no-population-gain}
\end{equation}
Thus persistent positive population gain requires a persistent displacement of the local procedure target from the evaluation regression.  This does not preclude finite-sample marginal gain through variance reduction.
\end{corollary}

\begin{example}[Positive and negative transfer in one supported family]
\label{ex:two-client-shift}
Let two equally weighted clients have a common covariate law and fitting regressions $m_1=0$ and $m_2=2b$, where $b$ is bounded, smooth, and nonzero.  Then the federated target is $m^{F}=b$.  If client 1 is evaluated under a law with regression $m_1^{Q}=b$, then
\begin{equation}
G_1^{P}(\alpha)=(1-\alpha^2)\norm{b}^2>0.
\label{eq:positive-transfer-example}
\end{equation}
Evaluating the same fitting configuration under client 1's unchanged law, for which $m_1^{Q}=0$, gives
\begin{equation}
G_1^{P}(\alpha)=-(1-\alpha)^2\norm{b}^2<0.
\label{eq:negative-transfer-example}
\end{equation}
Bounded response noise and an interior evaluation region make this a direct instance of the primitive Nadaraya--Watson assumptions; Appendix~\ref{app:worked-regime} verifies one such construction in detail.
\end{example}

Condition \eqref{eq:population-all-gain} is an oracle geometric statement.  The directly estimable distribution-level version is given next.

\subsection{Expected fitted gain and bias--variance conditions}
\label{subsec:expected-fitted-gain}
We next identify a second mechanism for positive gain: federation can improve prediction through variance reduction even when persistent target correction is absent. Define
\begin{align}
R_{L,j,t}&=\bm{E}_t^{\mathrm{fit}}\norm{L_{j,t}-m_{j,t}^{Q}}_{j,t}^2,\\
R_{F,j,t}&=\bm{E}_t^{\mathrm{fit}}\norm{F_t-m_{j,t}^{Q}}_{j,t}^2,\\
C_{LF,j,t}&=\bm{E}_t^{\mathrm{fit}}\inner{L_{j,t}-m_{j,t}^{Q}}{F_t-m_{j,t}^{Q}}_{j,t},\\
D_{LF,j,t}&=\bm{E}_t^{\mathrm{fit}}\norm{F_t-L_{j,t}}_{j,t}^2.
\end{align}

\begin{proposition}[Exact marginal gain]
\label{prop:marginal-gain-main}
For every $\alpha\in[0,1)$,
\begin{align}
\Gamma_{j,t}
&=\vartheta\{R_{L,j,t}-R_{F,j,t}+\alpha D_{LF,j,t}\}\\
&=2\vartheta\{R_{L,j,t}-C_{LF,j,t}\}-\vartheta^2D_{LF,j,t}.
\label{eq:marginal-gain-identity}
\end{align}
Consequently, some positive amount of borrowing helps if and only if
\begin{equation}
C_{LF,j,t}<R_{L,j,t},
\label{eq:some-borrowing-condition}
\end{equation}
and the optimal amount is
\begin{equation}
\vartheta_{j,t}^{\mathrm{opt}}
=\Pi_{[0,1]}\left\{\frac{R_{L,j,t}-C_{LF,j,t}}{D_{LF,j,t}}\right\},
\qquad D_{LF,j,t}>0.
\label{eq:optimal-borrowing}
\end{equation}
The sufficient condition $R_{F,j,t}<R_{L,j,t}$ guarantees positive gain for every $\alpha\in[0,1)$, equivalently for every nonzero borrowing amount $\vartheta\in(0,1]$.
\end{proposition}

Let $\overline L_{j,t}=\bm{E}_t^{\mathrm{fit}}L_{j,t}$ and $\overline F_t=\bm{E}_t^{\mathrm{fit}}F_t$.  Define the bias functions
\begin{equation}
\overline b_{L,j,t}=\overline L_{j,t}-m_{j,t}^{Q},
\qquad
\overline b_{F,j,t}=\overline F_t-m_{j,t}^{Q},
\end{equation}
and the integrated fitting variances and covariance
\begin{align}
V_{L,j,t}&=\bm{E}_t^{\mathrm{fit}}\norm{L_{j,t}-\overline L_{j,t}}_{j,t}^2,\\
V_{F,j,t}&=\bm{E}_t^{\mathrm{fit}}\norm{F_t-\overline F_t}_{j,t}^2,\\
C_{V,j,t}&=\bm{E}_t^{\mathrm{fit}}\inner{L_{j,t}-\overline L_{j,t}}{F_t-\overline F_t}_{j,t}.
\end{align}
Then \eqref{eq:some-borrowing-condition} is equivalent to
\begin{equation}
\norm{\overline b_{L,j,t}}_{j,t}^2+V_{L,j,t}
>\inner{\overline b_{L,j,t}}{\overline b_{F,j,t}}_{j,t}+C_{V,j,t}.
\label{eq:bias-variance-some}
\end{equation}
A stronger, transparent sufficient condition is
\begin{equation}
\norm{\overline b_{F,j,t}}_{j,t}^2+V_{F,j,t}
<\norm{\overline b_{L,j,t}}_{j,t}^2+V_{L,j,t}.
\label{eq:bias-variance-all}
\end{equation}
Condition \eqref{eq:bias-variance-all} says that the server's reduction in estimator variance exceeds any additional squared-bias cost.  It captures the setting in which a client's local data or procedure is subpar, while the remaining clients have lower variability and are not too far from the client's evaluation target.

The fitted gain components $A_{j,t}^{\mathrm{fit}}$ and $D_{j,t}^{\mathrm{fit}}$ can be estimated directly using the independent evaluation sample. Specifically, define
\begin{align}
\widehat A_{j,t}
&=\frac{1}{N_{j,t}^{\mathrm{eval}}}\sum_{r=1}^{N_{j,t}^{\mathrm{eval}}}
\{Y_{r,j,t}^{\mathrm{eval}}-L_{j,t}(X_{r,j,t}^{\mathrm{eval}})\}
\{F_t(X_{r,j,t}^{\mathrm{eval}})-L_{j,t}(X_{r,j,t}^{\mathrm{eval}})\},
\end{align}
and
\begin{align}
\widehat D_{j,t}
&=\frac{1}{N_{j,t}^{\mathrm{eval}}}\sum_{r=1}^{N_{j,t}^{\mathrm{eval}}}
\{F_t(X_{r,j,t}^{\mathrm{eval}})-L_{j,t}(X_{r,j,t}^{\mathrm{eval}})\}^2.
\end{align}
The resulting empirical gain curve is
\begin{align}
\widehat G_{j,t}(\vartheta)
&=2\vartheta\widehat A_{j,t}-\vartheta^2\widehat D_{j,t}.
\end{align}

The preceding results identify two mechanisms through which federated borrowing can improve client-specific prediction: correcting discrepancies between population procedure targets and reducing fitting variability. The empirical gain curve provides a direct way to assess predictive improvement using independent evaluation data.

\section{Inference for predictive gain}
\label{sec:gain-inference}

Statistical inference for predictive gain requires accounting for two distinct sources of uncertainty: the independent evaluation sample and the fitting samples used to construct the local and federated predictors. We first study inference for a single client, distinguishing conditional inference for the realized fitted predictors from procedure-level inference that incorporates fitting variability. We then extend the analysis to the client-indexed process, where the shared server estimator introduces dependence across clients.
\subsection{Conditional evaluation inference}
\label{subsec:conditional-evaluation}

For a fixed client $j$ and stage $t$, conditional on $\mathcal{F}_t^{\mathrm{fit}}$, the evaluation gain variables $W_{j,t}(Z_{r,j,t}^{\mathrm{eval}})$ are i.i.d. under $Q_{j,t}$. Define
\begin{align}
\sigma_{E,j,t}^2&=\Var\{W_{j,t}(Z)\mid\mathcal{F}_t^{\mathrm{fit}}\},\\
\beta_{E,j,t}&=
\frac{\bm{E}[|W_{j,t}(Z)-G_{j,t}^{\mathrm{fit}}|^3\mid\mathcal{F}_t^{\mathrm{fit}}]}
{\sigma_{E,j,t}^3}.
\end{align}
The detailed regular-conditional-distribution construction is given in Appendix~\ref{app:one-client-proof}.  The pathwise Berry--Esseen inequality gives
\begin{align}
\sup_x\left|\bm{P}\left[
\left.
\frac{\sqrt{N_{j,t}^{\mathrm{eval}}}
(\widehat G_{j,t}-G_{j,t}^{\mathrm{fit}})}
{\sigma_{E,j,t}}\le x
\right|\mathcal{F}_t^{\mathrm{fit}}\right]-\Phi(x)\right|
\le C_{\mathrm{BE}}\frac{\beta_{E,j,t}}
{\sqrt{N_{j,t}^{\mathrm{eval}}}}.
\label{eq:conditional-BE-main}
\end{align}
At fixed $t$, this implies almost-sure conditional weak convergence as the audit size diverges.  On the diagonal, the evaluation assumption makes $\beta_{E,j,t}=O_{\bm{P}}(1)$; hence $N_{j,t}^{\mathrm{eval}}\to\infty$ implies $\beta_{E,j,t}/\sqrt{N_{j,t}^{\mathrm{eval}}}=o_{\bm{P}}(1)$.  Sweeting's framework is used to formulate these conditional laws as random probability measures \citep{Sweeting1989}.  A one-sided conditional lower bound for the realized-fit gain is
\begin{equation}
\operatorname{LCB}^{\mathrm{fit}}_{j,t}(\delta)
=\widehat G_{j,t}-z_{1-\delta}
\frac{\widehat\sigma_{E,j,t}}{\sqrt{N_{j,t}^{\mathrm{eval}}}}.
\label{eq:fitted-gain-lcb-main}
\end{equation}
Here $\widehat\sigma_{E,j,t}^2$ denotes the sample variance of the observation-level evaluation gains $W_{j,t}(Z_{r,j,t}^{\mathrm{eval}})$.
Its coverage is conditional on the realized fitted predictors; it should not be interpreted as a confidence bound for $\Gamma_{j,t}$.

\subsection{One-client procedure-level theorem}
\label{subsec:one-client-theorem}

The fitting fluctuation $G_{j,t}^{\mathrm{fit}}-\Gamma_{j,t}$ is a scalar
functional of the local and federated Nadaraya--Watson estimators.  In the
nondegenerate first-order regime, its linear derivative is nondegenerate and the
centered quadratic gain term is of smaller order.

The following primitive conditions make explicit the regular regime used in
the procedure-level result.  Fix a client sequence $k_t\le K_t$, abbreviate
$Q_t=Q_{k_t,t}$, and specialize here to one-dimensional covariates and a
common bandwidth $h_t$.  The notation $\mathcal{X}$ denotes the compact
evaluation region.  In addition to kernel and moment regularity, the
conditions control the amount of information in the joint local--server fit.

\begin{assumption}[Kernel, support, and overlap]
\label{sg:ass:kernel}
The evaluation region $\mathcal X\subset\mathbb{R}$ is compact.  The kernel $K$ is
bounded, symmetric, compactly supported, and satisfies $\int K=1$.  With
$K_{h_t}(u)=h_t^{-1}K(u/h_t)$, the common bandwidth satisfies $h_t\downarrow0$.
The covariate densities under the client laws and under $Q_t$ are uniformly
bounded on a fixed neighborhood of $\mathcal X$.  The population local and
federated smoothed denominators are bounded below by a constant $c_0>0$ on
$\mathcal X$.  Empirical denominators are stabilized below $c_0/2$; the
stabilization is inactive with probability tending to one.
\end{assumption}

\begin{assumption}[Moments, smoothness, and clipping]
\label{sg:ass:moments}
For some $\delta>0$,
\[
\sup_{t,j,x}\bm{E}_{P_{j,t}}\{|Y|^{4+\delta}\mid X=x\}<\infty,
\qquad
\sup_{t,x}\bm{E}_{Q_t}\{|Y|^{3+\delta}\mid X=x\}<\infty.
\]
The densities, conditional first two moments, finite-bandwidth ratio targets,
and the coefficient functions defined in Appendix~\ref{app:one-client-proof} are uniformly bounded and
uniformly continuous on $\mathcal X$.  Fitted predictions are projected onto
a fixed deterministic clipping interval.  There is a constant
$c_{\mathrm{clip}}>0$ such that every finite-bandwidth local and federated
ratio target lies at distance at least $c_{\mathrm{clip}}$ from the boundary
of that interval.
\end{assumption}

To state the effective-information conditions, index the mutually
independent fitting observations by $a\in\mathcal{I}_t$.  Let $r_{a,L,t}$
and $r_{a,F,t}$ denote the coefficients with which observation $a$ enters
the local and federated kernel averages, respectively, and put
\begin{align}
\gamma_{a,t}&=|r_{a,L,t}|+|r_{a,F,t}|,\\
\mathfrak r_t&=\sum_{a\in\mathcal{I}_t}\gamma_{a,t}^{2},\\
\gamma_{\max,t}&=\max_{a\in\mathcal{I}_t}\gamma_{a,t}.
\end{align}
The scalar first-order fitting influence of observation $a$ is
$\lambda_{a,t}$, as defined in \eqref{sg:eq:lambda}.  Its oracle fitting
variance is $s_{\mathrm{fit},t}^{2}=\sum_{a\in\mathcal{I}_t}
\bm{E}_t^{\mathrm{fit}}\lambda_{a,t}^{2}$.

\begin{assumption}[Effective information and no dominant observation]
\label{sg:ass:rates}
As $t\to\infty$,
\begin{equation}
\frac{\mathfrak r_t}{h_t}\longrightarrow0,
\qquad
\frac{\sqrt{\mathfrak r_t}}{h_t}\longrightarrow0,
\qquad
\frac{\gamma_{\max,t}}{\sqrt{\mathfrak r_t}}\longrightarrow0.
\label{sg:eq:rate-assumptions}
\end{equation}
These conditions can be interpreted through the joint local--server effective fitting sample size $N_{\mathrm{eff},t}=\mathfrak r_t^{-1}$. The first two requirements are equivalently $N_{\mathrm{eff},t}h_t\to\infty$ and $\sqrt{N_{\mathrm{eff},t}}h_t\to\infty$, while the third prevents any single fitting observation from dominating. In general, $N_{\mathrm{eff},t}$ need not equal the total federated sample size $N_t$.
\end{assumption}

\begin{assumption}[Nondegenerate first-order information]
\label{sg:ass:nondegenerate}
There exist constants $0<c<C<\infty$ such that
\begin{equation}
c\,\mathfrak r_t
\le
s_{\mathrm{fit},t}^2
\le
C\,\mathfrak r_t
\label{sg:eq:sfit-order}
\end{equation}
for all sufficiently large $t$.
\end{assumption}
The nondegeneracy condition further ensures that the first-order fitting variance remains comparable to its natural scale $\mathfrak r_t$.

For the evaluation condition, let $\overline L_t=\bm{E}_t^{\mathrm{fit}}L_{k_t,t}$
and $\overline F_t=\bm{E}_t^{\mathrm{fit}}F_t$, and write $W_t^0$ for the
paired-gain function \eqref{eq:paired-gain-main} evaluated at these mean
predictors.  Set
\begin{equation}
v_{E,t}=\Var_{Q_t}\{W_t^0(Z)\}.
\end{equation}
In Appendix~\ref{app:one-client-proof}, the fitting observations are collected into a statistic $S_t$ satisfying $\mathcal F_t^{\mathrm{fit}}=\sigma(S_t)$. The quantity $\beta_t(S_t)$, formally defined in \eqref{sg:eq:beta}, is the conditional standardized third absolute moment of the paired gain and coincides with $\beta_{E,k_t,t}$ introduced in Subsection~\ref{subsec:conditional-evaluation}.
\begin{assumption}[Evaluation nondegeneracy]
\label{sg:ass:eval}
There are constants $0<c<C<\infty$ such that $c\le v_{E,t}\le C$.  The
standardized third moments $\beta_t(S_t)$ are tight.
\end{assumption}

The first four assumptions yield a nondegenerate first-order fitting
expansion, while the evaluation condition stabilizes the independent audit
fluctuation.  Consistent estimation of the fitting variance additionally
requires the cross-fitted nuisance condition in
Assumption~\ref{sg:ass:nuisance}, stated with the construction and proofs in
Appendix~\ref{app:one-client-proof}.

\begin{theorem}[Predictive-gain inference in the nondegenerate first-order regime]
\label{thm:one-client-main}
Fix a client sequence $k_t\le K_t$.  Under Assumptions~\ref{sg:ass:kernel}--\ref{sg:ass:eval}, the additional cross-fitted nuisance condition in Assumption~\ref{sg:ass:nuisance} of Appendix~\ref{app:one-client-proof}, and $N_{k_t,t}^{\mathrm{eval}}\to\infty$, there exist deterministic scales $s_{\mathrm{fit},t}$ and $s_{\mathrm{eval},t}$ and consistent estimators $\widehat s_{\mathrm{fit},t}$ and $\widehat s_{\mathrm{eval},t}$ such that
\begin{equation}
\frac{\widehat G_{k_t,t}-\Gamma_{k_t,t}}
{\{s_{\mathrm{fit},t}^2+s_{\mathrm{eval},t}^2\}^{1/2}}
\Longrightarrow N(0,1),
\label{eq:combined-gain-clt-main}
\end{equation}
\begin{equation}
\frac{\widehat s_{\mathrm{fit},t}^2+\widehat s_{\mathrm{eval},t}^2}
{s_{\mathrm{fit},t}^2+s_{\mathrm{eval},t}^2}
\xrightarrow{\bm{P}}1,
\end{equation}
and hence
\begin{equation}
\frac{\widehat G_{k_t,t}-\Gamma_{k_t,t}}
{\{\widehat s_{\mathrm{fit},t}^2+\widehat s_{\mathrm{eval},t}^2\}^{1/2}}
\Longrightarrow N(0,1).
\label{eq:studentized-gain-clt-main}
\end{equation}
The normalized fitting and evaluation fluctuations converge jointly to independent standard normal variables.
\end{theorem}
The theorem establishes a procedure-level normal approximation that incorporates both fitting and evaluation uncertainty. Although the evaluation sample is independent of the fitting data, the local and federated predictors are generally dependent because they share client observations. This dependence is accounted for in the fitting variance component, whereas the independent evaluation stage contributes a separate variance component. Their combination provides the appropriate uncertainty scale for inference on the marginal gain $\Gamma_{k_t,t}$.

For a deterministic threshold $\varepsilon_t$,
\begin{equation}
\bm{P}\{\widehat G_{k_t,t}>\varepsilon_t\}
=\Phi\left(
\frac{\Gamma_{k_t,t}-\varepsilon_t}
{\{s_{\mathrm{fit},t}^2+s_{\mathrm{eval},t}^2\}^{1/2}}
\right)+o(1).
\label{eq:positive-gain-probability-main}
\end{equation}
A one-sided lower confidence bound for the stage-$t$ marginal gain is
\begin{equation}
\operatorname{LCB}^{\mathrm{proc}}_{k_t,t}(\delta)
=\widehat G_{k_t,t}-z_{1-\delta}
\{\widehat s_{\mathrm{fit},t}^2+\widehat s_{\mathrm{eval},t}^2\}^{1/2}.
\label{eq:gain-lcb-main}
\end{equation}
This lower bound targets the repeated-fitting procedure gain $\Gamma_{k_t,t}$, not the particular realized fit.  The moving target should be retained.  If $\Gamma_{k_t,t}\to\Gamma$, replacement of $\Gamma_{k_t,t}$ by $\Gamma$ in a standardized margin additionally requires
\begin{equation}
\frac{\Gamma_{k_t,t}-\Gamma}{
\{s_{\mathrm{fit},t}^2+s_{\mathrm{eval},t}^2\}^{1/2}}
\longrightarrow0.
\label{eq:moving-gain-replacement}
\end{equation}
The preceding theorem applies to the nondegenerate first-order regime. When the first-order gain derivative vanishes, the linear approximation no longer captures the leading fitting fluctuation, and the quadratic term becomes relevant. This requires a separate boundary analysis based on a quadratic-martingale approximation.

\begin{theorem}[Exact-model boundary]
\label{thm:boundary-gain-main}
If the mean local and federated curves equal the evaluation regression, the gain derivative vanishes.  Under the Hall--de Jong \citep{Hall1984,deJong1987}, variance-estimation, and audit-stabilization conditions in Appendix~\ref{sg:thm:boundary-combined},
\begin{equation}
(\widehat G_{k_t,t}-\Gamma_{k_t,t})/\widehat s_{\mathrm{bd},t}\Longrightarrow N(0,1).
\end{equation}
This is a high-level quadratic-martingale result and may require additional communication.
\end{theorem}

\subsection{A worked feasible regime}
\label{subsec:worked-regime-main}
The preceding results distinguish the regular first-order regime from its exact-model boundary. We now provide a concrete example showing that the regular conditions are attainable under a simple federated Nadaraya--Watson model. Let two equally weighted clients have $X\sim\mathrm{Unif}[-1,1]$, fitting regressions $0$ and $2b$, and bounded centered noise; audit client 1 under regression $b$.  With $h_n=n^{-1/3}$ and independent pilot and audit samples of order $n$, the local and federated targets are $0$ and $b$, so
\begin{align}
G_1^P(\alpha)&=(1-\alpha^2)\norm{b}_{Q_{1,X}}^2>0,\\
\Gamma_{1,n}&\to(1-\alpha^2)\norm{b}_{Q_{1,X}}^2.
\end{align}
Appendix~\ref{app:worked-regime} verifies all regular fitting, audit, nuisance, variance, sub-gamma, and communication conditions.  Auditing the same fitted procedures under the original client law gives negative transfer.

\subsection{Client-indexed gain process}
\label{subsec:client-gain-process}

Single-client inference does not capture the dependence across clients induced by the common server fit. We therefore consider joint gain inference for fixed client blocks. Unlike coordinatewise normal approximations, the joint limit must retain the shared fitting influence of each observation on multiple clients. For fixed $r$, write
\begin{equation}
\widehat{\boldsymbol G}_{r,t}=(\widehat G_{1,t},\ldots,\widehat G_{r,t})^{\mathsf T},
\qquad
\boldsymbol\Gamma_{r,t}=(\Gamma_{1,t},\ldots,\Gamma_{r,t})^{\mathsf T},
\end{equation}
and let $D_{r,t}$ be the diagonal matrix of total standard deviations.  The off-diagonal covariance comes from the fitting samples and the common server estimator; independent evaluation streams add only diagonal covariance.

For the joint first-order approximation, write
$\boldsymbol G_{r,t}^{\mathrm{fit}}=(G_{1,t}^{\mathrm{fit}},\ldots,
G_{r,t}^{\mathrm{fit}})^{\mathsf T}$.  The centered fitting-gain vector admits
\begin{equation}
\boldsymbol G_{r,t}^{\mathrm{fit}}-\boldsymbol\Gamma_{r,t}
=\sum_{a\in\mathcal{I}_t}\bm\lambda_{a,t}^{(r)}
+\bm r_{r,t}^{\mathrm{fit}},
\end{equation}
where the vectors $\bm\lambda_{a,t}^{(r)}$ are independent and centered
across fitting observations, but their coordinates may be correlated because
they share the server fit.  For each fixed block, the key conditions are a
negligible standardized remainder, convergence of the normalized fitting
covariance $A_{r,t}\to A_r$, and the multivariate Lyapunov bound
\begin{equation}
\sum_{a\in\mathcal{I}_t}\bm{E}_t^{\mathrm{fit}}
\left\|D_{r,t}^{-1}\bm\lambda_{a,t}^{(r)}\right\|^{2+\delta}
\longrightarrow0
\qquad\text{for some }\delta>0.
\end{equation}
The independent audit streams separately give a conditional Gaussian limit
with diagonal covariance $B_r$, provided their standardized variances
stabilize.  The precise fitting-representation, remainder, Lyapunov, and
evaluation conditions are given in Assumptions~\ref{gp:ass:fit-representation},
\ref{gp:ass:fit-vector-clt}, and \ref{gp:ass:eval-vector} of
Appendix~\ref{app:client-process-proof}.  They are imposed for each fixed
$r$; no central limit theorem uniform over all $K_t$ clients is asserted.

\begin{theorem}[Client-indexed gain process]
\label{thm:client-process-main}
Under Assumptions~\ref{gp:ass:fit-representation}, \ref{gp:ass:fit-vector-clt}, and \ref{gp:ass:eval-vector} of Appendix~\ref{app:client-process-proof}, for every fixed $r$,
\begin{equation}
D_{r,t}^{-1}(\widehat{\boldsymbol G}_{r,t}-\boldsymbol\Gamma_{r,t})
\Longrightarrow N_r(0,R_r).
\label{eq:finite-block-gain-main}
\end{equation}
The matrix $R_r=A_r+B_r$ retains the non-diagonal fitting dependence created by the common server, while independent audit streams contribute the diagonal matrix $B_r$.  Under the aggregate cross-fitted covariance conditions in Assumption~\ref{gp:ass:cov-estimation} of Appendix~\ref{app:client-process-proof}, a consistent estimator of the stage-$t$ covariance is available for every fixed block.  The additional compatibility condition \eqref{gp:eq:compatibility} (Assumption~\ref{gp:ass:compatible}) ensures that these fixed-block limits form a consistent family, which can be embedded in the product topology; this is an organizational consequence rather than a separate inferential claim.
\end{theorem}

For deterministic thresholds $\varepsilon_{j,t}$, define the total standard deviation and standardized margin for client $j$ as
\begin{align}
s_{j,t}&=\left(s_{\mathrm{fit},j,t}^{2}+s_{\mathrm{eval},j,t}^{2}\right)^{1/2},\\
\gamma_{j,t}&=\frac{\Gamma_{j,t}-\varepsilon_{j,t}}{s_{j,t}}.
\end{align}
Here and throughout the subsequent gain-process analysis, $s_{j,t}$ denotes a scalar total standard deviation, distinct from the density-weighted second-moment function $s_{j,t}(z)$ used in the pointwise NW analysis. The coordinatewise normal-approximation error is defined by
\begin{align}
\delta_{j,t}
&=\sup_{u\in\mathbb{R}}\left|
\bm{P}\left\{
\frac{\widehat G_{j,t}-\Gamma_{j,t}}{s_{j,t}}\le u
\right\}-\Phi(u)
\right|.
\end{align}

Theorem~\ref{thm:client-process-main} gives Gaussian orthant limits for every fixed client block. For the moving all-client event, the sufficient summability condition
\begin{align}
\sum_{j=1}^{K_t}\Phi(-\gamma_{j,t})
+\sum_{j=1}^{K_t}\delta_{j,t}&\longrightarrow0
\label{eq:all-client-summability-main}
\end{align}
implies
\begin{align}
\bm{P}\{\widehat G_{j,t}>\varepsilon_{j,t}\text{ for every }j\le K_t\}
&\longrightarrow1.
\end{align}
This sufficient condition does not require independence across clients, but it requires the accumulated approximation errors and Gaussian lower-tail probabilities to vanish as $K_t$ grows.

\section{KL-robust predictive-gain certificates}
\label{sec:kl-robust}

Positive predictive gain under the nominal evaluation law does not necessarily persist when the client population changes. To assess the robustness of predictive improvement, we consider perturbations of the joint evaluation distribution within a Kullback--Leibler (KL) neighborhood. Conditional on the fitted predictors, we first characterize the worst-case expected gain and derive a variance-sensitive lower certificate. We then develop statistical inference for this certificate, accounting for both fitting and evaluation uncertainty.

\subsection{Exact robust value and variance-sensitive certificate}
\label{subsec:kl-certificate}
For a client-specific radius $\rho_{j,t}\ge0$, define the fitted KL-robust gain
\begin{equation}
\underline G_{j,t}^{\mathrm{fit}}(\rho_{j,t})
=\inf_{\substack{Q\ll Q_{j,t}\\ \operatorname{KL}(Q\|Q_{j,t})\le\rho_{j,t}}}
\bm{E}_Q W_{j,t}.
\label{eq:exact-kl-value-main}
\end{equation}
Assume $\bm{E}_{Q_{j,t}}|W_{j,t}|<\infty$ and
$\bm{E}_{Q_{j,t}}\exp(-\lambda W_{j,t})<\infty$ for at least one
$\lambda>0$.  Then, with terms having an infinite log-Laplace transform
interpreted as $-\infty$,
\begin{equation}
\underline G_{j,t}^{\mathrm{fit}}(\rho_{j,t})
=\sup_{\eta>0}\left\{-\eta\log\bm{E}_{Q_{j,t}}
\exp\left(-\frac{W_{j,t}}{\eta}\right)-\eta\rho_{j,t}\right\}.
\label{eq:kl-dual-main}
\end{equation}
For $\rho_{j,t}>0$, the nominal law $Q_{j,t}$ is a Slater point; the case
$\rho_{j,t}=0$ follows by the integrability assumption and the limit
$\eta\to\infty$.

Let
\begin{equation}
\mu_{j,t}=G_{j,t}^{\mathrm{fit}},
\qquad
v_{j,t}=\Var(W_{j,t}\mid\mathcal{F}_t^{\mathrm{fit}}).
\end{equation}

Let $c_{j,t}\ge0$ be a deterministic client-specific scale parameter governing the conditional lower tail of the paired gain. The following assumption specifies the sub-gamma condition required for the variance-sensitive certificate.

\begin{assumption}[Conditional sub-gamma lower tail]
\label{kl:ass:subgamma}
On a probability-one set, for every client $j\le K_t$ and every
$0\le\lambda<c_{j,t}^{-1}$,
\begin{equation}
\log\bm{E}\left[
\left.
\exp\{\lambda(\mu_{j,t}-W_{j,t}(Z))\}
\right|
\mathcal{F}_t^{\mathrm{fit}}
\right]
\le
\frac{\lambda^{2}v_{j,t}}{2(1-c_{j,t}\lambda)}.
\label{kl:eq:subgamma}
\end{equation}
When $c_{j,t}=0$, the domain in \eqref{kl:eq:subgamma} is $\lambda\ge0$.
\end{assumption}
Then
\begin{equation}
\underline G_{j,t}^{\mathrm{fit}}(\rho_{j,t})
\ge C_{j,t}^{\mathrm{fit}}(\rho_{j,t})
:=\mu_{j,t}-\sqrt{2\rho_{j,t}v_{j,t}}-c_{j,t}\rho_{j,t}.
\label{eq:kl-certificate-main}
\end{equation}
Thus certificate positivity is sufficient for positivity of the exact robust value.

\subsection{Certificate inference and client process}
\label{subsec:kl-inference}

The evaluation sample estimates $(\mu_{j,t},v_{j,t})$, and, away from $v_{j,t}=0$, the delta method yields an empirical certificate $\widehat C_{j,t}$.  Its fitting fluctuation depends jointly on the scalar fitting expansions of the gain mean and gain variance.  The corresponding joint mean--variance fitting condition is Assumption~\ref{kl:ass:mv-fitting} in Appendix~\ref{app:kl-process-proof}.

Specifically, write $W_{j,t,r}=W_{j,t}(Z_{r,j,t}^{\mathrm{eval}})$ for the observed paired gains. Define
\begin{align}
\widehat\mu_{j,t}
&=\frac{1}{N_{j,t}^{\mathrm{eval}}}
\sum_{r=1}^{N_{j,t}^{\mathrm{eval}}}W_{j,t,r},\\
\widehat v_{j,t}
&=\frac{1}{N_{j,t}^{\mathrm{eval}}}
\sum_{r=1}^{N_{j,t}^{\mathrm{eval}}}
(W_{j,t,r}-\widehat\mu_{j,t})^2,\\
\widehat C_{j,t}
&=\widehat\mu_{j,t}
-\sqrt{2\rho_{j,t}\widehat v_{j,t}}
-c_{j,t}\rho_{j,t}.
\end{align}

To state the regular fitting condition, write $\overline v_{j,t}=\bm{E}_t^{\mathrm{fit}}v_{j,t}$.  Let $s_{C,\mathrm{fit},j,t}^{2}$ denote the variance of the scalar fitting-certificate influence defined in \eqref{kl:eq:fit-scale-one-client}, and let $\mathfrak r_{j,t}$ denote the clientwise effective-information scale.  The following condition ensures that fitting uncertainty remains nondegenerate and comparable to that scale.

\begin{assumption}[Regular fitting-certificate regime]
\label{kl:ass:fit-regular}
For every client sequence covered by the theorem,
\begin{equation}
\inf_t\overline v_{j,t}>0,
\qquad
\sup_t\rho_{j,t}<\infty,
\qquad
c\mathfrak r_{j,t}
\le s_{C,\mathrm{fit},j,t}^{2}
\le C\mathfrak r_{j,t}
\label{kl:eq:fit-regular}
\end{equation}
for finite positive constants $c,C$.
\end{assumption}

The evaluation-side regularity requirements, including the conditional gain-variance floor, conditional influence-variance floor, and moment bound, are stated in Assumption~\ref{kl:ass:eval-regular} of Appendix~\ref{app:kl-process-proof}.  The finite-block covariance and Lyapunov conditions remain there as well.

\begin{theorem}[KL-certificate process in the regular positive-variance regime]
\label{thm:kl-process-main}
Under the conditional sub-gamma hypothesis (Assumption~\ref{kl:ass:subgamma}), the nondegenerate first-order assumptions (including Assumptions~\ref{kl:ass:mv-fitting} and \ref{kl:ass:fit-regular}), with $\overline v_{j,t}=\bm{E}_t^{\mathrm{fit}}v_{j,t}$, and the gain-variance floors
\[
\inf_t\min_{j\le r}\overline v_{j,t}>0,
\qquad
\bm{P}\!\left\{\min_{j\le r}v_{j,t}\ge c_v\right\}\longrightarrow1
\]
for some $c_v>0$, for every fixed $r$,
\begin{equation}
(D^C_{r,t})^{-1}
(\widehat{\boldsymbol C}_{r,t}-\boldsymbol{\mathcal C}_{r,t})
\Longrightarrow N_r(0,R^C_r),
\label{eq:finite-block-kl-main}
\end{equation}
where $\mathcal C_{j,t}=\bm{E}_t^{\mathrm{fit}}C_{j,t}^{\mathrm{fit}}$.  Compatible finite-dimensional limits define a Gaussian certificate process in the product topology.  The finite-block covariance matrices admit cross-fitted estimators under the aggregate nuisance and moment conditions in Appendix~\ref{app:kl-process-proof}; no general primitive bandwidth theorem is claimed for those nuisance estimators.
\end{theorem}
The detailed results and complete proofs underlying Theorem~\ref{thm:kl-process-main} are provided in Appendix~\ref{app:kl-process-proof}.

For fixed blocks, Theorem~\ref{thm:kl-process-main} gives Gaussian orthant probabilities for the marginal certificates $\mathcal C_{j,t}$.  A positive conditional lower confidence bound for $C_{j,t}^{\mathrm{fit}}$ instead supports robust benefit for the realized fitted pair.  Positivity of $\widehat C_{j,t}$ alone is not a certificate.  For all clients, the sufficient summability condition in Appendix~\ref{app:kl-process-proof} controls simultaneous empirical thresholds.

The certificate also yields a natural measure of how much distributional perturbation the predictive gain can tolerate. For $\mu>0$ and $v>0$, certificate positivity is governed by the critical radius solving
\begin{equation}
\mu=\sqrt{2\rho v}+c\rho.
\label{eq:cert-radius-balance-main}
\end{equation}
When $c=0$, $\rho^{\mathrm{cert}}=\mu^2/(2v)$; for $c>0$,
\begin{equation}
\rho^{\mathrm{cert}}
=\left\{\frac{\sqrt{2v+4c\mu}-\sqrt{2v}}{2c}\right\}^2.
\label{eq:cert-radius-main}
\end{equation}
Thus no universal sample-size radius follows from the regular theorem.  If a specified model verifies $\mu_t/q_t\to m>0$, $v_t/q_t\to\sigma_\Delta^2>0$, and $c_t\to c$, then for $\rho_t=rq_t$ the normalized certificate converges to $m-\sqrt{2r\sigma_\Delta^2}-cr$.

At a vanishing gain variance, $v\mapsto\sqrt{v}$ is nonsmooth.  For example, with $Y\sim N(0,\sigma^2)$, $\alpha=0$, and constant predictors $L=a$, $F=0$,
\begin{equation}
C^{\mathrm{fit}}(a)=a^2-2\sigma\sqrt{2\rho}|a|.
\label{eq:kl-absolute-value-boundary}
\end{equation}
The regular Gaussian certificate theorem therefore does not follow from the nominal quadratic boundary theorem.  A fresh client audit sample provides the empirical gain mean, variance, and scalar log-Laplace summaries; the fitting-covariance calculations require the additional fitting-side cross-fitting information stated in the Appendix.

\section{$L_2$-robust regression-function perturbations}
\label{sec:l2-robust}
We now consider a complementary robustness model in which the evaluation regression function is perturbed within an $L_2$ neighborhood, while the evaluation covariate distribution remains fixed. Unlike the KL formulation, which allows perturbations of the joint evaluation law, this model focuses on regression-function misspecification. We derive the exact least-favorable perturbation and then study inference for the resulting robust predictive gain.
\subsection{Exact least-favorable direction}
\label{subsec:l2-exact}
Let
\begin{equation}
d^{B}_{j,t}(x)=M_{j,t}(x)-L_{j,t}(x)=\vartheta\{F_t(x)-L_{j,t}(x)\}.
\end{equation}
For radius $\tau_{j,t}\ge0$, perturb the benchmark regression function within
\begin{equation}
\{h:\norm{h-m_{j,t}^{Q}}_{j,t}\le\tau_{j,t}\}.
\end{equation}
The least-favorable perturbation acts against the borrowing direction
$d^B_{j,t}$.
\begin{proposition}[Exact least-favorable regression function]
\label{prop:l2-main}
If $\norm{d^{B}_{j,t}}_{j,t}>0$, the least-favorable regression function is
\begin{equation}
m_{j,t}^{\circ}(x)
=m_{j,t}^{Q}(x)-\tau_{j,t}\frac{d^{B}_{j,t}(x)}{\norm{d^{B}_{j,t}}_{j,t}},
\label{eq:l2-direction-main}
\end{equation}
and the exact fitted robust value is
\begin{equation}
\mathcal R_{j,t}^{\mathrm{fit}}
=G_{j,t}^{\mathrm{fit}}-2\tau_{j,t}\norm{d^{B}_{j,t}}_{j,t}.
\label{eq:l2-value-main}
\end{equation}
If $\norm{d^B_{j,t}}_{j,t}=0$, the local and personalized predictors coincide $Q_{j,t,X}$-almost surely, and the gain is zero.
\end{proposition}

\subsection{Inference and client process}
\label{subsec:l2-inference}

Let $\overline d^{B}_{j,t}=\vartheta(\overline F_t-\overline L_{j,t})$.  The regular inference theory requires this deterministic mean borrowing direction to remain separated from zero, together with bounded robustness radii.

For feasible inference, let $\widehat\nu_{j,t}$ be the grid-based estimator of $\norm{d^B_{j,t}}_{j,t}$ constructed using estimated evaluation-covariate weights, as detailed in Appendix~\ref{app:l2-process-proof}. Define the empirical robust value
\begin{align}
\widehat{\mathcal R}_{j,t}
&=\widehat G_{j,t}-2\tau_{j,t}\widehat\nu_{j,t}.
\end{align}
\begin{assumption}[Persistent borrowing direction]
\label{l2:ass:persistent-direction}
For the client sequences covered by the theorem, there is a constant
$c_v>0$ such that
\begin{equation}
\inf_t\norm{\overline d^{B}_{j,t}}_{j,t}\ge c_v.
\label{l2:eq:direction-floor}
\end{equation}
The radii satisfy $\sup_{j,t}\tau_{j,t}<\infty$.
\end{assumption}

Under this condition, the norm has the scalar expansion
\begin{equation}
\norm{d^{B}_{j,t}}_{j,t}-\norm{\overline d^{B}_{j,t}}_{j,t}
=\frac{\inner{\overline d^{B}_{j,t}}{d^{B}_{j,t}-\overline d^{B}_{j,t}}_{j,t}}
{\norm{\overline d^{B}_{j,t}}_{j,t}}+r^V_{j,t},
\end{equation}
with a quadratic remainder.  Hence the robust-value influence equals the nominal gain influence minus twice the radius times the norm influence.

\begin{theorem}[Client-indexed $L_2$-robust gain process]
\label{thm:l2-process-main}
Under Assumption~\ref{l2:ass:persistent-direction} and the remaining fitting, evaluation, and finite-block conditions in Appendix~\ref{app:l2-process-proof}, for every fixed $r$,
\begin{equation}
(D^{\mathcal R}_{r,t})^{-1}
(\widehat{\boldsymbol{\mathcal R}}_{r,t}-\boldsymbol{\mathfrak R}_{r,t})
\Longrightarrow N_r(0,R^{\mathcal R}_r),
\label{eq:finite-block-l2-main}
\end{equation}
where $\mathfrak R_{j,t}=\bm{E}_t^{\mathrm{fit}}\mathcal R_{j,t}^{\mathrm{fit}}$.  Compatible covariance limits yield a Gaussian process in the product topology.  Studentization and grid implementation require the aggregate cross-fitting and grid-norm conditions in Appendix~\ref{app:l2-process-proof}; they are not presented as primitive consequences of the basic two-summary protocol.
\end{theorem}
When the mean borrowing direction vanishes, the regular norm expansion breaks down and requires a separate boundary analysis.
\begin{theorem}[Exact-zero mean borrowing direction]
\label{thm:l2-vanishing-main}
When the deterministic mean borrowing direction is zero at every stage, the oracle-norm robust value has a studentized Gaussian limit under the linear--quadratic martingale and audit conditions in Appendix~\ref{app:l2-process-proof}.  The natural local radius is $\tau_{j,t}\asymp\nu_{j,t}$, where $\nu_{j,t}^2=\bm{E}_t^{\mathrm{fit}}\norm{d^B_{j,t}}_{j,t}^2$; leverage-based simplifications require directional noncancellation.  Local-to-zero directions form a separate transition regime.
\end{theorem}
Estimating \(u_{j,t}=\norm{d^B_{j,t}}_{j,t}>0\) from the audit covariates adds
\(
-\frac{\tau_{j,t}}{u_{j,t}}
\left\{d^B_{j,t}(X)^2-u_{j,t}^2\right\}
\) to the gain influence, so its variance must include norm-estimation uncertainty
and the corresponding covariance. Grid or auxiliary estimates are ignorable only
when their error is negligible on the robust-value standard-error scale. At an exact zero mean direction, let $S^B_{j,t}=\norm{d^B_{j,t}}_{j,t}^2$. Oracle-to-feasible transfer requires
\(
\frac{
\tau_{j,t}
\left|\widehat\nu_{j,t}-\sqrt{S^B_{j,t}}\right|
}{
s_{\mathcal R,0,j,t}
}
\xrightarrow{\bm{P}} 0\). See Appendix~\ref{app:l2-process-proof}.

\section{Numerical Experiments}
\label{sec:supp-numerical}

This section presents numerical experiments examining the theoretical results developed above.  We separate the experiments according to their theoretical role: basic verification of the two-stage gain inference, the regular bias-correction/variance-rescue regime, exact-boundary behavior, client-indexed joint inference, and the KL and $L_2$ robust extensions.  Unless otherwise stated, fitting and audit samples are generated independently, and response errors are independent centered Gaussian variables with the standard deviations specified below.  Monte Carlo benchmarks are computed from independent replications and are therefore separated from the replications used to assess interval coverage and variance estimation.

\subsection{Experiment 0: Two-client verification of two-stage gain inference}
\label{sec:supp-exp0}

\paragraph{Design.}
We begin with a two-client experiment designed to isolate the distinction between realized-fit and procedure-level uncertainty and to verify the local--server covariance term in the fitting variance.  Both fitting clients have
\begin{align}
X \sim \operatorname{Unif}[-1,1],
\end{align}
equal aggregation weights $w_1=w_2=1/2$, and regression functions
\begin{align}
m_1^{\mathrm{fit}}(x)=0, \qquad m_2^{\mathrm{fit}}(x)=2b(x), \qquad b(x)=1+\frac{x}{2}.
\end{align}
Hence the population federated regression is
\begin{align}
m_F(x)=b(x).
\end{align}
The target client is client~1.  Its independent audit sample has
\begin{align}
X \sim \operatorname{Unif}[-0.5,0.5], \qquad m_1^Q(x)=\eta b(x),
\end{align}
where
\begin{align}
\eta\in\{0,0.25,1\}.
\end{align}
With local weight $\alpha=0.5$, these choices generate negative-, near-zero-, and positive-gain regimes, respectively.  The fitting and audit noise standard deviations are both $1$, each fitting client has $n_j=300$ observations, and the audit sample has size $500$.  We use the Epanechnikov kernel with
\begin{align}
h=n_j^{-1/3}=0.1494.
\end{align}
The audit support lies strictly inside the fitting support so that this experiment does not confound the inferential comparison with boundary-kernel effects.

For the ideal population regressions, the gain is
\begin{align}
G^P(\eta)=(1-\alpha)\{2\eta-(1-\alpha)\}\bm{E}_Q[b(X)^2].
\end{align}
The finite-sample procedure-level target
\begin{align}
\Gamma=\bm{E}_{\mathrm{fit}}[G^{\mathrm{fit}}]
\end{align}
is evaluated using $3000$ independent reference replications.  Coverage and variance calibration are then assessed using a separate set of $1000$ Monte Carlo replications.  To isolate the fitting-variance formula itself, this experiment uses the oracle finite-bandwidth nuisance quantities implied by the known data-generating process.

This design extends Example~\ref{ex:two-client-shift} across negative-,
near-zero-, and positive-gain regimes and directly examines the distinction
between conditional realized-fit inference in
\eqref{eq:fitted-gain-lcb-main} and procedure-level inference in
Theorem~\ref{thm:one-client-main}, including the shared local--server
covariance retained in the fitting variance.

\paragraph{Results.}
Table~\ref{tab:supp-exp0} summarizes the main inferential diagnostics.  The realized-fit interval, which conditions on the fitted local and federated predictors and therefore uses only evaluation uncertainty, has empirical coverage essentially equal to the nominal $95\%$ level in all three regimes.  For the marginal target $\Gamma$, incorporating both fitting and evaluation uncertainty also gives coverage close to $95\%$.  In contrast, treating $\Gamma$ as if only evaluation uncertainty were present produces substantial undercoverage.

The comparison between the correct and independence-based fitting-variance estimators also demonstrates the importance of retaining shared local--server fitting dependence.  The correct fitting-variance estimator is close to its Monte Carlo benchmark in all three regimes, with ratios between $0.938$ and $0.980$.  The independence approximation can instead substantially under- or over-estimate the fitting variance.  In particular, its Monte Carlo variance ratios are $0.626$, $0.740$, and $1.354$ across the negative-, near-zero-, and positive-gain regimes.  The sign of the estimated local--server covariance contribution changes across these regimes, explaining why omission of this term need not be conservative in a fixed direction.

\begin{table}[t]
\centering
\caption{Two-client verification of realized-fit and procedure-level gain inference.  ``Correct'' retains fitting uncertainty and local--server covariance; ``Indep.'' drops the shared-data covariance term; and ``Eval.'' uses evaluation variance only.  Variance ratios compare the mean estimated fitting variance with its Monte Carlo benchmark.}
\label{tab:supp-exp0}
\small
\setlength{\tabcolsep}{4.5pt}
\begin{tabular}{c l rr cccc cc}
\toprule
& & \multicolumn{2}{c}{Gain} & \multicolumn{4}{c}{95\% coverage} & \multicolumn{2}{c}{$\widehat V_{\rm fit}/V_{\rm fit}^{\rm MC}$} \\
\cmidrule(lr){3-4}\cmidrule(lr){5-8}\cmidrule(lr){9-10}
$\eta$ & Regime & $G^P$ & $\Gamma$ &
$G^{\rm fit}$ & Correct & Indep. & Eval. &
Correct & Indep. \\
\midrule
0.00 & Negative  & $-0.255$ & $-0.250$ & 0.950 & 0.938 & 0.901 & 0.756 & 0.980 & 0.626 \\
0.25 & Near-zero & $ 0.000$ & $ 0.005$ & 0.953 & 0.947 & 0.925 & 0.729 & 0.967 & 0.740 \\
1.00 & Positive  & $ 0.766$ & $ 0.769$ & 0.951 & 0.951 & 0.977 & 0.591 & 0.938 & 1.354 \\
\bottomrule
\end{tabular}
\end{table}

\subsection{Experiment 1: Bias correction and variance rescue}
\label{sec:num-exp1}

\paragraph{Design.}
We next study how federated gain transitions between bias correction and variance rescue in a heterogeneous multi-client system.  We use $K=30$ clients, with client~1 designated as the target client.  Each client contributes $n_j=200$ fitting observations, and an independent target-client audit sample of size $200$ is used for gain evaluation.  Since the fitting sample sizes are equal, the server aggregation weights are equal across clients.  The local weight is fixed at $\alpha=0.5$.

All covariates are supported on $[-1,1]$.  For client $j$, the fitting covariate distribution is a truncated normal distribution with standard deviation $0.6$.  The target-client mean is $\mu_1=0$, while the auxiliary-client means are equally spaced over $[-0.7,0.7]$.  The target audit covariate distribution is the same truncated normal distribution as that of the target fitting client.

The common baseline regression and two perturbation directions are
\begin{align}
f_0(x)=1.5\sin(\pi x)+0.5x,
\end{align}
\begin{align}
q(x)=1+x,
\end{align}
and
\begin{align}
r(x)=x^2-\frac{1}{3}.
\end{align}
The target fitting regression is
\begin{align}
m_1^{\mathrm{fit}}(x)=f_0(x)+b_Lq(x),
\end{align}
whereas auxiliary client $j$ has fitting regression
\begin{align}
m_j^{\mathrm{fit}}(x)=f_0(x)+b_Fq(x)+\delta_jr(x), \qquad j=2,\ldots,K.
\end{align}
The coefficients $\delta_j$ are generated once from a centered Gaussian sample with standard deviation $0.2$ and then recentered so that their empirical mean across auxiliary clients is exactly zero.  The audit regression is
\begin{align}
m_1^Q(x)=f_0(x).
\end{align}
Thus $b_L$ controls persistent target-local fitting bias relative to the audit population, while $b_F$ controls how much of the same bias direction is retained by the auxiliary clients.  We consider the grid
\begin{align}
b_L\in\{0.25,0.50,0.75\}, \qquad b_F\in\{0,0.10,0.25\}.
\end{align}

The response-noise standard deviation is $0.5$ for both target and auxiliary clients.  Local regressions are estimated by Nadaraya--Watson smoothing with the Epanechnikov kernel and bandwidth
\begin{align}
h_j=n_j^{-1/3}.
\end{align}
For each setting, $1000$ independent fitting replications are used to estimate the marginal gain and its Monte Carlo variance components, and a separate set of $1000$ fitting/audit replications is used to evaluate variance estimation and interval coverage.

This experiment therefore illustrates the bias--variance mechanism in
Proposition~\ref{prop:marginal-gain-main} and evaluates the procedure-level
inference of Theorem~\ref{thm:one-client-main} in the regular first-order
regime.

\paragraph{Gain decomposition.}
For each setting, we decompose the Monte Carlo marginal gain as
\begin{align}
\Gamma=\Delta_{\mathrm{bias}}+\Delta_{\mathrm{var}},
\end{align}
where
\begin{align}
\Delta_{\mathrm{bias}}=\| \bar m_L-m_Q\|_Q^2-\|\alpha\bar m_L+(1-\alpha)\bar m_F-m_Q\|_Q^2
\end{align}
is the reduction in squared bias and
\begin{align}
\Delta_{\mathrm{var}}=V_L-\{\alpha^2V_L+(1-\alpha)^2V_F+2\alpha(1-\alpha)C_{LF}\}
\end{align}
is the reduction in fitting variance.  Here $\bar m_L$ and $\bar m_F$ denote the Monte Carlo mean local and federated fitted regressions, and $V_L$, $V_F$, and $C_{LF}$ are their integrated fitting variances and covariance.  To summarize the dominant mechanism, we use
\begin{align}
D=\frac{\Delta_{\mathrm{var}}-\Delta_{\mathrm{bias}}}{|\Delta_{\mathrm{var}}|+|\Delta_{\mathrm{bias}}|},
\end{align}
so that $D<0$ indicates bias-correction dominance and $D>0$ indicates variance-rescue dominance.

\begin{figure}[t]
\centering
\begin{minipage}[t]{0.48\linewidth}
    \centering
    \includegraphics[width=\linewidth]{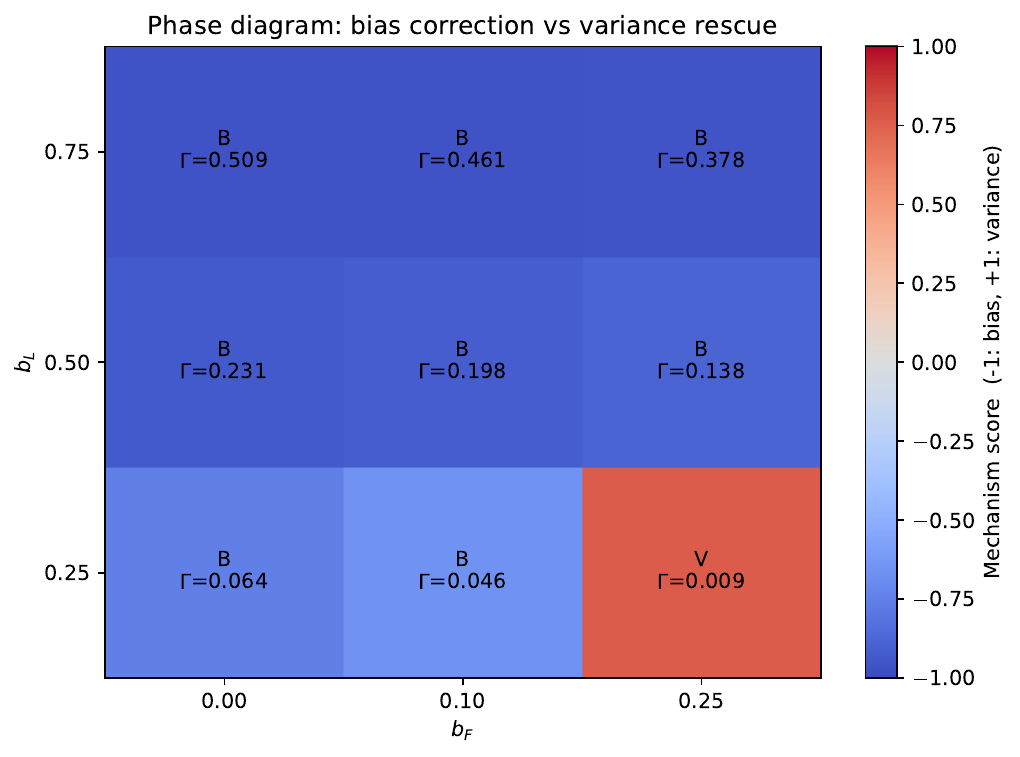}
    \vspace{-2mm}
    \textbf{(a)}
\end{minipage}
\hfill
\begin{minipage}[t]{0.48\linewidth}
    \centering
    \includegraphics[width=\linewidth]{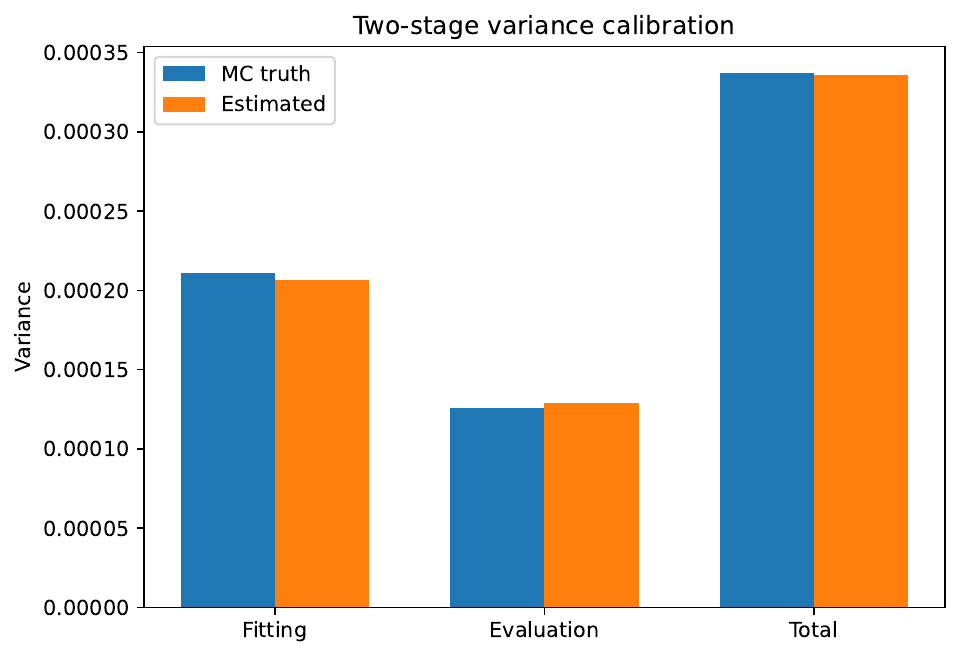}
    \vspace{-2mm}
    \textbf{(b)}
\end{minipage}
\caption{Regular-gain diagnostics for Experiment~1.  (a) Phase diagram of bias correction and variance rescue over the $(b_L,b_F)$ grid.  (b) Fitting, evaluation, and total variance calibration at $(b_L,b_F)=(0.25,0)$.}
\label{fig:num-exp1-summary}
\end{figure}

Table~\ref{tab:num-exp1-grid} reports the complete $3\times3$ grid, and Figure~\ref{fig:num-exp1-summary}(a) visualizes the corresponding phase transition.  The variance contribution remains remarkably stable, at approximately $7.9\times10^{-3}$ across all settings, whereas the bias contribution changes substantially with $(b_L,b_F)$.  Consequently, most settings are strongly bias-correction dominated, but when $(b_L,b_F)=(0.25,0.25)$ the remaining bias contribution is nearly zero and variance rescue becomes the dominant source of gain.  Positive federated gain can therefore persist even when borrowing no longer materially corrects target-local bias.

\begin{table}[t]
\centering
\caption{Complete grid for Experiment~1.  Gain components are reported on the original scale.  The mechanism score $D$ is negative under bias-correction dominance and positive under variance-rescue dominance.  The final two columns report empirical $95\%$ coverage for the correct two-stage interval and for an interval using evaluation variance only.}
\label{tab:num-exp1-grid}
\small
\setlength{\tabcolsep}{5pt}
\begin{tabular}{ccrrrrrrr}
\toprule
$b_L$ & $b_F$ & $\Gamma$ & $\Delta_{\rm bias}$ & $\Delta_{\rm var}$ & $D$ &
$\widehat V_{\rm fit}/V_{\rm fit}^{\rm MC}$ & 2-stage & Eval. \\
\midrule
0.25 & 0.00 & 0.0639 & 0.0560 & 0.0079 & $-0.753$ & 0.979 & 0.947 & 0.745 \\
0.25 & 0.10 & 0.0462 & 0.0383 & 0.0079 & $-0.658$ & 0.962 & 0.939 & 0.657 \\
0.25 & 0.25 & 0.0089 & 0.0011 & 0.0079 & $ 0.765$ & 0.917 & 0.926 & 0.521 \\
0.50 & 0.00 & 0.2308 & 0.2228 & 0.0079 & $-0.931$ & 1.025 & 0.946 & 0.773 \\
0.50 & 0.10 & 0.1979 & 0.1899 & 0.0079 & $-0.920$ & 1.025 & 0.945 & 0.738 \\
0.50 & 0.25 & 0.1379 & 0.1299 & 0.0079 & $-0.885$ & 1.023 & 0.950 & 0.641 \\
0.75 & 0.00 & 0.5086 & 0.5006 & 0.0080 & $-0.969$ & 1.035 & 0.950 & 0.814 \\
0.75 & 0.10 & 0.4606 & 0.4526 & 0.0080 & $-0.965$ & 1.035 & 0.946 & 0.795 \\
0.75 & 0.25 & 0.3779 & 0.3699 & 0.0080 & $-0.958$ & 1.037 & 0.943 & 0.747 \\
\bottomrule
\end{tabular}
\end{table}

\begin{figure}[t]
\centering
\begin{minipage}[t]{0.48\linewidth}
    \centering
    \includegraphics[width=\linewidth]{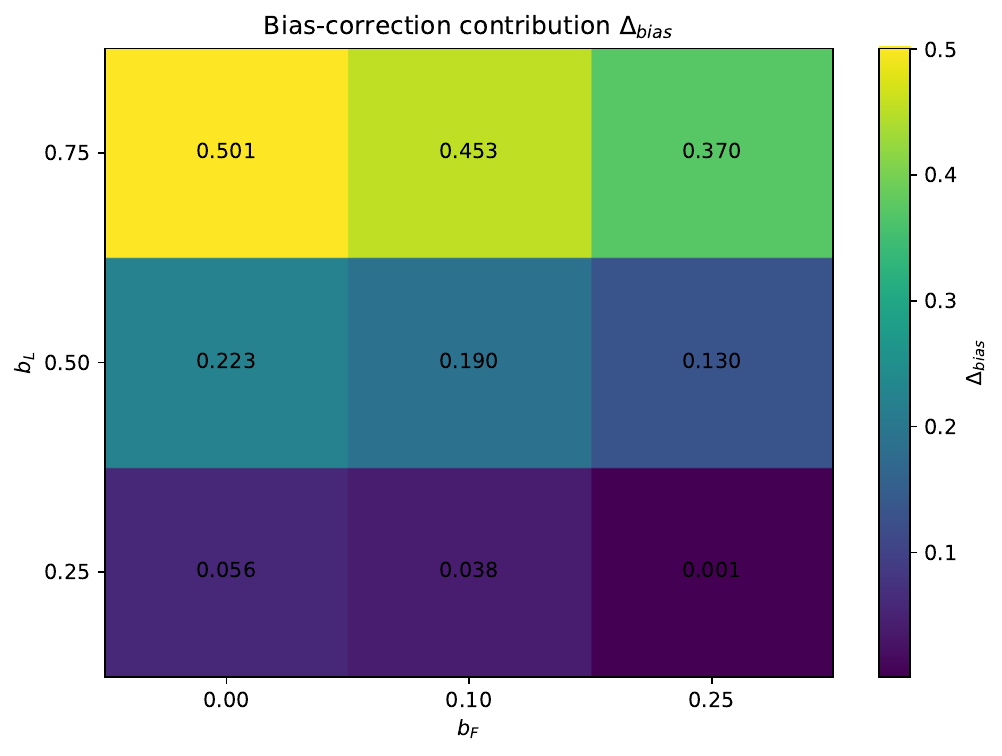}
    \vspace{-2mm}
    \textbf{(a)}
\end{minipage}
\hfill
\begin{minipage}[t]{0.48\linewidth}
    \centering
    \includegraphics[width=\linewidth]{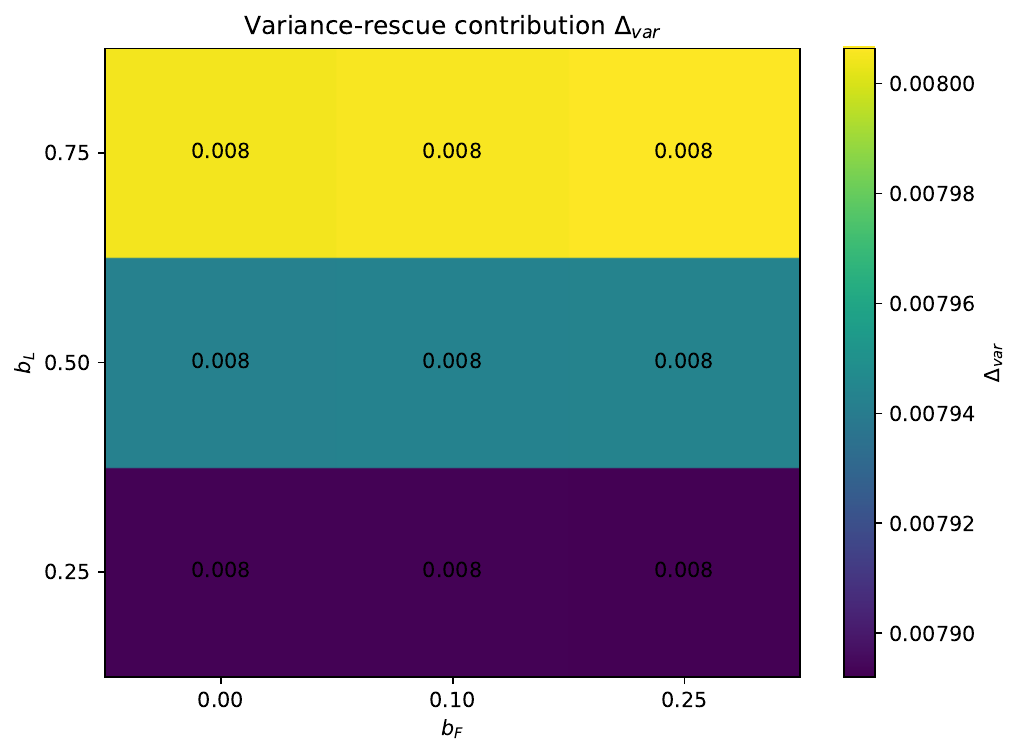}
    \vspace{-2mm}
    \textbf{(b)}
\end{minipage}
\caption{Decomposition of the marginal gain in Experiment~1.  (a) Bias-correction contribution $\Delta_{\rm bias}$.  (b) Variance-rescue contribution $\Delta_{\rm var}$.}
\label{fig:num-exp1-components}
\end{figure}

The separate gain components in Fig.~\ref{fig:num-exp1-components} clarify why the phase transition occurs.  Increasing $b_L$ substantially increases the amount of target-local bias that can be corrected by borrowing, whereas increasing $b_F$ reduces the discrepancy between the local and auxiliary bias directions and therefore decreases $\Delta_{\rm bias}$.  In contrast, $\Delta_{\rm var}$ changes little across the grid because the sample sizes, noise levels, bandwidths, and local weight are held fixed.  The observed phase transition is therefore generated primarily by attenuation of the bias-correction component rather than by a large change in the variance-rescue component.

\paragraph{Variance calibration and coverage.}
The fitting-variance estimator remains well calibrated across the full grid.  As reported in Table~\ref{tab:num-exp1-grid}, the ratio of the mean estimated fitting variance to its Monte Carlo benchmark ranges from $0.917$ to $1.037$.  Figure~\ref{fig:num-exp1-summary}(b) shows the corresponding fitting, evaluation, and total variance calibration at $(b_L,b_F)=(0.25,0)$.  The two-stage $95\%$ intervals attain empirical coverage between $0.926$ and $0.950$, whereas intervals using evaluation variance alone substantially under-cover, with coverage between $0.521$ and $0.814$.  Thus fitting randomness is essential for procedure-level inference even though gain is evaluated on an independent audit sample.

\subsection{Experiment 2: Exact-model quadratic boundary}
\label{sec:num-exp2}

\paragraph{Design.}
We next examine the exact-model quadratic-boundary regime of
Theorem~\ref{thm:boundary-gain-main}, with the detailed fitting and combined
limits given in Theorems~\ref{sg:thm:boundary-fitting-clt} and
\ref{sg:thm:boundary-combined}.  At the exact boundary, the
first-order gain derivative vanishes and the fitting fluctuation is governed
by the centered quadratic term rather than by the persistent-regime linear
approximation.  We again use $K=30$ clients on $[-1,1]$, with target-client
covariate mean $0$, auxiliary-client means equally spaced over
$[-0.7,0.7]$, and common covariate standard deviation $0.6$.

In contrast to Experiment~1, all underlying fitting and audit regression
functions are identical:
\begin{align}
m_j^{\mathrm{fit}}(x)=m_1^Q(x)=f_0(x)=1.5\sin(\pi x)+0.5x,
\qquad j=1,\ldots,K.
\end{align}
Thus the underlying regression functions have no persistent discrepancy.
At finite sample sizes, however, the local and federated
finite-bandwidth Nadaraya--Watson ratio targets need not coincide exactly
with $f_0$ under heterogeneous covariate laws.  Small residual first-order
terms may therefore remain at finite bandwidth, and the experiment assesses
whether the quadratic-boundary approximation captures the dominant fitting
fluctuation as these residual terms shrink with increasing sample size.

We consider three increasing fitting-sample regimes,
\begin{align}
(n_T,n_A)\in\{(30,120),(60,240),(120,480)\},
\end{align}
where $n_T$ denotes the target-client fitting sample size and $n_A$ the
sample size of each auxiliary client.  The server uses sample-size weights
and local weight $\alpha=0.5$.  The target and auxiliary response-noise
standard deviations are $1$ and $0.4$, respectively, while the independent
audit sample size is fixed at $200$.  Local regressions use the Epanechnikov
kernel with
\begin{align}
h_j=n_j^{-1/3}.
\end{align}
For each regime, $1000$ independent replications are used to estimate the
reference procedure-level gain and another $1000$ replications are used for
variance calibration and coverage.  To isolate the boundary approximation
itself, both the quadratic and persistent-linear calculations use oracle
finite-bandwidth nuisance quantities implied by the data-generating process.

\begin{figure}[t]
\centering
\begin{minipage}[t]{0.48\linewidth}
    \centering
    \includegraphics[width=\linewidth]{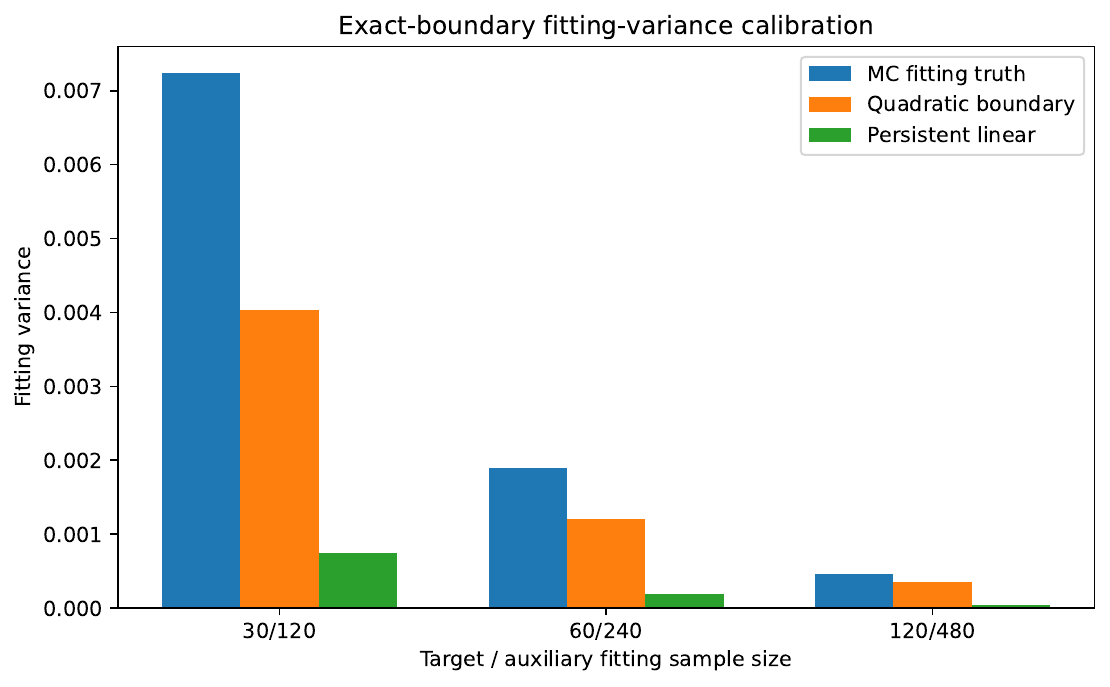}
    \vspace{-2mm}
    \textbf{(a)}
\end{minipage}
\hfill
\begin{minipage}[t]{0.48\linewidth}
    \centering
    \includegraphics[width=\linewidth]{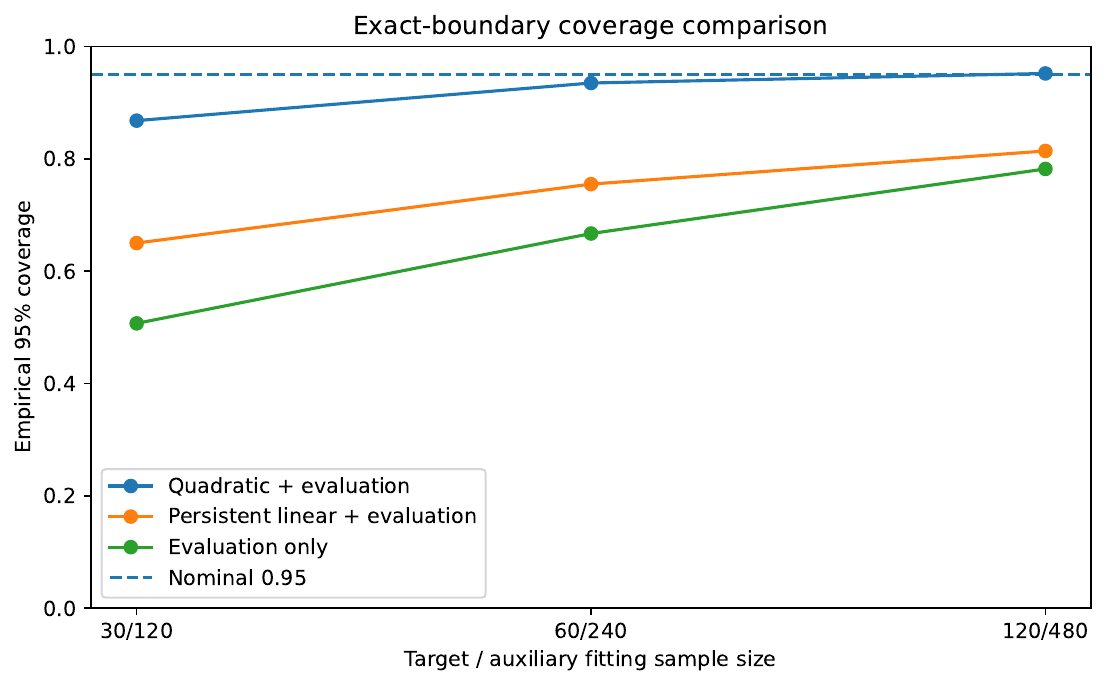}
    \vspace{-2mm}
    \textbf{(b)}
\end{minipage}
\caption{Exact-model boundary diagnostics.  (a) Monte Carlo fitting variance compared with the quadratic-boundary and persistent-linear variance estimates.  (b) Empirical $95\%$ procedure-level coverage using quadratic plus evaluation variance, persistent-linear plus evaluation variance, and evaluation variance alone.}
\label{fig:num-exp2-boundary}
\end{figure}

\paragraph{Boundary variance and coverage.}
Figure~\ref{fig:num-exp2-boundary} compares the quadratic boundary variance
with the persistent-regime linear approximation.  The distinction is
substantial.  Across the three regimes, the quadratic fitting-variance
estimate attains Monte Carlo variance ratios of $0.558$, $0.639$, and
$0.769$, respectively, showing improving finite-sample calibration as the
fitting samples increase.  The persistent-linear variance, by contrast,
captures only about one tenth of the Monte Carlo fitting variance, with
ratios $0.103$, $0.100$, and $0.094$. This persistent failure reflects the population exact-model boundary:
although finite-bandwidth smoothing can leave small residual first-order
terms, the first-order derivative vanishes in the boundary limit, so the
quadratic term is first order on the relevant scale and cannot be discarded.

The same distinction appears in procedure-level coverage.  Combining the
quadratic fitting variance with evaluation variance gives empirical $95\%$
coverages of $0.868$, $0.935$, and $0.952$ as the fitting sample sizes
increase.  The corresponding persistent-linear intervals cover only
$0.650$, $0.755$, and $0.814$, while evaluation-only intervals cover
$0.507$, $0.667$, and $0.782$.  The improvement toward nominal coverage is
consistent with the quadratic-boundary fitting and combined inference in
Theorems~\ref{sg:thm:boundary-fitting-clt} and
\ref{sg:thm:boundary-combined}.

\subsection{Experiment 3: Client-indexed joint inference}
\label{sec:num-exp3}

\paragraph{Design.}
We next examine the fixed-block client-indexed Gaussian approximation in
Theorem~\ref{thm:client-process-main}.  The experiment contains $K=10$
clients and focuses on the fixed block
\begin{align}
\mathcal I=\{1,\ldots,5\}.
\end{align}
The server uses equal aggregation weights and local weight
$\alpha=0.5$, so that the common federated estimator contributes
non-negligible cross-client dependence.

All fitting covariates follow truncated normal distributions supported on
$[-1,1]$.  For the five indexed
clients, the fitting sample sizes, covariate means, covariate standard
deviations, response-noise standard deviations, and regression coefficients
are
\begin{align*}
(n_j)_{j=1}^5&=(600,650,700,750,800), &
(\mu_j)_{j=1}^5&=(-0.35,-0.15,0,0.18,0.35),\\
(\sigma_{X,j})_{j=1}^5&=(0.48,0.58,0.62,0.52,0.66), &
(\sigma_{Y,j})_{j=1}^5&=(0.55,0.65,0.50,0.75,0.60),\\
(b_j^{\mathrm{fit}})_{j=1}^5&=(0.30,0.26,0.34,0.22,0.28), &
(\delta_j^{\mathrm{fit}})_{j=1}^5&=(-0.08,0.10,-0.05,0.12,-0.10).
\end{align*}
Their fitting regressions are
\begin{align}
m_j^{\mathrm{fit}}(x)
=
f_0(x)+b_j^{\mathrm{fit}}q(x)+\delta_j^{\mathrm{fit}}r(x).
\end{align}

The remaining five auxiliary clients each contribute $40$ fitting observations. Their covariate means are $-0.70$, $-0.35$, $0$, $0.35$, and $0.70$, with a common covariate standard deviation of $0.60$ and response-noise standard deviation of $1$. Their regression coefficients are
\begin{align*}
(b_j^{\mathrm{fit}})_{j=6}^{10}&=(0.02,0.04,0.06,0.04,0.02),\\
(\delta_j^{\mathrm{fit}})_{j=6}^{10}&=(-0.18,-0.08,0,0.08,0.18).
\end{align*}

The common baseline and perturbation directions are
\begin{align}
f_0(x)=1.5\sin(\pi x)+0.5x,\qquad
q(x)=1+x,\qquad
r(x)=x^2-\frac{1}{3},
\end{align}
and all local regressions use the Epanechnikov kernel with
$h_j=n_j^{-1/3}$.

The five indexed clients have independent audit samples whose covariates
follow truncated normal distributions supported on $[-0.6,0.6]$.  Their audit sample sizes, covariate means, covariate standard
deviations, response-noise standard deviations, and regression coefficients
are
\begin{align*}
(N_j^{\mathrm{eval}})_{j=1}^5&=(350,400,450,500,550), &
(\mu_j^Q)_{j=1}^5&=(-0.30,-0.10,0.08,0.22,0.30),\\
(\sigma_{X,j}^Q)_{j=1}^5&=(0.50,0.60,0.55,0.65,0.50), &
(\sigma_{Y,j}^Q)_{j=1}^5&=(0.65,0.55,0.70,0.60,0.75),\\
(b_j^Q)_{j=1}^5&=(-0.12,-0.10,-0.08,-0.06,-0.04), &
(\delta_j^Q)_{j=1}^5&=(0.02,-0.03,0.04,-0.02,0).
\end{align*}
Their audit regressions are
\begin{align}
m_j^Q(x)=f_0(x)+b_j^Qq(x)+\delta_j^Qr(x).
\end{align}
We use $1000$ independent fitting replications to obtain Monte Carlo
reference gains and fitting covariances, and a separate set of $1000$
fitting/audit replications to assess covariance calibration,
studentization, and marginal coverage.  The fitting-covariance calculation
uses oracle finite-bandwidth nuisance quantities so that the experiment
isolates the client-indexed covariance formula rather than nuisance
estimation error.  Gaussian orthant probabilities are evaluated using
$500{,}000$ Monte Carlo draws.

\paragraph{Marginal calibration.}
For each indexed client, define
\begin{align}
Z_j=\frac{\widehat G_j-\Gamma_j}{\widehat s_j}.
\end{align}
Table~\ref{tab:num-exp3-marginal} shows that the marginal studentization is
well calibrated despite substantial heterogeneity across clients.  The
means of $Z_j$ range from $-0.081$ to $0.007$, their empirical standard
deviations range from $0.967$ to $1.012$, and the empirical $95\%$
coverages range from $0.942$ to $0.961$.  The mean estimated total
variances are also close to their Monte Carlo benchmarks, with ratios
between $0.942$ and $0.995$.

\begin{table}[t]
\centering
\caption{Marginal calibration for the five-client block in Experiment~3.
The last column is the ratio of the mean estimated total variance to its
Monte Carlo benchmark.}
\label{tab:num-exp3-marginal}
\small
\setlength{\tabcolsep}{7pt}
\begin{tabular}{crrrrr}
\toprule
Client & $\Gamma_j$ & Mean$(Z_j)$ & SD$(Z_j)$ & Coverage & $\widehat V_j/V_j^{\rm MC}$ \\
\midrule
1 & 0.0491 & $-0.081$ & 0.972 & 0.961 & 0.945 \\
2 & 0.0280 & $-0.026$ & 0.967 & 0.955 & 0.983 \\
3 & 0.0891 & $-0.009$ & 0.983 & 0.955 & 0.995 \\
4 & 0.0196 & $-0.024$ & 1.007 & 0.942 & 0.942 \\
5 & 0.0658 & $ 0.007$ & 1.012 & 0.952 & 0.983 \\
\bottomrule
\end{tabular}
\end{table}

\begin{figure}[t]
\centering
\includegraphics[width=0.82\linewidth]
{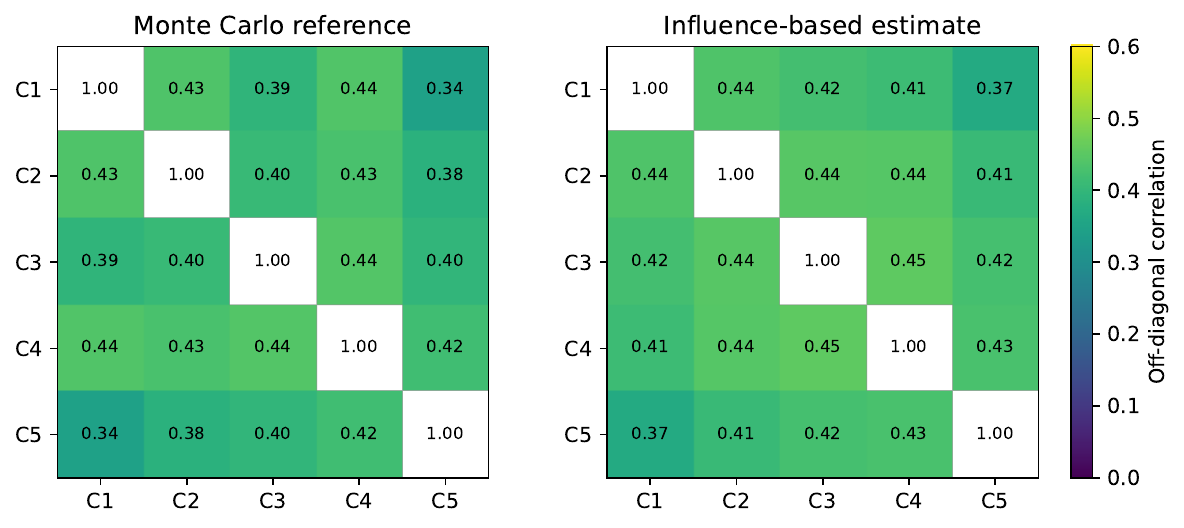}
\caption{Client-indexed correlation calibration in Experiment~3.
The left panel gives the Monte Carlo reference correlation matrix and the
right panel gives the oracle-nuisance influence-based estimate.  Diagonal
entries equal one; the color scale summarizes the off-diagonal correlations
generated by the shared federated estimator.}
\label{fig:num-exp3-corr}
\end{figure}

\paragraph{Cross-client dependence.}
The common federated estimator induces substantial positive dependence
between clientwise gains.  The Monte Carlo off-diagonal correlations range
from $0.342$ to $0.440$, so treating the coordinates as independent would
discard a material part of the joint uncertainty.  Figure~\ref{fig:num-exp3-corr}
compares the Monte Carlo correlation matrix with the oracle-nuisance
influence-based estimate obtained from the shared fitting data.  The agreement is close:
across the ten distinct off-diagonal entries, the mean absolute error is
$0.0205$, the root mean squared error is $0.0231$, and the maximum absolute
error is $0.0397$.  The empirical correlations of the studentized gains
are similarly close to the Monte Carlo reference, with mean absolute error
$0.0256$.

\paragraph{Fixed-block orthant probabilities.}
We finally compare empirical joint events with their Gaussian
approximations.  Table~\ref{tab:num-exp3-orthant} reports two events.  The
first is the directly interpretable event that all five estimated gains are
positive.  The second uses client-specific thresholds
\begin{align}
\varepsilon_j=\Gamma_j-\frac{1}{2}s_j,
\end{align}
so that every standardized population margin equals
\begin{align}
\gamma_j=\frac{\Gamma_j-\varepsilon_j}{s_j}=0.5.
\end{align}
This second event directly targets the fixed-block local-margin regime of
the Gaussian orthant result.

For the all-positive event, the Gaussian approximation is less accurate at the sample sizes considered, giving $0.889$ compared with the empirical probability of $0.921$. For the local-margin event, the empirical probability is $0.315$, compared
with $0.317$ under the Gaussian approximation using the Monte Carlo
covariance and $0.327$ using the estimated covariance.  By contrast,
forcing the coordinates to be independent gives only $0.158$.  Thus the
off-diagonal covariance is not merely a variance-estimation detail: it can
materially affect joint client-level conclusions. 

\begin{table}[t]
\centering
\caption{Fixed-block orthant probabilities in Experiment~3.  Gaussian
approximations are computed using the Monte Carlo covariance, the estimated
covariance, or a diagonal covariance that incorrectly treats clients as
independent.}
\label{tab:num-exp3-orthant}
\small
\setlength{\tabcolsep}{6pt}
\begin{tabular}{lcccc}
\toprule
Event & Empirical & Gaussian MC & Gaussian est. & Independent \\
\midrule
$\widehat G_j>0,\ \forall j$ &
0.921 & 0.889 & 0.896 & 0.877 \\
$\widehat G_j>\Gamma_j-0.5s_j,\ \forall j$ &
0.315 & 0.317 & 0.327 & 0.158 \\
\bottomrule
\end{tabular}
\end{table}

\subsection{Experiment 4: KL-robust gain and certificate inference}
\label{sec:num-exp4}

\paragraph{Design.}
We next evaluate the KL-robust gain and the variance-sensitive certificate in the regular regime.  The data-generating process is the same as the $(b_L,b_F)=(0.25,0)$ setting of Experiment~1, with $K=30$ clients and local weight $\alpha=0.5$.  Here each client has $n_j=400$ fitting observations and the independent target-client audit sample has size $500$.  All fitting response-noise standard deviations are $0.5$, and local regressions use the Epanechnikov kernel with
\begin{align}
h_j=n_j^{-1/3}.
\end{align}
Because all fitting sample sizes are equal, the server aggregation weights are equal across clients.  We use $1000$ independent fitting replications to construct the marginal certificate reference values and another $1000$ fitting/audit replications to evaluate the empirical KL-robust gain, certificate positivity, and confidence-bound coverage.

For a realized fitting sample, let $\mu$ and $v$ denote the conditional mean and variance of the paired gain under the target audit law.  The variance-sensitive KL certificate is
\begin{align}
C^{\mathrm{fit}}(\rho)=\mu-\sqrt{2\rho v}-c_{\rm sg}\rho,
\end{align}
which lower-bounds the exact fitted KL-robust gain under the sub-gamma condition in Proposition~\ref{kl:prop:certificate}. 
The corresponding procedure-level marginal certificate is
\begin{align}
\mathcal C(\rho)
=
\bm{E}_{\mathrm{fit}}\!\left[C^{\mathrm{fit}}(\rho)\right].
\end{align}
Following the local-radius scaling in the theory, we set
\begin{align}
\rho=c_\rho\frac{\mathfrak r}{h}, \qquad c_\rho\in\{0,0.25,0.5,1,2\},
\end{align}
where in this experiment
\begin{align}
\mathfrak r=0.00275, \qquad h=0.13572, \qquad \frac{\mathfrak r}{h}=0.02026.
\end{align}
The resulting KL radii are $0$, $0.00507$, $0.01013$, $0.02026$, and $0.04052$.

The certificate comparison corresponds to
Proposition~\ref{kl:prop:certificate}, while the confidence-bound experiment
evaluates the procedure-level certificate inference in
Theorem~\ref{thm:kl-process-main}.

\begin{figure}[t]
\centering
\includegraphics[width=0.68\linewidth]{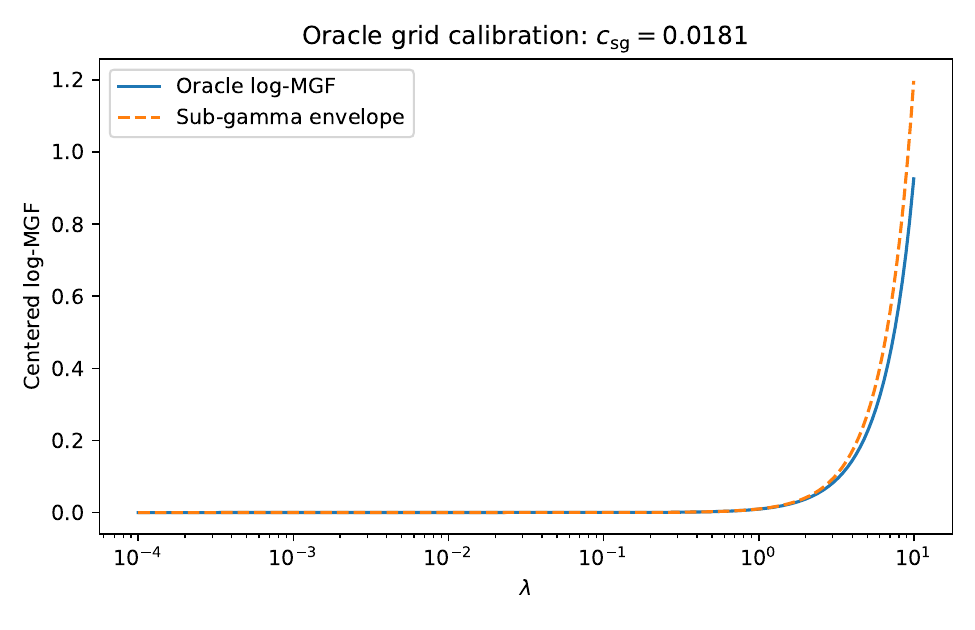}
\caption{Oracle grid calibration of the sub-gamma scale used in Experiment~4.  The calibrated envelope dominates the centered log-MGF over the displayed admissible grid.}
\label{fig:num-exp4-subgamma}
\end{figure}

\paragraph{Sub-gamma calibration.}
The sub-gamma scale is calibrated once from the known finite-bandwidth simulation law and then held fixed across all Monte Carlo replications.  Specifically, we evaluate the centered log-MGF on a dense $\lambda$ grid and choose the smallest grid-calibrated scale satisfying the sub-gamma envelope, followed by a $1.05$ safety factor.  This gives
\begin{align}
c_{\rm sg}=0.01810.
\end{align}
Figure~\ref{fig:num-exp4-subgamma} compares the oracle log-MGF with the resulting sub-gamma envelope.  This calculation is intended as an oracle grid-calibration diagnostic for the numerical experiment; it is not used as a numerical proof of the continuum condition in Assumption~\ref{kl:ass:subgamma}.

\begin{figure}[t]
\centering
\begin{minipage}[t]{0.32\linewidth}
    \centering
    \includegraphics[width=\linewidth]{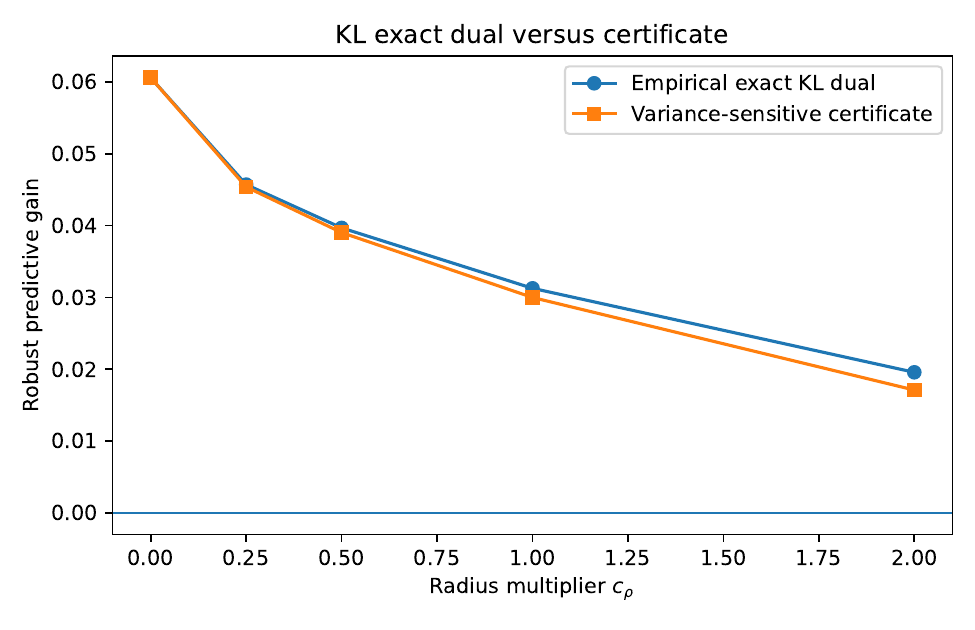}
    \vspace{-2mm}
    \textbf{(a)}
\end{minipage}
\hfill
\begin{minipage}[t]{0.32\linewidth}
    \centering
    \includegraphics[width=\linewidth]{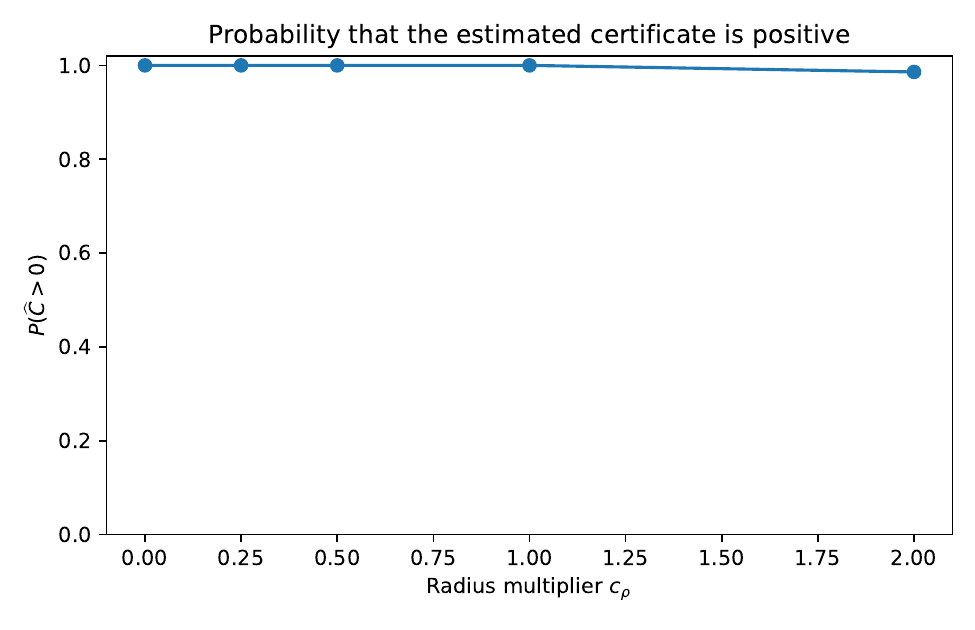}
    \vspace{-2mm}
    \textbf{(b)}
\end{minipage}
\hfill
\begin{minipage}[t]{0.32\linewidth}
    \centering
    \includegraphics[width=\linewidth]{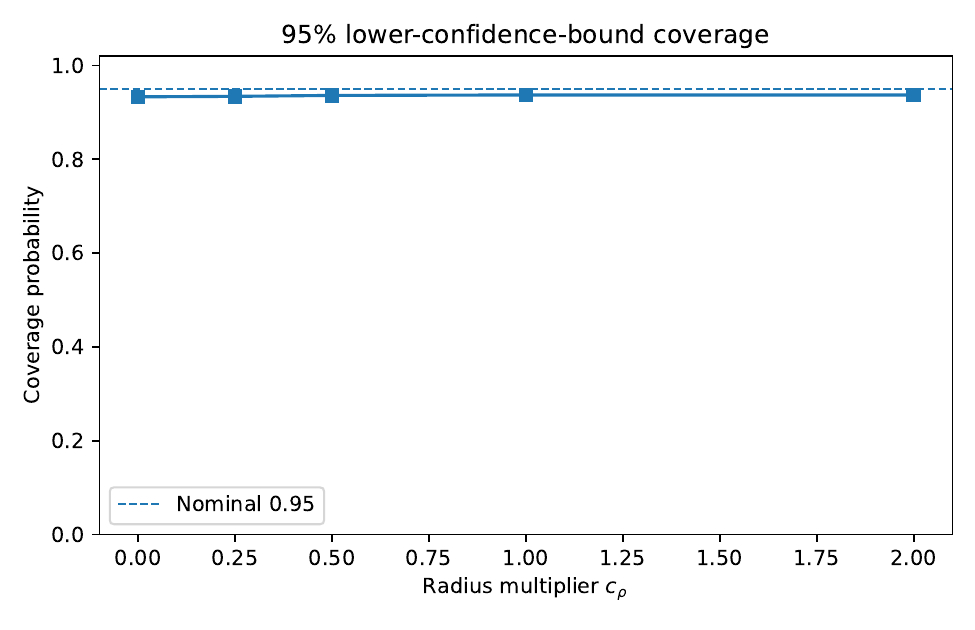}
    \vspace{-2mm}
    \textbf{(c)}
\end{minipage}
\caption{KL robustness and inference diagnostics for Experiment~4.  (a) Exact empirical KL-robust gain and the variance-sensitive certificate.  (b) Empirical probability that the estimated certificate is positive.  (c) Empirical coverage of the one-sided $95\%$ lower confidence bound for the procedure-level marginal certificate.}
\label{fig:num-exp4-robustness}
\end{figure}

\paragraph{Robustness and inference.}
The exact empirical KL dual and the variance-sensitive certificate decrease smoothly as the robustness radius increases.  Figure~\ref{fig:num-exp4-robustness}(a) shows that their Monte Carlo means are respectively $0.0606$ and $0.0606$ at $c_\rho=0$, $0.0457$ and $0.0454$ at $c_\rho=0.25$, and $0.0196$ and $0.0171$ at $c_\rho=2$.  Thus the certificate remains below and close to the exact robust value, with a mean gap increasing from $0$ to only $2.47\times10^{-3}$.  The empirical certificate never exceeds the empirical exact dual in any of the $1000$ Monte Carlo replications.  The certificate therefore gives a conservative but informative assessment of whether the borrowing benefit persists under moderate target-law shift.

\begin{table}[t]
\centering
\caption{KL robustness and inference diagnostics for Experiment~4.}
\label{tab:num-exp4-inference}
\small
\setlength{\tabcolsep}{7pt}
\begin{tabular}{cccc}
\toprule
$c_\rho$ & $\rho$ & $\Pr(\widehat C>0)$ & LCB coverage \\
\midrule
0.00 & 0.0000 & 1.000 & 0.933 \\
0.25 & 0.0051 & 1.000 & 0.934 \\
0.50 & 0.0101 & 1.000 & 0.936 \\
1.00 & 0.0203 & 1.000 & 0.937 \\
2.00 & 0.0405 & 0.986 & 0.937 \\
\bottomrule
\end{tabular}
\end{table}

Table~\ref{tab:num-exp4-inference} and Figure~\ref{fig:num-exp4-robustness}(b)--(c) summarize certificate positivity and one-sided inference.  The estimated certificate is positive in every replication for $c_\rho\le1$ and in $98.6\%$ of replications at $c_\rho=2$.  The one-sided $95\%$ lower confidence bound for the procedure-level marginal certificate $\mathcal C(\rho)$, using the combined fitting and evaluation uncertainty, has empirical coverage between $0.933$ and $0.937$ across the radius grid.

\subsection{Experiment 5: $L_2$ robustness in persistent and vanishing-direction regimes}
\label{sec:num-exp5}

This experiment illustrates the exact least-favorable geometry in Proposition~\ref{prop:l2-main}, examines estimation under persistent borrowing, and studies inference near the vanishing-direction boundary of Theorem~\ref{thm:l2-vanishing-main}.

\paragraph{Persistent-direction design.}
We first examine the regular $L_2$-robust regime in the same heterogeneous
client model used in Experiment~1, with $(b_L,b_F)=(0.25,0)$,
$K=30$, $\alpha=0.5$, and equal fitting sample sizes across clients.
The target fitting regression is
\begin{align}
m_1^{\mathrm{fit}}(x)=f_0(x)+0.25q(x),
\end{align}
the auxiliary regressions are
\begin{align}
m_j^{\mathrm{fit}}(x)=f_0(x)+\delta_jr(x), \qquad j=2,\ldots,K,
\end{align}
and the audit regression is $m_1^Q=f_0$.  The covariate distributions,
heterogeneity coefficients $\delta_j$, response-noise standard deviation
$0.5$, Epanechnikov kernel, and bandwidth rule $h_j=n_j^{-1/3}$ are the same
as in Experiment~1.

We consider common per-client fitting sample sizes
\begin{align}
n \in \{100,200,400,800\}.
\end{align}
Within each replication, an independent target-client pilot sample of size $n$ is used to estimate $m_1^Q$. For each $n$, 500 independent fitting replications are generated. At the baseline level $n=200$, an independent audit sample of size 500 is also used to compare the fitted robust value with its audit estimate.

For the fitted borrowing direction
\begin{align}
d^B(x)=(1-\alpha)\{F(x)-L(x)\},
\end{align}
the exact $L_2$-robust value is
\begin{align}
\mathcal R^{\mathrm{fit}}(\tau)
=
G^{\mathrm{fit}}
-
2\tau\norm{d^B}_{Q_X},
\end{align}
with least-favorable regression function
\begin{align}
m^\circ(x)
=
m_1^Q(x)
-
\tau\frac{d^B(x)}{\norm{d^B}_{Q_X}}.
\end{align}
The population borrowing-direction norm is $0.13356$, and we use
\begin{align}
\tau=c_\tau(0.13356), \qquad
c_\tau\in\{0,0.5,1,1.5,2\}.
\end{align}

\begin{figure}[t]
\centering
\begin{minipage}[t]{0.48\linewidth}
    \centering
    \includegraphics[width=\linewidth]{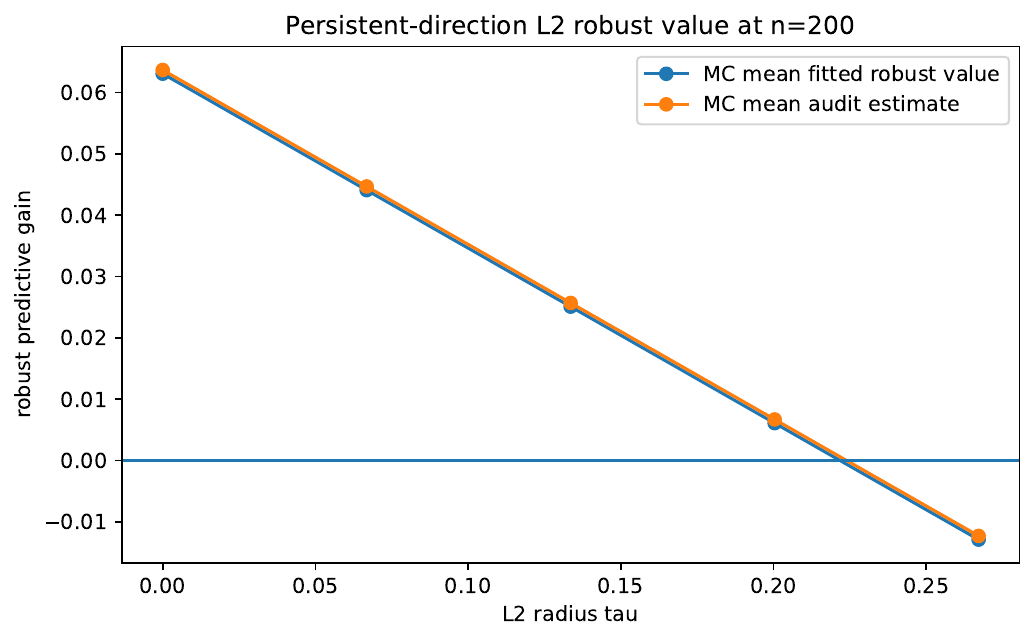}
    \vspace{-2mm}
    \textbf{(a)}
\end{minipage}
\hfill
\begin{minipage}[t]{0.48\linewidth}
    \centering
    \includegraphics[width=\linewidth]{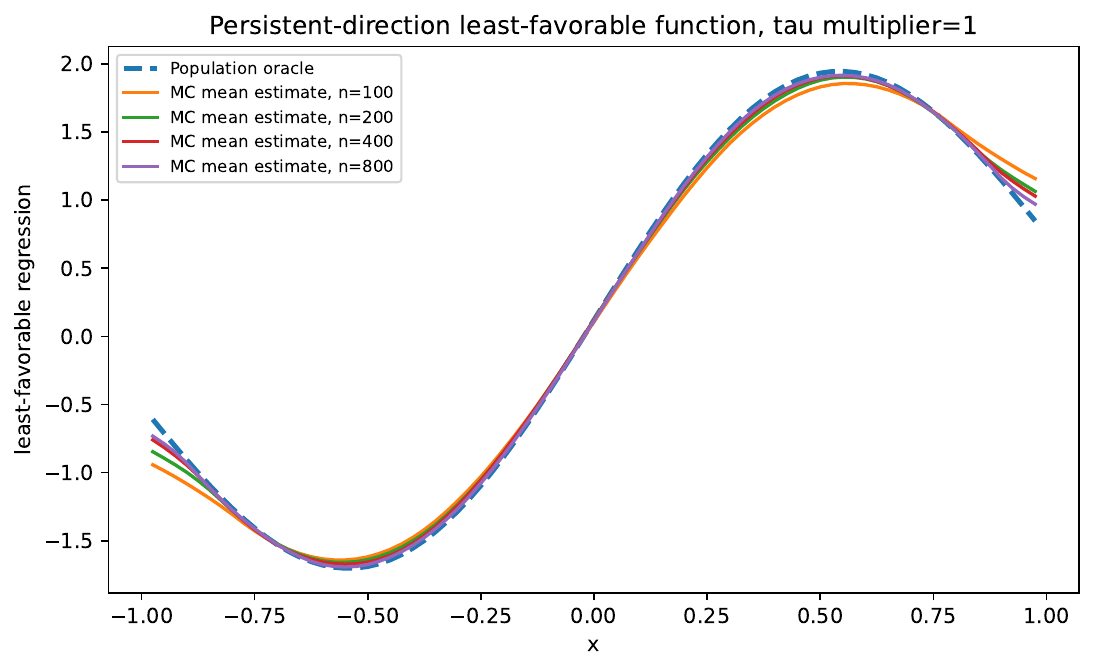}
    \vspace{-2mm}
    \textbf{(b)}
\end{minipage}
\caption{$L_2$ robustness in the persistent-direction regime.
(a) Monte Carlo mean fitted robust value and its audit estimate at $n=200$
over the radius grid.
(b) Convergence of the estimated least-favorable regression function toward
the population oracle at $c_\tau=1$.}
\label{fig:num-exp5-persistent}
\end{figure}

\paragraph{Persistent-direction results.}
Figure~\ref{fig:num-exp5-persistent}(a) shows the fitted and audit-based
robust values at the baseline sample size $n=200$.  The robust value
decreases monotonically with the $L_2$ radius, as implied by the exact
linear penalty in $\tau$.  The Monte Carlo mean fitted robust value decreases
from $0.0631$ at $\tau=0$ to $-0.0129$ at the largest radius, while the
corresponding audit estimates remain extremely close, with an average
audit-minus-fitted difference of approximately $6.0\times10^{-4}$ throughout
the radius grid.

Figure~\ref{fig:num-exp5-persistent}(b) examines the corresponding
least-favorable regression function at $c_\tau=1$.  The Monte Carlo mean
estimated curve moves toward the population least-favorable oracle as the
fitting sample size increases.  Table~\ref{tab:num-exp5-lf-error} quantifies
the remaining discrepancy: the mean $L_2$ error relative to the population
oracle decreases from $0.1680$ at $n=100$ to $0.1202$, $0.0910$, and
$0.0699$ at $n=200,400,800$, respectively.  These results illustrate both
the exact least-favorable geometry and the convergence of its feasible
estimator in the persistent-direction regime.

\begin{table}[t]
\centering
\caption{Convergence of the estimated least-favorable regression function in the persistent-direction regime.}
\label{tab:num-exp5-lf-error}
\small
\setlength{\tabcolsep}{8pt}
\begin{tabular}{cc}
\toprule
$n$ & Mean $L_2$ error \\
\midrule
100 & 0.1680 \\
200 & 0.1202 \\
400 & 0.0910 \\
800 & 0.0699 \\
\bottomrule
\end{tabular}
\end{table}

\paragraph{Vanishing-direction boundary.}
We next examine the exact-zero mean borrowing-direction regime of
Theorem~\ref{thm:l2-vanishing-main}, associated with the nondifferentiable
boundary
\begin{align}
\overline d^B=0.
\end{align}
The underlying fitting and audit regression functions are identical:
\begin{align}
m_j^{\mathrm{fit}}(x)=m_1^Q(x)=f_0(x)
=1.5\sin(\pi x)+0.5x,
\qquad j=1,\ldots,K.
\end{align}
Thus the underlying regression model is at the zero-direction boundary.
As in Experiment~2, however, finite-bandwidth Nadaraya--Watson smoothing
under heterogeneous covariate laws can leave a small stage-specific
borrowing direction at finite sample sizes; this residual direction shrinks
as the fitting samples increase.

The remaining design matches Experiment~2: $K=30$, $\alpha=0.5$, audit
sample size $200$, target and auxiliary response-noise standard deviations
$1$ and $0.4$, respectively, and fitting-sample regimes
\begin{align}
(n_T,n_A)\in\{(30,120),(60,240),(120,480)\}.
\end{align}
The server uses sample-size weights and all local bandwidths satisfy
$h_j=n_j^{-1/3}$.

At the zero-direction population limit, the borrowing-direction norm
functional is nondifferentiable.  Writing
\begin{align}
S^B=\norm{d^B}_{Q_X}^{2}, \qquad
\nu^2=\bm{E}_{\mathrm{fit}}S^B,
\end{align}
the square-root expansion introduces the stage-specific coefficient
\begin{align}
\kappa^B=\frac{\tau}{\nu}.
\end{align}
We choose radii on the boundary scale
\begin{align}
\tau=c_\tau\sqrt{\frac{\mathfrak r}{h_T}},
\qquad
c_\tau\in\{0,0.5,1,2\},
\end{align}
where $\mathfrak r$ denotes the selected-client leverage aggregate and
$h_T$ is the target-client bandwidth.  For the three sample-size regimes,
$\sqrt{\mathfrak r/h_T}$ equals $0.3259$, $0.2587$, and $0.2053$,
respectively.  Each regime uses $1000$ independent reference fitting
replications and another $1000$ fitting/audit replications for variance and
coverage diagnostics.

\begin{figure}[t]
\centering
\begin{minipage}[t]{0.48\linewidth}
    \centering
    \includegraphics[width=\linewidth]{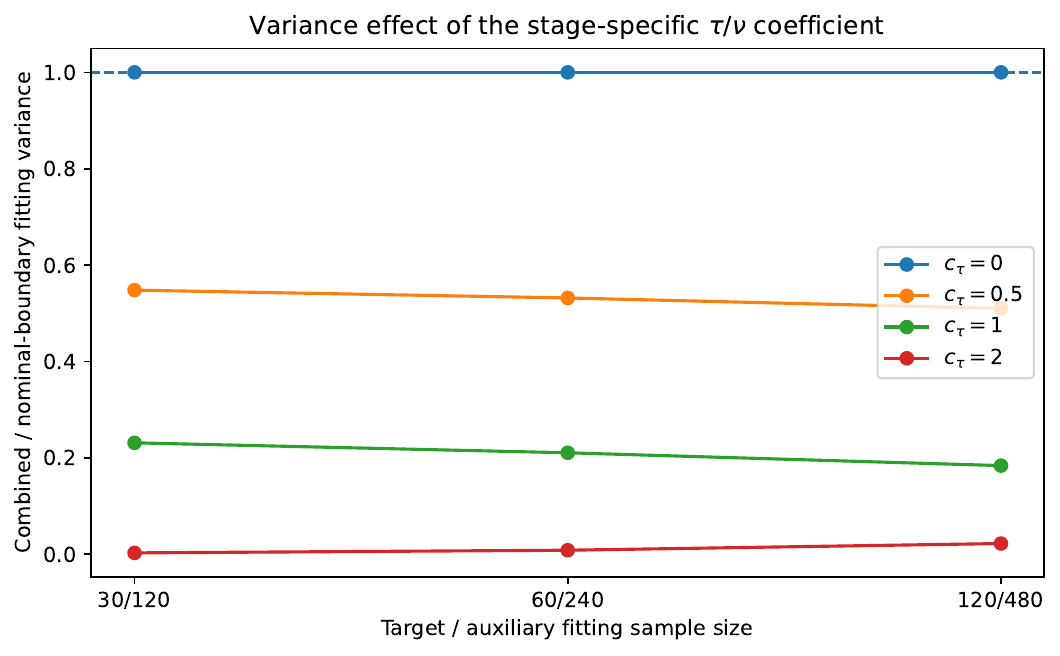}
    \vspace{-2mm}
    \textbf{(a)}
\end{minipage}
\hfill
\begin{minipage}[t]{0.48\linewidth}
    \centering
    \includegraphics[width=\linewidth]{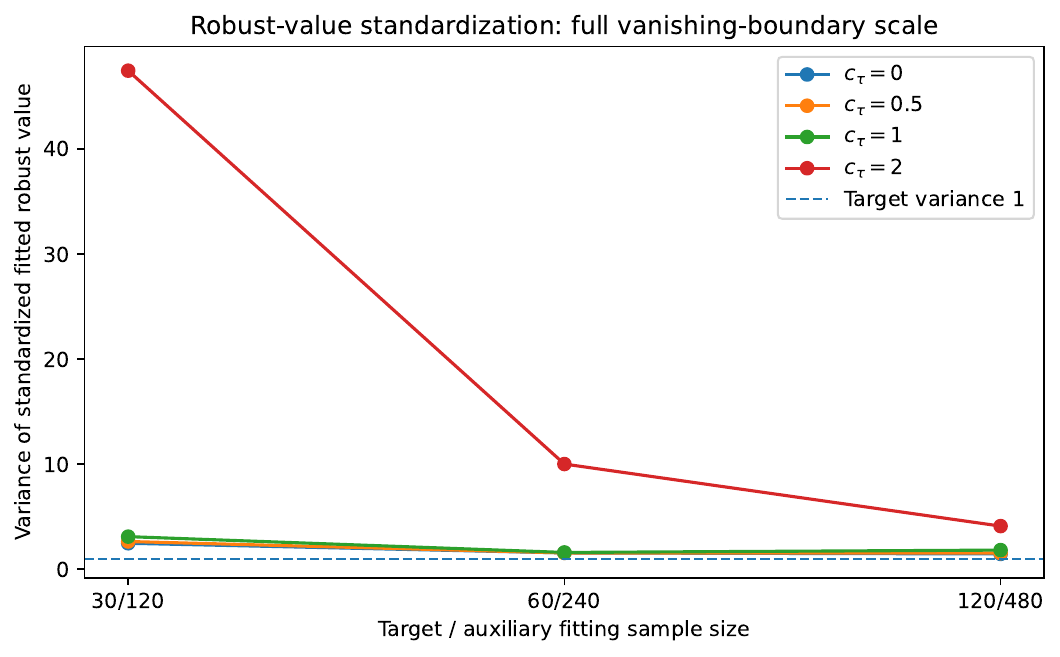}
    \vspace{-2mm}
    \textbf{(b)}
\end{minipage}
\caption{$L_2$ robustness at the vanishing-direction boundary.
(a) Fitting-variance effect of the stage-specific coefficient
$\kappa^B=\tau/\nu$, reported relative to the nominal gain-boundary
variance.
(b) Variance of the fitted robust value standardized by the full
vanishing-direction boundary scale.}
\label{fig:num-exp5-boundary}
\end{figure}

\paragraph{Boundary results.}
The stage-specific coefficient $\kappa^B$ materially changes the fitting
variance of the robust value.  Figure~\ref{fig:num-exp5-boundary}(a) reports
the ratio of the full vanishing-direction fitting variance to the nominal
gain-boundary variance.  At $c_\tau=0$ the two coincide, while for positive
radii the ratio decreases substantially.  For example, at $c_\tau=1$ the
ratios are approximately $0.231$, $0.211$, and $0.184$ across the three
sample-size regimes.  The direction term therefore cannot be ignored: its
covariance with the nominal gain component produces substantial
cancellation in this design.

Figure~\ref{fig:num-exp5-boundary}(b) examines the fitted robust value
standardized by the full vanishing-direction boundary scale.  The
standardization improves with increasing fitting sample size for moderate
radii, whereas the largest radius, $c_\tau=2$, remains more distorted at
the sample sizes considered.  This illustrates the finite-sample difficulty
created by the nondifferentiable borrowing-direction norm even after the
stage-specific variance correction is incorporated.

The combined fitting/evaluation inference is considerably more stable.
Across the largest sample-size regime $(n_T,n_A)=(120,480)$, the empirical
$95\%$ coverages are $0.929$, $0.937$, $0.941$, and $0.931$ for
$c_\tau=0,0.5,1,2$, respectively.  The mean square-root remainder also
decreases strongly with sample size; at $c_\tau=1$, it falls from
$6.80\times10^{-3}$ to $2.73\times10^{-3}$ and then
$1.43\times10^{-3}$.  Thus the experiment illustrates both the need for
the stage-specific vanishing-direction correction and the finite-sample
difficulty of the most aggressive robustness radii.



\section{Discussion}

This paper studies when a client can benefit from borrowing a federated predictor and how that gain can be quantified.  We distinguish the gain for the realized fitted predictors from the gain of the learning procedure under repeated fitting, and show that positive gain may come either from correcting a persistent target mismatch or from reducing estimation variability.  Because clients share the same server estimator, their gain statistics are dependent and this dependence must be retained in joint inference.  More broadly, a good federated predictor does not imply positive transfer for every client; the benefit of borrowing depends on the client's evaluation target and the amount of borrowing.  The KL and $L_2$ analyses extend this question to distributional robustness, with separate treatments for regular and boundary cases.  The current theory uses an independent audit sample and a prescribed borrowing weight; extending the framework to data-adaptive borrowing and other evaluation designs is a natural direction for future work.

\clearpage

\appendix 
\section*{Appendix}

\section{Pointwise Federated-Estimator Inference: Proofs}
\label{app:fnw-pointwise}

The pointwise regularity conditions are stated in Assumptions~\ref{fnw:ass:kernel-growth}--\ref{fnw:ass:undersmoothing} in Section~\ref{sec:fnw-pointwise}.  Throughout the following proofs, $x$ is a fixed query point.

\subsection{Basic notation}

Define the observation-level centered contribution
\begin{align}
\zeta_{i,j,t}(x)=\frac{w_{j,t}}{n_{j,t}B_t(x)}\left(\psi_{j,t}(x-X_{i,j,t})[Y_{i,j,t}-m_t^\star(x)]-\bm{E}_{P_{j,t}}\left[\psi_{j,t}(x-X)[Y-m_t^\star(x)]\right]\right).
\label{fnw:eq:zeta}
\end{align}
Then
\begin{align}
\Delta_t^{\mathrm{NW}}(x)-\bm{E}[\Delta_t^{\mathrm{NW}}(x)]=\sum_{j=1}^{K_t}\sum_{i=1}^{n_{j,t}}\zeta_{i,j,t}(x), \qquad V_t(x)=\sum_{j=1}^{K_t}\sum_{i=1}^{n_{j,t}}\bm{E}[\zeta_{i,j,t}(x)^2].
\label{fnw:eq:centered-sum}
\end{align}
For the variance expansion, define
\begin{align}
d_{j,t}^{(2)}(z;x)=\bm{E}_{P_{j,t}}([Y-m_t^\star(x)]^2\mid X=z), \qquad \Omega_{j,t}(z;x)=p_{j,t}(z)d_{j,t}^{(2)}(z;x).
\label{fnw:eq:variance-integrand}
\end{align}
At points where $p_{j,t}(z)=0$, any measurable version of the conditional
moment may be used; only the product $\Omega_{j,t}(z;x)$ enters the theory.

\subsection{Denominator stability and leading variance}

\begin{lemma}[Consistency of the federated denominator]
\label{fnw:lem:denominator}
Under Assumptions~\ref{fnw:ass:kernel-growth}--\ref{fnw:ass:pointwise},
\begin{align}
\frac{\widehat B_t(x)-B_t(x)}{B_t(x)}=o_p(1), \qquad \frac{B_t(x)}{\widehat B_t(x)}=1+o_p(1).
\label{fnw:eq:denominator-consistency}
\end{align}
\end{lemma}

\begin{proof}
Write
\begin{align}
\frac{\widehat B_t(x)-B_t(x)}{B_t(x)}=D_{1,t}(x)+D_{2,t}(x),
\end{align}
where
\begin{align}
D_{1,t}(x)=\frac{1}{B_t(x)}\sum_{j=1}^{K_t}w_{j,t}\left\{\widehat p_{j,t}(x)-\bm{E}[\widehat p_{j,t}(x)]\right\},
\end{align}
and
\begin{align}
D_{2,t}(x)=\frac{1}{B_t(x)}\sum_{j=1}^{K_t}w_{j,t}\left\{\bm{E}[\widehat p_{j,t}(x)]-p_{j,t}(x)\right\}.
\end{align}
We first control the smoothing-bias term.  For each active client $j$, a change of variables gives
\begin{align}
\bm{E}[\widehat p_{j,t}(x)]=\int_{\mathbb{R}}\mathcal K_{j,t}(u)p_{j,t}(x-h_{j,t}u)\,d u.
\end{align}
By a second-order Taylor expansion around $x$,
\begin{align}
p_{j,t}(x-h_{j,t}u)=p_{j,t}(x)-h_{j,t}up_{j,t}'(x)+\frac12h_{j,t}^2u^2p_{j,t}''(\xi_{j,t,u}),
\end{align}
where $\xi_{j,t,u}$ lies between $x$ and $x-h_{j,t}u$.  Since $\int_{\mathbb{R}}\mathcal K_{j,t}(u)\,d u=1$ and $\int_{\mathbb{R}}u\mathcal K_{j,t}(u)\,d u=0$,
\begin{align}
\bm{E}[\widehat p_{j,t}(x)]-p_{j,t}(x)=\frac12h_{j,t}^2\int_{\mathbb{R}}u^2\mathcal K_{j,t}(u)p_{j,t}''(\xi_{j,t,u})\,d u.
\end{align}
Assumptions~\ref{fnw:ass:kernel-growth} and \ref{fnw:ass:client-regularity} therefore imply
\begin{align}
\sup_{1\le j\le K_t}\left|\bm{E}[\widehat p_{j,t}(x)]-p_{j,t}(x)\right|\le\frac12C_{p,2}C_{K,2}\max_{1\le j\le K_t}h_{j,t}^2.
\end{align}
Since $B_t(x)\ge c_B$, $w_{j,t}\ge0$, and $\sum_{j=1}^{K_t}w_{j,t}=1$,
\begin{align}
|D_{2,t}(x)|\le\frac{C_{p,2}C_{K,2}}{2c_B}\max_{1\le j\le K_t}h_{j,t}^2\to0.
\end{align}
We next control the centered stochastic term.  Independence across clients and within-client observations gives
\begin{align}
\Var[D_{1,t}(x)]=\frac{1}{B_t(x)^2}\sum_{j=1}^{K_t}w_{j,t}^2\Var[\widehat p_{j,t}(x)].
\end{align}
For each client $j$,
\begin{align}
\Var[\widehat p_{j,t}(x)]\le\frac{1}{n_{j,t}}\bm{E}_{P_{j,t}}\left[\frac{1}{h_{j,t}^2}\mathcal K_{j,t}^2\left(\frac{x-X}{h_{j,t}}\right)\right].
\end{align}
A change of variables yields
\begin{align}
\bm{E}_{P_{j,t}}\left[\frac{1}{h_{j,t}^2}\mathcal K_{j,t}^2\left(\frac{x-X}{h_{j,t}}\right)\right]=\frac{1}{h_{j,t}}\int_{\mathbb{R}}\mathcal K_{j,t}(u)^2p_{j,t}(x-h_{j,t}u)\,d u.
\end{align}
Hence, by Assumptions~\ref{fnw:ass:kernel-growth} and \ref{fnw:ass:client-regularity},
\begin{align}
\Var[\widehat p_{j,t}(x)]\le\frac{C_{p,0}C_{K,R}}{n_{j,t}h_{j,t}}.
\end{align}
Using $B_t(x)\ge c_B$,
\begin{align}
\Var[D_{1,t}(x)]\le\frac{C_{p,0}C_{K,R}}{c_B^2}\sum_{j=1}^{K_t}\frac{w_{j,t}^2}{n_{j,t}h_{j,t}}.
\end{align}
Since $w_{j,t}\ge0$ and $\sum_{j=1}^{K_t}w_{j,t}=1$,
\begin{align}
\sum_{j=1}^{K_t}\frac{w_{j,t}^2}{n_{j,t}h_{j,t}}\le\frac{1}{\min_{1\le j\le K_t}n_{j,t}h_{j,t}}\sum_{j=1}^{K_t}w_{j,t}^2\le\frac{1}{\min_{1\le j\le K_t}n_{j,t}h_{j,t}}\to0.
\end{align}
Therefore $\Var[D_{1,t}(x)]\to0$, and Chebyshev's inequality implies $D_{1,t}(x)=o_p(1)$.  Combining the deterministic and stochastic terms gives
\begin{align}
\frac{\widehat B_t(x)-B_t(x)}{B_t(x)}=o_p(1).
\end{align}
Equivalently,
\begin{align}
\frac{\widehat B_t(x)}{B_t(x)}=1+o_p(1).
\end{align}
Since the map $z\mapsto z^{-1}$ is continuous at $z=1$, the continuous mapping theorem gives
\begin{align}
\frac{B_t(x)}{\widehat B_t(x)}=1+o_p(1).
\end{align}
This proves \eqref{fnw:eq:denominator-consistency}.
\end{proof}

\begin{lemma}[Uniform target and variance-integrand control]
\label{fnw:lem:variance-integrand}
Under Assumptions~\ref{fnw:ass:client-regularity} and \ref{fnw:ass:pointwise}, there are finite constants $C_m(x)$, $L_\Omega(x)$, and $C_\Omega(x)$ such that
\begin{align}
\sup_t|m_t^\star(x)|\le C_m(x).
\label{fnw:eq:target-bound}
\end{align}
Moreover, for all $z_1,z_2\in U_x$,
\begin{align}
|\Omega_{j,t}(z_1;x)-\Omega_{j,t}(z_2;x)|\le L_\Omega(x)|z_1-z_2|, \qquad \sup_{t,j,z}|\Omega_{j,t}(z;x)|\le C_\Omega(x).
\label{fnw:eq:omega-regularity}
\end{align}
\end{lemma}

\begin{proof}
We first establish the uniform bound for the federated target.  Conditional
Jensen gives
\[
|g_{j,t}(x)|^2
=
 p_{j,t}(x)^2|m_{j,t}(x)|^2
\le
 p_{j,t}(x)s_{j,t}(x)
\le C_{p,0}C_s.
\]
Consequently,
\[
|A_t(x)|
\le
\sum_{j=1}^{K_t}w_{j,t}|g_{j,t}(x)|
\le
(C_{p,0}C_s)^{1/2}.
\]
By \eqref{fnw:eq:denominator-floor},
\begin{align}
|m_t^\star(x)|=\frac{|A_t(x)|}{B_t(x)}
\le \frac{(C_{p,0}C_s)^{1/2}}{c_B}.
\end{align}
Thus \eqref{fnw:eq:target-bound} holds with
$C_m(x)=(C_{p,0}C_s)^{1/2}/c_B$.  No clientwise lower bound on
$p_{j,t}(x)$ is needed.

We next control the variance integrand.  Using $m_{j,t}(z)=\bm{E}_{P_{j,t}}[Y\mid X=z]$, $g_{j,t}(z)=p_{j,t}(z)m_{j,t}(z)$, and \eqref{fnw:eq:s-second-moment},
\begin{align}
\Omega_{j,t}(z;x)&=p_{j,t}(z)d_{j,t}^{(2)}(z;x)\\
&=p_{j,t}(z)\bm{E}_{P_{j,t}}([Y-m_t^\star(x)]^2\mid X=z)\\
&=s_{j,t}(z)-2m_t^\star(x)g_{j,t}(z)+m_t^\star(x)^2p_{j,t}(z).
\end{align}
For $z_1,z_2\in U_x$, Assumption~\ref{fnw:ass:client-regularity} and the preceding target bound give
\begin{align}
|\Omega_{j,t}(z_1;x)-\Omega_{j,t}(z_2;x)|\le\{L_s+2C_m(x)L_g+C_m(x)^2L_p\}|z_1-z_2|.
\end{align}
Hence one may take
\begin{align}
L_\Omega(x)=L_s+2C_m(x)L_g+C_m(x)^2L_p.
\end{align}
It remains to establish a global bound.  By the conditional Jensen inequality,
\begin{align}
|m_{j,t}(z)|^2\le\bm{E}_{P_{j,t}}[Y^2\mid X=z].
\end{align}
Therefore,
\begin{align}
|g_{j,t}(z)|^2=p_{j,t}(z)^2|m_{j,t}(z)|^2\le p_{j,t}(z)^2\bm{E}_{P_{j,t}}[Y^2\mid X=z]=p_{j,t}(z)s_{j,t}(z).
\end{align}
By Assumption~\ref{fnw:ass:client-regularity},
\begin{align}
\sup_{t,j,z}|g_{j,t}(z)|\le(C_{p,0}C_s)^{1/2}.
\end{align}
Recalling
\begin{align}
\Omega_{j,t}(z;x)=s_{j,t}(z)-2m_t^\star(x)g_{j,t}(z)+m_t^\star(x)^2p_{j,t}(z),
\end{align}
and using the uniform bounds on $s_{j,t}(z)$, $g_{j,t}(z)$, $p_{j,t}(z)$, and $m_t^\star(x)$, there is a finite constant $C_\Omega(x)$ independent of $j$ and $t$ such that
\begin{align}
\sup_{t,j,z}|\Omega_{j,t}(z;x)|\le C_\Omega(x).
\end{align}
This proves \eqref{fnw:eq:omega-regularity}.
\end{proof}

\begin{lemma}[Uniform client-level variance expansion]
\label{fnw:lem:variance-expansion}
Under Assumptions~\ref{fnw:ass:kernel-growth}--\ref{fnw:ass:pointwise}, there are remainders $\varepsilon_{j,t}(x)$ such that
\begin{align}
\Var_{P_{j,t}}\left(\psi_{j,t}(x-X)[Y-m_t^\star(x)]\right)=\frac{R(\mathcal K_{j,t})p_{j,t}(x)d_{j,t}^{(2)}(x;x)+\varepsilon_{j,t}(x)}{h_{j,t}},
\label{fnw:eq:client-variance-expansion}
\end{align}
where $R(\mathcal K_{j,t})=\int_{\mathbb{R}}\mathcal K_{j,t}(u)^2\,d u$ and
\begin{align}
\max_{1\le j\le K_t}|\varepsilon_{j,t}(x)|\to0.
\label{fnw:eq:variance-remainder-uniform}
\end{align}
\end{lemma}

\begin{proof}
We first expand the client-level second moment.  A change of variables gives
\begin{align}
\bm{E}_{P_{j,t}}\left[\psi_{j,t}(x-X)^2[Y-m_t^\star(x)]^2\right]=\frac{1}{h_{j,t}}\int_{\mathbb{R}}\mathcal K_{j,t}(u)^2\Omega_{j,t}(x-h_{j,t}u;x)\,d u.
\end{align}
Define
\begin{align}
\varepsilon_{j,t}^{(1)}(x)=\int_{\mathbb{R}}\mathcal K_{j,t}(u)^2\left\{\Omega_{j,t}(x-h_{j,t}u;x)-\Omega_{j,t}(x;x)\right\}\,d u.
\end{align}
Since $\Omega_{j,t}(x;x)=p_{j,t}(x)d_{j,t}^{(2)}(x;x)$,
\begin{align}
\bm{E}_{P_{j,t}}\left[\psi_{j,t}(x-X)^2[Y-m_t^\star(x)]^2\right]=\frac{R(\mathcal K_{j,t})p_{j,t}(x)d_{j,t}^{(2)}(x;x)+\varepsilon_{j,t}^{(1)}(x)}{h_{j,t}}.
\end{align}
Let
\begin{align}
\overline h_t=\max_{1\le j\le K_t}h_{j,t}.
\end{align}
Choose $\eta_x>0$ such that $(x-\eta_x,x+\eta_x)\subset U_x$.  Split the integral defining $\varepsilon_{j,t}^{(1)}(x)$ into the regions $|u|\le\eta_x/\overline h_t$ and $|u|>\eta_x/\overline h_t$.

On the first region, $x-h_{j,t}u\in U_x$.  By Lemma~\ref{fnw:lem:variance-integrand},
\begin{align}
\int_{|u|\le\eta_x/\overline h_t}\mathcal K_{j,t}(u)^2\left|\Omega_{j,t}(x-h_{j,t}u;x)-\Omega_{j,t}(x;x)\right|\,d u\le L_\Omega(x)\overline h_t\int_{\mathbb{R}}|u|\mathcal K_{j,t}(u)^2\,d u.
\end{align}
By Cauchy--Schwarz and Assumption~\ref{fnw:ass:kernel-growth},
\begin{align}
\sup_{t,j}\int_{\mathbb{R}}|u|\mathcal K_{j,t}(u)^2\,d u\le(C_{K,R}C_{K,2}^{(2)})^{1/2}<\infty.
\end{align}
Hence the inner-region contribution converges uniformly to zero.

On the second region, Lemma~\ref{fnw:lem:variance-integrand} gives
\begin{align}
\int_{|u|>\eta_x/\overline h_t}\mathcal K_{j,t}(u)^2\left|\Omega_{j,t}(x-h_{j,t}u;x)-\Omega_{j,t}(x;x)\right|\,d u\le2C_\Omega(x)\int_{|u|>\eta_x/\overline h_t}\mathcal K_{j,t}(u)^2\,d u.
\end{align}
Since $|u|>\eta_x/\overline h_t$ implies $1\le\overline h_t^2u^2/\eta_x^2$,
\begin{align}
\int_{|u|>\eta_x/\overline h_t}\mathcal K_{j,t}(u)^2\,d u\le\frac{\overline h_t^2}{\eta_x^2}\int_{\mathbb{R}}u^2\mathcal K_{j,t}(u)^2\,d u.
\end{align}
Therefore, by Assumption~\ref{fnw:ass:kernel-growth},
\begin{align}
\sup_{1\le j\le K_t}\int_{|u|>\eta_x/\overline h_t}\mathcal K_{j,t}(u)^2\,d u\le\frac{\overline h_t^2}{\eta_x^2}C_{K,2}^{(2)}\to0.
\end{align}
Combining the two regions yields
\begin{align}
\max_{1\le j\le K_t}|\varepsilon_{j,t}^{(1)}(x)|\to0.
\end{align}

We next control the squared first moment.  A change of variables gives
\begin{align}
\mu_{j,t}(x):=\bm{E}_{P_{j,t}}\left[\psi_{j,t}(x-X)[Y-m_t^\star(x)]\right]=\int_{\mathbb{R}}\mathcal K_{j,t}(u)\left\{g_{j,t}(x-h_{j,t}u)-m_t^\star(x)p_{j,t}(x-h_{j,t}u)\right\}\,d u.
\end{align}
By Lemma~\ref{fnw:lem:variance-integrand} and Assumptions~\ref{fnw:ass:kernel-growth} and \ref{fnw:ass:client-regularity}, there is a finite constant $C_\mu(x)$ independent of $j$ and $t$ such that
\begin{align}
\sup_{1\le j\le K_t}|\mu_{j,t}(x)|\le C_\mu(x).
\end{align}
Consequently,
\begin{align}
\max_{1\le j\le K_t}h_{j,t}\mu_{j,t}(x)^2\le C_\mu(x)^2\overline h_t\to0.
\end{align}
Define
\begin{align}
\varepsilon_{j,t}(x)=\varepsilon_{j,t}^{(1)}(x)-h_{j,t}\mu_{j,t}(x)^2.
\end{align}
Then
\begin{align}
\max_{1\le j\le K_t}|\varepsilon_{j,t}(x)|\to0.
\end{align}
Finally,
\begin{align}
\Var_{P_{j,t}}\left(\psi_{j,t}(x-X)[Y-m_t^\star(x)]\right)=\bm{E}_{P_{j,t}}\left[\psi_{j,t}(x-X)^2[Y-m_t^\star(x)]^2\right]-\mu_{j,t}(x)^2,
\end{align}
so substitution of the preceding second-moment expansion gives
\begin{align}
\Var_{P_{j,t}}\left(\psi_{j,t}(x-X)[Y-m_t^\star(x)]\right)=\frac{R(\mathcal K_{j,t})p_{j,t}(x)d_{j,t}^{(2)}(x;x)+\varepsilon_{j,t}(x)}{h_{j,t}}.
\end{align}
This proves \eqref{fnw:eq:client-variance-expansion}--\eqref{fnw:eq:variance-remainder-uniform}.
\end{proof}

\begin{theorem}[Order of the leading variance]
\label{fnw:thm:variance-order}
Under Assumptions~\ref{fnw:ass:kernel-growth}--\ref{fnw:ass:pointwise}, there are constants $0<c_V(x)\le C_V(x)<\infty$ such that, for all sufficiently large $t$,
\begin{align}
c_V(x)q_t\le V_t(x)\le C_V(x)q_t.
\label{fnw:eq:variance-order}
\end{align}
\end{theorem}

\begin{proof}
By \eqref{fnw:eq:centered-sum}, independence across clients and within-client observations gives
\begin{align}
V_t(x)=\frac{1}{B_t(x)^2}\sum_{j=1}^{K_t}\frac{w_{j,t}^2}{n_{j,t}}\Var_{P_{j,t}}\left(\psi_{j,t}(x-X)[Y-m_t^\star(x)]\right).
\end{align}
By Lemma~\ref{fnw:lem:variance-expansion},
\begin{align}
\Var_{P_{j,t}}\left(\psi_{j,t}(x-X)[Y-m_t^\star(x)]\right)=\frac{R(\mathcal K_{j,t})p_{j,t}(x)d_{j,t}^{(2)}(x;x)+\varepsilon_{j,t}(x)}{h_{j,t}},
\end{align}
with $\max_{j\le K_t}|\varepsilon_{j,t}(x)|\to0$.  Hence
\begin{align}
V_t(x)=\frac{1}{B_t(x)^2}\sum_{j=1}^{K_t}\frac{w_{j,t}^2}{n_{j,t}h_{j,t}}R(\mathcal K_{j,t})p_{j,t}(x)d_{j,t}^{(2)}(x;x)+\Delta_{V,t}(x),
\label{fnw:eq:variance-leading-term}
\end{align}
where
\begin{align}
\Delta_{V,t}(x)=\frac{1}{B_t(x)^2}\sum_{j=1}^{K_t}\frac{w_{j,t}^2}{n_{j,t}h_{j,t}}\varepsilon_{j,t}(x).
\end{align}
Since $B_t(x)\ge c_B$,
\begin{align}
|\Delta_{V,t}(x)|\le\frac{1}{c_B^2}\max_{1\le j\le K_t}|\varepsilon_{j,t}(x)|q_t=o(q_t).
\end{align}
It remains to compare the leading term in
\eqref{fnw:eq:variance-leading-term} with $q_t$.  The lower bound is exactly
Assumption~\ref{fnw:ass:pointwise}:
\begin{align}
\frac{1}{B_t(x)^2}\sum_{j=1}^{K_t}
\frac{w_{j,t}^2}{n_{j,t}h_{j,t}}
R(\mathcal K_{j,t})p_{j,t}(x)d_{j,t}^{(2)}(x;x)
\ge c_{V,0}(x)q_t.
\end{align}
For the upper bound, use
\[
p_{j,t}(x)d_{j,t}^{(2)}(x;x)
=
\Omega_{j,t}(x;x)
\le C_\Omega(x),
\]
Lemma~\ref{fnw:lem:variance-integrand}, the upper kernel-energy bound, and
$B_t(x)\ge c_B$.  Hence
\begin{align}
\frac{1}{B_t(x)^2}\sum_{j=1}^{K_t}
\frac{w_{j,t}^2}{n_{j,t}h_{j,t}}
R(\mathcal K_{j,t})p_{j,t}(x)d_{j,t}^{(2)}(x;x)
\le
\frac{C_{K,R}C_\Omega(x)}{c_B^2}q_t.
\end{align}
Thus one may take
$c_0(x)=c_{V,0}(x)$ and
$C_0(x)=C_{K,R}C_\Omega(x)/c_B^2$.
Since $\Delta_{V,t}(x)=o(q_t)$, for all sufficiently large $t$,
\begin{align}
|\Delta_{V,t}(x)|\le\frac{c_0(x)}{2}q_t.
\end{align}
Consequently,
\begin{align}
\frac{c_0(x)}{2}q_t\le V_t(x)\le\left\{C_0(x)+\frac{c_0(x)}{2}\right\}q_t.
\end{align}
Setting
\begin{align}
c_V(x)=\frac{c_0(x)}{2}, \qquad C_V(x)=C_0(x)+\frac{c_0(x)}{2}
\end{align}
proves \eqref{fnw:eq:variance-order}.
\end{proof}

\subsection{Pointwise consistency and central limit theorem}

\begin{lemma}[Smoothing-bias bounds]
\label{fnw:lem:bias}
Under Assumptions~\ref{fnw:ass:kernel-growth}--\ref{fnw:ass:pointwise},
\begin{align}
|\bm{E}[\Delta_t^{\mathrm{NW}}(x)]|\le C_{\mathrm{bias}}(x)\sum_{j=1}^{K_t}w_{j,t}h_{j,t}^2=o(1).
\label{fnw:eq:ordinary-bias}
\end{align}
If Assumption~\ref{fnw:ass:undersmoothing} also holds, then
\begin{align}
\bm{E}[\Delta_t^{\mathrm{NW}}(x)]=o(\sqrt{V_t(x)}).
\label{fnw:eq:stochastic-scale-bias}
\end{align}
\end{lemma}

\begin{proof}
Using $A_t(x)=m_t^\star(x)B_t(x)$ and the definition of $\Delta_t^{\mathrm{NW}}(x)$,
\begin{align}
\bm{E}[\Delta_t^{\mathrm{NW}}(x)]=\frac{1}{B_t(x)}\sum_{j=1}^{K_t}w_{j,t}\left(\bm{E}[\widehat g_{j,t}(x)]-g_{j,t}(x)-m_t^\star(x)\{\bm{E}[\widehat p_{j,t}(x)]-p_{j,t}(x)\}\right).
\end{align}
For the numerator term, a change of variables gives
\begin{align}
\bm{E}[\widehat g_{j,t}(x)]=\int_{\mathbb{R}}\mathcal K_{j,t}(u)g_{j,t}(x-h_{j,t}u)\,d u.
\end{align}
By a second-order Taylor expansion around $x$,
\begin{align}
g_{j,t}(x-h_{j,t}u)=g_{j,t}(x)-h_{j,t}ug_{j,t}'(x)+\frac12h_{j,t}^2u^2g_{j,t}''(\xi_{j,t,u}^{(g)}),
\end{align}
where $\xi_{j,t,u}^{(g)}$ lies between $x$ and $x-h_{j,t}u$.  Since $\int_{\mathbb{R}}\mathcal K_{j,t}(u)\,d u=1$ and $\int_{\mathbb{R}}u\mathcal K_{j,t}(u)\,d u=0$,
\begin{align}
\bm{E}[\widehat g_{j,t}(x)]-g_{j,t}(x)=\frac12h_{j,t}^2\int_{\mathbb{R}}u^2\mathcal K_{j,t}(u)g_{j,t}''(\xi_{j,t,u}^{(g)})\,d u.
\end{align}
Hence, by Assumptions~\ref{fnw:ass:kernel-growth} and \ref{fnw:ass:client-regularity},
\begin{align}
|\bm{E}[\widehat g_{j,t}(x)]-g_{j,t}(x)|\le\frac12C_{g,2}C_{K,2}h_{j,t}^2.
\end{align}
The same Taylor argument applied to $p_{j,t}$ gives
\begin{align}
|\bm{E}[\widehat p_{j,t}(x)]-p_{j,t}(x)|\le\frac12C_{p,2}C_{K,2}h_{j,t}^2.
\end{align}
By Lemma~\ref{fnw:lem:variance-integrand}, $|m_t^\star(x)|\le C_m(x)$ uniformly in $t$.  Together with $B_t(x)\ge c_B$,
\begin{align}
|\bm{E}[\Delta_t^{\mathrm{NW}}(x)]|\le\frac{C_{K,2}}{2c_B}\{C_{g,2}+C_m(x)C_{p,2}\}\sum_{j=1}^{K_t}w_{j,t}h_{j,t}^2.
\end{align}
Thus one may take
\begin{align}
C_{\mathrm{bias}}(x)=\frac{C_{K,2}}{2c_B}\{C_{g,2}+C_m(x)C_{p,2}\}.
\end{align}
Since $w_{j,t}\ge0$ and $\sum_{j=1}^{K_t}w_{j,t}=1$,
\begin{align}
\sum_{j=1}^{K_t}w_{j,t}h_{j,t}^2\le\max_{1\le j\le K_t}h_{j,t}^2\to0.
\end{align}
This proves \eqref{fnw:eq:ordinary-bias}.

If Assumption~\ref{fnw:ass:undersmoothing} also holds, Theorem~\ref{fnw:thm:variance-order} gives, for all sufficiently large $t$,
\begin{align}
V_t(x)\ge c_V(x)q_t.
\end{align}
Therefore,
\begin{align}
\frac{|\bm{E}[\Delta_t^{\mathrm{NW}}(x)]|}{\sqrt{V_t(x)}}\le\frac{C_{K,2}\{C_{g,2}+C_m(x)C_{p,2}\}}{2c_B\sqrt{c_V(x)}}\frac{\sum_{j=1}^{K_t}w_{j,t}h_{j,t}^2}{\sqrt{q_t}}\to0.
\end{align}
Hence \eqref{fnw:eq:stochastic-scale-bias} follows.
\end{proof}

\begin{lemma}[Pointwise consistency]
\label{fnw:lem:consistency}
Under Assumptions~\ref{fnw:ass:kernel-growth}--\ref{fnw:ass:pointwise},
\begin{align}
\widehat m_t(x)-m_t^\star(x)\xrightarrow[]{P}0.
\end{align}
\end{lemma}

\begin{proof}
By Theorem~\ref{fnw:thm:variance-order},
\begin{align}
V_t(x)\le C_V(x)q_t.
\end{align}
Moreover,
\begin{align}
q_t=\sum_{j=1}^{K_t}\frac{w_{j,t}^2}{n_{j,t}h_{j,t}}\le\frac{1}{\min_{1\le j\le K_t}n_{j,t}h_{j,t}}\sum_{j=1}^{K_t}w_{j,t}^2\le\frac{1}{\min_{1\le j\le K_t}n_{j,t}h_{j,t}}\to0.
\end{align}
Hence,
\begin{align}
\Delta_t^{\mathrm{NW}}(x)-\bm{E}[\Delta_t^{\mathrm{NW}}(x)]=O_p(\sqrt{V_t(x)})=o_p(1).
\end{align}
By Lemma~\ref{fnw:lem:bias},
\begin{align}
\bm{E}[\Delta_t^{\mathrm{NW}}(x)]=o(1).
\end{align}
Therefore,
\begin{align}
\Delta_t^{\mathrm{NW}}(x)=o_p(1).
\end{align}
By Lemma~\ref{fnw:lem:denominator},
\begin{align}
\frac{B_t(x)}{\widehat B_t(x)}=1+o_p(1).
\end{align}
Using the exact identity
\begin{align}
\widehat m_t(x)-m_t^\star(x)=\frac{B_t(x)}{\widehat B_t(x)}\Delta_t^{\mathrm{NW}}(x),
\end{align}
we obtain
\begin{align}
\widehat m_t(x)-m_t^\star(x)=\{1+o_p(1)\}o_p(1)=o_p(1).
\end{align}
Thus
\begin{align}
\widehat m_t(x)-m_t^\star(x)\xrightarrow[]{P}0.
\end{align}
\end{proof}

\begin{lemma}[Negligibility of the ratio remainder]
\label{fnw:lem:ratio-remainder}
Under Assumptions~\ref{fnw:ass:kernel-growth}--\ref{fnw:ass:pointwise} and Assumption~\ref{fnw:ass:undersmoothing},
\begin{align}
\operatorname{rem}_t^{\mathrm{NW}}(x)=o_p(\sqrt{V_t(x)}).
\label{fnw:eq:ratio-remainder-small}
\end{align}
\end{lemma}

\begin{proof}
By the exact ratio decomposition,
\begin{align}
\widehat m_t(x)-m_t^\star(x)
&=\Delta_t^{\mathrm{NW}}(x)+\operatorname{rem}_t^{\mathrm{NW}}(x),\notag\\
\operatorname{rem}_t^{\mathrm{NW}}(x)
&=\left\{\frac{B_t(x)}{\widehat B_t(x)}-1\right\}
  \Delta_t^{\mathrm{NW}}(x).
\label{fnw:eq:ratio-decomposition-supp}
\end{align}
By Lemma~\ref{fnw:lem:denominator},
\begin{align}
\frac{\widehat B_t(x)-B_t(x)}{B_t(x)}=o_p(1),
\end{align}
so
\begin{align}
\frac{\widehat B_t(x)}{B_t(x)}=1+o_p(1).
\end{align}
Since the map $z\mapsto z^{-1}$ is continuous at $z=1$,
\begin{align}
\frac{B_t(x)}{\widehat B_t(x)}=\left\{\frac{\widehat B_t(x)}{B_t(x)}\right\}^{-1}=1+o_p(1).
\end{align}
Hence,
\begin{align}
\frac{B_t(x)}{\widehat B_t(x)}-1=o_p(1).
\end{align}
Next decompose
\begin{align}
\frac{\Delta_t^{\mathrm{NW}}(x)}{\sqrt{V_t(x)}}=\frac{\Delta_t^{\mathrm{NW}}(x)-\bm{E}[\Delta_t^{\mathrm{NW}}(x)]}{\sqrt{V_t(x)}}+\frac{\bm{E}[\Delta_t^{\mathrm{NW}}(x)]}{\sqrt{V_t(x)}}.
\end{align}
By the definition of $V_t(x)$,
\begin{align}
\Var\left[\frac{\Delta_t^{\mathrm{NW}}(x)-\bm{E}[\Delta_t^{\mathrm{NW}}(x)]}{\sqrt{V_t(x)}}\right]=1.
\end{align}
Chebyshev's inequality therefore implies
\begin{align}
\frac{\Delta_t^{\mathrm{NW}}(x)-\bm{E}[\Delta_t^{\mathrm{NW}}(x)]}{\sqrt{V_t(x)}}=O_p(1).
\end{align}
By Lemma~\ref{fnw:lem:bias} and Assumption~\ref{fnw:ass:undersmoothing},
\begin{align}
\frac{\bm{E}[\Delta_t^{\mathrm{NW}}(x)]}{\sqrt{V_t(x)}}\to0.
\end{align}
Consequently,
\begin{align}
\frac{\Delta_t^{\mathrm{NW}}(x)}{\sqrt{V_t(x)}}=O_p(1).
\end{align}
It follows that
\begin{align}
\frac{\operatorname{rem}_t^{\mathrm{NW}}(x)}{\sqrt{V_t(x)}}=\left\{\frac{B_t(x)}{\widehat B_t(x)}-1\right\}\frac{\Delta_t^{\mathrm{NW}}(x)}{\sqrt{V_t(x)}}=o_p(1)O_p(1)=o_p(1).
\end{align}
Therefore,
\begin{align}
\operatorname{rem}_t^{\mathrm{NW}}(x)=o_p(\sqrt{V_t(x)}).
\end{align}
\end{proof}

\begin{lemma}[Lyapunov condition]
\label{fnw:lem:lyapunov}
Under Assumptions~\ref{fnw:ass:kernel-growth}--\ref{fnw:ass:higher-moment},
\begin{align}
\frac{\sum_{j=1}^{K_t}\sum_{i=1}^{n_{j,t}}\bm{E}[|\zeta_{i,j,t}(x)|^{2+\delta}]}{V_t(x)^{1+\delta/2}}\to0.
\label{fnw:eq:lyapunov}
\end{align}
\end{lemma}

\begin{proof}
Let
\begin{align}
\eta_{i,j,t}(x)=\psi_{j,t}(x-X_{i,j,t})[Y_{i,j,t}-m_t^\star(x)].
\label{fnw:eq:eta}
\end{align}
Then
\begin{align}
\zeta_{i,j,t}(x)=\frac{w_{j,t}}{n_{j,t}B_t(x)}\left\{\eta_{i,j,t}(x)-\bm{E}_{P_{j,t}}[\eta_{i,j,t}(x)]\right\}.
\end{align}
Let $p=2+\delta$.  By the inequality $|a+b|^p\le2^{p-1}(|a|^p+|b|^p)$,
\begin{align}
|\eta-\bm{E}[\eta]|^p\le2^{p-1}\{|\eta|^p+|\bm{E}[\eta]|^p\}.
\end{align}
Taking expectations and applying Jensen's inequality gives
\begin{align}
\bm{E}[|\eta-\bm{E}[\eta]|^p]\le2^p\bm{E}[|\eta|^p].
\end{align}
Therefore,
\begin{align}
\bm{E}[|\zeta_{i,j,t}(x)|^{2+\delta}]\le2^{2+\delta}\left\{\frac{w_{j,t}}{n_{j,t}B_t(x)}\right\}^{2+\delta}\bm{E}_{P_{j,t}}[|\eta_{i,j,t}(x)|^{2+\delta}].
\end{align}
By the law of iterated expectations,
\begin{align}
\bm{E}_{P_{j,t}}[|\eta_{i,j,t}(x)|^{2+\delta}]=\int_{\mathbb{R}}\frac{1}{h_{j,t}^{2+\delta}}\left|\mathcal K_{j,t}\left(\frac{x-z}{h_{j,t}}\right)\right|^{2+\delta}\bm{E}_{P_{j,t}}[|Y-m_t^\star(x)|^{2+\delta}\mid X=z]p_{j,t}(z)\,d z.
\end{align}
After the change of variables $u=(x-z)/h_{j,t}$,
\begin{align}
\bm{E}_{P_{j,t}}[|\eta_{i,j,t}(x)|^{2+\delta}]=\frac{1}{h_{j,t}^{1+\delta}}\int_{\mathbb{R}}|\mathcal K_{j,t}(u)|^{2+\delta}\bm{E}_{P_{j,t}}[|Y-m_t^\star(x)|^{2+\delta}\mid X=x-h_{j,t}u]p_{j,t}(x-h_{j,t}u)\,d u.
\end{align}
Using $|a-b|^{2+\delta}\le2^{1+\delta}(|a|^{2+\delta}+|b|^{2+\delta})$,
\begin{align}
p_{j,t}(z)\bm{E}_{P_{j,t}}[|Y-m_t^\star(x)|^{2+\delta}\mid X=z]\le2^{1+\delta}\left\{p_{j,t}(z)\bm{E}_{P_{j,t}}[|Y|^{2+\delta}\mid X=z]+p_{j,t}(z)|m_t^\star(x)|^{2+\delta}\right\}.
\end{align}
By Assumptions~\ref{fnw:ass:higher-moment} and \ref{fnw:ass:client-regularity}, together with Lemma~\ref{fnw:lem:variance-integrand}, there is a finite constant $C_\eta(x)$ independent of $j$, $t$, and $z$ such that
\begin{align}
p_{j,t}(z)\bm{E}_{P_{j,t}}[|Y-m_t^\star(x)|^{2+\delta}\mid X=z]\le C_\eta(x).
\end{align}
Hence,
\begin{align}
\bm{E}_{P_{j,t}}[|\eta_{i,j,t}(x)|^{2+\delta}]\le\frac{C_\eta(x)}{h_{j,t}^{1+\delta}}\int_{\mathbb{R}}|\mathcal K_{j,t}(u)|^{2+\delta}\,d u\le\frac{C_\eta(x)C_{K,2+\delta}}{h_{j,t}^{1+\delta}}.
\end{align}
Since $B_t(x)\ge c_B$, there is a finite constant $C_\zeta(x)$ independent of $j$ and $t$ such that
\begin{align}
\bm{E}[|\zeta_{i,j,t}(x)|^{2+\delta}]\le C_\zeta(x)\frac{w_{j,t}^{2+\delta}}{n_{j,t}^{2+\delta}h_{j,t}^{1+\delta}}.
\end{align}
Summing over observations and clients gives
\begin{align}
\sum_{j=1}^{K_t}\sum_{i=1}^{n_{j,t}}\bm{E}[|\zeta_{i,j,t}(x)|^{2+\delta}]\le C_\zeta(x)\sum_{j=1}^{K_t}\frac{w_{j,t}^{2+\delta}}{(n_{j,t}h_{j,t})^{1+\delta}}.
\end{align}
Define
\begin{align}
\underline\nu_t=\min_{1\le j\le K_t}n_{j,t}h_{j,t}
\end{align}
and
\begin{align}
\chi_{j,t}=\frac{w_{j,t}^2}{n_{j,t}h_{j,t}},
\end{align}
so that $q_t=\sum_{j=1}^{K_t}\chi_{j,t}$.  For each $j$,
\begin{align}
\frac{w_{j,t}^{2+\delta}}{(n_{j,t}h_{j,t})^{1+\delta}}=\chi_{j,t}^{1+\delta/2}(n_{j,t}h_{j,t})^{-\delta/2}.
\end{align}
Since $n_{j,t}h_{j,t}\ge\underline\nu_t$,
\begin{align}
\frac{w_{j,t}^{2+\delta}}{(n_{j,t}h_{j,t})^{1+\delta}}\le\underline\nu_t^{-\delta/2}\chi_{j,t}^{1+\delta/2}.
\end{align}
Therefore,
\begin{align}
\sum_{j=1}^{K_t}\frac{w_{j,t}^{2+\delta}}{(n_{j,t}h_{j,t})^{1+\delta}}\le\underline\nu_t^{-\delta/2}\sum_{j=1}^{K_t}\chi_{j,t}^{1+\delta/2}.
\end{align}
Since $1+\delta/2>1$ and $\chi_{j,t}\ge0$,
\begin{align}
\sum_{j=1}^{K_t}\chi_{j,t}^{1+\delta/2}\le\left\{\sum_{j=1}^{K_t}\chi_{j,t}\right\}^{1+\delta/2}=q_t^{1+\delta/2}.
\end{align}
Thus,
\begin{align}
\sum_{j=1}^{K_t}\sum_{i=1}^{n_{j,t}}\bm{E}[|\zeta_{i,j,t}(x)|^{2+\delta}]\le C_\zeta(x)\underline\nu_t^{-\delta/2}q_t^{1+\delta/2}.
\end{align}
By Theorem~\ref{fnw:thm:variance-order},
\begin{align}
V_t(x)\ge c_V(x)q_t,
\end{align}
so
\begin{align}
V_t(x)^{1+\delta/2}\ge c_V(x)^{1+\delta/2}q_t^{1+\delta/2}.
\end{align}
Consequently,
\begin{align}
\frac{\sum_{j=1}^{K_t}\sum_{i=1}^{n_{j,t}}\bm{E}[|\zeta_{i,j,t}(x)|^{2+\delta}]}{V_t(x)^{1+\delta/2}}\le\frac{C_\zeta(x)}{c_V(x)^{1+\delta/2}}\underline\nu_t^{-\delta/2}.
\end{align}
By Assumption~\ref{fnw:ass:kernel-growth}, $\underline\nu_t\to\infty$.  Therefore the right-hand side converges to zero, which proves \eqref{fnw:eq:lyapunov}.
\end{proof}

\begin{lemma}[Oracle pointwise central limit theorem]
\label{fnw:lem:oracle-clt}
Under Assumptions~\ref{fnw:ass:kernel-growth}--\ref{fnw:ass:undersmoothing},
\begin{align}
\frac{\widehat m_t(x)-m_t^\star(x)}{\sqrt{V_t(x)}}\xrightarrow[]{D}N(0,1).
\label{fnw:eq:oracle-clt-appendix}
\end{align}
\end{lemma}

\begin{proof}
For each stage $t$, the collection
\begin{align}
\{\zeta_{i,j,t}(x):1\le j\le K_t,\ 1\le i\le n_{j,t}\}
\end{align}
consists of independent mean-zero random variables.  Although both $K_t$ and the local sample sizes may vary with $t$, this collection forms the $t$th row of a triangular array with
\begin{align}
N_t=\sum_{j=1}^{K_t}n_{j,t}
\end{align}
entries.  By Lemma~\ref{fnw:lem:lyapunov},
\begin{align}
\frac{\sum_{j=1}^{K_t}\sum_{i=1}^{n_{j,t}}\bm{E}[|\zeta_{i,j,t}(x)|^{2+\delta}]}{V_t(x)^{1+\delta/2}}\to0.
\end{align}
Moreover, by \eqref{fnw:eq:centered-sum},
\begin{align}
V_t(x)=\sum_{j=1}^{K_t}\sum_{i=1}^{n_{j,t}}\bm{E}[\zeta_{i,j,t}(x)^2].
\end{align}
Therefore, the Lyapunov central limit theorem for triangular arrays gives
\begin{align}
\frac{\sum_{j=1}^{K_t}\sum_{i=1}^{n_{j,t}}\zeta_{i,j,t}(x)}{\sqrt{V_t(x)}}\xrightarrow[]{D}N(0,1).
\end{align}
Since
\begin{align}
\sum_{j=1}^{K_t}\sum_{i=1}^{n_{j,t}}\zeta_{i,j,t}(x)=\Delta_t^{\mathrm{NW}}(x)-\bm{E}[\Delta_t^{\mathrm{NW}}(x)],
\end{align}
we obtain
\begin{align}
\frac{\Delta_t^{\mathrm{NW}}(x)-\bm{E}[\Delta_t^{\mathrm{NW}}(x)]}{\sqrt{V_t(x)}}\xrightarrow[]{D}N(0,1).
\end{align}
By Lemma~\ref{fnw:lem:bias},
\begin{align}
\frac{\bm{E}[\Delta_t^{\mathrm{NW}}(x)]}{\sqrt{V_t(x)}}\to0.
\end{align}
Slutsky's theorem therefore yields
\begin{align}
\frac{\Delta_t^{\mathrm{NW}}(x)}{\sqrt{V_t(x)}}\xrightarrow[]{D}N(0,1).
\end{align}
Finally, the exact decomposition \eqref{fnw:eq:ratio-decomposition-supp} gives
\begin{align}
\frac{\widehat m_t(x)-m_t^\star(x)}{\sqrt{V_t(x)}}=\frac{\Delta_t^{\mathrm{NW}}(x)}{\sqrt{V_t(x)}}+\frac{\operatorname{rem}_t^{\mathrm{NW}}(x)}{\sqrt{V_t(x)}}.
\end{align}
By Lemma~\ref{fnw:lem:ratio-remainder}, the second term is $o_p(1)$.  Another application of Slutsky's theorem gives
\begin{align}
\frac{\widehat m_t(x)-m_t^\star(x)}{\sqrt{V_t(x)}}\xrightarrow[]{D}N(0,1).
\end{align}
This proves \eqref{fnw:eq:oracle-clt-appendix}.
\end{proof}

\subsection{Consistency of the distributed variance estimator}

Define the oracle residual contribution
\begin{align}
\eta_{i,j,t}(x)=\psi_{j,t}(x-X_{i,j,t})[Y_{i,j,t}-m_t^\star(x)].
\end{align}
Define the oracle empirical second-moment numerator and its population counterpart by
\begin{align}
\widetilde{\mathcal J}_t(x)=\sum_{j=1}^{K_t}\frac{w_{j,t}^2}{n_{j,t}^2}\sum_{i=1}^{n_{j,t}}\eta_{i,j,t}(x)^2, \qquad \mathcal J_t(x)=\sum_{j=1}^{K_t}\frac{w_{j,t}^2}{n_{j,t}}\bm{E}_{P_{j,t}}[\eta_{i,j,t}(x)^2].
\label{fnw:eq:J-oracle}
\end{align}
Also define the feasible numerator
\begin{align}
\widehat{\mathcal J}_t(x)=\sum_{j=1}^{K_t}\frac{w_{j,t}^2}{n_{j,t}^2}\sum_{i=1}^{n_{j,t}}\psi_{j,t}^2(x-X_{i,j,t})[Y_{i,j,t}-\widehat m_t(x)]^2.
\label{fnw:eq:J-hat}
\end{align}
Then $\widehat V_t(x)=\widehat{\mathcal J}_t(x)/\widehat B_t(x)^2$.

\begin{lemma}[Concentration of the oracle second moment]
\label{fnw:lem:oracle-second-moment}
Under Assumptions~\ref{fnw:ass:kernel-growth}--\ref{fnw:ass:higher-moment},
\begin{align}
\widetilde{\mathcal J}_t(x)-\mathcal J_t(x)=o_p(q_t).
\label{fnw:eq:J-concentration}
\end{align}
\end{lemma}

\begin{proof}
By decreasing $\delta$ if necessary, we may take $0<\delta\le2$, since Assumption~\ref{fnw:ass:higher-moment} also implies the corresponding bounds for every smaller positive exponent.  Define
\begin{align}
\nu=1+\frac{\delta}{2}.
\end{align}
Then $1<\nu\le2$ and $2\nu=2+\delta$.  For each $i$, $j$, and $t$, define
\begin{align}
\xi_{i,j,t}(x)=\frac{w_{j,t}^2}{n_{j,t}^2}\left\{\eta_{i,j,t}(x)^2-\bm{E}_{P_{j,t}}[\eta_{i,j,t}(x)^2]\right\}.
\end{align}
Then
\begin{align}
\widetilde{\mathcal J}_t(x)-\mathcal J_t(x)=\sum_{j=1}^{K_t}\sum_{i=1}^{n_{j,t}}\xi_{i,j,t}(x).
\end{align}
The variables $\xi_{i,j,t}(x)$ are independent and mean zero.  Since $1<\nu\le2$, the von Bahr--Esseen inequality gives
\begin{align}
\bm{E}\left[|\widetilde{\mathcal J}_t(x)-\mathcal J_t(x)|^\nu\right]\le C_\nu\sum_{j=1}^{K_t}\sum_{i=1}^{n_{j,t}}\bm{E}[|\xi_{i,j,t}(x)|^\nu].
\end{align}
Using the standard centered-moment inequality,
\begin{align}
\bm{E}\left[|\eta_{i,j,t}(x)^2-\bm{E}_{P_{j,t}}[\eta_{i,j,t}(x)^2]|^\nu\right]\le C_\nu\bm{E}_{P_{j,t}}[|\eta_{i,j,t}(x)|^{2+\delta}].
\end{align}
The calculation in Lemma~\ref{fnw:lem:lyapunov} gives a finite constant $C_\eta^\star(x)$ independent of $j$ and $t$ such that
\begin{align}
\bm{E}_{P_{j,t}}[|\eta_{i,j,t}(x)|^{2+\delta}]\le\frac{C_\eta^\star(x)}{h_{j,t}^{1+\delta}}.
\end{align}
Therefore, for some finite constant $C_J(x)$,
\begin{align}
\bm{E}\left[|\widetilde{\mathcal J}_t(x)-\mathcal J_t(x)|^\nu\right]\le C_J(x)\sum_{j=1}^{K_t}\frac{w_{j,t}^{2+\delta}}{(n_{j,t}h_{j,t})^{1+\delta}}.
\end{align}
Define
\begin{align}
\underline\nu_t=\min_{1\le j\le K_t}n_{j,t}h_{j,t}.
\end{align}
As in the proof of Lemma~\ref{fnw:lem:lyapunov},
\begin{align}
\sum_{j=1}^{K_t}\frac{w_{j,t}^{2+\delta}}{(n_{j,t}h_{j,t})^{1+\delta}}\le\underline\nu_t^{-\delta/2}q_t^{1+\delta/2}.
\end{align}
Since $\nu=1+\delta/2$,
\begin{align}
\bm{E}\left[|\widetilde{\mathcal J}_t(x)-\mathcal J_t(x)|^\nu\right]\le C_J(x)\underline\nu_t^{-\delta/2}q_t^\nu.
\end{align}
Hence, for every $\varepsilon>0$, Markov's inequality gives
\begin{align}
\bm{P}\left(|\widetilde{\mathcal J}_t(x)-\mathcal J_t(x)|>\varepsilon q_t\right)\le\frac{C_J(x)}{\varepsilon^\nu}\underline\nu_t^{-\delta/2}.
\end{align}
By Assumption~\ref{fnw:ass:kernel-growth},
\begin{align}
\underline\nu_t\to\infty.
\end{align}
Therefore,
\begin{align}
\widetilde{\mathcal J}_t(x)-\mathcal J_t(x)=o_p(q_t).
\end{align}
This proves \eqref{fnw:eq:J-concentration}.
\end{proof}

\begin{lemma}[Negligibility of plug-in centering]
\label{fnw:lem:plugin-centering}
Under Assumptions~\ref{fnw:ass:kernel-growth}--\ref{fnw:ass:pointwise},
\begin{align}
\widehat{\mathcal J}_t(x)-\widetilde{\mathcal J}_t(x)=o_p(q_t).
\label{fnw:eq:plugin-centering}
\end{align}
\end{lemma}

\begin{proof}
Define
\begin{align}
e_t(x)=\widehat m_t(x)-m_t^\star(x).
\end{align}
By Lemma~\ref{fnw:lem:consistency},
\begin{align}
e_t(x)=o_p(1).
\end{align}
For each observation,
\begin{align}
[Y-\widehat m_t(x)]^2-[Y-m_t^\star(x)]^2=-2e_t(x)[Y-m_t^\star(x)]+e_t(x)^2.
\end{align}
Therefore,
\begin{align}
|\widehat{\mathcal J}_t(x)-\widetilde{\mathcal J}_t(x)|\le2|e_t(x)|\mathcal A_t(x)+e_t(x)^2\mathcal C_t(x),
\end{align}
where
\begin{align}
\mathcal A_t(x)=\sum_{j=1}^{K_t}\frac{w_{j,t}^2}{n_{j,t}^2}\sum_{i=1}^{n_{j,t}}\psi_{j,t}^2(x-X_{i,j,t})|Y_{i,j,t}-m_t^\star(x)|
\end{align}
and
\begin{align}
\mathcal C_t(x)=\sum_{j=1}^{K_t}\frac{w_{j,t}^2}{n_{j,t}^2}\sum_{i=1}^{n_{j,t}}\psi_{j,t}^2(x-X_{i,j,t}).
\end{align}
For $\mathcal A_t(x)$, a change of variables gives
\begin{align}
&\bm{E}_{P_{j,t}}[\psi_{j,t}^2(x-X)|Y-m_t^\star(x)|]
\nonumber\\
&\qquad=\frac{1}{h_{j,t}}\int_{\mathbb{R}}\mathcal K_{j,t}(u)^2
p_{j,t}(x-h_{j,t}u)
\bm{E}_{P_{j,t}}[|Y-m_t^\star(x)|\mid X=x-h_{j,t}u]\,d u.
\end{align}
By the conditional Cauchy--Schwarz inequality,
\begin{align}
p_{j,t}(z)\bm{E}_{P_{j,t}}[|Y-m_t^\star(x)|\mid X=z]\le\{p_{j,t}(z)\Omega_{j,t}(z;x)\}^{1/2}.
\end{align}
By Assumption~\ref{fnw:ass:client-regularity} and Lemma~\ref{fnw:lem:variance-integrand}, there is a finite constant $C_A(x)$ independent of $j$ and $t$ such that
\begin{align}
\bm{E}_{P_{j,t}}[\psi_{j,t}^2(x-X)|Y-m_t^\star(x)|]\le\frac{C_A(x)}{h_{j,t}}.
\end{align}
Consequently,
\begin{align}
\bm{E}[\mathcal A_t(x)]\le C_A(x)\sum_{j=1}^{K_t}\frac{w_{j,t}^2}{n_{j,t}h_{j,t}}=C_A(x)q_t.
\end{align}
Hence,
\begin{align}
\mathcal A_t(x)=O_p(q_t).
\end{align}
Similarly,
\begin{align}
\bm{E}_{P_{j,t}}[\psi_{j,t}^2(x-X)]=\frac{1}{h_{j,t}}\int_{\mathbb{R}}\mathcal K_{j,t}(u)^2p_{j,t}(x-h_{j,t}u)\,d u.
\end{align}
By Assumptions~\ref{fnw:ass:kernel-growth} and \ref{fnw:ass:client-regularity},
\begin{align}
\bm{E}_{P_{j,t}}[\psi_{j,t}^2(x-X)]\le\frac{C_{p,0}C_{K,R}}{h_{j,t}}.
\end{align}
Therefore,
\begin{align}
\bm{E}[\mathcal C_t(x)]\le C_{p,0}C_{K,R}\sum_{j=1}^{K_t}\frac{w_{j,t}^2}{n_{j,t}h_{j,t}}=C_{p,0}C_{K,R}q_t.
\end{align}
Hence,
\begin{align}
\mathcal C_t(x)=O_p(q_t).
\end{align}
Combining these bounds with $e_t(x)=o_p(1)$ gives
\begin{align}
|\widehat{\mathcal J}_t(x)-\widetilde{\mathcal J}_t(x)|\le2o_p(1)O_p(q_t)+o_p(1)^2O_p(q_t)=o_p(q_t).
\end{align}
This proves \eqref{fnw:eq:plugin-centering}.
\end{proof}

\begin{lemma}[Ratio consistency of the distributed variance estimator]
\label{fnw:lem:variance-consistency}
Under Assumptions~\ref{fnw:ass:kernel-growth}--\ref{fnw:ass:higher-moment},
\begin{align}
\widehat V_t(x)-V_t(x)=o_p(q_t), \qquad \frac{\widehat V_t(x)}{V_t(x)}\xrightarrow[]{P}1.
\label{fnw:eq:variance-consistency-appendix}
\end{align}
\end{lemma}

\begin{proof}
Recall that
\begin{align}
B_t(x)^2V_t(x)=\sum_{j=1}^{K_t}\frac{w_{j,t}^2}{n_{j,t}}\Var_{P_{j,t}}[\eta_{i,j,t}(x)].
\end{align}
Define
\begin{align}
\mathcal H_t(x)=\sum_{j=1}^{K_t}\frac{w_{j,t}^2}{n_{j,t}}\left\{\bm{E}_{P_{j,t}}[\eta_{i,j,t}(x)]\right\}^2.
\end{align}
Since $\mathcal J_t(x)$ is the corresponding sum of second moments,
\begin{align}
B_t(x)^2V_t(x)=\mathcal J_t(x)-\mathcal H_t(x).
\end{align}
The squared-first-moment bound established in the proof of Lemma~\ref{fnw:lem:variance-expansion} gives
\begin{align}
\max_{1\le j\le K_t}h_{j,t}\left\{\bm{E}_{P_{j,t}}[\eta_{i,j,t}(x)]\right\}^2\to0.
\end{align}
Therefore,
\begin{align}
\mathcal H_t(x)=\sum_{j=1}^{K_t}\frac{w_{j,t}^2}{n_{j,t}h_{j,t}}h_{j,t}\left\{\bm{E}_{P_{j,t}}[\eta_{i,j,t}(x)]\right\}^2\le q_t\max_{1\le j\le K_t}h_{j,t}\left\{\bm{E}_{P_{j,t}}[\eta_{i,j,t}(x)]\right\}^2=o(q_t).
\end{align}
Hence,
\begin{align}
\mathcal J_t(x)-B_t(x)^2V_t(x)=o(q_t).
\end{align}
Now decompose
\begin{align}
\widehat{\mathcal J}_t(x)-B_t(x)^2V_t(x)=\{\widehat{\mathcal J}_t(x)-\widetilde{\mathcal J}_t(x)\}+\{\widetilde{\mathcal J}_t(x)-\mathcal J_t(x)\}+\{\mathcal J_t(x)-B_t(x)^2V_t(x)\}.
\end{align}
By Lemmas~\ref{fnw:lem:plugin-centering} and \ref{fnw:lem:oracle-second-moment}, together with the preceding deterministic bound,
\begin{align}
\widehat{\mathcal J}_t(x)-B_t(x)^2V_t(x)=o_p(q_t).
\end{align}
Since
\begin{align}
\widehat V_t(x)=\frac{\widehat{\mathcal J}_t(x)}{\widehat B_t(x)^2},
\end{align}
we have the exact decomposition
\begin{align}
\widehat V_t(x)-V_t(x)=\frac{\widehat{\mathcal J}_t(x)-B_t(x)^2V_t(x)}{\widehat B_t(x)^2}+V_t(x)\left\{\frac{B_t(x)^2}{\widehat B_t(x)^2}-1\right\}.
\end{align}
By Lemma~\ref{fnw:lem:denominator},
\begin{align}
\frac{B_t(x)^2}{\widehat B_t(x)^2}=1+o_p(1).
\end{align}
Moreover, Assumption~\ref{fnw:ass:pointwise} and Lemma~\ref{fnw:lem:denominator} imply
\begin{align}
\frac{1}{\widehat B_t(x)^2}=O_p(1).
\end{align}
Hence,
\begin{align}
\frac{\widehat{\mathcal J}_t(x)-B_t(x)^2V_t(x)}{\widehat B_t(x)^2}=o_p(q_t).
\end{align}
By Theorem~\ref{fnw:thm:variance-order},
\begin{align}
V_t(x)=O(q_t).
\end{align}
Therefore,
\begin{align}
V_t(x)\left\{\frac{B_t(x)^2}{\widehat B_t(x)^2}-1\right\}=o_p(q_t).
\end{align}
Combining the two terms yields
\begin{align}
\widehat V_t(x)-V_t(x)=o_p(q_t).
\end{align}
Finally, Theorem~\ref{fnw:thm:variance-order} gives
\begin{align}
V_t(x)\ge c_V(x)q_t
\end{align}
for all sufficiently large $t$.  Therefore,
\begin{align}
\left|\frac{\widehat V_t(x)}{V_t(x)}-1\right|\le\frac{|\widehat V_t(x)-V_t(x)|}{c_V(x)q_t}=o_p(1).
\end{align}
Thus,
\begin{align}
\frac{\widehat V_t(x)}{V_t(x)}\xrightarrow[]{P}1.
\end{align}
This proves \eqref{fnw:eq:variance-consistency-appendix}.
\end{proof}

\begin{proof}[Proof of Theorem~\ref{fnw:thm:pointwise-inference}]
Under Assumptions~\ref{fnw:ass:kernel-growth}--\ref{fnw:ass:pointwise}, Lemma~\ref{fnw:lem:consistency} gives
\begin{align}
\widehat m_t(x)-m_t^\star(x)\xrightarrow[]{P}0,
\end{align}
This proves consistency.  Under the additional higher-moment and undersmoothing assumptions, Lemma~\ref{fnw:lem:oracle-clt} gives
\begin{align}
\frac{\widehat m_t(x)-m_t^\star(x)}{\sqrt{V_t(x)}}\xrightarrow[]{D}N(0,1),
\end{align}
This proves the oracle-normalized limit.  Under Assumption~\ref{fnw:ass:higher-moment}, Lemma~\ref{fnw:lem:variance-consistency} gives
\begin{align}
\frac{\widehat V_t(x)}{V_t(x)}\xrightarrow[]{P}1,
\end{align}
This proves variance ratio consistency.  Finally,
\begin{align}
\frac{\widehat m_t(x)-m_t^\star(x)}{\sqrt{\widehat V_t(x)}}=\frac{\widehat m_t(x)-m_t^\star(x)}{\sqrt{V_t(x)}}\left\{\frac{V_t(x)}{\widehat V_t(x)}\right\}^{1/2}.
\end{align}
The first factor converges in distribution to $N(0,1)$ and the second converges in probability to one.  Slutsky's theorem therefore yields
\begin{align}
\frac{\widehat m_t(x)-m_t^\star(x)}{\sqrt{\widehat V_t(x)}}\xrightarrow[]{D}N(0,1),
\end{align}
which proves the studentized conclusion.
\end{proof}

\section{One-client gain inference: detailed assumptions and proofs}
\label{app:one-client-proof}

\subsection{Triangular-array setup and gain targets}

For each stage $t$, let $K_t$ clients participate.  Fix an arbitrary client
sequence $k_t\le K_t$.  The observations used to construct the local and
federated predictors are collected in a fitting statistic $S_t$ and generate
\[
\mathcal{F}_t^{\mathrm{fit}}=\sigma(S_t).
\]
The selected client's fitted local predictor and the fitted server predictor
are denoted by
\[
L_t(x)=\widehat m_{k_t,L,t}(x),
\qquad
F_t(x)=\widehat m_t(x).
\]
The same observations from client $k_t$ may enter both $L_t$ and $F_t$.
Fitting samples from distinct clients, and any additional local-only sample,
are independent.

Let $Q_t$ be the prescribed evaluation law for the selected client coordinate,
with covariate marginal $Q_{t,X}$ and regression function
\[
m_t^Q(x)=\bm{E}_{Q_t}[Y\mid X=x].
\]
For $f,g\in L^2(Q_{t,X})$, write
\[
\inner{f}{g}_t=\int f(x)g(x)\,Q_{t,X}(\,d x),
\qquad
\norm{f}_t^2=\inner{f}{f}_t.
\]
Fix $\alpha\in[0,1)$ and put $\vartheta=1-\alpha\in(0,1]$.  The fitted
personalized predictor is
\[
M_t(x)=\alpha L_t(x)+\vartheta F_t(x).
\]
For $z=(x,y)$, define the paired gain
\begin{align}
W_t(z)
&=\{y-L_t(x)\}^2-\{y-M_t(x)\}^2 \label{sg:eq:paired-gain}\\
&=2\vartheta\{y-L_t(x)\}\{F_t(x)-L_t(x)\}
-\vartheta^2\{F_t(x)-L_t(x)\}^2. \nonumber
\end{align}
The gain of the predictors actually fitted at stage $t$ is
\begin{equation}
G_t^{\mathrm{fit}}
=
\int W_t(z)Q_t(\,d z)
=
\bm{E}\{W_t(Z)\mid\mathcal{F}_t^{\mathrm{fit}}\}.
\label{sg:eq:fitted-gain}
\end{equation}
Let $\bm{P}_t^{\mathrm{fit}}$ be the joint law of all fitting samples and let
$\bm{E}_t^{\mathrm{fit}}$ denote expectation under this law.  The deterministic
stage-$t$ marginal gain is
\begin{equation}
\Gamma_t=\bm{E}_t^{\mathrm{fit}}G_t^{\mathrm{fit}}.
\label{sg:eq:marginal-gain}
\end{equation}

Client $k_t$ also has an evaluation sample
\[
Z_{t,1}^{\mathrm{eval}},\ldots,Z_{t,N_t^{\mathrm{eval}}}^{\mathrm{eval}},
\qquad N_t^{\mathrm{eval}}=N_{k_t,t}^{\mathrm{eval}},
\]
independent of $\mathcal{F}_t^{\mathrm{fit}}$ and i.i.d. from $Q_t$.  The empirical gain
is
\begin{equation}
\widehat G_t
=
\frac1{N_t^{\mathrm{eval}}}\sum_{r=1}^{N_t^{\mathrm{eval}}}W_t(Z_{t,r}^{\mathrm{eval}}).
\label{sg:eq:empirical-gain}
\end{equation}
The exact two-stage decomposition is
\begin{equation}
\widehat G_t-\Gamma_t
=
\underbrace{G_t^{\mathrm{fit}}-\Gamma_t}_{H_t^{\mathrm{fit}}}
+
\underbrace{\widehat G_t-G_t^{\mathrm{fit}}}_{E_t^{\mathrm{eval}}}.
\label{sg:eq:two-stage}
\end{equation}

Conditionally on the fitted predictors, define
\begin{align}
A_t^{\mathrm{fit}}
&=
\bm{E}\!\left[
\{Y-L_t(X)\}\{F_t(X)-L_t(X)\}
\mid\mathcal{F}_t^{\mathrm{fit}}
\right],\label{sg:eq:Afit}\\
D_t^{\mathrm{fit}}
&=
\bm{E}\!\left[
\{F_t(X)-L_t(X)\}^2
\mid\mathcal{F}_t^{\mathrm{fit}}
\right].\label{sg:eq:Dfit}
\end{align}
Then
\begin{equation}
G_t^{\mathrm{fit}}
=2\vartheta A_t^{\mathrm{fit}}
-\vartheta^2D_t^{\mathrm{fit}}.
\label{sg:eq:gain-AD}
\end{equation}
Thus, for the prescribed borrowing amount $\vartheta$,
\[
G_t^{\mathrm{fit}}>0
\quad\Longleftrightarrow\quad
2A_t^{\mathrm{fit}}>\vartheta D_t^{\mathrm{fit}},
\]
and, when $D_t^{\mathrm{fit}}>0$, some sufficiently small positive amount of
borrowing is beneficial if and only if $A_t^{\mathrm{fit}}>0$.

\subsection{Primitive setup and leverage notation}

The primitive kernel, moment, effective-information, first-order nondegeneracy,
and evaluation assumptions are stated in Section~\ref{sec:gain-inference}
(Assumptions~\ref{sg:ass:kernel}--\ref{sg:ass:eval}).  The theorem is
stated for one-dimensional covariates and a common bandwidth; this
specialization is used only in the proof of the fitting-sample result.

To describe the leverage structure, index all mutually independent fitting
observations by $a\in\mathcal{I}_t$.  Let $r_{a,L,t}$ and $r_{a,F,t}$ be the coefficients
with which observation $a$ enters the local and federated empirical averages.
For example, $r_{a,L,t}=N_{L,t}^{-1}$ for an observation entering the updated
local fit and $r_{a,F,t}=w_{j,t}/n_{j,t}$ for an observation from client $j$
entering the server fit.  Put
\[
\gamma_{a,t}=|r_{a,L,t}|+|r_{a,F,t}|,
\qquad
\mathfrak r_t=\sum_{a\in\mathcal{I}_t}\gamma_{a,t}^2,
\qquad
\gamma_{\max,t}=\max_{a\in\mathcal{I}_t}\gamma_{a,t}.
\]

Let
\[
\overline L_t=\bm{E}_t^{\mathrm{fit}}L_t,
\qquad
\overline F_t=\bm{E}_t^{\mathrm{fit}}F_t,
\]
and define
\[
b_{L,t}=\overline L_t-m_t^Q,
\qquad
b_{F,t}=\overline F_t-m_t^Q.
\]
The derivative coefficient functions of the gain are
\begin{align}
c_{L,t}
&=2\vartheta\{(1+\alpha)b_{L,t}-\alpha b_{F,t}\},\label{sg:eq:cL}\\
c_{F,t}
&=-2\vartheta\{\alpha b_{L,t}+\vartheta b_{F,t}\}.\label{sg:eq:cF}
\end{align}

\subsection{Fitting-sample linearization}

Let $p_{L,t}^h,m_{L,t}^h$ denote the selected client's finite-bandwidth local
denominator and ratio target.  Let $B_t^h,m_{F,t}^h$ denote the corresponding
federated denominator and ratio target.  For an observation $a$ from client
$j(a)$, define centered residual functions
\begin{align}
R_{a,L,t}(x)
&=K_{h_t}(x-X_a)\{Y_a-m_{L,t}^h(x)\},\label{sg:eq:RaL}\\
R_{a,F,t}(x)
&=K_{h_t}(x-X_a)\{Y_a-m_{F,t}^h(x)\}
-\mu_{j(a),F,t}(x),\label{sg:eq:RaF}
\end{align}
where
\[
\mu_{j,F,t}(x)
=
\bm{E}_{P_{j,t}}
\left[K_{h_t}(x-X)\{Y-m_{F,t}^h(x)\}\right].
\]
For local observations, the corresponding local centering is zero by the
definition of $m_{L,t}^h$.  Define the function-valued first-order terms
\begin{align}
\phi_{a,L,t}(x)
&=r_{a,L,t}\frac{R_{a,L,t}(x)}{p_{L,t}^h(x)},\label{sg:eq:phiL}\\
\phi_{a,F,t}(x)
&=r_{a,F,t}\frac{R_{a,F,t}(x)}{B_t^h(x)}.\label{sg:eq:phiF}
\end{align}
They are centered under the law of observation $a$.  Define the scalar
influence summand
\begin{equation}
\lambda_{a,t}
=
\inner{c_{L,t}}{\phi_{a,L,t}}_t
+
\inner{c_{F,t}}{\phi_{a,F,t}}_t.
\label{sg:eq:lambda}
\end{equation}
For shared observations from client $k_t$, both terms in \eqref{sg:eq:lambda} are
present, so their covariance is retained automatically.

Let
\begin{equation}
s_{\mathrm{fit},t}^2
=
\sum_{a\in\mathcal{I}_t}\bm{E}_t^{\mathrm{fit}}\lambda_{a,t}^2.
\label{sg:eq:sfit}
\end{equation}

\begin{lemma}[Integrated ratio linearization]
\label{sg:lem:ratio-linearization}
Under Assumptions~\ref{sg:ass:kernel}--\ref{sg:ass:rates},
\begin{align}
\inner{c_{L,t}}{L_t-\overline L_t}_t
&=
\sum_{a\in\mathcal{I}_t}\inner{c_{L,t}}{\phi_{a,L,t}}_t
+R_{L,t}^{\mathrm{rat}},\label{sg:eq:local-linearization}\\
\inner{c_{F,t}}{F_t-\overline F_t}_t
&=
\sum_{a\in\mathcal{I}_t}\inner{c_{F,t}}{\phi_{a,F,t}}_t
+R_{F,t}^{\mathrm{rat}},\label{sg:eq:fed-linearization}
\end{align}
where
\begin{equation}
R_{L,t}^{\mathrm{rat}}
=o_{\bm{P}}(\sqrt{\mathfrak r_t}),
\qquad
R_{F,t}^{\mathrm{rat}}
=o_{\bm{P}}(\sqrt{\mathfrak r_t}),
\qquad
\bm{E}_t^{\mathrm{fit}}|R_{L,t}^{\mathrm{rat}}|
+
\bm{E}_t^{\mathrm{fit}}|R_{F,t}^{\mathrm{rat}}|
=
O\!\left(\frac{\mathfrak r_t}{h_t}\right).
\label{sg:eq:ratio-remainder-first-mean}
\end{equation}
The same conclusions hold when $c_{L,t}$ and $c_{F,t}$ are replaced by any
deterministic coefficient functions $q_{L,t}$ and $q_{F,t}$ whose sup norms
are uniformly bounded.
\end{lemma}

\begin{proof}
We prove the federated assertion; the local assertion is the same argument
with one client law.  Let $\widehat B_t$ be the raw empirical denominator,
let
\[
D_t=\widehat B_t\vee(c_0/2),
\qquad
\widetilde F_t=\frac{\widehat A_t}{D_t},
\]
and write $\delta B_t=\widehat B_t-B_t^h$.  With
\[
S_t=(\widehat A_t-A_t^h)-m_{F,t}^h(\widehat B_t-B_t^h),
\]
the globally valid stabilized-ratio identity is
\begin{equation}
\widetilde F_t-m_{F,t}^h
=
\frac{S_t}{B_t^h}
-
\frac{S_t(D_t-B_t^h)}{B_t^hD_t}
-
\frac{m_{F,t}^h(D_t-\widehat B_t)}{D_t}.
\label{sg:eq:stabilized-ratio-identity}
\end{equation}
The first term is the sum of the centered functions in \eqref{sg:eq:phiF}.
Since $B_t^h\ge c_0$ and $D_t\ge c_0/2$,
\[
|D_t-B_t^h|\le|\delta B_t|.
\]
Moreover, $D_t-\widehat B_t=0$ off the floor event, while on that event
$|\delta B_t|\ge c_0/2$ and therefore
\[
0\le D_t-\widehat B_t
\le |\delta B_t|
\le \frac{2}{c_0}|\delta B_t|^2.
\]
Thus, uniformly in $x$,
\begin{equation}
\left|
\widetilde F_t-m_{F,t}^h-\frac{S_t}{B_t^h}
\right|
\le C|S_t|\,|\delta B_t|+C|\delta B_t|^2.
\label{sg:eq:stabilized-ratio-global-bound}
\end{equation}
Assumptions~\ref{sg:ass:kernel}--\ref{sg:ass:moments} give
\[
\bm{E}_t^{\mathrm{fit}}\norm{\delta B_t}_t^2
+
\bm{E}_t^{\mathrm{fit}}\norm{S_t}_t^2
\le C\frac{\mathfrak r_t}{h_t}.
\]
Hence, for any deterministic bounded coefficient $q_{F,t}$,
Cauchy--Schwarz and \eqref{sg:eq:stabilized-ratio-global-bound} yield
\begin{equation}
\bm{E}_t^{\mathrm{fit}}
\left|
\inner{q_{F,t}}{
\widetilde F_t-m_{F,t}^h-S_t/B_t^h}_t
\right|
\le C\norm{q_{F,t}}_\infty\frac{\mathfrak r_t}{h_t}.
\label{sg:eq:stabilized-ratio-first-mean}
\end{equation}
This bound includes the denominator-floor event and is therefore valid before
centering by the fitting expectation.

It remains to account for clipping.  Let $\Pi$ be the projection onto the
fixed clipping interval, so that $F_t=\Pi(\widetilde F_t)$.  Because
$m_{F,t}^h$ lies at least $c_{\mathrm{clip}}$ inside that interval,
\[
|F_t(x)-\widetilde F_t(x)|
\le
|\widetilde F_t(x)-m_{F,t}^h(x)|
\mathbf{1}\{|\widetilde F_t(x)-m_{F,t}^h(x)|\ge c_{\mathrm{clip}}\}
\le
c_{\mathrm{clip}}^{-1}
|\widetilde F_t(x)-m_{F,t}^h(x)|^2.
\]
The integrated second-moment bound therefore gives an additional absolute
first-mean remainder of order $O(\mathfrak r_t/h_t)$.  After subtracting
the fitting expectation, the full stabilized-and-clipped ratio remainder is
$O_{\bm{P}}(\mathfrak r_t/h_t)$ and has absolute first mean of the same order.
Since $\sqrt{\mathfrak r_t}/h_t\to0$, the remainder is
$o_{\bm{P}}(\sqrt{\mathfrak r_t})$.  This proves the federated conclusion and
\eqref{sg:eq:ratio-remainder-first-mean}; the local argument is identical.
\end{proof}

\begin{lemma}[Integrated second-moment bounds for the fitted ratios]
\label{sg:lem:integrated-second-moments}
Under Assumptions~\ref{sg:ass:kernel}--\ref{sg:ass:rates},
\begin{equation}
\bm{E}_t^{\mathrm{fit}}\norm{L_t-\overline L_t}_t^2
+
\bm{E}_t^{\mathrm{fit}}\norm{F_t-\overline F_t}_t^2
\le
C\frac{\mathfrak r_t}{h_t}.
\label{sg:eq:integrated-second-moments}
\end{equation}
\end{lemma}

\begin{proof}
We give the federated argument.  Write the empirical numerator and denominator
as $A_t^h+\delta A_t$ and $B_t^h+\delta B_t$, respectively, and let
$m_{F,t}^h=A_t^h/B_t^h$.  The stabilized denominator is bounded below by
$c_0/2$, while $B_t^h\ge c_0$.  Since the clipping map is nonexpansive and
$m_{F,t}^h$ is uniformly bounded,
\[
|F_t(x)-m_{F,t}^h(x)|
\le
C\{|\delta A_t(x)|+|\delta B_t(x)|\}
\]
for all $x\in\mathcal X$.  Standard kernel second-moment calculations give
\[
\bm{E}_t^{\mathrm{fit}}\norm{\delta A_t}_t^2
+
\bm{E}_t^{\mathrm{fit}}\norm{\delta B_t}_t^2
\le C\frac{\mathfrak r_t}{h_t}.
\]
Hence
$\bm{E}_t^{\mathrm{fit}}\norm{F_t-m_{F,t}^h}_t^2
\le C\mathfrak r_t/h_t$.  Because the mean minimizes squared distance,
\[
\bm{E}_t^{\mathrm{fit}}\norm{F_t-\overline F_t}_t^2
\le
\bm{E}_t^{\mathrm{fit}}\norm{F_t-m_{F,t}^h}_t^2.
\]
The local assertion follows from the same calculation with the selected
client's numerator and denominator.
\end{proof}

\begin{lemma}[Negligibility of the centered quadratic gain]
\label{sg:lem:quadratic-negligible}
Let
\[
e_{L,t}=L_t-\overline L_t,
\qquad
e_{F,t}=F_t-\overline F_t,
\]
and
\[
\mathcal Q_t
=
\norm{e_{L,t}}_t^2
-
\norm{\alpha e_{L,t}+\vartheta e_{F,t}}_t^2.
\]
Under Assumptions~\ref{sg:ass:kernel}--\ref{sg:ass:rates} and
\ref{sg:ass:nondegenerate},
\begin{equation}
\frac{
\mathcal Q_t-\bm{E}_t^{\mathrm{fit}}\mathcal Q_t
}{s_{\mathrm{fit},t}}
\xrightarrow{\bm{P}}0.
\label{sg:eq:quadratic-negligible}
\end{equation}
\end{lemma}

\begin{proof}
There is a constant $C$ depending only on $\alpha$ such that
\[
|\mathcal Q_t|
\le
C\{\norm{e_{L,t}}_t^2+\norm{e_{F,t}}_t^2\}.
\]
Lemma~\ref{sg:lem:integrated-second-moments} therefore gives
\[
\bm{E}_t^{\mathrm{fit}}|\mathcal Q_t|
\le C\frac{\mathfrak r_t}{h_t},
\qquad
\mathcal Q_t=O_{\bm{P}}\!\left(\frac{\mathfrak r_t}{h_t}\right).
\]
Consequently,
\[
\mathcal Q_t-\bm{E}_t^{\mathrm{fit}}\mathcal Q_t
=
O_{\bm{P}}\!\left(\frac{\mathfrak r_t}{h_t}\right).
\]
Assumption~\ref{sg:ass:nondegenerate} gives
$s_{\mathrm{fit},t}\ge c\sqrt{\mathfrak r_t}$, and hence
\[
\frac{
\mathcal Q_t-\bm{E}_t^{\mathrm{fit}}\mathcal Q_t
}{s_{\mathrm{fit},t}}
=
O_{\bm{P}}\!\left(\frac{\sqrt{\mathfrak r_t}}{h_t}\right)
\xrightarrow{\bm{P}}0
\]
by Assumption~\ref{sg:ass:rates}.
\end{proof}

\begin{theorem}[Fitting-sample gain CLT]
\label{sg:thm:fitting-clt}
Under Assumptions~\ref{sg:ass:kernel}--\ref{sg:ass:nondegenerate},
\begin{equation}
\frac{G_t^{\mathrm{fit}}-\Gamma_t}{s_{\mathrm{fit},t}}
\Longrightarrow N(0,1).
\label{sg:eq:fitting-clt}
\end{equation}
\end{theorem}

\begin{proof}
The exact centered gain expansion is
\[
G_t^{\mathrm{fit}}-\Gamma_t
=
\inner{c_{L,t}}{e_{L,t}}_t
+
\inner{c_{F,t}}{e_{F,t}}_t
+
\{\mathcal Q_t-\bm{E}_t^{\mathrm{fit}}\mathcal Q_t\}.
\]
Lemmas~\ref{sg:lem:ratio-linearization}--\ref{sg:lem:quadratic-negligible} give
\[
G_t^{\mathrm{fit}}-\Gamma_t
=
\sum_{a\in\mathcal{I}_t}\lambda_{a,t}
+o_{\bm{P}}(s_{\mathrm{fit},t}).
\]
The summands are independent and centered.  Kernel integration and bounded
coefficient functions give, for some constant $C$,
\[
\bm{E}_t^{\mathrm{fit}}|\lambda_{a,t}|^{2+\delta}
\le C\gamma_{a,t}^{2+\delta}.
\]
Hence
\[
\frac{
\sum_a\bm{E}_t^{\mathrm{fit}}|\lambda_{a,t}|^{2+\delta}
}{s_{\mathrm{fit},t}^{2+\delta}}
\le
C\left(
\frac{\gamma_{\max,t}}{\sqrt{\mathfrak r_t}}
\right)^\delta
\longrightarrow0.
\]
The Lyapunov central limit theorem gives
$\sum_a\lambda_{a,t}/s_{\mathrm{fit},t}\Longrightarrow N(0,1)$, and Slutsky's
theorem completes the proof.
\end{proof}

\subsection{A consistent estimator of the fitting variance}

The theoretical influence summands in \eqref{sg:eq:lambda} use the mean fitted
curves $\overline L_t$ and $\overline F_t$.  These repeated-fitting means are
not observed directly.  We therefore first replace them by the estimable
finite-bandwidth ratio targets.  Put
\[
\widetilde b_{L,t}=m_{L,t}^h-m_t^Q,
\qquad
\widetilde b_{F,t}=m_{F,t}^h-m_t^Q,
\]
and define
\begin{align}
\widetilde c_{L,t}
&=2\vartheta\{(1+\alpha)\widetilde b_{L,t}
-\alpha\widetilde b_{F,t}\},\\
\widetilde c_{F,t}
&=-2\vartheta\{\alpha\widetilde b_{L,t}
+\vartheta\widetilde b_{F,t}\}.
\end{align}
The corresponding proxy influence is
\begin{equation}
\widetilde\lambda_{a,t}
=
\inner{\widetilde c_{L,t}}{\phi_{a,L,t}}_t
+
\inner{\widetilde c_{F,t}}{\phi_{a,F,t}}_t.
\label{sg:eq:lambda-proxy}
\end{equation}

\begin{lemma}[Finite-bandwidth proxy equivalence]
\label{sg:lem:proxy-equivalence}
Under Assumptions~\ref{sg:ass:kernel}--\ref{sg:ass:rates} and
\ref{sg:ass:nondegenerate},
\begin{equation}
\frac{
\sum_{a\in\mathcal{I}_t}
\bm{E}_t^{\mathrm{fit}}
(\widetilde\lambda_{a,t}-\lambda_{a,t})^2
}{s_{\mathrm{fit},t}^2}
\longrightarrow0.
\label{sg:eq:proxy-equivalence}
\end{equation}
\end{lemma}

\begin{proof}
By Jensen's inequality and the proof of
Lemma~\ref{sg:lem:integrated-second-moments},
\[
\norm{\overline L_t-m_{L,t}^h}_t^2
+
\norm{\overline F_t-m_{F,t}^h}_t^2
\le
C\frac{\mathfrak r_t}{h_t}
\longrightarrow0.
\]
Hence
\[
\norm{\widetilde c_{L,t}-c_{L,t}}_t
+
\norm{\widetilde c_{F,t}-c_{F,t}}_t
\longrightarrow0.
\]
The single-kernel integration bound gives, for deterministic coefficient
functions $g_{L,t},g_{F,t}$,
\[
\sum_a\bm{E}_t^{\mathrm{fit}}
\left|
\inner{g_{L,t}}{\phi_{a,L,t}}_t
+
\inner{g_{F,t}}{\phi_{a,F,t}}_t
\right|^2
\le
C\mathfrak r_t
(\norm{g_{L,t}}_t^2+\norm{g_{F,t}}_t^2).
\]
Apply this with the two coefficient differences and use
$s_{\mathrm{fit},t}^2\ge c\mathfrak r_t$ to obtain
\eqref{sg:eq:proxy-equivalence}.
\end{proof}

Let $\widehat\lambda_{a,t}$ be \eqref{sg:eq:lambda-proxy} evaluated with
leave-one-out or finite-fold cross-fitted estimates of the finite-bandwidth
denominator and ratio targets, the evaluation regression $m_t^Q$, and the
evaluation integral.  Define
\begin{equation}
\widehat s_{\mathrm{fit},t}^2
=
\sum_{a\in\mathcal{I}_t}\widehat\lambda_{a,t}^2.
\label{sg:eq:fit-var-estimator}
\end{equation}

\begin{assumption}[Cross-fitted nuisance stability]
\label{sg:ass:nuisance}
There is a deterministic sequence $a_t\downarrow0$.  For every observation
$a$, let $\widehat\eta_{a,t}$ denote the leave-out or foldwise nuisance
estimate used in $\widehat\lambda_{a,t}$, constructed without observation
$a$.  There is an event
$\mathcal H_{a,t}\in\sigma(\widehat\eta_{a,t})$ on which all estimated
denominators are bounded below and the nuisance estimate differs from the
finite-bandwidth targets in \eqref{sg:eq:lambda-proxy} by at most $a_t$ in
the required sup and integration norms.  The simultaneous event
\[
\mathcal H_t=\bigcap_{a\in\mathcal{I}_t}\mathcal H_{a,t}
\]
satisfies $\bm{P}(\mathcal H_t)\to1$.  An independent pilot sample may be
used to estimate $m_t^Q$ and the $Q_t$-integrals; cross-fitting is used only
for fitting-side nuisances and does not replace the independent audit sample.
\end{assumption}

\begin{lemma}[Localization of cross-fitted plug-in errors]
\label{sg:lem:localization}
Let $\Delta_{q,t}$ be observationwise or pairwise plug-in errors, and let
$\mathcal H_{q,t}\in\sigma(\widehat\eta_{q,t})$ be nuisance-specific good
events whose intersection has probability tending to one.  If, on
$\mathcal H_{q,t}$,
\[
\bm{E}_t^{\mathrm{fit}}
\left(\Delta_{q,t}^2\mid\widehat\eta_{q,t}\right)
\le C\kappa_{q,t}^2a_t^2,
\]
then
\begin{equation}
\sum_q\Delta_{q,t}^2
=O_{\bm{P}}\left(a_t^2\sum_q\kappa_{q,t}^2\right).
\label{sg:eq:localized-plugin-generic}
\end{equation}
No moment bound is required on the complement of the simultaneous good event.
\end{lemma}

\begin{proof}
For $M>0$, split the probability of the event in
\eqref{sg:eq:localized-plugin-generic} according to the simultaneous good
event.  On that event, its indicator is bounded by every nuisance-specific
good-event indicator.  Termwise conditioning, the tower property, and
Markov's inequality give an upper bound $C/M$ for the localized probability;
the complement probability tends to zero.  The argument does not condition
on the complete collection of overlapping leave-out nuisances.
\end{proof}

\begin{lemma}[Aggregate plug-in bound]
\label{sg:lem:plugin-bound}
Under Assumptions~\ref{sg:ass:kernel}--\ref{sg:ass:nondegenerate} and \ref{sg:ass:nuisance},
\begin{equation}
\sum_{a\in\mathcal{I}_t}
(\widehat\lambda_{a,t}-\widetilde\lambda_{a,t})^2
=
O_{\bm{P}}(\mathfrak r_t a_t^2).
\label{sg:eq:plugin-bound}
\end{equation}
Consequently,
\begin{equation}
\frac{
\sum_a(\widehat\lambda_{a,t}-\lambda_{a,t})^2
}{s_{\mathrm{fit},t}^2}
\xrightarrow{\bm{P}}0.
\label{sg:eq:total-plugin-bound}
\end{equation}
\end{lemma}

\begin{proof}
For each $a$, condition only on $\widehat\eta_{a,t}$.  It is independent of
observation $a$.  On $\mathcal H_{a,t}$, the denominator floors and
Lipschitz dependence of the integrated influence give
\[
\bm{E}_t^{\mathrm{fit}}
\left[
(\widehat\lambda_{a,t}-\widetilde\lambda_{a,t})^2
\mid \widehat\eta_{a,t}
\right]
\le C\gamma_{a,t}^2a_t^2.
\]
Lemma~\ref{sg:lem:localization}, with
$\kappa_{a,t}=\gamma_{a,t}$, proves \eqref{sg:eq:plugin-bound}.  Combine it
with Lemma~\ref{sg:lem:proxy-equivalence}, the triangle inequality in
$\ell^2(\mathcal{I}_t)$, and $s_{\mathrm{fit},t}^2\asymp\mathfrak r_t$ to obtain
\eqref{sg:eq:total-plugin-bound}.
\end{proof}

\begin{theorem}[Fitting-variance consistency]
\label{sg:thm:fit-var-consistency}
Under Assumptions~\ref{sg:ass:kernel}--\ref{sg:ass:nondegenerate} and \ref{sg:ass:nuisance},
\begin{equation}
\frac{\widehat s_{\mathrm{fit},t}^2}{s_{\mathrm{fit},t}^2}
\xrightarrow{\bm{P}}1.
\label{sg:eq:fit-var-consistency}
\end{equation}
\end{theorem}

\begin{proof}
First define the oracle empirical estimator
\[
\widetilde s_{\mathrm{fit},t}^2
=
\sum_a\lambda_{a,t}^2.
\]
It is unbiased for $s_{\mathrm{fit},t}^2$.  Moreover,
\[
\Var_t^{\mathrm{fit}}(\widetilde s_{\mathrm{fit},t}^2)
\le
\sum_a\bm{E}_t^{\mathrm{fit}}\lambda_{a,t}^4
\le
C\sum_a\gamma_{a,t}^4.
\]
Assumptions~\ref{sg:ass:rates} and \ref{sg:ass:nondegenerate} imply
\[
\frac{\sum_a\gamma_{a,t}^4}{s_{\mathrm{fit},t}^4}
\le
C\frac{\gamma_{\max,t}^2}{\mathfrak r_t}
\longrightarrow0.
\]
Thus $\widetilde s_{\mathrm{fit},t}^2/s_{\mathrm{fit},t}^2\xrightarrow{\bm{P}}1$.

By Lemma~\ref{sg:lem:plugin-bound},
\[
\frac{\sum_a(\widehat\lambda_{a,t}-\lambda_{a,t})^2}
{s_{\mathrm{fit},t}^2}
\xrightarrow{\bm{P}}0.
\]
Cauchy--Schwarz gives
\[
\left|
\sum_a\widehat\lambda_{a,t}^2-
\sum_a\lambda_{a,t}^2
\right|
\le
2\left(\sum_a\lambda_{a,t}^2\right)^{1/2}
\left\{\sum_a(\widehat\lambda_{a,t}-\lambda_{a,t})^2\right\}^{1/2}
+
\sum_a(\widehat\lambda_{a,t}-\lambda_{a,t})^2.
\]
After division by $s_{\mathrm{fit},t}^2$, the right side converges to zero in
probability.  Combining the oracle and plug-in results proves
\eqref{sg:eq:fit-var-consistency}.
\end{proof}

\begin{remark}[Scope of feasible nominal studentization]
Assumption~\ref{sg:ass:nuisance} is an aggregate cross-fitting condition around
estimable finite-bandwidth targets.  The preceding theorem is therefore a
statistically feasible high-level variance result.  It is not, without an
additional model-specific nuisance-rate lemma, a primitive consequence of the
basic numerator--denominator summaries and bandwidth conditions.
\end{remark}

\subsection{Conditional evaluation inference}

For fixed $t$, collect the fitting observations into a random element $S_t$ in
a standard Borel space and set $\mathcal{G}_t^0=\sigma(S_t)$.  Write the paired gain as
a jointly measurable map $w_t(s;z)$.  Define
\begin{align}
\mu_t(s)&=\int w_t(s;z)Q_t(\,d z),\label{sg:eq:eval-mu}\\
\sigma_{E,t}^2(s)&=\int\{w_t(s;z)-\mu_t(s)\}^2Q_t(\,d z),\label{sg:eq:eval-sigma}\\
\beta_t(s)&=
\frac{\int|w_t(s;z)-\mu_t(s)|^3Q_t(\,d z)}{\sigma_{E,t}^3(s)}.
\label{sg:eq:beta}
\end{align}
For $m\ge1$, define the regular conditional probability kernel
\[
\Pi_{t,m}(s,B)
=
Q_t^{\otimes m}
\left\{
\frac{\sum_{r=1}^m\{w_t(s;Z_r)-\mu_t(s)\}}
{\sqrt m\,\sigma_{E,t}(s)}
\in B
\right\}.
\]
It is a version of the conditional law of the standardized evaluation error
given $\mathcal{G}_t^0$.

\begin{theorem}[Conditional Berry--Esseen and Sweeting convergence]
\label{sg:thm:conditional-BE}
For every fixed $t$, every $m\ge1$, and every fitting realization satisfying
$0<\sigma_{E,t}^2(s)<\infty$ and $\beta_t(s)<\infty$,
\begin{equation}
d_K\{\Pi_{t,m}(s,\cdot),\Phi\}
\le
C_{\mathrm{BE}}\frac{\beta_t(s)}{\sqrt m}.
\label{sg:eq:conditional-BE}
\end{equation}
Consequently, for the diagonal $m=N_t^{\mathrm{eval}}$, if
$N_t^{\mathrm{eval}}\to\infty$ and $\beta_t(S_t)$ is tight, then
\[
\frac{\beta_t(S_t)}{\sqrt{N_t^{\mathrm{eval}}}}\xrightarrow{\bm{P}}0,
\]
and the random conditional laws converge in probability to $N(0,1)$ in the
sense of \citet{Sweeting1989}.  Moreover, for every deterministic or
$\mathcal{F}_t^{\mathrm{fit}}$-measurable threshold $\varepsilon_t$,
\begin{align}
\bigg|&
\bm{P}\{\widehat G_t>\varepsilon_t\mid\mathcal{F}_t^{\mathrm{fit}}\}
-
\Phi\!\left(
\frac{\sqrt{N_t^{\mathrm{eval}}}\{G_t^{\mathrm{fit}}-\varepsilon_t\}}
{\sigma_{E,t}(S_t)}
\right)
\bigg|
\nonumber\\
&\qquad\le
C_{\mathrm{BE}}\frac{\beta_t(S_t)}{\sqrt{N_t^{\mathrm{eval}}}}
\label{sg:eq:conditional-positive}
\end{align}
with probability tending to one.
\end{theorem}

\begin{proof}
For fixed $s$, the standardized paired gains are i.i.d., centered, have unit
variance, and have third absolute moment $\beta_t(s)$.  The classical
Berry--Esseen theorem \citep{Petrov1975} under the deterministic product measure
$Q_t^{\otimes m}$ gives \eqref{sg:eq:conditional-BE}.  The conditional law is a
measurable random probability measure because it was constructed as the
kernel $\Pi_{t,m}$.  The diagonal conclusion follows from the probability mode
of conditional weak convergence in \citet{Sweeting1989}.  The final bound is
obtained by evaluating the uniform Kolmogorov bound at the random standardized
threshold.
\end{proof}

The conditional studentized audit theorem also gives the realized-fit lower
confidence bound
\begin{equation}
\operatorname{LCB}^{\mathrm{fit}}_t(\delta)
=
\widehat G_t-z_{1-\delta}
\frac{\widehat\sigma_{E,t}}{\sqrt{N_t^{\mathrm{eval}}}},
\label{sg:eq:fitted-gain-lcb}
\end{equation}
whose coverage is conditional on the realized fitted predictors.  It targets
$G_t^{\mathrm{fit}}$, whereas
$\operatorname{LCB}^{\mathrm{proc}}_t$ in
Theorem~\ref{sg:thm:standalone} targets $\Gamma_t$.

Define the deterministic oracle evaluation variance
\begin{equation}
v_{E,t}
=
\Var_{Q_t}\{W_t^0(Z)\},
\label{sg:eq:vE}
\end{equation}
where $W_t^0$ is the paired gain obtained by replacing $(L_t,F_t)$ with
$(\overline L_t,\overline F_t)$.

\begin{lemma}[Evaluation-variance stabilization]
\label{sg:lem:eval-stabilization}
Under Assumptions~\ref{sg:ass:kernel}--\ref{sg:ass:eval} and \ref{sg:ass:nuisance},
\begin{equation}
\frac{\sigma_{E,t}^2(S_t)}{v_{E,t}}\xrightarrow{\bm{P}}1.
\label{sg:eq:eval-stabilization}
\end{equation}
\end{lemma}

\begin{proof}
The gain map is locally Lipschitz in $(L,F)$ under clipping:
\[
|W_t(L,F;Z)-W_t(L',F';Z)|
\le
C(1+|Y|)
\{|L-L'|+|F-F'|\}.
\]
Assumptions~\ref{sg:ass:kernel}--\ref{sg:ass:rates} imply
$\norm{L_t-\overline L_t}_{L^4(Q_{t,X})}+
\norm{F_t-\overline F_t}_{L^4(Q_{t,X})}\xrightarrow{\bm{P}}0$.
The response moment bound and Cauchy--Schwarz therefore yield
\[
\bm{E}_{Q_t}\{|W_t-W_t^0|^2\mid\mathcal{F}_t^{\mathrm{fit}}\}\xrightarrow{\bm{P}}0.
\]
The difference between the corresponding first and second conditional moments
converges to zero in probability.  Since $v_{E,t}$ is bounded below, the ratio
conclusion follows.
\end{proof}

Let $\widehat\sigma_{E,t}^2$ denote the usual sample variance of the
observation-level evaluation gains.

\begin{lemma}[Conditional sample-variance consistency]
\label{sg:lem:eval-sample-variance}
Under Assumption~\ref{sg:ass:eval}, if $N_t^{\mathrm{eval}}\to\infty$, then the sample
variance of the observation-level evaluation gains satisfies
\begin{equation}
\frac{\widehat\sigma_{E,t}^2}{\sigma_{E,t}^2(S_t)}\xrightarrow{\bm{P}}1,
\qquad
\frac{\widehat\sigma_{E,t}^2}{v_{E,t}}\xrightarrow{\bm{P}}1.
\label{sg:eq:eval-sample-variance}
\end{equation}
\end{lemma}

\begin{proof}
Condition on $\mathcal{F}_t^{\mathrm{fit}}$ and put
$X_{t,r}=\{W_t(Z_{t,r}^{\mathrm{eval}})-G_t^{\mathrm{fit}}\}/
\sigma_{E,t}(S_t)$.  Then the $X_{t,r}$ are conditionally iid with mean zero,
variance one, and conditional third absolute moment $\beta_t(S_t)$.  With
$Y_{t,r}=X_{t,r}^2-1$, the inequality
$|x^2-1|^{3/2}\le C(1+|x|^3)$ gives
\[
\bm{E}(|Y_{t,r}|^{3/2}\mid\mathcal{F}_t^{\mathrm{fit}})
\le C\{1+\beta_t(S_t)\}.
\]
The conditional von Bahr--Esseen inequality therefore yields
\[
\bm{E}\left[
\left.
\left|
\frac1{N_t^{\mathrm{eval}}}\sum_{r=1}^{N_t^{\mathrm{eval}}}Y_{t,r}
\right|^{3/2}
\right|\mathcal{F}_t^{\mathrm{fit}}
\right]
\le
\frac{C\{1+\beta_t(S_t)\}}{\sqrt{N_t^{\mathrm{eval}}}}.
\]
Because $\beta_t(S_t)$ is tight, the centered empirical second moment converges
to one in probability.  Conditional Chebyshev gives
$(N_t^{\mathrm{eval}})^{-1}\sum_rX_{t,r}\xrightarrow{\bm{P}}0$, and hence its square is negligible.  The
usual identity between the sample variance, the empirical second moment, and
the squared sample mean proves the first ratio in
\eqref{sg:eq:eval-sample-variance}.  The second follows from
Lemma~\ref{sg:lem:eval-stabilization}.
\end{proof}

Let
\[
s_{\mathrm{eval},t}^2=\frac{v_{E,t}}{N_t^{\mathrm{eval}}},
\qquad
s_t^2=s_{\mathrm{fit},t}^2+s_{\mathrm{eval},t}^2.
\]
Define
\[
\widehat s_t^2
=
\widehat s_{\mathrm{fit},t}^2
+
\frac{\widehat\sigma_{E,t}^2}{N_t^{\mathrm{eval}}}.
\]

\subsection{Standalone theorem}

\begin{theorem}[Predictive-gain inference in the nondegenerate first-order regime]
\label{sg:thm:standalone}
Under Assumptions~\ref{sg:ass:kernel}--\ref{sg:ass:eval} and \ref{sg:ass:nuisance}, suppose $N_t^{\mathrm{eval}}\to\infty$.
Then:
\begin{enumerate}[label=(\roman*)]
\item
\[
\left(
\frac{G_t^{\mathrm{fit}}-\Gamma_t}{s_{\mathrm{fit},t}},
\frac{\widehat G_t-G_t^{\mathrm{fit}}}{s_{\mathrm{eval},t}}
\right)
\Longrightarrow
(Z_1,Z_2),
\]
where $Z_1,Z_2$ are independent standard normal variables.

\item
\begin{equation}
\frac{\widehat G_t-\Gamma_t}{s_t}
\Longrightarrow N(0,1).
\label{sg:eq:combined-clt}
\end{equation}

\item
\begin{equation}
\frac{\widehat s_t^2}{s_t^2}\xrightarrow{\bm{P}}1,
\qquad
\frac{\widehat G_t-\Gamma_t}{\widehat s_t}
\Longrightarrow N(0,1).
\label{sg:eq:studentized-clt}
\end{equation}

\item For every deterministic threshold $\varepsilon_t$,
\begin{equation}
\bm{P}\{\widehat G_t>\varepsilon_t\}
=
\Phi\!\left(
\frac{\Gamma_t-\varepsilon_t}{s_t}
\right)+o(1).
\label{sg:eq:positive-probability}
\end{equation}

\item For $\delta\in(0,1)$,
\[
\operatorname{LCB}^{\mathrm{proc}}_t(\delta)
=
\widehat G_t-z_{1-\delta}\widehat s_t
\]
satisfies
\begin{equation}
\bm{P}\{\Gamma_t\ge\operatorname{LCB}_t(\delta)\}
\longrightarrow1-\delta.
\label{sg:eq:coverage}
\end{equation}
\end{enumerate}
\end{theorem}

\begin{proof}
Theorem~\ref{sg:thm:fitting-clt} gives the first coordinate in part (i).
Theorem~\ref{sg:thm:conditional-BE}, Lemma~\ref{sg:lem:eval-stabilization}, and
Sweeting's probability mode give the conditional convergence of the second
coordinate to $N(0,1)$.  Since the first coordinate is
$\mathcal{F}_t^{\mathrm{fit}}$-measurable, for $u,v\in\mathbb{R}$,
\[
\begin{aligned}
&\bm{E}\exp\left\{
iu\frac{G_t^{\mathrm{fit}}-\Gamma_t}{s_{\mathrm{fit},t}}
+iv\frac{\widehat G_t-G_t^{\mathrm{fit}}}{s_{\mathrm{eval},t}}
\right\}\\
&\quad=
\bm{E}\left[
\exp\left\{
iu\frac{G_t^{\mathrm{fit}}-\Gamma_t}{s_{\mathrm{fit},t}}
\right\}
\bm{E}\left[
\left.
\exp\left\{
iv\frac{\widehat G_t-G_t^{\mathrm{fit}}}{s_{\mathrm{eval},t}}
\right\}
\right|\mathcal{F}_t^{\mathrm{fit}}
\right]
\right].
\end{aligned}
\]
The inner conditional characteristic function converges in probability to
$e^{-v^2/2}$ and is bounded by one, so the convergence also holds in $L^1$.
The outer characteristic function converges to $e^{-u^2/2}$.  This proves
part (i), including asymptotic independence.

For part (ii), write the normalized total error as a weighted sum of the two
coordinates in part (i), with weights
$s_{\mathrm{fit},t}/s_t$ and $s_{\mathrm{eval},t}/s_t$.  Every subsequence has
a further subsequence along which these weights converge; the limiting
weighted sum is standard normal because the squared weights sum to one.  This
proves \eqref{sg:eq:combined-clt} without requiring a limit of the variance
fraction.

The fitting-variance estimator is consistent by
Theorem~\ref{sg:thm:fit-var-consistency}.  Lemma~\ref{sg:lem:eval-sample-variance}
gives $\widehat\sigma_{E,t}^2/v_{E,t}\xrightarrow{\bm{P}}1$.  Therefore
$\widehat s_t^2/s_t^2\xrightarrow{\bm{P}}1$, and part (iii) follows from Slutsky's theorem.
Parts (iv) and (v) follow from the continuity of the standard normal
distribution and the studentized limit.
\end{proof}

\begin{corollary}[Probability of positive marginal gain]
\label{sg:cor:positive}
If $\Gamma_t/s_t\to\gamma\in\mathbb{R}$, then
\[
\bm{P}\{\widehat G_t>0\}\longrightarrow\Phi(\gamma).
\]
If $\Gamma_t/s_t\to+\infty$, the probability tends to one; if it tends to
$-\infty$, the probability tends to zero.
\end{corollary}

The theorem is formulated for the moving target $\Gamma_t$.  If
$\Gamma_t\to\Gamma$, replacement by $\Gamma$ inside a standardized margin
is valid only when
\begin{equation}
\frac{\Gamma_t-\Gamma}{s_t}\longrightarrow0.
\label{sg:eq:moving-target-replacement}
\end{equation}
Ordinary convergence is insufficient near the zero-gain boundary.

\subsection{Exact-model quadratic boundary}
\label{sg:subsec:quadratic-boundary}

We now treat the boundary excluded from the nondegenerate first-order
theorem \citep{HardleMammen1993}.  Assume
\begin{equation}
\overline L_t=\overline F_t=m_t^Q
\qquad Q_{t,X}\text{-a.e.}
\label{sg:eq:exact-model-boundary}
\end{equation}
so that $b_{L,t}=b_{F,t}=0$ and both first-order gain coefficients vanish.
Then
\begin{equation}
G_t^{\mathrm{fit}}
=
\norm{e_{L,t}}_t^2
-
\norm{\alpha e_{L,t}+\vartheta e_{F,t}}_t^2
=\mathcal Q_t.
\label{sg:eq:boundary-gain-quadratic}
\end{equation}

For pairs $\bm u=(u_L,u_F)$ and $\bm v=(v_L,v_F)$ in
$L^2(Q_{t,X})^2$, define the symmetric bilinear form
\begin{equation}
\mathcal B_{\alpha,t}(\bm u,\bm v)
=
\inner{u_L}{v_L}_t
-
\inner{\alpha u_L+\vartheta u_F}
{\alpha v_L+\vartheta v_F}_t.
\label{sg:eq:boundary-bilinear-form}
\end{equation}
Put
\[
\bm\phi_{a,t}=(\phi_{a,L,t},\phi_{a,F,t}),
\qquad
\bm S_t=\sum_{a\in\mathcal{I}_t}\bm\phi_{a,t}.
\]
For $a<b$, define the canonical quadratic kernel
\begin{equation}
u_{ab,t}
=
2\mathcal B_{\alpha,t}(\bm\phi_{a,t},\bm\phi_{b,t}),
\qquad
U_t=\sum_{a<b}u_{ab,t},
\label{sg:eq:boundary-off-diagonal}
\end{equation}
and the centered diagonal term
\begin{equation}
D_t
=
\sum_{a\in\mathcal{I}_t}
\left[
\mathcal B_{\alpha,t}(\bm\phi_{a,t},\bm\phi_{a,t})
-
\bm{E}_t^{\mathrm{fit}}
\mathcal B_{\alpha,t}(\bm\phi_{a,t},\bm\phi_{a,t})
\right].
\label{sg:eq:boundary-diagonal}
\end{equation}
Because each $\bm\phi_{a,t}$ is centered and the fitting observations are
independent, $u_{ab,t}$ is canonical in each argument and
\begin{equation}
s_{Q,t}^{2}
:=\Var_t^{\mathrm{fit}}(U_t)
=
\sum_{a<b}\bm{E}_t^{\mathrm{fit}}u_{ab,t}^{2}.
\label{sg:eq:boundary-fitting-variance}
\end{equation}

\begin{assumption}[High-level quadratic-boundary Nadaraya--Watson array]
\label{sg:ass:quadratic-boundary}
In addition to \eqref{sg:eq:exact-model-boundary}, the fitted ratio errors
admit
\begin{equation}
e_{L,t}=S_{L,t}+r_{L,t}^{\mathrm{rat}},
\qquad
e_{F,t}=S_{F,t}+r_{F,t}^{\mathrm{rat}},
\label{sg:eq:boundary-ratio-expansion}
\end{equation}
where
\begin{align}
\norm{r_{L,t}^{\mathrm{rat}}}_t
+\norm{r_{F,t}^{\mathrm{rat}}}_t
&=O_{\bm{P}}(\mathfrak r_t/h_t),
\label{sg:eq:boundary-ratio-rate}\\
\bm{E}_t^{\mathrm{fit}}
\{\norm{r_{L,t}^{\mathrm{rat}}}_t^2
+\norm{r_{F,t}^{\mathrm{rat}}}_t^2\}
&=O\!\left((\mathfrak r_t/h_t)^2\right).
\label{sg:eq:boundary-ratio-second-mean}
\end{align}
There are constants $0<c<C<\infty$ such that
\begin{equation}
c\frac{\mathfrak r_t^2}{h_t}
\le s_{Q,t}^{2}
\le C\frac{\mathfrak r_t^2}{h_t}.
\label{sg:eq:boundary-variance-order}
\end{equation}
The centered diagonal term is negligible:
\begin{equation}
\frac{D_t}{s_{Q,t}}\xrightarrow{\bm{P}}0.
\label{sg:eq:boundary-diagonal-negligible}
\end{equation}
Finally, after fixing any deterministic ordering of $\mathcal{I}_t$, let
\[
\mathcal H_{a,t}
=\sigma(\bm\phi_{b,t}:b\le a),
\qquad
M_{a,t}=\sum_{b<a}u_{ba,t}.
\]
The quadratic martingale satisfies, for every $\epsilon>0$,
\begin{align}
\frac1{s_{Q,t}^{2}}
\sum_a
\bm{E}_t^{\mathrm{fit}}(M_{a,t}^{2}\mid\mathcal H_{a-1,t})
&\xrightarrow{\bm{P}}1,
\label{sg:eq:boundary-pqv}\\
\frac1{s_{Q,t}^{2}}
\sum_a
\bm{E}_t^{\mathrm{fit}}
\{M_{a,t}^{2}\mathbf{1}(|M_{a,t}|>\epsilon s_{Q,t})
\mid\mathcal H_{a-1,t}\}
&\xrightarrow{\bm{P}}0.
\label{sg:eq:boundary-lindeberg}
\end{align}
\end{assumption}

\begin{remark}[Scope and sufficient conditions]
Assumption~\ref{sg:ass:quadratic-boundary} is a high-level Hall--de Jong input \citep{Hall1984,deJong1987},
not a primitive theorem derived here from the basic kernel conditions.  The
ratio remainder rates follow from the exact stabilized ratio identity and
fourth-moment kernel bounds.  A convenient sufficient condition for
\eqref{sg:eq:boundary-diagonal-negligible} is
\[
\frac{\gamma_{\max,t}^{2}}{\mathfrak r_t h_t}\longrightarrow0.
\]
Conditions \eqref{sg:eq:boundary-pqv}--\eqref{sg:eq:boundary-lindeberg} are
the Hall--de Jong conditions for the clean generalized quadratic form.  They
follow, for example, if the predictable quadratic variation has variance
$o(s_{Q,t}^{4})$ and
$\sum_a\bm{E}_t^{\mathrm{fit}}M_{a,t}^{4}=o(s_{Q,t}^{4})$.
For sample-size weights these conditions reduce to the usual no-dominant-
observation and kernel-overlap requirements.
\end{remark}

\begin{lemma}[A checkable Hall--de Jong criterion]
\label{sg:lem:boundary-checkable-clt}
Suppose
\begin{align}
\frac{\Var_t^{\mathrm{fit}}\!\left\{
\sum_a\bm{E}_t^{\mathrm{fit}}(M_{a,t}^2\mid\mathcal H_{a-1,t})
\right\}}{s_{Q,t}^{4}}
&\longrightarrow0,
\label{sg:eq:boundary-pqv-variance}\\
\frac{\sum_a\bm{E}_t^{\mathrm{fit}}M_{a,t}^{4}}{s_{Q,t}^{4}}
&\longrightarrow0.
\label{sg:eq:boundary-martingale-fourth}
\end{align}
Then \eqref{sg:eq:boundary-pqv}--\eqref{sg:eq:boundary-lindeberg} hold.
\end{lemma}

\begin{proof}
The expectation of the predictable quadratic variation equals
$\Var_t^{\mathrm{fit}}(U_t)=s_{Q,t}^{2}$, so
\eqref{sg:eq:boundary-pqv} follows from
\eqref{sg:eq:boundary-pqv-variance} and Chebyshev's inequality.  For every
$\epsilon>0$,
\[
\sum_a\bm{E}_t^{\mathrm{fit}}
\{M_{a,t}^{2}\mathbf{1}(|M_{a,t}|>\epsilon s_{Q,t})
\mid\mathcal H_{a-1,t}\}
\le
\frac1{\epsilon^2s_{Q,t}^{2}}
\sum_a\bm{E}_t^{\mathrm{fit}}
(M_{a,t}^{4}\mid\mathcal H_{a-1,t}).
\]
After division by $s_{Q,t}^{2}$, Markov's inequality and
\eqref{sg:eq:boundary-martingale-fourth} prove
\eqref{sg:eq:boundary-lindeberg}.
\end{proof}

\begin{lemma}[Quadratic approximation of the boundary gain]
\label{sg:lem:boundary-quadratic-approximation}
Under Assumptions~\ref{sg:ass:kernel}--\ref{sg:ass:rates} and
\ref{sg:ass:quadratic-boundary},
\begin{equation}
G_t^{\mathrm{fit}}-\Gamma_t
=
D_t+U_t+r_{Q,t},
\qquad
\frac{r_{Q,t}}{s_{Q,t}}\xrightarrow{\bm{P}}0,
\label{sg:eq:boundary-quadratic-approximation}
\end{equation}
and the same normalized remainder converges to zero in first mean.
\end{lemma}

\begin{proof}
By bilinearity,
\[
|\mathcal B_{\alpha,t}(\bm S_t+\bm r_t,
\bm S_t+\bm r_t)
-\mathcal B_{\alpha,t}(\bm S_t,\bm S_t)|
\le
C\norm{\bm S_t}_t\norm{\bm r_t}_t
+C\norm{\bm r_t}_t^2,
\]
where $\bm r_t=(r_{L,t}^{\mathrm{rat}},r_{F,t}^{\mathrm{rat}})$ and the
product norm is used.  Lemma~\ref{sg:lem:integrated-second-moments} gives
$\norm{\bm S_t}_t=O_{\bm{P}}(\sqrt{\mathfrak r_t/h_t})$, while
\eqref{sg:eq:boundary-ratio-rate} gives
$\norm{\bm r_t}_t=O_{\bm{P}}(\mathfrak r_t/h_t)$.  Hence the displayed
difference is
\[
O_{\bm{P}}
\left(\frac{\mathfrak r_t^{3/2}}{h_t^{3/2}}
+\frac{\mathfrak r_t^2}{h_t^2}\right)
=o_{\bm{P}}(\mathfrak r_t/\sqrt{h_t})
=o_{\bm{P}}(s_{Q,t})
\]
by $\sqrt{\mathfrak r_t}/h_t\to0$ and
$\mathfrak r_t/h_t\to0$.  The first-mean statement follows from the
corresponding first-mean ratio bounds and Cauchy--Schwarz.  Expanding
$\mathcal B_{\alpha,t}(\bm S_t,\bm S_t)$ into diagonal and off-diagonal
terms gives \eqref{sg:eq:boundary-quadratic-approximation}.
\end{proof}

\begin{theorem}[Exact-model boundary fitting CLT]
\label{sg:thm:boundary-fitting-clt}
Under Assumptions~\ref{sg:ass:kernel}--\ref{sg:ass:rates} and
\ref{sg:ass:quadratic-boundary},
\begin{equation}
\frac{G_t^{\mathrm{fit}}-\Gamma_t}{s_{Q,t}}
\Longrightarrow N(0,1).
\label{sg:eq:boundary-fitting-clt}
\end{equation}
\end{theorem}

\begin{proof}
The variables $M_{a,t}$ form a martingale-difference array and
$U_t=\sum_aM_{a,t}$.  Conditions
\eqref{sg:eq:boundary-pqv}--\eqref{sg:eq:boundary-lindeberg} and the
martingale central limit theorem \citep{HallHeyde1980} give $U_t/s_{Q,t}\Longrightarrow N(0,1)$.
Lemma~\ref{sg:lem:boundary-quadratic-approximation} and
\eqref{sg:eq:boundary-diagonal-negligible} complete the proof.
\end{proof}

For feasible studentization, let $\widehat u_{ab,t}$ be
\eqref{sg:eq:boundary-off-diagonal} evaluated with leave-two-out or foldwise
cross-fitted estimates of all ratio and evaluation-law nuisances, and define
\begin{equation}
\widehat s_{Q,t}^{2}
=
\sum_{a<b}\widehat u_{ab,t}^{2}.
\label{sg:eq:boundary-variance-estimator}
\end{equation}

\begin{assumption}[Quadratic-boundary variance estimation]
\label{sg:ass:boundary-variance-estimation}
With
\begin{align}
\mathcal V_{Q,t}
&=
\sum_{a<b}\bm{E}_t^{\mathrm{fit}}u_{ab,t}^{4}
+
\sum_a
\left[
\sum_{b\ne a}
\{\bm{E}_t^{\mathrm{fit}}u_{a\wedge b,a\vee b,t}^{4}\}^{1/2}
\right]^2,
\label{sg:eq:boundary-fourth-envelope}
\end{align}
assume $\mathcal V_{Q,t}/s_{Q,t}^{4}\to0$.  The pairwise nuisance estimate
used in $\widehat u_{ab,t}$ is constructed without observations $a$ and $b$.
On events whose probabilities tend to one, its denominators are bounded below
and it differs from the finite-bandwidth nuisance target by at most a
deterministic $a_t\downarrow0$ in the sup and integration norms entering the
bilinear kernel.
\end{assumption}

\begin{lemma}[Pairwise quadratic plug-in bound]
\label{sg:lem:boundary-pair-plugin}
Under Assumptions~\ref{sg:ass:kernel}--\ref{sg:ass:rates},
\ref{sg:ass:quadratic-boundary}, and
\ref{sg:ass:boundary-variance-estimation},
\begin{equation}
\sum_{a<b}(\widehat u_{ab,t}-u_{ab,t})^2
=
O_{\bm{P}}\left(\frac{\mathfrak r_t^2}{h_t}a_t^2\right),
\qquad
\frac{
\sum_{a<b}(\widehat u_{ab,t}-u_{ab,t})^2
}{s_{Q,t}^{2}}
\xrightarrow{\bm{P}}0.
\label{sg:eq:boundary-pair-plugin}
\end{equation}
\end{lemma}

\begin{proof}
Condition on the leave-two-out nuisance estimate used for pair $(a,b)$.  The
kernel-overlap bound and Lipschitz dependence on the nuisances give
\[
\bm{E}_t^{\mathrm{fit}}
\left[
(\widehat u_{ab,t}-u_{ab,t})^2
\mid\widehat\eta_t^{(-a,-b)}
\right]
\le
C\frac{\gamma_{a,t}^2\gamma_{b,t}^2}{h_t}a_t^2.
\]
Summing over pairs and using
$\sum_{a<b}\gamma_{a,t}^2\gamma_{b,t}^2
\le\mathfrak r_t^2/2$ gives the first assertion by Markov's inequality.
The variance order \eqref{sg:eq:boundary-variance-order} gives the second.
\end{proof}

\begin{proposition}[Boundary fitting-variance consistency]
\label{sg:prop:boundary-variance-consistency}
Under Assumptions~\ref{sg:ass:kernel}--\ref{sg:ass:rates},
\ref{sg:ass:quadratic-boundary}, and
\ref{sg:ass:boundary-variance-estimation},
\begin{equation}
\frac{\widehat s_{Q,t}^{2}}{s_{Q,t}^{2}}\xrightarrow{\bm{P}}1.
\label{sg:eq:boundary-variance-consistency}
\end{equation}
\end{proposition}

\begin{proof}
The oracle estimator $\sum_{a<b}u_{ab,t}^{2}$ is unbiased for $s_{Q,t}^{2}$.
Pairs with disjoint index sets are independent, and Cauchy--Schwarz controls
pairs sharing one index.  Therefore its variance is bounded by a universal
constant times $\mathcal V_{Q,t}$, which gives oracle ratio consistency.  The
plug-in conclusion follows from \eqref{sg:eq:boundary-pair-plugin} and the
same $\ell^2$ Cauchy--Schwarz argument used in
Theorem~\ref{sg:thm:fit-var-consistency}.
\end{proof}

\begin{assumption}[Boundary evaluation stabilization]
\label{sg:ass:boundary-evaluation}
Let
\begin{equation}
v_{\partial E,t}
=
4\vartheta^2
\bm{E}_t^{\mathrm{fit}}
\int
\{F_t(x)-L_t(x)\}^{2}\sigma_{Q,t}^{2}(x)
\,Q_{t,X}(\,d x),
\label{sg:eq:boundary-evaluation-oracle}
\end{equation}
where $\sigma_{Q,t}^{2}(x)=\Var_{Q_t}(Y\mid X=x)$.  There are constants
$0<c<C<\infty$ such that
\begin{equation}
c\frac{\mathfrak r_t}{h_t}
\le v_{\partial E,t}
\le C\frac{\mathfrak r_t}{h_t},
\qquad
\frac{\sigma_{E,t}^{2}(S_t)}{v_{\partial E,t}}\xrightarrow{\bm{P}}1.
\label{sg:eq:boundary-evaluation-stabilization}
\end{equation}
Moreover, $N_t^{\mathrm{eval}}\to\infty$ and
$\beta_t(S_t)/\sqrt{N_t^{\mathrm{eval}}}\xrightarrow{\bm{P}}0$.
\end{assumption}

\begin{remark}
At the exact-model boundary the mean-fit gain is identically zero, so the
persistent-regime oracle variance $\Var_{Q_t}\{W_t^0(Z)\}$ is also zero.
The leading conditional evaluation variance instead comes from the noise term
$2\vartheta\{F_t(X)-L_t(X)\}\varepsilon$.  Assumption
\ref{sg:ass:boundary-evaluation} stabilizes this random variance on its
natural order $\mathfrak r_t/h_t$.
\end{remark}

Set
\[
s_{\partial E,t}^{2}=\frac{v_{\partial E,t}}{N_t^{\mathrm{eval}}},
\qquad
s_{\mathrm{bd},t}^{2}=s_{Q,t}^{2}+s_{\partial E,t}^{2},
\qquad
\widehat s_{\mathrm{bd},t}^{2}
=\widehat s_{Q,t}^{2}+\widehat\sigma_{E,t}^{2}/N_t^{\mathrm{eval}}.
\]

\begin{theorem}[Combined exact-model boundary gain inference]
\label{sg:thm:boundary-combined}
Under Theorem~\ref{sg:thm:boundary-fitting-clt},
Assumptions~\ref{sg:ass:boundary-variance-estimation} and
\ref{sg:ass:boundary-evaluation},
\begin{equation}
\frac{\widehat G_t-\Gamma_t}{s_{\mathrm{bd},t}}
\Longrightarrow N(0,1),
\qquad
\frac{\widehat s_{\mathrm{bd},t}^{2}}{s_{\mathrm{bd},t}^{2}}
\xrightarrow{\bm{P}}1,
\label{sg:eq:boundary-combined}
\end{equation}
and hence
\begin{equation}
\frac{\widehat G_t-\Gamma_t}{\widehat s_{\mathrm{bd},t}}
\Longrightarrow N(0,1).
\label{sg:eq:boundary-studentized}
\end{equation}
The normalized fitting and evaluation terms are asymptotically independent.
\end{theorem}

\begin{proof}
The fitting term is $\mathcal{F}_t^{\mathrm{fit}}$-measurable and satisfies
Theorem~\ref{sg:thm:boundary-fitting-clt}.  The pathwise Berry--Esseen bound,
Assumption~\ref{sg:ass:boundary-evaluation}, and
$\sigma_{E,t}^2(S_t)/v_{\partial E,t}\xrightarrow{\bm{P}}1$ give the conditional Sweeting
limit for the evaluation term on the scale $s_{\partial E,t}$.  Conditioning
in the joint characteristic function proves asymptotic independence.  The
subsequence argument for the deterministic variance fractions gives the first
limit.  The conditional von Bahr--Esseen proof of
Lemma~\ref{sg:lem:eval-sample-variance} applies verbatim with
$\beta_t(S_t)/\sqrt{N_t^{\mathrm{eval}}}\xrightarrow{\bm{P}}0$, yielding
$\widehat\sigma_{E,t}^2/v_{\partial E,t}\xrightarrow{\bm{P}}1$.  Together with
Proposition~\ref{sg:prop:boundary-variance-consistency}, this proves
studentization.
\end{proof}

For a fixed client block $1,\ldots,r$, let
\[
\bm u^{(r)}_{ab,t}
=(u_{1,ab,t},\ldots,u_{r,ab,t})^{\mathsf{T}},
\qquad
\bm M^{(r)}_{a,t}=\sum_{b<a}\bm u^{(r)}_{ba,t},
\]
and define the quadratic fitting covariance
\begin{equation}
\Sigma^{\mathrm{bd},\mathrm{fit}}_{r,t}
=
\sum_{a<b}\bm{E}_t^{\mathrm{fit}}
\left[
\bm u^{(r)}_{ab,t}\bm u^{(r)\mathsf{T}}_{ab,t}
\right].
\label{sg:eq:boundary-block-fit-covariance}
\end{equation}
For client $j\le r$, let
$s_{\partial E,j,t}^{2}=v_{\partial E,j,t}/N_{j,t}^{\mathrm{eval}}$ be the
client-specific analogue of \eqref{sg:eq:boundary-evaluation-oracle}, and put
\[
s_{\mathrm{bd},j,t}^{2}
=\Sigma^{\mathrm{bd},\mathrm{fit}}_{jj,t}
+s_{\partial E,j,t}^{2},
\qquad
D^{\mathrm{bd}}_{r,t}
=\operatorname{diag}(s_{\mathrm{bd},1,t},\ldots,s_{\mathrm{bd},r,t}).
\]
Define
\[
A^{\mathrm{bd}}_{r,t}
=(D^{\mathrm{bd}}_{r,t})^{-1}
\Sigma^{\mathrm{bd},\mathrm{fit}}_{r,t}
(D^{\mathrm{bd}}_{r,t})^{-1},
\]
and let $B^{\mathrm{bd}}_{r,t}$ be the diagonal matrix with entries
$s_{\partial E,j,t}^{2}/s_{\mathrm{bd},j,t}^{2}$.

\begin{assumption}[Fixed-block quadratic-boundary process]
\label{sg:ass:boundary-block}
For every fixed $r$:
\begin{enumerate}[label=(\roman*)]
\item the vector quadratic approximation and diagonal-negligibility conditions
hold coordinatewise, with the remainder negligible after multiplication by
$(D^{\mathrm{bd}}_{r,t})^{-1}$;
\item $A^{\mathrm{bd}}_{r,t}\to A^{\mathrm{bd}}_r$ and
$B^{\mathrm{bd}}_{r,t}\to B^{\mathrm{bd}}_r$ for nonnegative definite
matrices, with $B^{\mathrm{bd}}_r$ diagonal;
\item for every $\bm q\in\mathbb{R}^r$, the scalar martingale increments
$\bm q^{\mathsf{T}}(D^{\mathrm{bd}}_{r,t})^{-1}\bm M^{(r)}_{a,t}$ satisfy the
predictable quadratic-variation and Lindeberg conditions with limiting
variance $\bm q^{\mathsf{T}}A^{\mathrm{bd}}_r\bm q$;
\item the client evaluation samples are independent across clients and the
boundary evaluation-stabilization and Berry--Esseen conditions hold for every
coordinate;
\item with
\[
\widehat\Sigma^{\mathrm{bd},\mathrm{fit}}_{r,t}
=
\sum_{a<b}
\widehat{\bm u}^{(r)}_{ab,t}
\widehat{\bm u}^{(r)\mathsf{T}}_{ab,t},
\]
the cross-fitted block covariance estimator and the diagonal evaluation
variance estimators satisfy
\begin{equation}
\left\|
(D^{\mathrm{bd}}_{r,t})^{-1}
(\widehat\Sigma^{\mathrm{bd},\mathrm{fit}}_{r,t}
-\Sigma^{\mathrm{bd},\mathrm{fit}}_{r,t})
(D^{\mathrm{bd}}_{r,t})^{-1}
\right\|\xrightarrow{\bm{P}}0,
\label{sg:eq:boundary-block-covariance-consistency}
\end{equation}
and $\widehat s_{\mathrm{bd},j,t}^{2}/s_{\mathrm{bd},j,t}^{2}\xrightarrow{\bm{P}}1$.
Writing
\[
\widehat D^{\mathrm{bd}}_{r,t}
=\operatorname{diag}(\widehat s_{\mathrm{bd},1,t},\ldots,
       \widehat s_{\mathrm{bd},r,t}),
\]
and
\[
\widehat R^{\mathrm{bd}}_{r,t}
=(\widehat D^{\mathrm{bd}}_{r,t})^{-1}
\left\{
\widehat\Sigma^{\mathrm{bd},\mathrm{fit}}_{r,t}
+\operatorname{diag}(\widehat\sigma_{E,1,t}^2/N_{1,t}^{\mathrm{eval}},\ldots,
       \widehat\sigma_{E,r,t}^2/N_{r,t}^{\mathrm{eval}})
\right\}
(\widehat D^{\mathrm{bd}}_{r,t})^{-1},
\]
assume the displayed evaluation variance estimators are the coordinatewise
ones entering the preceding ratio consistency.
\end{enumerate}
\end{assumption}

\begin{theorem}[Fixed-block quadratic-boundary process]
\label{sg:thm:boundary-fixed-block}
Under Assumption~\ref{sg:ass:boundary-block},
\begin{equation}
(D^{\mathrm{bd}}_{r,t})^{-1}
(\widehat{\boldsymbol G}_{r,t}-\boldsymbol\Gamma_{r,t})
\Longrightarrow
N_r\{0,R^{\mathrm{bd}}_r\},
\qquad
R^{\mathrm{bd}}_r=A^{\mathrm{bd}}_r+B^{\mathrm{bd}}_r.
\label{sg:eq:boundary-fixed-block-clt}
\end{equation}
The same conclusion holds with the consistent coordinatewise scales
$\widehat D^{\mathrm{bd}}_{r,t}$, and
$\widehat R^{\mathrm{bd}}_{r,t}\xrightarrow{\bm{P}} R^{\mathrm{bd}}_r$.  If the fixed-block limits are compatible, the standardized
sequence converges in $\mathbb{R}^{\mathbb N}$ with the product topology.  Fixed-block
orthant limits follow, while the moving all-client event requires the same
separate coordinatewise summability condition as in
Corollary~\ref{gp:cor:all-client}.
\end{theorem}

\begin{proof}
Apply the martingale central limit theorem to every fixed scalar contrast and
use Cram\'er--Wold for the fitting block.  The conditional product law of the
independent evaluation samples gives the diagonal evaluation limit.
Conditioning in the joint characteristic function yields asymptotic
independence.  Assumption~\ref{sg:ass:boundary-block}(v) gives multivariate
Slutsky and covariance consistency.  Product-topology tightness and the
orthant and summability conclusions follow exactly as in the persistent
client-process theorem.
\end{proof}

\begin{remark}[Communication scope]
The block estimators above are theoretical cross-fitted estimators based on
observation-pair kernels.  The paper does not claim that they can be
reconstructed from the basic numerator--denominator server summaries without
additional client computation or communication.
\end{remark}
\subsection{Interpretation}

The theorem targets the deterministic moving stage-$t$ marginal gain
$\Gamma_t$.  It permits all client laws, sample sizes, weights, bandwidths,
and the selected client index $k_t$ to vary with $t$ and does not require
$\Gamma_t$ to converge.  A limiting value $\Gamma$ may replace $\Gamma_t$
inside a standardized margin only under the additional rate condition
$(\Gamma_t-\Gamma)/s_t\to0$ in \eqref{sg:eq:moving-target-replacement}.

The fitting variance retains the dependence between the selected client's
local fit and the common server fit through the shared-observation influence
summands.  The evaluation variance is conditional on the fitted models and is
estimated from the independent second-stage sample.  Asymptotic independence
of the two normalized fluctuations is proved from conditional convergence; it
is not assumed.

\section{A reusable two-stage client-array assembly theorem}
\label{app:two-stage-assembly}

The nominal, KL-certificate, and $L_2$ applications use the same probabilistic
assembly after their distinct scalar fitting expansions have been established.
The following theorem records that common step.

\begin{theorem}[Two-stage client-array assembly]
\label{gen:thm:two-stage-assembly}
For each stage $t$ and fixed block size $r$, suppose
\[
\widehat{\bm T}_{r,t}-\bm\theta_{r,t}
=\bm H_{r,t}+\bm E_{r,t},
\]
where $\bm H_{r,t}$ is $\mathcal{F}_t^{\mathrm{fit}}$-measurable.  Let
$D_{r,t}$ be a deterministic diagonal matrix with positive entries.  Assume:
\begin{enumerate}[label=(\roman*)]
\item $D_{r,t}^{-1}\bm H_{r,t}\Longrightarrow N_r(0,A_r)$;
\item conditionally on $\mathcal{F}_t^{\mathrm{fit}}$,
$D_{r,t}^{-1}\bm E_{r,t}$ converges in Sweeting's probability mode to
$N_r(0,B_r)$, where $B_r$ is deterministic;
\item the conditional characteristic functions in (ii) are bounded by one;
\item stage-$t$ covariance estimators satisfy
\[
\left\|D_{r,t}^{-1}\widehat\Sigma_{r,t}D_{r,t}^{-1}
-R_{r,t}\right\|\xrightarrow{\bm{P}}0,
\qquad R_{r,t}\to R_r:=A_r+B_r.
\]
\end{enumerate}
Then
\begin{equation}
D_{r,t}^{-1}(\widehat{\bm T}_{r,t}-\bm\theta_{r,t})
\Longrightarrow N_r(0,R_r),
\label{gen:eq:assembled-clt}
\end{equation}
the fitting and evaluation Gaussian limits are independent, and the
studentized version obtained from the diagonal of $\widehat\Sigma_{r,t}$ has
the same limit.  Compatible fixed-block limits define a Gaussian sequence in
$\mathbb{R}^{\mathbb N}$ with the product topology.  Fixed-block orthant limits
follow from \eqref{gen:eq:assembled-clt}; a moving all-client event requires a
separate coordinatewise summability bound.
\end{theorem}

\begin{proof}
Condition on $\mathcal{F}_t^{\mathrm{fit}}$ in the joint characteristic function.
The conditional factor converges in probability to the deterministic Gaussian
factor and, being bounded by one, also converges in first mean.  Multiplying by
the fitting characteristic function proves asymptotic independence and
\eqref{gen:eq:assembled-clt}.  Covariance approximation and multivariate
Slutsky give studentization.  Coordinatewise tightness and compatibility yield
tightness and uniqueness in the countable product topology.
\end{proof}

\section{Client-indexed nominal gain process: detailed assumptions and proofs}
\label{app:client-process-proof}

\subsection{Client-indexed setup}

For each stage $t$, let $K_t$ clients participate, with $K_t\to\infty$.
Client $j\le K_t$ has a fitted local predictor $L_{j,t}$, and all clients share
the fitted federated predictor $F_t$.  All fitting observations generate the
sigma-field
\[
\mathcal{F}_t^{\mathrm{fit}}.
\]
The client gain coordinates need not be independent: a client's observations
may enter both its local and server predictors, and the same fitted server
predictor enters every client coordinate.

Fix a prescribed local weight $\alpha\in[0,1)$ and put
\[
\vartheta=1-\alpha.
\]
Let $Q_{j,t}$ be the evaluation law for client $j$, and define the personalized
predictor
\[
M_{j,t}(x)=\alpha L_{j,t}(x)+\vartheta F_t(x).
\]
For $z=(x,y)$, the paired gain is
\begin{equation}
W_{j,t}(z)
=
\{y-L_{j,t}(x)\}^{2}-\{y-M_{j,t}(x)\}^{2}.
\label{gp:eq:paired-gain}
\end{equation}
The gain of the fitted predictors is
\begin{equation}
G_{j,t}^{\mathrm{fit}}
=
\int W_{j,t}(z)Q_{j,t}(\,d z)
=
\bm{E}\{W_{j,t}(Z)\mid\mathcal{F}_t^{\mathrm{fit}}\},
\label{gp:eq:fitted-gain}
\end{equation}
and the deterministic stage-$t$ marginal gain is
\begin{equation}
\Gamma_{j,t}
=
\bm{E}_t^{\mathrm{fit}}G_{j,t}^{\mathrm{fit}},
\label{gp:eq:marginal-gain}
\end{equation}
where $\bm{E}_t^{\mathrm{fit}}$ denotes expectation under the joint law of all
stage-$t$ fitting samples.

Client $j$ has a fresh evaluation sample
\[
Z_{j,t,1}^{\mathrm{eval}},\ldots,Z_{j,t,N_{j,t}^{\mathrm{eval}}}^{\mathrm{eval}}
\overset{\mathrm{iid}}{\sim}Q_{j,t}.
\]
The evaluation samples are independent across clients and independent of
$\mathcal{F}_t^{\mathrm{fit}}$.  Define
\begin{equation}
\widehat G_{j,t}
=
\frac1{N_{j,t}^{\mathrm{eval}}}
\sum_{r=1}^{N_{j,t}^{\mathrm{eval}}}
W_{j,t}(Z_{j,t,r}^{\mathrm{eval}}).
\label{gp:eq:empirical-gain}
\end{equation}
The exact two-stage decomposition is
\begin{equation}
\widehat G_{j,t}-\Gamma_{j,t}
=
H_{j,t}+E_{j,t},
\qquad
H_{j,t}=G_{j,t}^{\mathrm{fit}}-\Gamma_{j,t},
\qquad
E_{j,t}=\widehat G_{j,t}-G_{j,t}^{\mathrm{fit}}.
\label{gp:eq:two-stage}
\end{equation}
Here $H_{j,t}$ is $\mathcal{F}_t^{\mathrm{fit}}$-measurable and
\[
\bm{E}(E_{j,t}\mid\mathcal{F}_t^{\mathrm{fit}})=0.
\]
Consequently,
\begin{equation}
\Cov(H_{j,t},E_{\ell,t})=0
\qquad
\text{for all }j,\ell\le K_t.
\label{gp:eq:cross-orthogonality}
\end{equation}
This is orthogonality, not unconditional independence.

\subsection{Finite-dimensional fitting limit}

Fix $r\ge1$.  For all sufficiently large $t$, $r\le K_t$.  Write
\[
\bm H_{r,t}=(H_{1,t},\ldots,H_{r,t})^{\mathsf{T}}.
\]
The one-client gain theorem in the nondegenerate first-order regime gives,
for each fixed client coordinate, a scalar influence representation.  The
joint version is stated below.

\begin{assumption}[Finite-block fitting representation]
\label{gp:ass:fit-representation}
For every fixed $r$, there are independent, centered random vectors
\[
\bm\lambda_{a,t}^{(r)}
=(\lambda_{1,a,t},\ldots,\lambda_{r,a,t})^{\mathsf{T}},
\qquad a\in\mathcal{I}_t,
\]
and a remainder vector $\bm r_{r,t}^{\mathrm{fit}}$ such that
\begin{equation}
\bm H_{r,t}
=
\sum_{a\in\mathcal{I}_t}\bm\lambda_{a,t}^{(r)}
+
\bm r_{r,t}^{\mathrm{fit}}.
\label{gp:eq:fit-vector-representation}
\end{equation}
The component $\lambda_{j,a,t}$ is the scalar influence of fitting observation
$a$ on client $j$'s gain.  It contains the local contribution, when observation
$a$ enters client $j$'s local fit, and the common server contribution, when it
enters the federated fit.
\end{assumption}

Let
\begin{equation}
\Sigma_{r,t}^{\mathrm{fit}}
=
\sum_{a\in\mathcal{I}_t}
\bm{E}_t^{\mathrm{fit}}
\left[
\bm\lambda_{a,t}^{(r)}
\bm\lambda_{a,t}^{(r)\mathsf{T}}
\right].
\label{gp:eq:fit-covariance}
\end{equation}
The off-diagonal entries of \eqref{gp:eq:fit-covariance} retain the dependence
created by the common server estimator.

For each client, let $s_{\mathrm{fit},j,t}^{2}$ be the fitting variance from
the one-client theorem.  Define the conditional variance of one evaluation
gain by
\begin{equation}
\sigma_{E,j,t}^{2}
=
\Var\{W_{j,t}(Z_{j,t,1}^{\mathrm{eval}})\mid\mathcal{F}_t^{\mathrm{fit}}\}.
\label{gp:eq:conditional-eval-variance}
\end{equation}
Suppose there is a deterministic $v_{E,j,t}>0$ satisfying
\begin{equation}
\frac{\sigma_{E,j,t}^{2}}{v_{E,j,t}}\xrightarrow{\bm{P}}1,
\qquad
s_{\mathrm{eval},j,t}^{2}=\frac{v_{E,j,t}}{N_{j,t}^{\mathrm{eval}}}.
\label{gp:eq:eval-stabilization}
\end{equation}
Set
\begin{equation}
s_{j,t}^{2}
=
s_{\mathrm{fit},j,t}^{2}+s_{\mathrm{eval},j,t}^{2},
\qquad
D_{r,t}=\operatorname{diag}(s_{1,t},\ldots,s_{r,t}).
\label{gp:eq:total-scales}
\end{equation}

Define the normalized fitting covariance
\begin{equation}
A_{r,t}
=
D_{r,t}^{-1}
\Sigma_{r,t}^{\mathrm{fit}}
D_{r,t}^{-1}.
\label{gp:eq:normalized-fit-covariance}
\end{equation}

\begin{assumption}[Finite-block covariance and Lyapunov conditions]
\label{gp:ass:fit-vector-clt}
For every fixed $r$, there is a symmetric nonnegative definite matrix $A_r$
such that
\begin{equation}
A_{r,t}\longrightarrow A_r.
\label{gp:eq:A-convergence}
\end{equation}
For some $\delta>0$,
\begin{equation}
\sum_{a\in\mathcal{I}_t}
\bm{E}_t^{\mathrm{fit}}
\left\|
D_{r,t}^{-1}\bm\lambda_{a,t}^{(r)}
\right\|^{2+\delta}
\longrightarrow0,
\label{gp:eq:vector-lyapunov}
\end{equation}
and
\begin{equation}
\left\|
D_{r,t}^{-1}\bm r_{r,t}^{\mathrm{fit}}
\right\|
\xrightarrow{\bm{P}}0.
\label{gp:eq:vector-remainder}
\end{equation}
\end{assumption}

\begin{remark}
For fixed $r$, condition \eqref{gp:eq:vector-lyapunov} follows from the
coordinatewise Lyapunov conditions
\[
\sum_{a\in\mathcal{I}_t}
\bm{E}_t^{\mathrm{fit}}
\left|
\frac{\lambda_{j,a,t}}{s_{j,t}}
\right|^{2+\delta}
\longrightarrow0,
\qquad j=1,\ldots,r,
\]
because
\[
\norm{\bm x}^{2+\delta}
\le
r^{\delta/2}
\sum_{j=1}^{r}|x_j|^{2+\delta}.
\]
The primitive one-client Nadaraya--Watson assumptions imply these conditions
for every fixed block in the nondegenerate first-order regime.
\end{remark}

\begin{theorem}[Finite-block fitting CLT]
\label{gp:thm:finite-fit-clt}
Under Assumptions~\ref{gp:ass:fit-representation} and
\ref{gp:ass:fit-vector-clt},
\begin{equation}
D_{r,t}^{-1}\bm H_{r,t}
\Longrightarrow
N_r(\bm0,A_r).
\label{gp:eq:finite-fit-clt}
\end{equation}
\end{theorem}

\begin{proof}
Fix $\bm u\in\mathbb{R}^r$.  By \eqref{gp:eq:fit-vector-representation},
\[
\bm u^{\mathsf{T}}D_{r,t}^{-1}\bm H_{r,t}
=
\sum_{a\in\mathcal{I}_t}
\bm u^{\mathsf{T}}D_{r,t}^{-1}\bm\lambda_{a,t}^{(r)}
+
\bm u^{\mathsf{T}}D_{r,t}^{-1}\bm r_{r,t}^{\mathrm{fit}}.
\]
The scalar summands are independent and centered.  Their variance converges to
$\bm u^{\mathsf{T}}A_r\bm u$ by \eqref{gp:eq:A-convergence}.  The Lyapunov condition
follows from \eqref{gp:eq:vector-lyapunov}, and the remainder vanishes by
\eqref{gp:eq:vector-remainder}.  The scalar Lyapunov theorem and Slutsky's theorem
give
\[
\bm u^{\mathsf{T}}D_{r,t}^{-1}\bm H_{r,t}
\Longrightarrow
N(0,\bm u^{\mathsf{T}}A_r\bm u).
\]
Cram\'er--Wold proves \eqref{gp:eq:finite-fit-clt}.
\end{proof}

\subsection{Finite-dimensional conditional evaluation limit}

Let
\[
\bm E_{r,t}=(E_{1,t},\ldots,E_{r,t})^{\mathsf{T}}.
\]
Because the evaluation streams are independent across clients, their regular
conditional distribution given $\mathcal{F}_t^{\mathrm{fit}}$ is the product of the
client-specific conditional probability kernels.

Define
\begin{equation}
B_{r,t}
=
\operatorname{diag}\left(
\frac{s_{\mathrm{eval},1,t}^{2}}{s_{1,t}^{2}},\ldots,
\frac{s_{\mathrm{eval},r,t}^{2}}{s_{r,t}^{2}}
\right).
\label{gp:eq:Brt}
\end{equation}

\begin{assumption}[Evaluation conditional convergence]
\label{gp:ass:eval-vector}
For every fixed $r$, $B_{r,t}\to B_r$ for a diagonal nonnegative definite
matrix $B_r$.  For each $j\le r$, the regular conditional law satisfies
\begin{equation}
\mathcal{L}\left(
\left.
\frac{E_{j,t}}{s_{\mathrm{eval},j,t}}
\right|
\mathcal{F}_t^{\mathrm{fit}}
\right)
\overset{\bm{P}}{\Longrightarrow}
N(0,1),
\label{gp:eq:coordinate-eval-sweeting}
\end{equation}
where the arrow denotes Sweeting's probability mode of conditional weak
convergence.  When $s_{\mathrm{eval},j,t}/s_{j,t}\to0$, the corresponding
coordinate is interpreted as degenerating at zero after total
standardization.
\end{assumption}

\begin{theorem}[Finite-block conditional evaluation limit]
\label{gp:thm:finite-eval-clt}
Under Assumption~\ref{gp:ass:eval-vector},
\begin{equation}
\mathcal{L}\left(
\left.
D_{r,t}^{-1}\bm E_{r,t}
\right|
\mathcal{F}_t^{\mathrm{fit}}
\right)
\overset{\bm{P}}{\Longrightarrow}
N_r(\bm0,B_r).
\label{gp:eq:finite-eval-clt}
\end{equation}
\end{theorem}

\begin{proof}
Fix $\bm v\in\mathbb{R}^r$.  Conditional independence gives
\begin{align*}
&\bm{E}\left[
\left.
\exp\{i\bm v^{\mathsf{T}}D_{r,t}^{-1}\bm E_{r,t}\}
\right|
\mathcal{F}_t^{\mathrm{fit}}
\right]\\
&\qquad=
\prod_{j=1}^{r}
\bm{E}\left[
\left.
\exp\left
\{i v_j E_{j,t}/s_{j,t}\right\}
\right|
\mathcal{F}_t^{\mathrm{fit}}
\right].
\end{align*}
Write $b_{j,t}=s_{\mathrm{eval},j,t}/s_{j,t}$.  By
\eqref{gp:eq:coordinate-eval-sweeting}, the $j$th factor converges in probability
to $\exp(-v_j^2 b_j^2/2)$ whenever $b_{j,t}\to b_j$.  Since $r$ is fixed, the
product converges in probability to
\[
\exp\left(-\frac12\bm v^{\mathsf{T}}B_r\bm v\right).
\]
Sweeting's random-measure characterization, applied on $\mathbb{R}^r$, gives
\eqref{gp:eq:finite-eval-clt}.
\end{proof}

\subsection{Finite-dimensional gain theorem}

Define
\begin{equation}
R_{r,t}=A_{r,t}+B_{r,t},
\qquad
R_r=A_r+B_r.
\label{gp:eq:R-matrices}
\end{equation}
The diagonal entries of $R_{r,t}$ equal one by the definition of $s_{j,t}$.
Thus $R_r$ is a correlation matrix, possibly singular.

\begin{lemma}[Asymptotic independence of fitting and evaluation blocks]
\label{gp:lem:asymptotic-independence}
Under the assumptions of Theorems~\ref{gp:thm:finite-fit-clt} and
\ref{gp:thm:finite-eval-clt},
\begin{equation}
\left(
D_{r,t}^{-1}\bm H_{r,t},
D_{r,t}^{-1}\bm E_{r,t}
\right)
\Longrightarrow
(\bm Z_{F,r},\bm Z_{E,r}),
\label{gp:eq:joint-block-limit}
\end{equation}
where
\[
\bm Z_{F,r}\sim N_r(\bm0,A_r),
\qquad
\bm Z_{E,r}\sim N_r(\bm0,B_r),
\]
and the two Gaussian vectors are independent.
\end{lemma}

\begin{proof}
For $\bm u,\bm v\in\mathbb{R}^r$, use that $\bm H_{r,t}$ is
$\mathcal{F}_t^{\mathrm{fit}}$-measurable:
\begin{align*}
&\bm{E}\exp\left\{
 i\bm u^{\mathsf{T}}D_{r,t}^{-1}\bm H_{r,t}
+i\bm v^{\mathsf{T}}D_{r,t}^{-1}\bm E_{r,t}
\right\}\\
&\quad=
\bm{E}\left[
\exp\left\{i\bm u^{\mathsf{T}}D_{r,t}^{-1}\bm H_{r,t}\right\}
\bm{E}\left[
\left.
\exp\left\{i\bm v^{\mathsf{T}}D_{r,t}^{-1}\bm E_{r,t}\right\}
\right|
\mathcal{F}_t^{\mathrm{fit}}
\right]
\right].
\end{align*}
The inner conditional characteristic function converges in probability to
$\exp(-\bm v^{\mathsf{T}}B_r\bm v/2)$ and is bounded by one, so the convergence
also holds in $L^1$.  The finite-block fitting CLT then gives the limiting
characteristic function
\[
\exp\left(-\frac12\bm u^{\mathsf{T}}A_r\bm u\right)
\exp\left(-\frac12\bm v^{\mathsf{T}}B_r\bm v\right).
\]
This proves \eqref{gp:eq:joint-block-limit}.
\end{proof}

\begin{theorem}[Finite-dimensional client gain CLT]
\label{gp:thm:finite-gain-clt}
Under the preceding assumptions,
\begin{equation}
D_{r,t}^{-1}
\left(
\widehat{\bm G}_{r,t}-\bm\Gamma_{r,t}
\right)
\Longrightarrow
N_r(\bm0,R_r),
\label{gp:eq:finite-gain-clt}
\end{equation}
where
\[
\widehat{\bm G}_{r,t}
=(\widehat G_{1,t},\ldots,\widehat G_{r,t})^{\mathsf{T}},
\qquad
\bm\Gamma_{r,t}
=(\Gamma_{1,t},\ldots,\Gamma_{r,t})^{\mathsf{T}}.
\]
\end{theorem}

\begin{proof}
Apply Theorem~\ref{gen:thm:two-stage-assembly} with
$\widehat{\bm T}_{r,t}=\widehat{\bm G}_{r,t}$,
$\bm\theta_{r,t}=\bm\Gamma_{r,t}$, and the fitting and evaluation inputs
from Theorems~\ref{gp:thm:finite-fit-clt} and
\ref{gp:thm:finite-eval-clt}.
\end{proof}

\subsection{Consistent finite-block covariance estimation}

Let $\widehat\lambda_{j,a,t}$ be the cross-fitted feasible influence summand
from the one-client theorem.  Define
\begin{equation}
\widehat\Sigma_{j\ell,t}^{\mathrm{fit}}
=
\sum_{a\in\mathcal{I}_t}
\widehat\lambda_{j,a,t}\widehat\lambda_{\ell,a,t}.
\label{gp:eq:fit-cross-cov-estimator}
\end{equation}
Let $\widehat\sigma_{E,j,t}^{2}$ be the sample variance of the observation-level
evaluation gains, and define
\begin{equation}
\widehat s_{\mathrm{eval},j,t}^{2}
=
\frac{\widehat\sigma_{E,j,t}^{2}}{N_{j,t}^{\mathrm{eval}}}.
\label{gp:eq:eval-var-estimator}
\end{equation}
Set
\begin{align}
\widehat\Sigma_{r,t}
&=
\left(
\widehat\Sigma_{j\ell,t}^{\mathrm{fit}}
\right)_{1\le j,\ell\le r}
+
\operatorname{diag}\left(
\widehat s_{\mathrm{eval},1,t}^{2},\ldots,
\widehat s_{\mathrm{eval},r,t}^{2}
\right),
\label{gp:eq:total-cov-estimator}\\
\widehat D_{r,t}
&=
\operatorname{diag}\left(
\widehat\Sigma_{11,t}^{1/2},\ldots,
\widehat\Sigma_{rr,t}^{1/2}
\right),
\label{gp:eq:Dhat}\\
\widehat R_{r,t}
&=
\widehat D_{r,t}^{-1}
\widehat\Sigma_{r,t}
\widehat D_{r,t}^{-1}.
\label{gp:eq:Rhat}
\end{align}

\begin{assumption}[Finite-block covariance estimation]
\label{gp:ass:cov-estimation}
For every fixed $r$ and every $j,\ell\le r$,
\begin{align}
&\frac{
\sum_{a\in\mathcal{I}_t}
\bm{E}_t^{\mathrm{fit}}
(\lambda_{j,a,t}^{2}\lambda_{\ell,a,t}^{2})
}{s_{j,t}^{2}s_{\ell,t}^{2}}
\longrightarrow0,
\label{gp:eq:oracle-cross-cov-condition}\\
&\frac{
\sum_{a\in\mathcal{I}_t}
(\widehat\lambda_{j,a,t}-\lambda_{j,a,t})^{2}
}{s_{j,t}^{2}}
\xrightarrow{\bm{P}}0,
\label{gp:eq:plugin-j}\\
&\frac{\widehat s_{\mathrm{eval},j,t}^{2}}
{s_{\mathrm{eval},j,t}^{2}}
\xrightarrow{\bm{P}}1.
\label{gp:eq:eval-var-consistency}
\end{align}
If $s_{\mathrm{eval},j,t}^{2}/s_{j,t}^{2}\to0$, the final ratio in
\eqref{gp:eq:eval-var-consistency} may be replaced by
$|\widehat s_{\mathrm{eval},j,t}^{2}-s_{\mathrm{eval},j,t}^{2}|/s_{j,t}^{2}\xrightarrow{\bm{P}}0$.
\end{assumption}

\begin{theorem}[Finite-block covariance approximation and studentization]
\label{gp:thm:cov-consistency}
Under Assumption~\ref{gp:ass:cov-estimation}, for every fixed $r$,
\begin{equation}
\left\|
D_{r,t}^{-1}\widehat\Sigma_{r,t}D_{r,t}^{-1}-R_{r,t}
\right\|\xrightarrow{\bm{P}}0,
\qquad
\widehat D_{r,t}D_{r,t}^{-1}\xrightarrow{\bm{P}} I_r.
\label{gp:eq:stage-covariance-approximation}
\end{equation}
If, in addition, $R_{r,t}\to R_r$, then
\begin{equation}
\widehat R_{r,t}\xrightarrow{\bm{P}} R_r.
\label{gp:eq:R-consistency}
\end{equation}
Under the hypotheses of Theorem~\ref{gp:thm:finite-gain-clt} as well,
\begin{equation}
\widehat D_{r,t}^{-1}
(\widehat{\bm G}_{r,t}-\bm\Gamma_{r,t})
\Longrightarrow
N_r(\bm0,R_r).
\label{gp:eq:studentized-vector-clt}
\end{equation}
\end{theorem}

\begin{proof}
First define the oracle empirical fitting covariance
\[
\widetilde\Sigma_{j\ell,t}^{\mathrm{fit}}
=
\sum_a\lambda_{j,a,t}\lambda_{\ell,a,t}.
\]
It is unbiased for $\Sigma_{j\ell,t}^{\mathrm{fit}}$.  Independence across
$a$ gives
\[
\Var\left(
\widetilde\Sigma_{j\ell,t}^{\mathrm{fit}}
\right)
\le
\sum_a
\bm{E}_t^{\mathrm{fit}}
(\lambda_{j,a,t}^{2}\lambda_{\ell,a,t}^{2}).
\]
Condition \eqref{gp:eq:oracle-cross-cov-condition} therefore gives
\[
\frac{
\widetilde\Sigma_{j\ell,t}^{\mathrm{fit}}
-
\Sigma_{j\ell,t}^{\mathrm{fit}}
}{s_{j,t}s_{\ell,t}}
\xrightarrow{\bm{P}}0.
\]

For the plug-in difference, Cauchy--Schwarz gives
\begin{align*}
&\left|
\sum_a\widehat\lambda_{j,a,t}\widehat\lambda_{\ell,a,t}
-
\sum_a\lambda_{j,a,t}\lambda_{\ell,a,t}
\right|\\
&\quad\le
\left\|
\widehat{\bm\lambda}_{j,t}-\bm\lambda_{j,t}
\right\|_2
\left\|
\widehat{\bm\lambda}_{\ell,t}
\right\|_2
+
\left\|
\bm\lambda_{j,t}
\right\|_2
\left\|
\widehat{\bm\lambda}_{\ell,t}-\bm\lambda_{\ell,t}
\right\|_2.
\end{align*}
The aggregate plug-in conditions and the oracle variance consistency imply
that the preceding display, divided by $s_{j,t}s_{\ell,t}$, converges to zero
in probability.  The evaluation covariance is diagonal and is consistently
estimated by \eqref{gp:eq:eval-var-estimator}.  Hence \eqref{gp:eq:stage-covariance-approximation} holds.  If
$R_{r,t}\to R_r$, it implies \eqref{gp:eq:R-consistency}.  Under the oracle
finite-dimensional CLT, \eqref{gp:eq:studentized-vector-clt} follows by
multivariate Slutsky.
\end{proof}

\subsection{The client-indexed gain process}

Let
\[
\mathbb{R}^{\infty}=\prod_{j=1}^{\infty}\mathbb{R}
\]
with the product topology, metrized by
\begin{equation}
d(\bm x,\bm y)
=
\sum_{j=1}^{\infty}
2^{-j}(|x_j-y_j|\wedge1).
\label{gp:eq:product-metric}
\end{equation}
Define the standardized gain process
\begin{equation}
T_{j,t}
=
\begin{cases}
(\widehat G_{j,t}-\Gamma_{j,t})/s_{j,t},&j\le K_t,\\
0,&j>K_t,
\end{cases}
\qquad
\bm T_t=(T_{1,t},T_{2,t},\ldots).
\label{gp:eq:gain-process}
\end{equation}

\begin{assumption}[Compatible covariance limits]
\label{gp:ass:compatible}
For every fixed $r$, the matrices $R_{r,t}$ converge to $R_r$, and the limits
are compatible:
\begin{equation}
R_{r+1}[1{:}r,1{:}r]=R_r.
\label{gp:eq:compatibility}
\end{equation}
\end{assumption}

By compatibility and the Gaussian extension theorem, there exists a centered
Gaussian random sequence
\[
\bm Z=(Z_1,Z_2,\ldots)
\]
whose first $r$ coordinates have covariance $R_r$ for every $r$.

\begin{theorem}[Client-indexed gain-process CLT]
\label{gp:thm:process-clt}
Suppose the hypotheses of Theorem~\ref{gp:thm:finite-gain-clt} hold for every
fixed client block and Assumption~\ref{gp:ass:compatible} holds.  Then
\begin{equation}
\bm T_t\Longrightarrow\bm Z
\qquad
\text{in }\mathbb{R}^{\infty}
\text{ with the product topology}.
\label{gp:eq:process-clt}
\end{equation}
If, in addition, Assumption~\ref{gp:ass:cov-estimation} holds for every fixed
block, the same conclusion holds with $s_{j,t}$ replaced by the consistent
studentizing scales $\widehat s_{j,t}$, coordinatewise.
\end{theorem}

\begin{proof}
For every fixed $r$, Theorem~\ref{gp:thm:finite-gain-clt} gives convergence of the
first $r$ coordinates of $\bm T_t$ to $N_r(\bm0,R_r)$.

It remains to prove tightness in the product space.  For each fixed $j$, the
family $\{\mathcal{L}(T_{j,t}):t\ge1\}$ is tight.  Indeed, its tail in $t$ converges to
$N(0,1)$ and the finitely many preceding distributions are tight.  Given
$\eta>0$, choose compact intervals $C_j\subset\mathbb{R}$ satisfying
\[
\sup_t\bm{P}\{T_{j,t}\notin C_j\}<\eta2^{-j},
\qquad j\ge1.
\]
Then
\[
C=\prod_{j=1}^{\infty}C_j
\]
is compact in the product topology, and the union bound gives
\[
\sup_t\bm{P}\{\bm T_t\notin C\}
\le
\sum_{j=1}^{\infty}
\sup_t\bm{P}\{T_{j,t}\notin C_j\}
<\eta.
\]
Thus the sequence of laws of $\bm T_t$ is tight.  Every subsequential limit has
the compatible Gaussian finite-dimensional distributions $\{N_r(0,R_r)\}$,
so the limit is unique and equals the law of $\bm Z$.  This proves
\eqref{gp:eq:process-clt}.

For studentization, finite-dimensional convergence follows from
Theorem~\ref{gp:thm:cov-consistency} and multivariate Slutsky.  The same
tightness argument applies coordinatewise.
\end{proof}

\subsection{Positive-gain probabilities}

Let $r$ be fixed and let $\varepsilon_{j,t}$ be deterministic thresholds.
Define the standardized margins
\begin{equation}
\gamma_{j,t}
=
\frac{\Gamma_{j,t}-\varepsilon_{j,t}}{s_{j,t}},
\qquad j=1,\ldots,r.
\label{gp:eq:margins}
\end{equation}

\begin{corollary}[Finite-block positive-gain probability]
\label{gp:cor:finite-positive}
Suppose
\[
\gamma_{j,t}\longrightarrow\gamma_j,
\qquad j=1,\ldots,r.
\]
Then
\begin{align}
&\bm{P}\left\{
\widehat G_{j,t}>\varepsilon_{j,t}
\text{ for }j=1,\ldots,r
\right\}
\nonumber\\
&\qquad\longrightarrow
\bm{P}\left\{
Z_j>-\gamma_j
\text{ for }j=1,\ldots,r
\right\}.
\label{gp:eq:finite-orthant}
\end{align}
The right side is the Gaussian orthant probability determined by $R_r$.
\end{corollary}

\begin{proof}
The event in \eqref{gp:eq:finite-orthant} can be written as
\[
\{T_{j,t}>-\gamma_{j,t},\ j=1,\ldots,r\}.
\]
The boundary of the limiting orthant is contained in the union of the
hyperplanes $\{z_j=-\gamma_j\}$.  Every coordinate $Z_j$ has variance one, so
this boundary has zero Gaussian probability.  The finite-dimensional CLT and
the portmanteau theorem give the result.
\end{proof}

\begin{remark}[Local margins]
The orthant limit is a theoretical local-power statement.  When
$\gamma_{j,t}$ converges to a finite value, the standardized margin is not
consistently estimated by replacing $\Gamma_{j,t}$ with $\widehat G_{j,t}$,
because their difference is itself of order $s_{j,t}$.  Feasible confidence
statements should therefore use the studentized gain vector or coordinatewise
confidence bounds, rather than a naive plug-in estimate of the local orthant
probability.
\end{remark}

For the moving event involving every client $j\le K_t$, product-topology
convergence alone is insufficient.  A useful sufficient condition follows
without requiring cross-client independence.

Define
\begin{equation}
\delta_{j,t}
=
\sup_{x\in\mathbb{R}}
\left|
\bm{P}(T_{j,t}\le x)-\Phi(x)
\right|.
\label{gp:eq:coordinate-error}
\end{equation}

\begin{corollary}[All-client positivity by summability]
\label{gp:cor:all-client}
Let
\[
\gamma_{j,t}
=
\frac{\Gamma_{j,t}-\varepsilon_{j,t}}{s_{j,t}},
\qquad j\le K_t.
\]
Then
\begin{align}
&\bm{P}\left\{
\widehat G_{j,t}>\varepsilon_{j,t}
\text{ for every }j\le K_t
\right\}
\nonumber\\
&\quad\ge
1-
\sum_{j=1}^{K_t}
\left\{
\Phi(-\gamma_{j,t})+\delta_{j,t}
\right\}.
\label{gp:eq:all-client-lower-bound}
\end{align}
Consequently, if
\begin{equation}
\sum_{j=1}^{K_t}
\Phi(-\gamma_{j,t})
+
\sum_{j=1}^{K_t}\delta_{j,t}
\longrightarrow0,
\label{gp:eq:summability}
\end{equation}
then
\begin{equation}
\bm{P}\left\{
\widehat G_{j,t}>\varepsilon_{j,t}
\text{ for every }j\le K_t
\right\}
\longrightarrow1.
\label{gp:eq:all-client-one}
\end{equation}
\end{corollary}

\begin{proof}
By the union bound,
\begin{align*}
&1-
\bm{P}\left\{
\widehat G_{j,t}>\varepsilon_{j,t}
\text{ for every }j\le K_t
\right\}\\
&\qquad\le
\sum_{j=1}^{K_t}
\bm{P}\{T_{j,t}\le-\gamma_{j,t}\}\\
&\qquad\le
\sum_{j=1}^{K_t}
\{\Phi(-\gamma_{j,t})+\delta_{j,t}\}.
\end{align*}
This proves \eqref{gp:eq:all-client-lower-bound}; condition
\eqref{gp:eq:summability} gives \eqref{gp:eq:all-client-one}.
\end{proof}

\subsection{Interpretation and scope}

The covariance matrix $R_r$ separates the two sources of cross-client
behavior:
\[
R_r=A_r+B_r.
\]
The fitting matrix $A_r$ is generally non-diagonal because the same federated
estimator enters every client gain.  The evaluation matrix $B_r$ is diagonal
because the fresh evaluation streams are independent across clients.  The
finite-block theorem therefore preserves the server commonality rather than
assuming independent client gains.

The process theorem is deliberately stated in the product topology.  It gives
joint limits for every fixed collection of clients and a coherent Gaussian
random sequence.  It does not, without an additional uniform condition,
convert the moving all-client event into an infinite-dimensional continuity
set.  Corollary~\ref{gp:cor:all-client} supplies a direct summability route for
that growing event.

The theorem inherits the nondegenerate first-order scope of the one-client
fitting theorem.  At the exact-model boundary, where the derivative of the
gain vanishes, the centered quadratic fitting term must be retained in every
client coordinate.  The finite-dimensional and sequence-space arguments
remain valid after replacing the linear influence vectors by the corresponding
linear-plus-clean-quadratic martingale arrays, but that boundary requires its
own one-client input theorem.

\section{Client-indexed KL certificate process: detailed assumptions and proofs}
\label{app:kl-process-proof}

\subsection{Clientwise fitted certificate}

For each stage $t$, let $K_t$ clients participate.  Client $j\le K_t$ has a
fitted local predictor $L_{j,t}$, and all clients share the fitted federated
predictor $F_t$.  The fitting samples generate the sigma-field
\[
\mathcal{F}_t^{\mathrm{fit}}.
\]
Fix a prescribed local weight $\alpha\in[0,1)$ and put
\[
\vartheta=1-\alpha.
\]
Let $Q_{j,t}$ be the evaluation law for client $j$, and define
\[
M_{j,t}(x)=\alpha L_{j,t}(x)+\vartheta F_t(x).
\]
For $z=(x,y)$, the paired predictive gain is
\begin{equation}
W_{j,t}(z)
=
\{y-L_{j,t}(x)\}^{2}-\{y-M_{j,t}(x)\}^{2}.
\label{kl:eq:paired-gain}
\end{equation}
Conditional on $\mathcal{F}_t^{\mathrm{fit}}$, define
\begin{align}
\mu_{j,t}
&=
\int W_{j,t}(z)Q_{j,t}(\,d z),
\label{kl:eq:conditional-mean}\\
v_{j,t}
&=
\int\{W_{j,t}(z)-\mu_{j,t}\}^{2}Q_{j,t}(\,d z).
\label{kl:eq:conditional-variance}
\end{align}
Thus $\mu_{j,t}$ is the gain of the realized fitted predictors and $v_{j,t}$
is the evaluation-law variance of one realized paired gain.

Let $\rho_{j,t}\ge0$ and $c_{j,t}\ge0$ be deterministic, client-specific
robustness parameters.  Define the fitted variance-sensitive certificate
\begin{equation}
C_{j,t}^{\mathrm{fit}}
=
\mu_{j,t}-\sqrt{2\rho_{j,t}v_{j,t}}-c_{j,t}\rho_{j,t},
\label{kl:eq:fitted-certificate}
\end{equation}
and its deterministic stage-$t$ marginal value
\begin{equation}
\mathcal C_{j,t}
=
\bm{E}_t^{\mathrm{fit}}C_{j,t}^{\mathrm{fit}},
\label{kl:eq:marginal-certificate}
\end{equation}
where $\bm{E}_t^{\mathrm{fit}}$ denotes expectation under the joint law of all
stage-$t$ fitting samples.

\subsubsection{Why the certificate is robustly sufficient}

For fixed fitted predictors, define the exact fitted KL-robust gain 
\begin{equation}
\underline G_{j,t}^{\mathrm{fit}}(\rho_{j,t})
=
\inf_{\substack{Q\ll Q_{j,t}\\
\operatorname{KL}(Q\|Q_{j,t})\le\rho_{j,t}}}
\int W_{j,t}(z)Q(\,d z).
\label{kl:eq:exact-robust-gain}
\end{equation}

\begin{proposition}[Exact fitted KL dual]
\label{kl:prop:exact-dual}
Fix the fitted predictors.  Suppose
$\bm{E}_{Q_{j,t}}|W_{j,t}|<\infty$ and
$\bm{E}_{Q_{j,t}}\exp(-\lambda_0W_{j,t})<\infty$ for some $\lambda_0>0$.
Then, for every $\rho_{j,t}\ge0$,
\begin{equation}
\underline G_{j,t}^{\mathrm{fit}}(\rho_{j,t})
=
\sup_{\eta>0}
\left\{
-\eta\log\bm{E}_{Q_{j,t}}\exp(-W_{j,t}/\eta)
-\eta\rho_{j,t}
\right\},
\label{kl:eq:exact-dual}
\end{equation}
where a term with an infinite log-Laplace transform is read as $-\infty$.
For $\rho_{j,t}>0$, $Q_{j,t}$ is a Slater point.  When
$\rho_{j,t}=0$, both sides equal $\bm{E}_{Q_{j,t}}W_{j,t}$.
\end{proposition}

\begin{proof}
For every $\eta>0$ and every $Q\ll Q_{j,t}$, the Gibbs variational
inequality gives
\[
\bm{E}_QW_{j,t}+\eta\operatorname{KL}(Q\|Q_{j,t})
\ge
-\eta\log\bm{E}_{Q_{j,t}}e^{-W_{j,t}/\eta}.
\]
Taking the infimum over the KL ball and then the supremum over $\eta$ gives
one inequality.  For $\rho_{j,t}>0$, the density one is strictly feasible;
convex duality for the integral functional therefore gives equality under the
stated exponential-integrability condition.  If $\rho_{j,t}=0$, the only
feasible law is $Q_{j,t}$, while integrability of $W_{j,t}$ implies
$-\eta\log\bm{E} e^{-W_{j,t}/\eta}\to\bm{E} W_{j,t}$ as
$\eta\to\infty$.
\end{proof}

The conditional sub-gamma condition used below is stated as Assumption~\ref{kl:ass:subgamma} in Section~\ref{sec:kl-robust}.

\begin{proposition}[Variance-sensitive KL certificate]
\label{kl:prop:certificate}
Under Assumption~\ref{kl:ass:subgamma},
\begin{equation}
\underline G_{j,t}^{\mathrm{fit}}(\rho_{j,t})
\ge
C_{j,t}^{\mathrm{fit}}.
\label{kl:eq:certificate-lower-bound}
\end{equation}
Consequently,
\[
C_{j,t}^{\mathrm{fit}}>0
\quad\Longrightarrow\quad
\underline G_{j,t}^{\mathrm{fit}}(\rho_{j,t})>0.
\]
\end{proposition}

\begin{proof}
For every $Q\ll Q_{j,t}$ and every admissible $\lambda>0$, the entropy
inequality gives
\[
\bm{E}_Q(\mu_{j,t}-W_{j,t})
\le
\frac{\operatorname{KL}(Q\|Q_{j,t})+
\log\bm{E}_{Q_{j,t}}e^{\lambda(\mu_{j,t}-W_{j,t})}}{\lambda}.
\]
Using \eqref{kl:eq:subgamma} and $\operatorname{KL}(Q\|Q_{j,t})\le\rho_{j,t}$,
\[
\bm{E}_QW_{j,t}
\ge
\mu_{j,t}
-
\frac{\rho_{j,t}}{\lambda}
-
\frac{\lambda v_{j,t}}{2(1-c_{j,t}\lambda)}.
\]
For $v_{j,t}>0$, choose
\[
\lambda
=
\frac{\sqrt{2\rho_{j,t}/v_{j,t}}}
{1+c_{j,t}\sqrt{2\rho_{j,t}/v_{j,t}}}.
\]
The two penalty terms then sum to
$\sqrt{2\rho_{j,t}v_{j,t}}+c_{j,t}\rho_{j,t}$.  The cases
$\rho_{j,t}=0$ or $v_{j,t}=0$ follow by continuity.  Taking the infimum over
$Q$ proves \eqref{kl:eq:certificate-lower-bound}.
\end{proof}

\begin{proposition}[Certified radius and local calibration]
\label{kl:prop:certified-radius}
Fix the fitted predictors and suppose $\mu_{j,t}>0$ and $v_{j,t}>0$.
The certificate $C_{j,t}^{\mathrm{fit}}(\rho)$ is strictly decreasing in
$\rho\ge0$.  Its critical radius
\[
\rho_{j,t}^{\mathrm{cert}}
=
\sup\{\rho\ge0:C_{j,t}^{\mathrm{fit}}(\rho)>0\}
\]
is
\begin{equation}
\rho_{j,t}^{\mathrm{cert}}
=
\frac{\mu_{j,t}^{2}}{2v_{j,t}}
\qquad\text{when }c_{j,t}=0,
\label{kl:eq:certified-radius-subgaussian}
\end{equation}
and, when $c_{j,t}>0$,
\begin{equation}
\rho_{j,t}^{\mathrm{cert}}
=
\left\{
\frac{
\sqrt{2v_{j,t}+4c_{j,t}\mu_{j,t}}-\sqrt{2v_{j,t}}
}{2c_{j,t}}
\right\}^{2}.
\label{kl:eq:certified-radius-subgamma}
\end{equation}
If $\mu_{j,t}\le0$, no positive radius is certified.

More generally, let $q_t\downarrow0$ and suppose, in probability,
\[
\frac{\mu_{j,t}}{q_t}\longrightarrow m>0,
\qquad
\frac{v_{j,t}}{q_t}\longrightarrow\sigma_\Delta^2>0,
\qquad
c_{j,t}\longrightarrow c\ge0.
\]
For $\rho_{j,t}=rq_t$,
\begin{equation}
\frac{C_{j,t}^{\mathrm{fit}}(rq_t)}{q_t}
\xrightarrow{\bm{P}}
m-\sqrt{2r\sigma_\Delta^2}-cr.
\label{kl:eq:local-radius-calibration}
\end{equation}
Thus a radius of order $q_t$ is justified only after the mean and gain
variance are both verified to have order $q_t$.
\end{proposition}

\begin{proof}
For fixed $\mu>0$ and $v>0$, solve
$\mu=\sqrt{2\rho v}+c\rho$.  When $c=0$ this gives
\eqref{kl:eq:certified-radius-subgaussian}.  When $c>0$, put
$x=\sqrt\rho$ and take the positive root of
$cx^2+\sqrt{2v}\,x-\mu=0$, giving
\eqref{kl:eq:certified-radius-subgamma}.  Equation
\eqref{kl:eq:local-radius-calibration} follows by direct division by $q_t$.
\end{proof}

\begin{remark}[Vanishing gain variance is a nonsmooth boundary]
\label{kl:rem:vanishing-variance}
The certificate remains valid when $v_{j,t}=0$, but the Gaussian delta-method
inference below does not.  For example, let $Y\sim N(0,\sigma^2)$,
$\alpha=0$, and take constant predictors $L=a$ and $F=0$.  Then
\[
W=a^2-2aY,
\qquad
\mu(a)=a^2,
\qquad
v(a)=4\sigma^2a^2,
\]
so, for a Gaussian lower tail,
\begin{equation}
C^{\mathrm{fit}}(a)
=a^2-2\sigma\sqrt{2\rho}\,|a|.
\label{kl:eq:absolute-value-boundary}
\end{equation}
The leading boundary term is an absolute-value functional, not the nominal
quadratic gain.  A nominal Hall--de Jong boundary theorem therefore does not
supply a KL-certificate boundary theorem.  Radius calibration and boundary
limit theory are separate questions.
\end{remark}

\subsection{Empirical certificate and two-stage decomposition}

Client $j$ has a fresh evaluation sample
\[
Z_{j,t,1}^{\mathrm{eval}},\ldots,Z_{j,t,N_{j,t}^{\mathrm{eval}}}^{\mathrm{eval}}
\overset{\mathrm{iid}}{\sim}Q_{j,t}.
\]
The evaluation samples are independent across clients and independent of
$\mathcal{F}_t^{\mathrm{fit}}$.  Put
\[
W_{j,t,r}=W_{j,t}(Z_{j,t,r}^{\mathrm{eval}}).
\]
Define
\begin{align}
\widehat\mu_{j,t}
&=
\frac1{N_{j,t}^{\mathrm{eval}}}\sum_{r=1}^{N_{j,t}^{\mathrm{eval}}}W_{j,t,r},
\label{kl:eq:mu-hat}\\
\widehat v_{j,t}
&=
\frac1{N_{j,t}^{\mathrm{eval}}}\sum_{r=1}^{N_{j,t}^{\mathrm{eval}}}
(W_{j,t,r}-\widehat\mu_{j,t})^{2},
\label{kl:eq:v-hat}\\
\widehat C_{j,t}
&=
\widehat\mu_{j,t}
-
\sqrt{2\rho_{j,t}\widehat v_{j,t}}
-
 c_{j,t}\rho_{j,t}.
\label{kl:eq:C-hat}
\end{align}
The exact two-stage decomposition is
\begin{equation}
\widehat C_{j,t}-\mathcal C_{j,t}
=
H^C_{j,t}+E^C_{j,t},
\qquad
H^C_{j,t}=C_{j,t}^{\mathrm{fit}}-\mathcal C_{j,t},
\qquad
E^C_{j,t}=\widehat C_{j,t}-C_{j,t}^{\mathrm{fit}}.
\label{kl:eq:two-stage}
\end{equation}
Here $H^C_{j,t}$ is $\mathcal{F}_t^{\mathrm{fit}}$-measurable.  The evaluation
term is not exactly conditionally centered because of the nonlinear square
root, but its first-order influence representation is conditionally centered.

\subsection{One-client conditional evaluation theorem}

Let
\[
X_{j,t,r}=W_{j,t,r}-\mu_{j,t},
\qquad
\chi^E_{j,t}=\sqrt{\frac{\rho_{j,t}}{2v_{j,t}}}
\]
on the event $v_{j,t}>0$.  Define the conditional certificate influence
variable
\begin{equation}
\xi_{j,t,r}
=
X_{j,t,r}
-
\chi^E_{j,t}\{X_{j,t,r}^{2}-v_{j,t}\},
\label{kl:eq:eval-influence}
\end{equation}
and its conditional variance
\begin{equation}
\omega_{C,j,t}^{2}
=
\Var(\xi_{j,t,1}\mid\mathcal{F}_t^{\mathrm{fit}}).
\label{kl:eq:omega-C}
\end{equation}

\begin{assumption}[Regular conditional certificate regime]
\label{kl:ass:eval-regular}
For every fixed client block $1,\ldots,r$,
$\min_{j\le r}N_{j,t}^{\mathrm{eval}}\to\infty$ and
$\sup_{j\le r,t}\rho_{j,t}<\infty$.  There are constants
$0<c_v<C_v<\infty$, $c_\omega>0$, and $C_6<\infty$ such that
\begin{equation}
\bm{P}\left\{
\min_{j\le r}v_{j,t}\ge c_v,
\quad
\max_{j\le r}v_{j,t}\le C_v,
\quad
\min_{j\le r}\omega_{C,j,t}^{2}\ge c_\omega,
\quad
\max_{j\le r}
\bm{E}(|X_{j,t,1}|^6\mid\mathcal{F}_t^{\mathrm{fit}})\le C_6
\right\}
\longrightarrow1.
\label{kl:eq:eval-regular}
\end{equation}
\end{assumption}

\begin{theorem}[Conditional KL-certificate delta theorem]
\label{kl:thm:one-client-eval}
Under Assumption~\ref{kl:ass:eval-regular}, for every fixed client $j$,
\begin{equation}
E^C_{j,t}
=
\frac1{N_{j,t}^{\mathrm{eval}}}\sum_{r=1}^{N_{j,t}^{\mathrm{eval}}}\xi_{j,t,r}
+r^E_{j,t},
\qquad
\frac{\sqrt{N_{j,t}^{\mathrm{eval}}}\,r^E_{j,t}}{\omega_{C,j,t}}
\xrightarrow{\bm{P}}0.
\label{kl:eq:eval-linearization}
\end{equation}
Moreover, in Sweeting's probability mode,
\begin{equation}
\mathcal{L}\left(
\left.
\frac{\sqrt{N_{j,t}^{\mathrm{eval}}}E^C_{j,t}}{\omega_{C,j,t}}
\right|
\mathcal{F}_t^{\mathrm{fit}}
\right)
\overset{\bm{P}}{\Longrightarrow}
N(0,1).
\label{kl:eq:one-client-eval-clt}
\end{equation}
\end{theorem}

\begin{proof}
Let $m=N_{j,t}^{\mathrm{eval}}$, $\overline X=m^{-1}\sum_{r=1}^mX_{j,t,r}$, and suppress
$(j,t)$, writing $\omega_C^2=\omega_{C,j,t}^2$.  The sample-variance identity is
\begin{equation}
\widehat v-v
=
\frac1m\sum_{r=1}^m(X_r^2-v)-\overline X^2.
\label{kl:eq:sample-variance-identity}
\end{equation}
Put $\Delta_v=\widehat v-v$.  Since $\widehat v\ge0$ and $v>0$, the square-root
identity
\[
\sqrt{\widehat v}-\sqrt v-\frac{\Delta_v}{2\sqrt v}
=
-\frac{\Delta_v^2}
{2\sqrt v(\sqrt{\widehat v}+\sqrt v)^2}
\]
holds without restricting to a Taylor neighborhood.  Therefore
\begin{equation}
\widehat C-C^{\mathrm{fit}}
=
(\widehat\mu-\mu)-\chi^E\Delta_v+R_m,
\qquad
0\le R_m
\le
\frac{\chi^E}{v}\Delta_v^2.
\label{kl:eq:global-square-root-remainder}
\end{equation}
Substitution of \eqref{kl:eq:sample-variance-identity} gives the exact
linearization
\[
\widehat C-C^{\mathrm{fit}}
=
\frac1m\sum_{r=1}^m\xi_r
+
\chi^E\overline X^2+R_m.
\]
On the event in \eqref{kl:eq:eval-regular}, conditional moments through order
six are uniformly bounded, $v$ is bounded above and below, $\chi^E$ is bounded,
and $\omega_C$ is bounded below.  Consequently,
\[
\bm{E}\!\left[
\left.
\left\{\frac1m\sum_{r=1}^m(X_r^2-v)\right\}^{2}
\right|\mathcal{F}_t^{\mathrm{fit}}
\right]\le\frac{C}{m},
\qquad
\bm{E}(\overline X^2\mid\mathcal{F}_t^{\mathrm{fit}})=\frac{v}{m},
\]
and the standard fourth-moment formula for an iid sample mean gives
$\bm{E}(\overline X^4\mid\mathcal{F}_t^{\mathrm{fit}})\le C/m^2$.  Hence
\[
\bm{E}(\Delta_v^2\mid\mathcal{F}_t^{\mathrm{fit}})\le\frac{C}{m}
\]
and, by \eqref{kl:eq:global-square-root-remainder},
\[
\bm{E}\left[
\left.
\frac{\sqrt m}{\omega_C}
\{\chi^E\overline X^2+R_m\}
\right|\mathcal{F}_t^{\mathrm{fit}}
\right]
\le\frac{C}{\sqrt m}.
\]
Conditional Markov's inequality proves
\eqref{kl:eq:eval-linearization}.

Conditionally on $\mathcal{F}_t^{\mathrm{fit}}$, the variables $\xi_r$ are iid,
centered, and have variance $\omega_C^2$.  Moreover,
\[
|\xi_r|^3
\le
C\{1+|X_r|^6\}
\]
on the event in \eqref{kl:eq:eval-regular}.  Thus their conditional
standardized third moments are uniformly bounded.  The pathwise
Berry--Esseen theorem gives a Kolmogorov error of order $m^{-1/2}$ for the
linear term, while the preceding remainder bound is negligible in conditional
probability.  Interpreting the conditional distributions as random probability
measures and using Sweeting's probability mode yields
\eqref{kl:eq:one-client-eval-clt}.
\end{proof}

For studentization, define
\begin{equation}
\widehat\chi^E_{j,t}
=
\mathbf{1}\{\widehat v_{j,t}>0\}
\sqrt{\frac{\rho_{j,t}}{2\widehat v_{j,t}}},
\label{kl:eq:chi-hat}
\end{equation}
so that it is defined on the zero empirical-variance event as well, and
\begin{equation}
\widehat\xi_{j,t,r}
=
(W_{j,t,r}-\widehat\mu_{j,t})
-
\widehat\chi^E_{j,t}
\{(W_{j,t,r}-\widehat\mu_{j,t})^2-\widehat v_{j,t}\}.
\label{kl:eq:xi-hat}
\end{equation}
Set
\begin{equation}
\widehat\omega_{C,j,t}^{2}
=
\frac1{N_{j,t}^{\mathrm{eval}}}\sum_{r=1}^{N_{j,t}^{\mathrm{eval}}}\widehat\xi_{j,t,r}^{2},
\qquad
\widehat s_{C,\mathrm{eval},j,t}^{2}
=
\frac{\widehat\omega_{C,j,t}^{2}}{N_{j,t}^{\mathrm{eval}}}.
\label{kl:eq:eval-var-estimator}
\end{equation}

\begin{corollary}[Conditional certificate studentization]
\label{kl:cor:eval-studentization}
Under Assumption~\ref{kl:ass:eval-regular},
\begin{equation}
\frac{\widehat\omega_{C,j,t}^{2}}{\omega_{C,j,t}^{2}}\xrightarrow{\bm{P}}1,
\qquad
\mathcal{L}\left(
\left.
\frac{\sqrt{N_{j,t}^{\mathrm{eval}}}E^C_{j,t}}{\widehat\omega_{C,j,t}}
\right|\mathcal{F}_t^{\mathrm{fit}}
\right)
\overset{\bm{P}}{\Longrightarrow}N(0,1).
\label{kl:eq:eval-studentization}
\end{equation}
\end{corollary}

\begin{proof}
Suppress $(j,t)$ and again condition on the fitting sigma-field, writing $\omega_C^2=\omega_{C,j,t}^2$ and $\widehat\omega_C^2=\widehat\omega_{C,j,t}^2$.  First,
$\widehat\mu-\mu=\overline X\to0$ and $\widehat v-v\to0$ in conditional
probability, uniformly on the event in \eqref{kl:eq:eval-regular}.  Hence
$\widehat\chi^E-\chi^E\to0$ in conditional probability; the convention in
\eqref{kl:eq:chi-hat} is immaterial because
$\bm{P}(\widehat v=0\mid\mathcal{F}_t^{\mathrm{fit}})\to0$ on that event.

For the oracle influence variables, the conditional von Bahr--Esseen
inequality with exponent $3/2$ gives
\[
\bm{E}\left[
\left.
\left|
\frac1m\sum_{r=1}^m(\xi_r^2-\omega_C^2)
\right|^{3/2}
\right|\mathcal{F}_t^{\mathrm{fit}}
\right]
\le
\frac{C\,\bm{E}(|\xi_1|^3\mid\mathcal{F}_t^{\mathrm{fit}})}{\sqrt m}
\le\frac{C}{\sqrt m}.
\]
Thus $m^{-1}\sum_r\xi_r^2\to\omega_C^2$ in conditional probability.  Next, direct
expansion of \eqref{kl:eq:xi-hat} gives
\[
\frac1m\sum_{r=1}^m(\widehat\xi_r-\xi_r)^2
\le
C\left[
\overline X^2\left\{1+\frac1m\sum_{r=1}^mX_r^2\right\}
+
(\widehat v-v)^2
+
(\widehat\chi^E-\chi^E)^2
\left\{1+\frac1m\sum_{r=1}^mX_r^4\right\}
\right].
\]
The right side converges to zero in conditional probability.  Indeed, the
conditional sixth-moment bound and another von Bahr--Esseen argument imply
that the empirical second and fourth moments are conditionally bounded in
probability.  Cauchy--Schwarz now shows
\[
\left|
\frac1m\sum_{r=1}^m\widehat\xi_r^2
-
\frac1m\sum_{r=1}^m\xi_r^2
\right|
\longrightarrow0
\]
in conditional probability.  Therefore
$\widehat\omega_C^2/\omega_C^2\to1$ in conditional probability, and hence also
unconditionally in probability.  The variance floor and Sweeting's
conditional mapping theorem, applied to
Theorem~\ref{kl:thm:one-client-eval}, prove the studentized conclusion.
\end{proof}

For a one-sided conditional confidence level $1-\delta$, define
\begin{equation}
\operatorname{LCB}^{C,\mathrm{fit}}_{j,t}(\delta)
=
\widehat C_{j,t}
-z_{1-\delta}
\frac{\widehat\omega_{C,j,t}}
{\sqrt{N_{j,t}^{\mathrm{eval}}}}.
\label{kl:eq:fitted-certificate-lcb}
\end{equation}
Then, under Assumption~\ref{kl:ass:eval-regular},
\[
\bm{P}\left[
\left.
C_{j,t}^{\mathrm{fit}}
\ge
\operatorname{LCB}^{C,\mathrm{fit}}_{j,t}(\delta)
\right|\mathcal{F}_t^{\mathrm{fit}}
\right]
=1-\delta+o_{\bm{P}}(1).
\]
Consequently, positivity of this lower confidence bound supports, for the
realized fitted pair, the pathwise implication in
Proposition~\ref{kl:prop:certificate}.  Positivity of the empirical certificate
alone does not provide that uncertainty statement.

\subsection{One-client fitting expansion of the gain mean and variance}

Let $m^Q_{j,t}(x)=\bm{E}_{Q_{j,t}}(Y\mid X=x)$ and
$\sigma_{Q,j,t}^{2}(x)=\Var_{Q_{j,t}}(Y\mid X=x)$.  Write
$\Gamma_{j,t}=\bm{E}_t^{\mathrm{fit}}\mu_{j,t}$, and let
$\mathfrak r_{j,t}$ denote the leverage aggregate from the one-client
construction when client $j$ is selected.  For deterministic functions
$(l,f)$, put
\[
d=f-l,
\qquad
a=m^Q_{j,t}-l,
\qquad
\delta_{j,t}(l,f)=2\vartheta ad-\vartheta^2d^2.
\]
Writing $Y=m^Q_{j,t}(X)+\varepsilon$ with
$\bm{E}(\varepsilon\mid X)=0$ gives
\begin{align}
\mu_{j,t}(l,f)
&=
\int\delta_{j,t}(l,f;x)Q_{j,t,X}(\,d x),
\label{kl:eq:mean-functional}\\
v_{j,t}(l,f)
&=
\int\{\delta_{j,t}(l,f;x)-\mu_{j,t}(l,f)\}^2Q_{j,t,X}(\,d x)
\nonumber\\
&\quad+
4\vartheta^2\int d(x)^2\sigma_{Q,j,t}^2(x)Q_{j,t,X}(\,d x).
\label{kl:eq:variance-functional}
\end{align}

Let
\[
\overline L_{j,t}=\bm{E}_t^{\mathrm{fit}}L_{j,t},
\qquad
\overline F_t=\bm{E}_t^{\mathrm{fit}}F_t,
\]
and define
\[
a^0_{j,t}=m^Q_{j,t}-\overline L_{j,t},
\qquad
d^0_{j,t}=\overline F_t-\overline L_{j,t},
\]
\[
\delta^0_{j,t}
=2\vartheta a^0_{j,t}d^0_{j,t}
-\vartheta^2(d^0_{j,t})^2,
\qquad
\mu^0_{j,t}=\int\delta^0_{j,t}\,d Q_{j,t,X}.
\]
The derivative coefficient functions are
\begin{align}
c_{\mu,L,j,t}
&=-2\vartheta\{a^0_{j,t}+\alpha d^0_{j,t}\},
\label{kl:eq:c-mu-L}\\
c_{\mu,F,j,t}
&=2\vartheta\{a^0_{j,t}-\vartheta d^0_{j,t}\},
\label{kl:eq:c-mu-F}\\
c_{v,L,j,t}
&=2(\delta^0_{j,t}-\mu^0_{j,t})c_{\mu,L,j,t}
-8\vartheta^2d^0_{j,t}\sigma_{Q,j,t}^2,
\label{kl:eq:c-v-L}\\
c_{v,F,j,t}
&=2(\delta^0_{j,t}-\mu^0_{j,t})c_{\mu,F,j,t}
+8\vartheta^2d^0_{j,t}\sigma_{Q,j,t}^2.
\label{kl:eq:c-v-F}
\end{align}

\begin{lemma}[Differentiability of the gain mean and variance]
\label{kl:lem:mean-variance-differentiability}
Suppose the fitted and mean curves are clipped at a fixed level,
$\sup_{j,t}\norm{m^Q_{j,t}}_\infty<\infty$, and
$\sup_{j,t}\norm{\sigma_{Q,j,t}^2}_\infty<\infty$.  For perturbations
$(u_L,u_F)$ whose sup norms are uniformly bounded,
\begin{align}
\mu_{j,t}(\overline L_{j,t}+u_L,\overline F_t+u_F)-\mu^0_{j,t}
&=
\inner{c_{\mu,L,j,t}}{u_L}_{j,t}
+
\inner{c_{\mu,F,j,t}}{u_F}_{j,t}
+R_{\mu,j,t},
\label{kl:eq:mu-taylor}\\
v_{j,t}(\overline L_{j,t}+u_L,\overline F_t+u_F)-v^0_{j,t}
&=
\inner{c_{v,L,j,t}}{u_L}_{j,t}
+
\inner{c_{v,F,j,t}}{u_F}_{j,t}
+R_{v,j,t},
\label{kl:eq:v-taylor}
\end{align}
where $v^0_{j,t}=v_{j,t}(\overline L_{j,t},\overline F_t)$ and
\begin{equation}
|R_{\mu,j,t}|+|R_{v,j,t}|
\le
C\{\norm{u_L}_{j,t}^{2}+\norm{u_F}_{j,t}^{2}\}.
\label{kl:eq:mv-remainder}
\end{equation}
\end{lemma}

\begin{proof}
Put $s=u_F-u_L$ and suppress $(j,t)$.  A direct expansion of the pointwise
mean gain gives
\[
\delta(\overline L+u_L,\overline F+u_F)
=
\delta^0
+
h
+
q,
\]
where
\[
h=c_{\mu,L}u_L+c_{\mu,F}u_F,
\qquad
q=-2\vartheta u_Ls-\vartheta^2s^2.
\]
Thus $|q|\le C(u_L^2+u_F^2)$ pointwise, which proves
\eqref{kl:eq:mu-taylor} and the asserted bound for $R_\mu$.

For the variance functional, write
$\widetilde\delta=\delta^0+h+q$.  Expanding its variance gives
\begin{align*}
\Var(\widetilde\delta)-\Var(\delta^0)
={}&
2\Cov(\delta^0,h)
+
2\Cov(\delta^0,q)
+
\Var(h)\\
&+
2\Cov(h,q)
+
\Var(q).
\end{align*}
The first term is the derivative of the first term in
\eqref{kl:eq:variance-functional}.  The derivative of the conditional-noise
term is
\[
8\vartheta^2
\int d^0s\,\sigma_Q^2\,\,d Q_X.
\]
Together these two linear terms equal
$\inner{c_{v,L}}{u_L}+\inner{c_{v,F}}{u_F}$.

The remaining contribution from the conditional-noise term is
$4\vartheta^2\int s^2\sigma_Q^2\,\,d Q_X$.  Since $\delta^0$ is uniformly
bounded, $|h|\le C(|u_L|+|u_F|)$, $|q|\le C(u_L^2+u_F^2)$, and the
perturbations are uniformly bounded, each of
$\Cov(\delta^0,q)$, $\Var(h)$, $\Cov(h,q)$, $\Var(q)$, and the last displayed
noise term is bounded in absolute value by
\[
C\{\norm{u_L}_{j,t}^{2}+\norm{u_F}_{j,t}^{2}\}.
\]
This proves \eqref{kl:eq:v-taylor} and
\eqref{kl:eq:mv-remainder}.
\end{proof}

For client $j$, let $\phi_{j,a,L,t}$ and $\phi_{a,F,t}$ be the local and
server first-order ratio influences from \eqref{sg:eq:phiL}--
\eqref{sg:eq:phiF}, applied with the selected client equal to $j$.  Define
\begin{align}
\lambda_{\mu,j,a,t}
&=
\inner{c_{\mu,L,j,t}}{\phi_{j,a,L,t}}_{j,t}
+
\inner{c_{\mu,F,j,t}}{\phi_{a,F,t}}_{j,t},
\label{kl:eq:lambda-mu}\\
\lambda_{v,j,a,t}
&=
\inner{c_{v,L,j,t}}{\phi_{j,a,L,t}}_{j,t}
+
\inner{c_{v,F,j,t}}{\phi_{a,F,t}}_{j,t}.
\label{kl:eq:lambda-v}
\end{align}
Put
\begin{equation}
\bm\lambda^{MV}_{j,a,t}
=
(\lambda_{\mu,j,a,t},\lambda_{v,j,a,t})^{\mathsf{T}},
\qquad
\Sigma^{MV}_{j,t}
=
\sum_{a\in\mathcal{I}_t}
\bm{E}_t^{\mathrm{fit}}
\left[
\bm\lambda^{MV}_{j,a,t}\bm\lambda^{MV\mathsf{T}}_{j,a,t}
\right].
\label{kl:eq:mv-covariance}
\end{equation}

\begin{assumption}[Persistent joint mean--variance information]
\label{kl:ass:mv-fitting}
For every client sequence covered by the theorem, the assumptions of the
one-client nominal-gain theorem hold with evaluation law $Q_{j,t}$, and the
coefficient functions in \eqref{kl:eq:c-mu-L}--\eqref{kl:eq:c-v-F} are
uniformly bounded.  There is a finite constant $C$ such that
\begin{equation}
\lambda_{\max}(\Sigma^{MV}_{j,t})
\le C\mathfrak r_{j,t}
\label{kl:eq:mv-information-upper}
\end{equation}
for all sufficiently large $t$.  The no-dominant-observation condition holds
with the same moment exponent as in Assumption~\ref{sg:ass:moments}.
\end{assumption}

\begin{theorem}[Joint fitting expansion for the gain mean and variance]
\label{kl:thm:mv-fitting-clt}
Under Assumption~\ref{kl:ass:mv-fitting}, with
$\overline v_{j,t}=\bm{E}_t^{\mathrm{fit}}v_{j,t}$,
\begin{equation}
\begin{pmatrix}
\mu_{j,t}-\Gamma_{j,t}\\
v_{j,t}-\overline v_{j,t}
\end{pmatrix}
=
\sum_{a\in\mathcal{I}_t}\bm\lambda^{MV}_{j,a,t}
+
\bm r^{MV}_{j,t},
\qquad
\frac{\norm{\bm r^{MV}_{j,t}}}{\sqrt{\mathfrak r_{j,t}}}
\xrightarrow{\bm{P}}0.
\label{kl:eq:mv-representation}
\end{equation}
If, in addition,
\begin{equation}
\frac{\Sigma^{MV}_{j,t}}{\mathfrak r_{j,t}}
\longrightarrow \Omega^{MV}_j
\label{kl:eq:mv-covariance-limit}
\end{equation}
for a symmetric nonnegative definite matrix $\Omega^{MV}_j$, then
\begin{equation}
\frac1{\sqrt{\mathfrak r_{j,t}}}
\begin{pmatrix}
\mu_{j,t}-\Gamma_{j,t}\\
v_{j,t}-\overline v_{j,t}
\end{pmatrix}
\Longrightarrow N_2(0,\Omega^{MV}_j).
\label{kl:eq:mv-fitting-clt}
\end{equation}
More generally, let $\bm a_{j,t}\in\mathbb{R}^2$ be deterministic with
$\sup_t\norm{\bm a_{j,t}}<\infty$, and put
\begin{equation}
s_{a,j,t}^{2}
=
\bm a_{j,t}^{\mathsf{T}}\Sigma^{MV}_{j,t}\bm a_{j,t}.
\label{kl:eq:contrast-variance}
\end{equation}
Whenever
\begin{equation}
0<c\mathfrak r_{j,t}
\le s_{a,j,t}^{2}
\le C\mathfrak r_{j,t}
\label{kl:eq:contrast-nondegeneracy}
\end{equation}
for finite positive constants $c,C$,
\begin{equation}
\frac{
\bm a_{j,t}^{\mathsf{T}}
\begin{pmatrix}
\mu_{j,t}-\Gamma_{j,t}\\
v_{j,t}-\overline v_{j,t}
\end{pmatrix}
}{s_{a,j,t}}
\Longrightarrow N(0,1).
\label{kl:eq:contrast-clt}
\end{equation}
Thus the bivariate covariance may be singular; only the contrast used by the
certificate must be nondegenerate.
\end{theorem}

\begin{proof}
Use the generic coefficient form of
Lemma~\ref{sg:lem:ratio-linearization} for each of the four deterministic
bounded coefficient functions in
\eqref{kl:eq:c-mu-L}--\eqref{kl:eq:c-v-F}.  This gives the four integrated
linear terms as the corresponding sums in
\eqref{kl:eq:lambda-mu}--\eqref{kl:eq:lambda-v}, with scalar ratio
remainders that are $O_{\bm{P}}(\mathfrak r_{j,t}/h_t)$ and whose absolute
first moments have the same order.

Apply Lemma~\ref{kl:lem:mean-variance-differentiability} with
$u_L=L_{j,t}-\overline L_{j,t}$ and
$u_F=F_t-\overline F_t$.  By
Lemma~\ref{sg:lem:integrated-second-moments},
\[
\bm{E}_t^{\mathrm{fit}}
\{|R_{\mu,j,t}|+|R_{v,j,t}|\}
\le
C\frac{\mathfrak r_{j,t}}{h_t},
\]
and Markov's inequality gives the same stochastic order.  After subtracting
fitting expectations, all ratio and functional remainders remain
$O_{\bm{P}}(\mathfrak r_{j,t}/h_t)$.  Because
$\sqrt{\mathfrak r_{j,t}}/h_t\to0$, the vector remainder divided by
$\sqrt{\mathfrak r_{j,t}}$ is negligible.  This proves
\eqref{kl:eq:mv-representation}.

The observation-level vectors are independent and centered.  Bounded
coefficient functions and kernel integration give
\[
\sum_{a\in\mathcal{I}_t}
\bm{E}_t^{\mathrm{fit}}
\norm{\bm\lambda^{MV}_{j,a,t}}^{2+\delta}
\le
C\sum_a\gamma_{a,t}^{2+\delta}
\le
C\gamma_{\max,t}^{\delta}
\mathfrak r_{j,t}.
\]
Consequently,
\[
\frac{
\sum_a\bm{E}_t^{\mathrm{fit}}
\norm{\bm\lambda^{MV}_{j,a,t}}^{2+\delta}
}{\mathfrak r_{j,t}^{1+\delta/2}}
\le
C\left(
\frac{\gamma_{\max,t}}{\sqrt{\mathfrak r_{j,t}}}
\right)^{\delta}
\longrightarrow0.
\]
Cram\'er--Wold, the vector Lyapunov theorem, and
\eqref{kl:eq:mv-covariance-limit} prove
\eqref{kl:eq:mv-fitting-clt}.  For the moving scalar contrast, the summands
$\bm a_{j,t}^{\mathsf{T}}\bm\lambda^{MV}_{j,a,t}$ have total variance
$s_{a,j,t}^{2}$.  Boundedness of $\bm a_{j,t}$ and
\eqref{kl:eq:contrast-nondegeneracy} give the same Lyapunov bound after
normalization by $s_{a,j,t}$.  Slutsky's theorem then proves
\eqref{kl:eq:contrast-clt}.
\end{proof}
Define
\begin{equation}
\chi^F_{j,t}=\sqrt{\frac{\rho_{j,t}}{2\overline v_{j,t}}},
\qquad
\lambda^C_{j,a,t}
=
\lambda_{\mu,j,a,t}-\chi^F_{j,t}\lambda_{v,j,a,t},
\label{kl:eq:certificate-fit-influence}
\end{equation}
and
\begin{equation}
s_{C,\mathrm{fit},j,t}^{2}
=
\sum_{a\in\mathcal{I}_t}
\bm{E}_t^{\mathrm{fit}}(\lambda^C_{j,a,t})^2.
\label{kl:eq:fit-scale-one-client}
\end{equation}

The regular fitting-certificate condition is Assumption~\ref{kl:ass:fit-regular}, stated in Section~\ref{sec:kl-robust}.

\begin{corollary}[One-client fitting CLT for the KL certificate]
\label{kl:cor:one-client-fit}
Under Theorem~\ref{kl:thm:mv-fitting-clt} and
Assumption~\ref{kl:ass:fit-regular},
\begin{equation}
C_{j,t}^{\mathrm{fit}}-\mathcal C_{j,t}
=
\sum_{a\in\mathcal{I}_t}\lambda^C_{j,a,t}+r^C_{j,t},
\qquad
\frac{r^C_{j,t}}{s_{C,\mathrm{fit},j,t}}\xrightarrow{\bm{P}}0,
\label{kl:eq:certificate-fit-representation-one}
\end{equation}
and
\begin{equation}
\frac{C_{j,t}^{\mathrm{fit}}-\mathcal C_{j,t}}
{s_{C,\mathrm{fit},j,t}}
\Longrightarrow N(0,1).
\label{kl:eq:one-client-fit-clt}
\end{equation}
\end{corollary}

\begin{proof}
Apply Taylor's theorem to
$g_{j,t}(\mu,v)=\mu-\sqrt{2\rho_{j,t}v}-c_{j,t}\rho_{j,t}$ at
$(\Gamma_{j,t},\overline v_{j,t})$.  Its derivative is
$(1,-\chi^F_{j,t})$, and, on the deterministic variance floor in
Assumption~\ref{kl:ass:fit-regular},
\[
\left|
g_{j,t}(\mu,v)-g_{j,t}(\Gamma_{j,t},\overline v_{j,t})
-(\mu-\Gamma_{j,t})
+\chi^F_{j,t}(v-\overline v_{j,t})
\right|
\le
C(v-\overline v_{j,t})^2.
\]
Theorem~\ref{kl:thm:mv-fitting-clt} gives
$v_{j,t}-\overline v_{j,t}=O_{\bm{P}}(\sqrt{\mathfrak r_{j,t}})$, so the
random Taylor remainder is $O_{\bm{P}}(\mathfrak r_{j,t})$ and is negligible
relative to $s_{C,\mathrm{fit},j,t}$.

To center at $\mathcal C_{j,t}$, we also need a first-mean bound.  On the
clipped set the variance functional is Lipschitz in the two fitted curves:
\[
|v_{j,t}(L_{j,t},F_t)-v^0_{j,t}|
\le
C\{\norm{L_{j,t}-\overline L_{j,t}}_{j,t}
+\norm{F_t-\overline F_t}_{j,t}\}.
\]
Hence
\[
\bm{E}_t^{\mathrm{fit}}
(v_{j,t}-\overline v_{j,t})^2
\le
C\frac{\mathfrak r_{j,t}}{h_t}
\]
by Lemma~\ref{sg:lem:integrated-second-moments}.  The expected absolute
Taylor remainder is therefore
$O(\mathfrak r_{j,t}/h_t)=o(\sqrt{\mathfrak r_{j,t}})$ by
Assumption~\ref{sg:ass:rates}.  Subtracting fitting expectations now gives
\eqref{kl:eq:certificate-fit-representation-one}.

Finally, the scalar Lyapunov condition for
$\lambda_{\mu,j,a,t}-\chi^F_{j,t}\lambda_{v,j,a,t}$ follows from
Assumption~\ref{kl:ass:mv-fitting}, the boundedness of $\chi^F_{j,t}$, and
Assumption~\ref{kl:ass:fit-regular}.  The scalar triangular-array CLT and
Slutsky's theorem prove \eqref{kl:eq:one-client-fit-clt}.
\end{proof}

\subsection{Feasible certificate influences under cross-fitting}

The theoretical certificate influence uses the repeated-fitting mean curves
$\overline L_{j,t}$ and $\overline F_t$ and the repeated-fitting variance
$\overline v_{j,t}$.  As in the nominal-gain variance construction, these
objects are replaced by finite-bandwidth targets before nuisance estimation.
Let $m^h_{j,L,t}$ and $m^h_{F,t}$ denote the client-$j$ local and federated
finite-bandwidth ratio targets.  Put
\[
\widetilde a_{j,t}=m^Q_{j,t}-m^h_{j,L,t},
\qquad
\widetilde d_{j,t}=m^h_{F,t}-m^h_{j,L,t},
\]
\[
\widetilde\delta_{j,t}
=2\vartheta\widetilde a_{j,t}\widetilde d_{j,t}
-\vartheta^2\widetilde d_{j,t}^{2},
\qquad
\widetilde\mu_{j,t}
=\int\widetilde\delta_{j,t}\,\,d Q_{j,t,X},
\]
and
\begin{align}
\widetilde v_{j,t}
&=
\int(\widetilde\delta_{j,t}-\widetilde\mu_{j,t})^2
\,\,d Q_{j,t,X}
+4\vartheta^2
\int\widetilde d_{j,t}^{2}\sigma_{Q,j,t}^{2}
\,\,d Q_{j,t,X}.
\label{kl:eq:finite-bandwidth-gain-variance}
\end{align}
Define the finite-bandwidth coefficient functions
\begin{align}
\widetilde c_{\mu,L,j,t}
&=-2\vartheta(\widetilde a_{j,t}+\alpha\widetilde d_{j,t}),\\
\widetilde c_{\mu,F,j,t}
&=2\vartheta(\widetilde a_{j,t}-\vartheta\widetilde d_{j,t}),\\
\widetilde c_{v,L,j,t}
&=2(\widetilde\delta_{j,t}-\widetilde\mu_{j,t})
\widetilde c_{\mu,L,j,t}
-8\vartheta^2\widetilde d_{j,t}\sigma_{Q,j,t}^{2},\\
\widetilde c_{v,F,j,t}
&=2(\widetilde\delta_{j,t}-\widetilde\mu_{j,t})
\widetilde c_{\mu,F,j,t}
+8\vartheta^2\widetilde d_{j,t}\sigma_{Q,j,t}^{2}.
\label{kl:eq:finite-bandwidth-coefficients}
\end{align}
The corresponding proxy influences are
\begin{align}
\widetilde\lambda_{\mu,j,a,t}
&=
\inner{\widetilde c_{\mu,L,j,t}}{\phi_{j,a,L,t}}_{j,t}
+
\inner{\widetilde c_{\mu,F,j,t}}{\phi_{a,F,t}}_{j,t},\\
\widetilde\lambda_{v,j,a,t}
&=
\inner{\widetilde c_{v,L,j,t}}{\phi_{j,a,L,t}}_{j,t}
+
\inner{\widetilde c_{v,F,j,t}}{\phi_{a,F,t}}_{j,t}.
\label{kl:eq:finite-bandwidth-mv-influences}
\end{align}
On the event $\widetilde v_{j,t}>0$, put
\[
\widetilde\chi^F_{j,t}
=
\sqrt{\frac{\rho_{j,t}}{2\widetilde v_{j,t}}},
\qquad
\widetilde\lambda^C_{j,a,t}
=
\widetilde\lambda_{\mu,j,a,t}
-
\widetilde\chi^F_{j,t}\widetilde\lambda_{v,j,a,t}.
\]
On its complement, define these quantities arbitrarily.  The next lemma shows that this exceptional case is absent for all sufficiently large $t$ under the regular certificate variance floor.

\begin{lemma}[Finite-bandwidth certificate proxy equivalence]
\label{kl:lem:certificate-proxy}
Under Assumptions~\ref{sg:ass:kernel}--\ref{sg:ass:rates},
Assumption~\ref{kl:ass:mv-fitting}, and
Assumption~\ref{kl:ass:fit-regular},
\begin{equation}
|\widetilde v_{j,t}-\overline v_{j,t}|\longrightarrow0
\label{kl:eq:variance-proxy-consistency}
\end{equation}
and
\begin{equation}
\frac{
\sum_{a\in\mathcal{I}_t}
\bm{E}_t^{\mathrm{fit}}
(\widetilde\lambda^C_{j,a,t}-\lambda^C_{j,a,t})^2
}{s_{C,\mathrm{fit},j,t}^{2}}
\longrightarrow0.
\label{kl:eq:certificate-proxy-equivalence}
\end{equation}
\end{lemma}

\begin{proof}
The proof of Lemma~\ref{sg:lem:proxy-equivalence} gives
\[
\norm{\overline L_{j,t}-m^h_{j,L,t}}_{j,t}^{2}
+
\norm{\overline F_t-m^h_{F,t}}_{j,t}^{2}
\le C\frac{\mathfrak r_{j,t}}{h_t}
\longrightarrow0.
\]
The clipped gain-variance functional is locally Lipschitz in its two curve
arguments.  Hence
\[
|v^0_{j,t}-\widetilde v_{j,t}|\longrightarrow0,
\qquad
v^0_{j,t}=v_{j,t}(\overline L_{j,t},\overline F_t).
\]
Moreover, the differentiability expansion and centeredness of the fitting
errors give
\[
|\overline v_{j,t}-v^0_{j,t}|
\le
C\bm{E}_t^{\mathrm{fit}}
\{\norm{L_{j,t}-\overline L_{j,t}}_{j,t}^{2}
+\norm{F_t-\overline F_t}_{j,t}^{2}\}
\le C\frac{\mathfrak r_{j,t}}{h_t}.
\]
This proves \eqref{kl:eq:variance-proxy-consistency}.  The deterministic
variance floor in Assumption~\ref{kl:ass:fit-regular} then implies
$\widetilde\chi^F_{j,t}-\chi^F_{j,t}\to0$.

The four coefficient differences in
\eqref{kl:eq:finite-bandwidth-coefficients} converge to zero in
$L^2(Q_{j,t,X})$.  The single-kernel integration inequality used in
Lemma~\ref{sg:lem:proxy-equivalence} therefore gives
\[
\sum_a\bm{E}_t^{\mathrm{fit}}
\left\|
\begin{pmatrix}
\widetilde\lambda_{\mu,j,a,t}-\lambda_{\mu,j,a,t}\\
\widetilde\lambda_{v,j,a,t}-\lambda_{v,j,a,t}
\end{pmatrix}
\right\|^2
=o(\mathfrak r_{j,t}).
\]
Also,
$\sum_a\bm{E}_t^{\mathrm{fit}}\lambda_{v,j,a,t}^{2}=O(\mathfrak r_{j,t})$.
Combining these bounds with
$\widetilde\chi^F_{j,t}-\chi^F_{j,t}\to0$ yields
\[
\sum_a\bm{E}_t^{\mathrm{fit}}
(\widetilde\lambda^C_{j,a,t}-\lambda^C_{j,a,t})^2
=o(\mathfrak r_{j,t}).
\]
Since $s_{C,\mathrm{fit},j,t}^{2}\ge c\mathfrak r_{j,t}$,
\eqref{kl:eq:certificate-proxy-equivalence} follows.
\end{proof}

Let $\widehat\lambda^C_{j,a,t}$ be
$\widetilde\lambda^C_{j,a,t}$ with all finite-bandwidth denominators, ratio
targets, $m^Q_{j,t}$, $\sigma_{Q,j,t}^{2}$, and $Q_{j,t,X}$-integrals
replaced by leave-one-out or finite-fold cross-fitted estimates.
For a fixed client coordinate, let $\gamma_{j,a,t}$ denote the sum of the
absolute local and federated coefficients with which observation $a$ enters
that coordinate, and put
$\mathfrak r_{j,t}=\sum_a\gamma_{j,a,t}^2$ and
$\gamma_{j,\max,t}=\max_a\gamma_{j,a,t}$.

\begin{assumption}[Cross-fitted certificate nuisance stability]
\label{kl:ass:primitive-nuisance}
For every fixed client block and every fitting observation $a$, let
$\widehat\eta^C_{j,a,t}$ be the nuisance estimate used in
$\widehat\lambda^C_{j,a,t}$, constructed without observation $a$.  There is an
event
\[
\mathcal H^C_{j,a,t}\in\sigma(\widehat\eta^C_{j,a,t})
\]
on which all estimated denominators and estimated versions of
$\widetilde v_{j,t}$ are bounded below and every required nuisance error is at
most a deterministic $a_t\downarrow0$ in the sup norm and in each
$Q_{j,t,X}$-integration norm appearing above.  The simultaneous event
\[
\mathcal H^C_{r,t}
=
\bigcap_{j\le r}\bigcap_{a\in\mathcal{I}_t}\mathcal H^C_{j,a,t}
\]
satisfies $\bm{P}(\mathcal H^C_{r,t})\to1$ for every fixed $r$.
Independent pilot information is used for $m^Q_{j,t}$,
$\sigma_{Q,j,t}^{2}$, and the evaluation-law integrals; the audit sample used
for gain inference is not reused for these fitting-side nuisances.
\end{assumption}

\begin{remark}[No universal primitive nuisance rate]
\label{kl:rem:no-universal-primitive-rate}
Assumption~\ref{kl:ass:primitive-nuisance} is intentionally high level.  A
uniform primitive rate depends on the smoothness order, the order of the pilot
kernel, the tails or truncation of the response and squared-response pilot
variables, and the empirical-process complexity of the observation-indexed
integration classes.  The paper therefore does not claim that a single
bandwidth formula verifies this assumption under the baseline polynomial
moment conditions.  Model-specific bounded, clipped, sub-exponential, or
truncated pilot constructions may be used to verify the assumption.
\end{remark}

\begin{lemma}[Aggregate certificate plug-in bound]
\label{kl:lem:primitive-plugin}
Under Assumptions~\ref{kl:ass:primitive-nuisance} and
\ref{kl:ass:fit-regular}, and the conditions of
Lemma~\ref{kl:lem:certificate-proxy}, for every fixed client
$j$,
\begin{equation}
\sum_{a\in\mathcal{I}_t}
(\widehat\lambda^C_{j,a,t}-\widetilde\lambda^C_{j,a,t})^2
=
O_{\bm{P}}(\mathfrak r_{j,t}a_t^2).
\label{kl:eq:primitive-plugin-bound}
\end{equation}
Consequently,
\begin{equation}
\frac{
\sum_a(\widehat\lambda^C_{j,a,t}-\lambda^C_{j,a,t})^2
}{s_{C,\mathrm{fit},j,t}^{2}}
\xrightarrow{\bm{P}}0.
\label{kl:eq:primitive-total-plugin}
\end{equation}
\end{lemma}

\begin{proof}
For each $a$, condition only on $\widehat\eta^C_{j,a,t}$.  It is independent
of observation $a$.  On $\mathcal H^C_{j,a,t}$, the denominator and variance
floors and the Lipschitz dependence of the integrated influence give
\[
\bm{E}_t^{\mathrm{fit}}
\left[
(\widehat\lambda^C_{j,a,t}-\widetilde\lambda^C_{j,a,t})^2
\mid\widehat\eta^C_{j,a,t}
\right]
\mathbf{1}\{\mathcal H^C_{j,a,t}\}
\le C\gamma_{j,a,t}^{2}a_t^2
\mathbf{1}\{\mathcal H^C_{j,a,t}\}.
\]
Apply the localization lemma, Lemma~\ref{sg:lem:localization}, with the
simultaneous event $\mathcal H^C_{r,t}$.  This proves
\eqref{kl:eq:primitive-plugin-bound} without imposing any bound on the
complement of the good event.  Equation
\eqref{kl:eq:primitive-total-plugin} follows from
Lemma~\ref{kl:lem:certificate-proxy}, the triangle inequality in
$\ell^2(\mathcal{I}_t)$, and certificate-contrast nondegeneracy.
\end{proof}

\begin{lemma}[Oracle certificate cross-covariance concentration]
\label{kl:lem:oracle-cross-covariance}
Under Assumption~\ref{kl:ass:fit-regular}, suppose that, for every fixed block
and some finite $C$,
\begin{equation}
\bm{E}_t^{\mathrm{fit}}|\lambda^C_{j,a,t}|^4
\le C\gamma_{j,a,t}^{4},
\qquad
\frac{\gamma_{j,\max,t}}{\sqrt{\mathfrak r_{j,t}}}
\longrightarrow0.
\label{kl:eq:certificate-fourth-bound}
\end{equation}
Then, for every fixed $j,\ell$,
\begin{equation}
\frac{
\sum_a\bm{E}_t^{\mathrm{fit}}
\{(\lambda^C_{j,a,t})^2(\lambda^C_{\ell,a,t})^2\}
}{s_{C,j,t}^{2}s_{C,\ell,t}^{2}}
\longrightarrow0.
\label{kl:eq:derived-oracle-cross-covariance}
\end{equation}
\end{lemma}

\begin{proof}
Cauchy--Schwarz and \eqref{kl:eq:certificate-fourth-bound} give
\[
\sum_a\bm{E}_t^{\mathrm{fit}}
\{(\lambda^C_{j,a,t})^2(\lambda^C_{\ell,a,t})^2\}
\le
C
\left(\sum_a\gamma_{j,a,t}^4\right)^{1/2}
\left(\sum_a\gamma_{\ell,a,t}^4\right)^{1/2}.
\]
Since
$\sum_a\gamma_{j,a,t}^4
\le\gamma_{j,\max,t}^{2}\mathfrak r_{j,t}$ and
$s_{C,j,t}^{2}\ge s_{C,\mathrm{fit},j,t}^{2}
\ge c\mathfrak r_{j,t}$, division by the total scales proves the result.
\end{proof}
\subsection{Finite-dimensional fitting limit}

Fix $r\ge1$.  For all sufficiently large $t$, $r\le K_t$.  Write
\[
\bm H^C_{r,t}
=(H^C_{1,t},\ldots,H^C_{r,t})^{\mathsf{T}}.
\]
For each client $j\le r$, use the certificate influences
\eqref{kl:eq:certificate-fit-influence} and the one-client remainder from
Corollary~\ref{kl:cor:one-client-fit}.  Put
\[
\bm\lambda^{C,(r)}_{a,t}
=
(\lambda^C_{1,a,t},\ldots,\lambda^C_{r,a,t})^{\mathsf{T}}.
\]

\begin{lemma}[Finite-block certificate fitting representation]
\label{kl:lem:fit-representation}
Suppose Theorem~\ref{kl:thm:mv-fitting-clt} and
Assumption~\ref{kl:ass:fit-regular} hold for every client in a fixed block.
Then the vectors $\bm\lambda^{C,(r)}_{a,t}$ are independent and centered, and
there is a remainder vector $\bm r^C_{r,t}$ such that
\begin{equation}
\bm H^C_{r,t}
=
\sum_{a\in\mathcal{I}_t}\bm\lambda^{C,(r)}_{a,t}
+
\bm r^C_{r,t}.
\label{kl:eq:fit-vector-representation}
\end{equation}
Moreover, if $D^C_{r,t}$ is the diagonal total scale defined below, then
\begin{equation}
\|(D^C_{r,t})^{-1}\bm r^C_{r,t}\|\xrightarrow{\bm{P}}0.
\label{kl:eq:block-remainder-derived}
\end{equation}
\end{lemma}

\begin{proof}
Apply Corollary~\ref{kl:cor:one-client-fit} simultaneously to the finitely many
coordinates.  The same fitting observation index $a$ is retained across
coordinates, so the common-server dependence is preserved.  For each coordinate,
$|r^C_{j,t}|/s_{C,j,t}\le |r^C_{j,t}|/s_{C,\mathrm{fit},j,t}\xrightarrow{\bm{P}}0$ because
the total scale is at least the fitting scale.  A finite sum of such
coordinatewise bounds is negligible in Euclidean norm.
\end{proof}

Define the fitting covariance matrix
\begin{equation}
\Sigma^{C,\mathrm{fit}}_{r,t}
=
\sum_{a\in\mathcal{I}_t}
\bm{E}_t^{\mathrm{fit}}
\left[
\bm\lambda^{C,(r)}_{a,t}
\bm\lambda^{C,(r)\mathsf{T}}_{a,t}
\right].
\label{kl:eq:fit-covariance}
\end{equation}
For client $j$, put
\begin{equation}
s_{C,\mathrm{fit},j,t}^{2}
=
\Sigma^{C,\mathrm{fit}}_{jj,t}.
\label{kl:eq:fit-scale}
\end{equation}

Let $W^0_{j,t}$ be the paired gain obtained by replacing
$(L_{j,t},F_t)$ with $(\overline L_{j,t},\overline F_t)$, and put
\begin{align}
\mu^0_{E,j,t}&=\bm{E}_{Q_{j,t}}W^0_{j,t},\\
v^0_{E,j,t}&=\Var_{Q_{j,t}}(W^0_{j,t}),\\
\chi^{E,0}_{j,t}&=\sqrt{\frac{\rho_{j,t}}{2v^0_{E,j,t}}},\\
\xi^0_{j,t}&=(W^0_{j,t}-\mu^0_{E,j,t})
-\chi^{E,0}_{j,t}
\{(W^0_{j,t}-\mu^0_{E,j,t})^2-v^0_{E,j,t}\},\\
\overline\omega_{C,j,t}^{2}&=\Var_{Q_{j,t}}(\xi^0_{j,t}).
\label{kl:eq:oracle-eval-variance}
\end{align}

\begin{lemma}[Certificate evaluation-variance stabilization]
\label{kl:lem:eval-stabilization}
Suppose, for every fixed client block, the conditional sixth moments in
Assumption~\ref{kl:ass:eval-regular} are bounded, the deterministic quantities
$v^0_{E,j,t}$ and $\overline\omega_{C,j,t}^{2}$ are bounded below, and
\begin{equation}
\sup_{j\le r,t}\bm{E}_{Q_{j,t}}|Y|^6<\infty.
\label{kl:eq:Y-sixth-moment}
\end{equation}
Also assume
\[
\norm{L_{j,t}-\overline L_{j,t}}_{L^6(Q_{j,t,X})}
+
\norm{F_t-\overline F_t}_{L^6(Q_{j,t,X})}
\xrightarrow{\bm{P}}0.
\]
Under the clipping condition, this displayed convergence follows from the
integrated second-moment bound whenever
$\mathfrak r_{j,t}/h_t\to0$.
Then
\begin{equation}
\frac{\omega_{C,j,t}^{2}}{\overline\omega_{C,j,t}^{2}}\xrightarrow{\bm{P}}1.
\label{kl:eq:eval-stabilization-ratio}
\end{equation}
\end{lemma}

\begin{proof}
Under clipping, the paired-gain map is Lipschitz in $(L,F)$ with envelope
$C(1+|Y|)$.  The stated $L^6$ convergence and the conditional sixth-moment
bound imply convergence, in probability, of the conditional moments of
$W_{j,t}$ through order four to those of $W^0_{j,t}$.  In particular,
$v_{j,t}-v^0_{E,j,t}\xrightarrow{\bm{P}}0$.  The deterministic variance floor therefore gives
$\chi^E_{j,t}-\chi^{E,0}_{j,t}\xrightarrow{\bm{P}}0$.

Because $\xi$ is a quadratic polynomial in the centered gain with coefficients
that remain bounded on the variance floor, the same moment convergence yields
\[
\bm{E}_{Q_{j,t}}
\{(\xi_{j,t,1}-\xi^0_{j,t})^2\mid\mathcal{F}_t^{\mathrm{fit}}\}
\xrightarrow{\bm{P}}0.
\]
Consequently, both the conditional means and second moments of
$\xi_{j,t,1}$ converge to those of $\xi^0_{j,t}$.  Since the former mean is
zero and the deterministic variance
$\overline\omega_{C,j,t}^{2}$ is bounded below,
$\omega_{C,j,t}^{2}/\overline\omega_{C,j,t}^{2}\xrightarrow{\bm{P}}1$.
\end{proof}

Define
\begin{equation}
s_{C,\mathrm{eval},j,t}^{2}
=
\frac{\overline\omega_{C,j,t}^{2}}{N_{j,t}^{\mathrm{eval}}}.
\label{kl:eq:eval-stabilization}
\end{equation}
Set
\begin{equation}
s_{C,j,t}^{2}
=
s_{C,\mathrm{fit},j,t}^{2}
+
s_{C,\mathrm{eval},j,t}^{2},
\qquad
D^C_{r,t}
=
\operatorname{diag}(s_{C,1,t},\ldots,s_{C,r,t}).
\label{kl:eq:total-scales}
\end{equation}
Define
\begin{equation}
A^C_{r,t}
=
(D^C_{r,t})^{-1}
\Sigma^{C,\mathrm{fit}}_{r,t}
(D^C_{r,t})^{-1}.
\label{kl:eq:A-matrix}
\end{equation}

\begin{assumption}[Finite-block covariance and Lyapunov conditions]
\label{kl:ass:fit-vector-clt}
For every fixed $r$, there is a symmetric nonnegative definite matrix
$A^C_r$ such that
\begin{equation}
A^C_{r,t}\longrightarrow A^C_r.
\label{kl:eq:A-convergence}
\end{equation}
For some $\delta>0$,
\begin{equation}
\sum_{a\in\mathcal{I}_t}
\bm{E}_t^{\mathrm{fit}}
\left\|
(D^C_{r,t})^{-1}\bm\lambda^{C,(r)}_{a,t}
\right\|^{2+\delta}
\longrightarrow0.
\label{kl:eq:vector-lyapunov}
\end{equation}
\end{assumption}

\begin{theorem}[Finite-block fitting CLT for KL certificates]
\label{kl:thm:finite-fit-clt}
Under Lemma~\ref{kl:lem:fit-representation} and
Assumption~\ref{kl:ass:fit-vector-clt},
\begin{equation}
(D^C_{r,t})^{-1}\bm H^C_{r,t}
\Longrightarrow
N_r(\bm0,A^C_r).
\label{kl:eq:finite-fit-clt}
\end{equation}
\end{theorem}

\begin{proof}
Fix $\bm u\in\mathbb{R}^r$.  By \eqref{kl:eq:fit-vector-representation},
\[
\bm u^{\mathsf{T}}(D^C_{r,t})^{-1}\bm H^C_{r,t}
=
\sum_{a\in\mathcal{I}_t}
\bm u^{\mathsf{T}}(D^C_{r,t})^{-1}\bm\lambda^{C,(r)}_{a,t}
+
o_{\bm{P}}(1).
\]
The scalar summands are independent and centered.  Their variance converges
to $\bm u^{\mathsf{T}}A^C_r\bm u$, and
\eqref{kl:eq:vector-lyapunov} gives their Lyapunov condition.  The scalar
Lyapunov theorem and Cram\'er--Wold yield \eqref{kl:eq:finite-fit-clt}.
\end{proof}

\subsection{Finite-dimensional conditional evaluation limit}

Let
\[
\bm E^C_{r,t}=(E^C_{1,t},\ldots,E^C_{r,t})^{\mathsf{T}}.
\]
Define
\begin{equation}
B^C_{r,t}
=
\operatorname{diag}\left(
\frac{s_{C,\mathrm{eval},1,t}^{2}}{s_{C,1,t}^{2}},\ldots,
\frac{s_{C,\mathrm{eval},r,t}^{2}}{s_{C,r,t}^{2}}
\right).
\label{kl:eq:B-matrix}
\end{equation}

\begin{assumption}[Evaluation variance stabilization]
\label{kl:ass:eval-vector}
For every fixed $r$, $B^C_{r,t}\to B^C_r$ for a diagonal nonnegative
definite matrix $B^C_r$, and the conditions of
Lemma~\ref{kl:lem:eval-stabilization} hold for every $j\le r$.
If $s_{C,\mathrm{eval},j,t}/s_{C,j,t}\to0$, the corresponding coordinate is
interpreted as degenerating at zero after total standardization.
\end{assumption}

\begin{theorem}[Finite-block conditional evaluation limit]
\label{kl:thm:finite-eval-clt}
Under Assumptions~\ref{kl:ass:eval-regular} and
\ref{kl:ass:eval-vector},
\begin{equation}
\mathcal{L}\left(
\left.
(D^C_{r,t})^{-1}\bm E^C_{r,t}
\right|
\mathcal{F}_t^{\mathrm{fit}}
\right)
\overset{\bm{P}}{\Longrightarrow}
N_r(\bm0,B^C_r).
\label{kl:eq:finite-eval-clt}
\end{equation}
\end{theorem}

\begin{proof}
Conditional on $\mathcal{F}_t^{\mathrm{fit}}$, the $r$ evaluation samples are
independent.  For $\bm v\in\mathbb{R}^r$,
\begin{align*}
&\bm{E}\left[
\left.
\exp\{i\bm v^{\mathsf{T}}(D^C_{r,t})^{-1}\bm E^C_{r,t}\}
\right|
\mathcal{F}_t^{\mathrm{fit}}
\right]\\
&\qquad=
\prod_{j=1}^{r}
\bm{E}\left[
\left.
\exp\{i v_j E^C_{j,t}/s_{C,j,t}\}
\right|
\mathcal{F}_t^{\mathrm{fit}}
\right].
\end{align*}
Write $b^C_{j,t}=s_{C,\mathrm{eval},j,t}/s_{C,j,t}$.  By
Theorem~\ref{kl:thm:one-client-eval} and
$\omega_{C,j,t}^2/\overline\omega_{C,j,t}^2\xrightarrow{\bm{P}}1$, the $j$th factor converges in
probability to $\exp\{-v_j^2(b^C_j)^2/2\}$ whenever
$b^C_{j,t}\to b^C_j$.  Since $r$ is
fixed, the product converges in probability to
$\exp(-\bm v^{\mathsf{T}}B^C_r\bm v/2)$.  Sweeting's random-measure
characterization on $\mathbb{R}^r$ gives \eqref{kl:eq:finite-eval-clt}.
\end{proof}

\subsection{Finite-dimensional certificate theorem}

Define
\begin{equation}
R^C_{r,t}=A^C_{r,t}+B^C_{r,t},
\qquad
R^C_r=A^C_r+B^C_r.
\label{kl:eq:R-matrices}
\end{equation}
The diagonal entries of $R^C_{r,t}$ equal one; hence $R^C_r$ is a correlation
matrix, possibly singular.

\begin{lemma}[Asymptotic independence]
\label{kl:lem:asymptotic-independence}
Under the assumptions of Theorems~\ref{kl:thm:finite-fit-clt} and
\ref{kl:thm:finite-eval-clt},
\begin{equation}
\left(
(D^C_{r,t})^{-1}\bm H^C_{r,t},
(D^C_{r,t})^{-1}\bm E^C_{r,t}
\right)
\Longrightarrow
(\bm Z^C_{F,r},\bm Z^C_{E,r}),
\label{kl:eq:joint-block-limit}
\end{equation}
where
\[
\bm Z^C_{F,r}\sim N_r(\bm0,A^C_r),
\qquad
\bm Z^C_{E,r}\sim N_r(\bm0,B^C_r),
\]
and the two Gaussian vectors are independent.
\end{lemma}

\begin{proof}
For $\bm u,\bm v\in\mathbb{R}^r$, use that $\bm H^C_{r,t}$ is
$\mathcal{F}_t^{\mathrm{fit}}$-measurable:
\begin{align*}
&\bm{E}\exp\left\{
 i\bm u^{\mathsf{T}}(D^C_{r,t})^{-1}\bm H^C_{r,t}
+i\bm v^{\mathsf{T}}(D^C_{r,t})^{-1}\bm E^C_{r,t}
\right\}\\
&\quad=
\bm{E}\left[
\exp\left\{i\bm u^{\mathsf{T}}(D^C_{r,t})^{-1}\bm H^C_{r,t}\right\}
\bm{E}\left[
\left.
\exp\left\{i\bm v^{\mathsf{T}}(D^C_{r,t})^{-1}\bm E^C_{r,t}\right\}
\right|
\mathcal{F}_t^{\mathrm{fit}}
\right]
\right].
\end{align*}
The inner conditional characteristic function converges in probability to
$\exp(-\bm v^{\mathsf{T}}B^C_r\bm v/2)$ and is bounded by one, so convergence
also holds in $L^1$.  The fitting CLT then gives the product limiting
characteristic function, proving \eqref{kl:eq:joint-block-limit}.
\end{proof}

\begin{theorem}[Finite-dimensional KL-certificate CLT]
\label{kl:thm:finite-certificate-clt}
Under the preceding assumptions,
\begin{equation}
(D^C_{r,t})^{-1}
\left(
\widehat{\bm C}_{r,t}-\bm{\mathcal C}_{r,t}
\right)
\Longrightarrow
N_r(\bm0,R^C_r),
\label{kl:eq:finite-certificate-clt}
\end{equation}
where
\[
\widehat{\bm C}_{r,t}
=(\widehat C_{1,t},\ldots,\widehat C_{r,t})^{\mathsf{T}},
\qquad
\bm{\mathcal C}_{r,t}
=(\mathcal C_{1,t},\ldots,\mathcal C_{r,t})^{\mathsf{T}}.
\]
\end{theorem}

\begin{proof}
Apply the two-stage client-array assembly theorem,
Theorem~\ref{gen:thm:two-stage-assembly}, with
$\bm H_{r,t}=\bm H^C_{r,t}$,
$\bm E_{r,t}=\bm E^C_{r,t}$,
$A_r=A^C_r$, and $B_r=B^C_r$.
\end{proof}

\subsection{Consistent finite-block covariance estimation}

Let $\widehat\lambda^C_{j,a,t}$ be a cross-fitted feasible estimate of
\eqref{kl:eq:certificate-fit-influence}, and define
\begin{equation}
\widehat\Sigma^{C,\mathrm{fit}}_{j\ell,t}
=
\sum_{a\in\mathcal{I}_t}
\widehat\lambda^C_{j,a,t}
\widehat\lambda^C_{\ell,a,t}.
\label{kl:eq:fit-cross-cov-estimator}
\end{equation}
For the evaluation component, use
$\widehat s_{C,\mathrm{eval},j,t}^{2}$ from
\eqref{kl:eq:eval-var-estimator}.
Define
\begin{align}
\widehat\Sigma^C_{r,t}
&=
\left(
\widehat\Sigma^{C,\mathrm{fit}}_{j\ell,t}
\right)_{1\le j,\ell\le r}
+
\operatorname{diag}\left(
\widehat s_{C,\mathrm{eval},1,t}^{2},\ldots,
\widehat s_{C,\mathrm{eval},r,t}^{2}
\right),
\label{kl:eq:total-cov-estimator}\\
\widehat D^C_{r,t}
&=
\operatorname{diag}\left(
(\widehat\Sigma^C_{11,t})^{1/2},\ldots,
(\widehat\Sigma^C_{rr,t})^{1/2}
\right),
\label{kl:eq:Dhat}\\
\widehat R^C_{r,t}
&=
(\widehat D^C_{r,t})^{-1}
\widehat\Sigma^C_{r,t}
(\widehat D^C_{r,t})^{-1}.
\label{kl:eq:Rhat}
\end{align}

\begin{lemma}[Primitive certificate covariance conditions]
\label{kl:lem:cov-estimation}
Fix $r$.  Suppose Assumptions~\ref{sg:ass:kernel}--\ref{sg:ass:rates},
\ref{kl:ass:mv-fitting}, \ref{kl:ass:fit-regular},
\ref{kl:ass:primitive-nuisance}, \ref{kl:ass:eval-regular}, and
\ref{kl:ass:eval-vector} hold uniformly over $j\le r$; suppose also that the
finite-bandwidth proxy conditions of Lemma~\ref{kl:lem:certificate-proxy}
and the coordinatewise no-dominant-observation conditions hold.
Then, for every $j,\ell\le r$,
\begin{align}
&
\frac{
\sum_{a\in\mathcal{I}_t}
\bm{E}_t^{\mathrm{fit}}
\{(\lambda^C_{j,a,t})^{2}(\lambda^C_{\ell,a,t})^{2}\}
}{
s_{C,j,t}^{2}s_{C,\ell,t}^{2}
}
\longrightarrow0,
\label{kl:eq:oracle-cross-cov-condition}\\
&
\frac{
\sum_{a\in\mathcal{I}_t}
(\widehat\lambda^C_{j,a,t}-\lambda^C_{j,a,t})^{2}
}{s_{C,j,t}^{2}}
\xrightarrow{\bm{P}}0,
\label{kl:eq:plugin-condition}\\
&
\frac{
\widehat s_{C,\mathrm{eval},j,t}^{2}
}{s_{C,\mathrm{eval},j,t}^{2}}
\xrightarrow{\bm{P}}1.
\label{kl:eq:eval-var-consistency}
\end{align}
If $s_{C,\mathrm{eval},j,t}^{2}/s_{C,j,t}^{2}\to0$, the last display may be
replaced by
$|\widehat s_{C,\mathrm{eval},j,t}^{2}-s_{C,\mathrm{eval},j,t}^{2}|/
 s_{C,j,t}^{2}\xrightarrow{\bm{P}}0$.
\end{lemma}

\begin{proof}
The bounded coefficient functions, the $4+\delta$ fitting moment, and the
single-kernel integration bound imply
$\bm{E}_t^{\mathrm{fit}}|\lambda^C_{j,a,t}|^4
\le C\gamma_{j,a,t}^4$ uniformly over every fixed block.  Lemma~\ref{kl:lem:oracle-cross-covariance}
therefore proves \eqref{kl:eq:oracle-cross-cov-condition}.  Lemma~\ref{kl:lem:primitive-plugin}
proves \eqref{kl:eq:plugin-condition}.  Finally,
Corollary~\ref{kl:cor:eval-studentization} and
Lemma~\ref{kl:lem:eval-stabilization} give
\eqref{kl:eq:eval-var-consistency}; when the evaluation contribution is
negligible on the total scale, the stated absolute version follows directly.
\end{proof}
\begin{theorem}[Finite-block covariance approximation and studentization]
\label{kl:thm:cov-consistency}
Under Lemma~\ref{kl:lem:cov-estimation}, for every fixed $r$,
\begin{equation}
\left\|
(D^C_{r,t})^{-1}\widehat\Sigma^C_{r,t}(D^C_{r,t})^{-1}
-R^C_{r,t}
\right\|
\xrightarrow{\bm{P}}0,
\qquad
\widehat D^C_{r,t}(D^C_{r,t})^{-1}\xrightarrow{\bm{P}} I_r.
\label{kl:eq:stage-covariance-approximation}
\end{equation}
If, in addition, $R^C_{r,t}\to R^C_r$, then
\begin{equation}
\widehat R^C_{r,t}\xrightarrow{\bm{P}} R^C_r.
\label{kl:eq:R-consistency}
\end{equation}
If the finite-dimensional certificate CLT,
Theorem~\ref{kl:thm:finite-certificate-clt}, also holds, then
\begin{equation}
(\widehat D^C_{r,t})^{-1}
(\widehat{\bm C}_{r,t}-\bm{\mathcal C}_{r,t})
\Longrightarrow
N_r(\bm0,R^C_r).
\label{kl:eq:studentized-vector-clt}
\end{equation}
\end{theorem}

\begin{proof}
Define the oracle empirical fitting covariance
\[
\widetilde\Sigma^{C,\mathrm{fit}}_{j\ell,t}
=
\sum_a\lambda^C_{j,a,t}\lambda^C_{\ell,a,t}.
\]
It is unbiased for $\Sigma^{C,\mathrm{fit}}_{j\ell,t}$.  Independence across
$a$ and \eqref{kl:eq:oracle-cross-cov-condition} give
\[
\frac{
\widetilde\Sigma^{C,\mathrm{fit}}_{j\ell,t}
-
\Sigma^{C,\mathrm{fit}}_{j\ell,t}
}{s_{C,j,t}s_{C,\ell,t}}
\xrightarrow{\bm{P}}0.
\]
The plug-in difference is controlled by Cauchy--Schwarz:
\begin{align*}
&\left|
\sum_a\widehat\lambda^C_{j,a,t}\widehat\lambda^C_{\ell,a,t}
-
\sum_a\lambda^C_{j,a,t}\lambda^C_{\ell,a,t}
\right|\\
&\quad\le
\norm{\widehat{\bm\lambda}^C_{j,t}-\bm\lambda^C_{j,t}}_2
\norm{\widehat{\bm\lambda}^C_{\ell,t}}_2
+
\norm{\bm\lambda^C_{j,t}}_2
\norm{\widehat{\bm\lambda}^C_{\ell,t}-\bm\lambda^C_{\ell,t}}_2.
\end{align*}
Lemma~\ref{kl:lem:cov-estimation} therefore gives consistency of every fixed
fitting covariance entry relative to its stage-$t$ scale.  The evaluation
covariance is diagonal and consistently estimated by
\eqref{kl:eq:eval-var-estimator}.  This proves
\eqref{kl:eq:stage-covariance-approximation}.  The fixed-limit conclusion
then follows from $R^C_{r,t}\to R^C_r$, and the studentized CLT follows from
Theorem~\ref{kl:thm:finite-certificate-clt} and multivariate Slutsky.
\end{proof}

\subsection{The client-indexed certificate process}

Let
\[
\mathbb{R}^\infty=\prod_{j=1}^{\infty}\mathbb{R}
\]
with the product topology, metrized by
\begin{equation}
d(\bm x,\bm y)
=
\sum_{j=1}^{\infty}2^{-j}(|x_j-y_j|\wedge1).
\label{kl:eq:product-metric}
\end{equation}
Define
\begin{equation}
T^C_{j,t}
=
\begin{cases}
(\widehat C_{j,t}-\mathcal C_{j,t})/s_{C,j,t},&j\le K_t,\\[1.2ex]
0,&j>K_t,
\end{cases}
\qquad
\bm T^C_t=(T^C_{1,t},T^C_{2,t},\ldots).
\label{kl:eq:certificate-process}
\end{equation}

\begin{assumption}[Compatible covariance limits]
\label{kl:ass:compatible}
For every fixed $r$, the matrices $R^C_{r,t}$ converge to $R^C_r$, and the
limits are compatible:
\begin{equation}
R^C_{r+1}[1{:}r,1{:}r]=R^C_r.
\label{kl:eq:compatibility}
\end{equation}
\end{assumption}

By compatibility, there exists a centered Gaussian random sequence
\[
\bm Z^C=(Z^C_1,Z^C_2,\ldots)
\]
whose first $r$ coordinates have covariance $R^C_r$ for every $r$.

\begin{theorem}[Client-indexed KL-certificate process CLT]
\label{kl:thm:process-clt}
Under the assumptions of Theorem~\ref{kl:thm:finite-certificate-clt} for every
fixed client block and Assumption~\ref{kl:ass:compatible},
\begin{equation}
\bm T^C_t\Longrightarrow\bm Z^C
\qquad
\text{in }\mathbb{R}^\infty
\text{ with the product topology}.
\label{kl:eq:process-clt}
\end{equation}
If, in addition, the conditions of
Theorem~\ref{kl:thm:cov-consistency} hold for every fixed block, the same
conclusion holds with $s_{C,j,t}$ replaced coordinatewise by the consistent
studentizing scales.
\end{theorem}

\begin{proof}
Every fixed finite block converges by
Theorem~\ref{kl:thm:finite-certificate-clt}.  For each fixed $j$, the family of
marginal laws $\{\mathcal{L}(T^C_{j,t}):t\ge1\}$ is tight.  Given $\eta>0$, choose
compact intervals $I_j\subset\mathbb{R}$ such that
\[
\sup_t\bm{P}\{T^C_{j,t}\notin I_j\}<\eta2^{-j}.
\]
Then $I=\prod_{j\ge1}I_j$ is compact in the product topology and
\[
\sup_t\bm{P}\{\bm T^C_t\notin I\}
\le
\sum_{j\ge1}\sup_t\bm{P}\{T^C_{j,t}\notin I_j\}
<\eta.
\]
Thus the laws are tight.  Every subsequential limit has the compatible
Gaussian finite-dimensional distributions $\{N_r(0,R^C_r)\}$, so the limit is
unique and equals the law of $\bm Z^C$.
\end{proof}

\subsection{Robust-positive certificate probabilities}

Let $r$ be fixed and let $\varepsilon_{j,t}$ be deterministic thresholds.
Define
\begin{equation}
\gamma^C_{j,t}
=
\frac{\mathcal C_{j,t}-\varepsilon_{j,t}}{s_{C,j,t}},
\qquad j=1,\ldots,r.
\label{kl:eq:margins}
\end{equation}

\begin{corollary}[Finite-block robust-positive probability]
\label{kl:cor:finite-positive}
Suppose
\[
\gamma^C_{j,t}\longrightarrow\gamma^C_j,
\qquad j=1,\ldots,r.
\]
Then
\begin{align}
&\bm{P}\left\{
\widehat C_{j,t}>\varepsilon_{j,t}
\text{ for }j=1,\ldots,r
\right\}
\nonumber\\
&\qquad\longrightarrow
\bm{P}\left\{
Z^C_j>-\gamma^C_j
\text{ for }j=1,\ldots,r
\right\}.
\label{kl:eq:finite-orthant}
\end{align}
For $\varepsilon_{j,t}=0$, this is the limiting probability that every
empirical variance-sensitive KL certificate in the fixed block is positive.
\end{corollary}

\begin{proof}
The event is
$\{T^C_{j,t}>-\gamma^C_{j,t},\ j=1,\ldots,r\}$.  Its limiting boundary is a
finite union of coordinate hyperplanes.  Every $Z^C_j$ has variance one, so
the boundary has zero Gaussian probability.  Apply the finite-dimensional
CLT and the portmanteau theorem.
\end{proof}

\begin{remark}[Interpretation]
The event $C_{j,t}^{\mathrm{fit}}>0$ is pathwise sufficient for positivity of
the exact fitted KL-robust gain by Proposition~\ref{kl:prop:certificate}.  The
quantity $\mathcal C_{j,t}$ is instead the repeated-fitting mean of the fitted
certificate.  Thus the process theorem describes the sampling behavior of
empirical certificates around their stage-$t$ marginal centers; it does not
turn positivity of $\mathcal C_{j,t}$ into an almost-sure pathwise robust
guarantee for every fitted sample.
\end{remark}

For the moving event involving every $j\le K_t$, product-topology convergence
alone is insufficient.  Define
\begin{equation}
\delta^C_{j,t}
=
\sup_{x\in\mathbb{R}}
\left|
\bm{P}(T^C_{j,t}\le x)-\Phi(x)
\right|.
\label{kl:eq:coordinate-error}
\end{equation}

\begin{corollary}[All-client certificate positivity by summability]
\label{kl:cor:all-client}
For deterministic thresholds $\varepsilon_{j,t}$, put
\[
\gamma^C_{j,t}
=
\frac{\mathcal C_{j,t}-\varepsilon_{j,t}}{s_{C,j,t}}.
\]
Then
\begin{align}
&\bm{P}\left\{
\widehat C_{j,t}>\varepsilon_{j,t}
\text{ for every }j\le K_t
\right\}
\nonumber\\
&\quad\ge
1-
\sum_{j=1}^{K_t}
\left\{
\Phi(-\gamma^C_{j,t})+\delta^C_{j,t}
\right\}.
\label{kl:eq:all-client-lower-bound}
\end{align}
Consequently, if
\begin{equation}
\sum_{j=1}^{K_t}\Phi(-\gamma^C_{j,t})
+
\sum_{j=1}^{K_t}\delta^C_{j,t}
\longrightarrow0,
\label{kl:eq:summability}
\end{equation}
then
\begin{equation}
\bm{P}\left\{
\widehat C_{j,t}>\varepsilon_{j,t}
\text{ for every }j\le K_t
\right\}
\longrightarrow1.
\label{kl:eq:all-client-one}
\end{equation}
\end{corollary}

\begin{proof}
By the union bound,
\begin{align*}
&1-
\bm{P}\left\{
\widehat C_{j,t}>\varepsilon_{j,t}
\text{ for every }j\le K_t
\right\}\\
&\qquad\le
\sum_{j=1}^{K_t}
\bm{P}\{T^C_{j,t}\le-\gamma^C_{j,t}\}\\
&\qquad\le
\sum_{j=1}^{K_t}
\{\Phi(-\gamma^C_{j,t})+\delta^C_{j,t}\}.
\end{align*}
\end{proof}

\subsection{Interpretation and scope}

The covariance decomposition
\[
R^C_r=A^C_r+B^C_r
\]
separates fitting dependence from evaluation uncertainty.  The matrix $A^C_r$
is generally non-diagonal because the same federated predictor enters every
client certificate and because each client's certificate depends jointly on
the fitted gain mean and fitted gain variance.  The matrix $B^C_r$ is diagonal
because the fresh client evaluation streams are independent.

The theorem allows client-specific radii $\rho_{j,t}$ and sub-gamma scales
$c_{j,t}$, including local robustness radii that shrink with the client
information scale.  It does not require a common fixed radius across the
growing federation.

The result concerns the variance-sensitive lower certificate, not the exact
KL dual value.  Under Assumption~\ref{kl:ass:subgamma}, certificate positivity is
sufficient for positivity of the exact fitted robust gain.  Inference for the
exact dual value would require an additional envelope theorem for the random
log-Laplace optimization problem.

The certificate-process theorem is a regular positive-gain-variance result.
The exact robust value and the lower certificate remain well defined at
$v_{j,t}=0$, but the map $v\mapsto\sqrt{2\rho v}$ is not differentiable there
when $\rho>0$.  At an exact-model or locally exact boundary the leading
certificate fluctuation may therefore be a norm or absolute-value functional
of the first-order fitting error.  The nominal quadratic-gain theorem does
not by itself provide this missing scalar boundary input.  The finite-block
assembly can be reused only after a separate nonsmooth one-client theorem is
established.  Local radius calibration, as in
Proposition~\ref{kl:prop:certified-radius}, does not by itself imply a
Gaussian boundary limit.

\section{Client-indexed $L_2$-robust gain process: detailed assumptions and proofs}
\label{app:l2-process-proof}

\subsection{Fitted $L_2$-robust value}

At stage $t$, let $K_t$ clients participate.  Client $j\le K_t$ has a fitted
local predictor $L_{j,t}$, and all clients share the fitted federated predictor
$F_t$.  The fitting samples generate the sigma-field
\[
\mathcal{F}_t^{\mathrm{fit}}.
\]
Fix a prescribed local weight $\alpha\in[0,1)$ and put
\[
\vartheta=1-\alpha.
\]
Let $Q_{j,t}$ be the evaluation law for client $j$, with covariate marginal
$Q_{j,t,X}$ and regression function
\[
m_{j,t}^{0}(x)=\bm{E}_{Q_{j,t}}[Y\mid X=x].
\]
For $f,g\in L^2(Q_{j,t,X})$, write
\[
\inner{f}{g}_{j,t}
=
\int f(x)g(x)Q_{j,t,X}(\,d x),
\qquad
\norm{f}_{j,t}^{2}=\inner{f}{f}_{j,t}.
\]
The personalized predictor and borrowing direction are
\begin{align}
M_{j,t}(x)
&=
\alpha L_{j,t}(x)+\vartheta F_t(x),
\label{l2:eq:personalized}\\
d^{B}_{j,t}(x)
&=
M_{j,t}(x)-L_{j,t}(x)
=
\vartheta\{F_t(x)-L_{j,t}(x)\}.
\label{l2:eq:borrowing-direction}
\end{align}
For $z=(x,y)$, define the paired predictive gain
\begin{equation}
W_{j,t}(z)
=
\{y-L_{j,t}(x)\}^{2}-\{y-M_{j,t}(x)\}^{2}.
\label{l2:eq:paired-gain}
\end{equation}
Its fitted conditional mean is
\begin{equation}
G_{j,t}^{\mathrm{fit}}
=
\int W_{j,t}(z)Q_{j,t}(\,d z)
=
\bm{E}\{W_{j,t}(Z)\mid\mathcal{F}_t^{\mathrm{fit}}\}.
\label{l2:eq:fitted-gain}
\end{equation}

For a candidate regression function $h\in L^2(Q_{j,t,X})$, put
\begin{equation}
D_{j,t}(h)
=
\int
\left[
\{h(x)-L_{j,t}(x)\}^{2}
-
\{h(x)-M_{j,t}(x)\}^{2}
\right]Q_{j,t,X}(\,d x).
\label{l2:eq:function-gain}
\end{equation}
Because the conditional response variance cancels,
\[
G_{j,t}^{\mathrm{fit}}=D_{j,t}(m_{j,t}^{0}).
\]
Let $\tau_{j,t}\ge0$ be deterministic and define the fitted function-level
robust value
\begin{equation}
\mathcal R_{j,t}^{\mathrm{fit}}
=
\inf_{\norm{h-m_{j,t}^{0}}_{j,t}\le\tau_{j,t}}
D_{j,t}(h).
\label{l2:eq:fitted-robust-value-definition}
\end{equation}

\begin{proposition}[Exact least-favorable regression function]
\label{l2:prop:exact-lf}
For every fitted pair $(L_{j,t},F_t)$,
\begin{equation}
\mathcal R_{j,t}^{\mathrm{fit}}
=
G_{j,t}^{\mathrm{fit}}
-
2\tau_{j,t}\norm{d^{B}_{j,t}}_{j,t}.
\label{l2:eq:exact-robust-value}
\end{equation}
If $\norm{d^{B}_{j,t}}_{j,t}>0$, the unique least-favorable direction is
\begin{equation}
m_{j,t}^{\circ}
=
m_{j,t}^{0}
-
\tau_{j,t}
\frac{d^{B}_{j,t}}{\norm{d^{B}_{j,t}}_{j,t}}.
\label{l2:eq:least-favorable-function}
\end{equation}
If $d^{B}_{j,t}=0$ in $L^2(Q_{j,t,X})$, the local and personalized predictors
coincide and the robust value is zero.
\end{proposition}

\begin{proof}
Using $M_{j,t}=L_{j,t}+d^{B}_{j,t}$,
\[
D_{j,t}(h)
=
2\inner{h}{d^{B}_{j,t}}_{j,t}
+\norm{L_{j,t}}_{j,t}^{2}
-\norm{M_{j,t}}_{j,t}^{2}.
\]
Hence
\[
D_{j,t}(h)-D_{j,t}(m_{j,t}^{0})
=
2\inner{h-m_{j,t}^{0}}{d^{B}_{j,t}}_{j,t}.
\]
Cauchy--Schwarz gives
\[
\inf_{\norm{h-m_{j,t}^{0}}_{j,t}\le\tau_{j,t}}
\inner{h-m_{j,t}^{0}}{d^{B}_{j,t}}_{j,t}
=
-\tau_{j,t}\norm{d^{B}_{j,t}}_{j,t},
\]
with optimizer \eqref{l2:eq:least-favorable-function} when the direction is
nonzero.  This proves \eqref{l2:eq:exact-robust-value}.
\end{proof}

Let
\begin{equation}
\mathfrak R_{j,t}
=
\bm{E}_t^{\mathrm{fit}}\mathcal R_{j,t}^{\mathrm{fit}}
\label{l2:eq:marginal-robust-value}
\end{equation}
be the deterministic stage-$t$ marginal robust value, where
$\bm{E}_t^{\mathrm{fit}}$ denotes expectation under the joint law of all
stage-$t$ fitting samples.

\subsection{Persistent-direction fitting expansion}

Define the mean fitted curves
\[
\overline L_{j,t}
=
\bm{E}_t^{\mathrm{fit}}L_{j,t},
\qquad
\overline F_t
=
\bm{E}_t^{\mathrm{fit}}F_t,
\]
and the deterministic borrowing direction
\begin{equation}
\overline d^{B}_{j,t}
=
\vartheta\{\overline F_t-\overline L_{j,t}\}.
\label{l2:eq:mean-borrowing-direction}
\end{equation}
Write
\[
e_{L,j,t}=L_{j,t}-\overline L_{j,t},
\qquad
e_{F,t}=F_t-\overline F_t,
\qquad
e_{B,j,t}=d^{B}_{j,t}-\overline d^{B}_{j,t}
=
\vartheta(e_{F,t}-e_{L,j,t}).
\]

The nonvanishing mean-direction condition is Assumption~\ref{l2:ass:persistent-direction}, stated in Section~\ref{sec:l2-robust}.

\begin{lemma}[Scalar expansion of the borrowing norm]
\label{l2:lem:norm-expansion}
On the event
$\norm{e_{B,j,t}}_{j,t}\le\norm{\overline d^{B}_{j,t}}_{j,t}/2$,
\begin{equation}
\norm{d^{B}_{j,t}}_{j,t}
-\norm{\overline d^{B}_{j,t}}_{j,t}
=
\frac{
\inner{\overline d^{B}_{j,t}}{e_{B,j,t}}_{j,t}
}{\norm{\overline d^{B}_{j,t}}_{j,t}
}
+
r^{V}_{j,t},
\label{l2:eq:norm-linearization}
\end{equation}
where
\begin{equation}
|r^{V}_{j,t}|
\le
\frac{2\norm{e_{B,j,t}}_{j,t}^{2}}
{\norm{\overline d^{B}_{j,t}}_{j,t}}.
\label{l2:eq:norm-remainder}
\end{equation}
\end{lemma}

\begin{proof}
Let $u=\overline d^{B}_{j,t}$ and $e=e_{B,j,t}$.  Since
\[
\norm{u+e}-\norm{u}
=
\frac{2\inner{u}{e}+\norm{e}^{2}}
{\norm{u+e}+\norm{u}},
\]
subtracting $\inner{u}{e}/\norm{u}$ and simplifying gives a remainder
bounded by $2\norm{e}^{2}/\norm{u}$ whenever
$\norm{e}\le\norm{u}/2$.
\end{proof}

The nominal gain theorem supplies scalar observation-level fitting influences.
The next assumption records only the input needed here.

\begin{assumption}[Nominal gain and predictor-error representations]
\label{l2:ass:nominal-input}
For each fixed client $j$, there are mutually independent fitting observations
indexed by $a\in\mathcal{I}_t$, centered scalar influences $\lambda^G_{j,a,t}$,
centered function-valued influences
$\phi_{j,a,L,t},\phi_{a,F,t}\in L^2(Q_{j,t,X})$, and remainders such that
\begin{align}
G_{j,t}^{\mathrm{fit}}
-
\Gamma_{j,t}
&=
\sum_{a\in\mathcal{I}_t}\lambda^G_{j,a,t}
+
r^G_{j,t},
\label{l2:eq:nominal-gain-representation}\\
e_{L,j,t}
&=
\sum_{a\in\mathcal{I}_t}\phi_{j,a,L,t}
+
r_{j,L,t},
\label{l2:eq:local-error-representation}\\
e_{F,t}
&=
\sum_{a\in\mathcal{I}_t}\phi_{a,F,t}
+
r_{F,t}.
\label{l2:eq:server-error-representation}
\end{align}
Here $\Gamma_{j,t}=\bm{E}_t^{\mathrm{fit}}G_{j,t}^{\mathrm{fit}}$.
The same fitting observation may have nonzero local and server influences,
thereby retaining the local--server dependence.
\end{assumption}

Define
\begin{equation}
\psi^V_{j,a,t}
=
\vartheta\{\phi_{a,F,t}-\phi_{j,a,L,t}\},
\qquad
\lambda^V_{j,a,t}
=
\frac{
\inner{\overline d^{B}_{j,t}}{\psi^V_{j,a,t}}_{j,t}
}{\norm{\overline d^{B}_{j,t}}_{j,t}
}.
\label{l2:eq:norm-influence}
\end{equation}
The robust-value influence is
\begin{equation}
\lambda^{\mathcal R}_{j,a,t}
=
\lambda^G_{j,a,t}
-
2\tau_{j,t}\lambda^V_{j,a,t}.
\label{l2:eq:robust-influence}
\end{equation}
Let
\begin{equation}
s_{\mathcal R,\mathrm{fit},j,t}^{2}
=
\sum_{a\in\mathcal{I}_t}
\bm{E}_t^{\mathrm{fit}}
\left[
\{\lambda^{\mathcal R}_{j,a,t}\}^{2}
\right].
\label{l2:eq:robust-fit-variance}
\end{equation}

\begin{assumption}[Remainder and Lyapunov conditions]
\label{l2:ass:fit-clt}
For each client sequence under consideration,
$s_{\mathcal R,\mathrm{fit},j,t}>0$ and, for some $\delta>0$,
\begin{equation}
\frac{
\sum_{a\in\mathcal{I}_t}
\bm{E}_t^{\mathrm{fit}}
|\lambda^{\mathcal R}_{j,a,t}|^{2+\delta}
}{
s_{\mathcal R,\mathrm{fit},j,t}^{2+\delta}
}
\longrightarrow0.
\label{l2:eq:robust-lyapunov}
\end{equation}
Moreover,
\begin{align}
\frac{|r^G_{j,t}|}{s_{\mathcal R,\mathrm{fit},j,t}}
&\xrightarrow{\bm{P}}0,
\label{l2:eq:gain-remainder-small}\\
\frac{
\left|
\inner{\overline d^{B}_{j,t}}
{\vartheta(r_{F,t}-r_{j,L,t})}_{j,t}
\right|
}{
\norm{\overline d^{B}_{j,t}}_{j,t}
 s_{\mathcal R,\mathrm{fit},j,t}
}
&\xrightarrow{\bm{P}}0,
\label{l2:eq:norm-linear-remainder-small}\\
\frac{
\tau_{j,t}\norm{e_{B,j,t}}_{j,t}^{2}
}{
\norm{\overline d^{B}_{j,t}}_{j,t}
 s_{\mathcal R,\mathrm{fit},j,t}
}
&\xrightarrow{\bm{P}}0.
\label{l2:eq:norm-quadratic-remainder-small}
\end{align}
Finally,
$\norm{e_{B,j,t}}_{j,t}/\norm{\overline d^{B}_{j,t}}_{j,t}\xrightarrow{\bm{P}}0$.
The quantities in \eqref{l2:eq:gain-remainder-small}--
\eqref{l2:eq:norm-quadratic-remainder-small}, with absolute values where
needed, also converge to zero in first mean after the same normalization.
\end{assumption}

\begin{theorem}[One-client fitting CLT for the $L_2$-robust value]
\label{l2:thm:one-client-fit}
Under Assumptions~\ref{l2:ass:persistent-direction}, \ref{l2:ass:nominal-input}, and \ref{l2:ass:fit-clt},
\begin{equation}
\frac{
\mathcal R_{j,t}^{\mathrm{fit}}-
\mathfrak R_{j,t}
}{
s_{\mathcal R,\mathrm{fit},j,t}
}
\Longrightarrow N(0,1).
\label{l2:eq:one-client-fit-clt}
\end{equation}
\end{theorem}

\begin{proof}
By Proposition~\ref{l2:prop:exact-lf},
\[
\mathcal R_{j,t}^{\mathrm{fit}}
=
G_{j,t}^{\mathrm{fit}}
-
2\tau_{j,t}\norm{d^{B}_{j,t}}_{j,t}.
\]
Assumption~\ref{l2:ass:persistent-direction} and the final condition in
Assumption~\ref{l2:ass:fit-clt} imply that the event in
Lemma~\ref{l2:lem:norm-expansion} has probability tending to one.  Combining
\eqref{l2:eq:norm-linearization} with
\eqref{l2:eq:local-error-representation}--\eqref{l2:eq:server-error-representation}
gives
\[
\norm{d^{B}_{j,t}}_{j,t}-\norm{\overline d^{B}_{j,t}}_{j,t}
=
\sum_{a\in\mathcal{I}_t}\lambda^V_{j,a,t}
+
r^V_{j,t},
\]
where the remainder is negligible after division by
$s_{\mathcal R,\mathrm{fit},j,t}$ by
\eqref{l2:eq:norm-linear-remainder-small}--\eqref{l2:eq:norm-quadratic-remainder-small}.
Combining this with \eqref{l2:eq:nominal-gain-representation}, subtracting
expectations, and using the corresponding first-mean remainder bounds gives
\[
\mathcal R_{j,t}^{\mathrm{fit}}-
\mathfrak R_{j,t}
=
\sum_{a\in\mathcal{I}_t}\lambda^{\mathcal R}_{j,a,t}
+
o_{\bm{P}}(s_{\mathcal R,\mathrm{fit},j,t}).
\]
The summands are independent and centered.  The Lyapunov condition
\eqref{l2:eq:robust-lyapunov} gives the scalar triangular-array central limit
theorem, and Slutsky's theorem proves \eqref{l2:eq:one-client-fit-clt}.
\end{proof}

\subsection{Consistent fitting-variance estimation}

Define the oracle empirical fitting variance
\begin{equation}
\widetilde s_{\mathcal R,\mathrm{fit},j,t}^{2}
=
\sum_{a\in\mathcal{I}_t}
\{\lambda^{\mathcal R}_{j,a,t}\}^{2}.
\label{l2:eq:oracle-fit-variance}
\end{equation}

\begin{assumption}[Oracle and plug-in variance conditions]
\label{l2:ass:variance-estimation}
The fourth moments satisfy
\begin{equation}
\frac{
\sum_{a\in\mathcal{I}_t}
\bm{E}_t^{\mathrm{fit}}
|\lambda^{\mathcal R}_{j,a,t}|^{4}
}{
s_{\mathcal R,\mathrm{fit},j,t}^{4}
}
\longrightarrow0.
\label{l2:eq:oracle-fourth}
\end{equation}
Cross-fitted estimates $\widehat\lambda^{\mathcal R}_{j,a,t}$ satisfy
\begin{equation}
\frac{
\sum_{a\in\mathcal{I}_t}
(\widehat\lambda^{\mathcal R}_{j,a,t}
-\lambda^{\mathcal R}_{j,a,t})^{2}
}{
s_{\mathcal R,\mathrm{fit},j,t}^{2}
}
\xrightarrow{\bm{P}}0.
\label{l2:eq:plugin-aggregate}
\end{equation}
\end{assumption}

When the aggregate plug-in condition is verified from leave-out or foldwise
nuisance estimates, the conditional mean-square bounds are imposed only on
nuisance-specific good events and are summed using
Lemma~\ref{sg:lem:localization}.  Conditioning on the full collection of
overlapping leave-out estimates is neither required nor valid.

Define
\begin{equation}
\widehat s_{\mathcal R,\mathrm{fit},j,t}^{2}
=
\sum_{a\in\mathcal{I}_t}
\{\widehat\lambda^{\mathcal R}_{j,a,t}\}^{2}.
\label{l2:eq:feasible-fit-variance}
\end{equation}

\begin{proposition}[Ratio consistency of the fitting variance]
\label{l2:prop:fit-variance-consistency}
Under Assumption~\ref{l2:ass:variance-estimation},
\begin{equation}
\frac{
\widehat s_{\mathcal R,\mathrm{fit},j,t}^{2}
}{
s_{\mathcal R,\mathrm{fit},j,t}^{2}
}
\xrightarrow{\bm{P}}1.
\label{l2:eq:fit-variance-ratio}
\end{equation}
\end{proposition}

\begin{proof}
Independence gives
\[
\Var_t^{\mathrm{fit}}
\left(
\widetilde s_{\mathcal R,\mathrm{fit},j,t}^{2}
\right)
\le
\sum_a
\bm{E}_t^{\mathrm{fit}}|\lambda^{\mathcal R}_{j,a,t}|^{4}.
\]
Hence \eqref{l2:eq:oracle-fourth} implies
$\widetilde s_{\mathcal R,\mathrm{fit},j,t}^{2}/
 s_{\mathcal R,\mathrm{fit},j,t}^{2}\xrightarrow{\bm{P}}1$.
For the feasible estimator, Cauchy--Schwarz yields
\begin{align*}
&\left|
\sum_a(\widehat\lambda^{\mathcal R}_{j,a,t})^2
-
\sum_a(\lambda^{\mathcal R}_{j,a,t})^2
\right|\\
&\quad\le
2
\left\{\sum_a(\lambda^{\mathcal R}_{j,a,t})^2\right\}^{1/2}
\left\{\sum_a(\widehat\lambda^{\mathcal R}_{j,a,t}
-\lambda^{\mathcal R}_{j,a,t})^2\right\}^{1/2}
+
\sum_a(\widehat\lambda^{\mathcal R}_{j,a,t}
-\lambda^{\mathcal R}_{j,a,t})^2.
\end{align*}
Divide by $s_{\mathcal R,\mathrm{fit},j,t}^{2}$ and use
\eqref{l2:eq:plugin-aggregate}.
\end{proof}

\begin{remark}[Scope of persistent-direction studentization]
Assumption~\ref{l2:ass:variance-estimation} is an aggregate plug-in condition
for the robust-value influences.  Likewise, the grid result below assumes the
required norm-approximation rate.  These are feasible high-level results; a
primitive theorem tying them to grid spacing, density-weight estimation,
client sample sizes, and communication cost is not asserted here.
\end{remark}

\subsection{Evaluation-sample contribution}

Client $j$ has a fresh evaluation sample
\[
Z_{j,t,1}^{\mathrm{eval}},\ldots,
Z_{j,t,N_{j,t}^{\mathrm{eval}}}^{\mathrm{eval}}
\overset{\mathrm{iid}}{\sim}Q_{j,t},
\]
independent across clients and independent of $\mathcal{F}_t^{\mathrm{fit}}$.  Define
\begin{equation}
\widehat G_{j,t}
=
\frac1{N_{j,t}^{\mathrm{eval}}}
\sum_{r=1}^{N_{j,t}^{\mathrm{eval}}}
W_{j,t}(Z_{j,t,r}^{\mathrm{eval}})
\label{l2:eq:empirical-gain}
\end{equation}
and the empirical robust value with the fitted borrowing norm held fixed,
\begin{equation}
\widetilde{\mathcal R}_{j,t}
=
\widehat G_{j,t}
-
2\tau_{j,t}\norm{d^{B}_{j,t}}_{j,t}.
\label{l2:eq:oracle-norm-empirical-robust}
\end{equation}
Then the evaluation fluctuation is exactly
\begin{equation}
\widetilde{\mathcal R}_{j,t}
-
\mathcal R_{j,t}^{\mathrm{fit}}
=
\widehat G_{j,t}-G_{j,t}^{\mathrm{fit}}.
\label{l2:eq:same-eval-fluctuation}
\end{equation}

Conditionally on $\mathcal{F}_t^{\mathrm{fit}}$, define
\begin{equation}
\sigma_{E,j,t}^{2}
=
\Var\{W_{j,t}(Z_{j,t,1}^{\mathrm{eval}})
\mid\mathcal{F}_t^{\mathrm{fit}}\},
\label{l2:eq:conditional-eval-variance}
\end{equation}
and the standardized conditional third moment
\begin{equation}
\beta_{E,j,t}
=
\frac{
\bm{E}[|W_{j,t}-G_{j,t}^{\mathrm{fit}}|^{3}
\mid\mathcal{F}_t^{\mathrm{fit}}]
}{
\sigma_{E,j,t}^{3}
}.
\label{l2:eq:conditional-third}
\end{equation}

The regular conditional law is constructed as in the nominal-gain theorem:
the fitting samples are collected in a standard-Borel statistic, the fitted
gain map is jointly measurable, and the conditional distribution is the
pushforward of $Q_{j,t}^{\otimes N_{j,t}^{\mathrm{eval}}}$ through the standardized empirical
gain map.  Applying the classical Berry--Esseen theorem pathwise gives
\begin{equation}
\sup_{x\in\mathbb{R}}
\left|
\bm{P}\left[
\left.
\frac{
\sqrt{N_{j,t}^{\mathrm{eval}}}
(\widetilde{\mathcal R}_{j,t}-
 \mathcal R_{j,t}^{\mathrm{fit}})
}{
\sigma_{E,j,t}
}
\le x
\right|
\mathcal{F}_t^{\mathrm{fit}}
\right]
-
\Phi(x)
\right|
\le
C_{\mathrm{BE}}
\frac{\beta_{E,j,t}}{\sqrt{N_{j,t}^{\mathrm{eval}}}}.
\label{l2:eq:conditional-BE}
\end{equation}
Consequently, if the right side tends to zero in probability, the random
conditional laws converge in Sweeting's probability mode to $N(0,1)$
\citep{Sweeting1989}.

Assume there is deterministic $v_{E,j,t}>0$ such that
\begin{equation}
\frac{\sigma_{E,j,t}^{2}}{v_{E,j,t}}\xrightarrow{\bm{P}}1,
\qquad
s_{\mathcal R,\mathrm{eval},j,t}^{2}
=
\frac{v_{E,j,t}}{N_{j,t}^{\mathrm{eval}}}.
\label{l2:eq:eval-stabilization}
\end{equation}
Then
\begin{equation}
\mathcal{L}\left(
\left.
\frac{
\widetilde{\mathcal R}_{j,t}-
\mathcal R_{j,t}^{\mathrm{fit}}
}{
s_{\mathcal R,\mathrm{eval},j,t}
}
\right|
\mathcal{F}_t^{\mathrm{fit}}
\right)
\overset{\bm{P}}{\Longrightarrow}
N(0,1).
\label{l2:eq:conditional-eval-limit}
\end{equation}

Let
\begin{equation}
\widehat\sigma_{E,j,t}^{2}
=
\frac1{N_{j,t}^{\mathrm{eval}}-1}
\sum_{r=1}^{N_{j,t}^{\mathrm{eval}}}
\{W_{j,t}(Z_{j,t,r}^{\mathrm{eval}})-\widehat G_{j,t}\}^{2}.
\label{l2:eq:eval-variance-estimator}
\end{equation}
Under the corresponding conditional Lyapunov condition,
$\widehat\sigma_{E,j,t}^{2}/\sigma_{E,j,t}^{2}\xrightarrow{\bm{P}}1$.

\subsection{Same-audit-sample estimation of the borrowing norm}

The oracle-norm estimator above treats $\norm{d^B_{j,t}}_{j,t}$ as fixed
conditional on the fit.  A common implementation estimates the norm from the
same audit covariates used for the paired gain.  Put
\[
u_{j,t}=\norm{d^B_{j,t}}_{j,t}>0,
\qquad
\widehat u_{j,t}^{2}
=
\frac1{N_{j,t}^{\mathrm{eval}}}
\sum_{r=1}^{N_{j,t}^{\mathrm{eval}}}
\{d^B_{j,t}(X_{j,t,r}^{\mathrm{eval}})\}^{2},
\]
and define
\begin{equation}
\widehat{\mathcal R}^{\mathrm{same}}_{j,t}
=
\widehat G_{j,t}-2\tau_{j,t}\widehat u_{j,t}.
\label{l2:eq:same-sample-robust-value}
\end{equation}
Conditionally on $\mathcal{F}_t^{\mathrm{fit}}$, let
\begin{equation}
\xi^{\mathcal R,E}_{j,t}(X,Y)
=
W_{j,t}(X,Y)-G_{j,t}^{\mathrm{fit}}
-
\frac{\tau_{j,t}}{u_{j,t}}
\left[
\{d^B_{j,t}(X)\}^{2}-u_{j,t}^{2}
\right],
\label{l2:eq:same-sample-eval-influence}
\end{equation}
and put
\begin{equation}
\omega_{\mathcal R,E,j,t}^{2}
=
\Var\{\xi^{\mathcal R,E}_{j,t}(X,Y)\mid\mathcal{F}_t^{\mathrm{fit}}\}.
\label{l2:eq:same-sample-eval-variance}
\end{equation}

\begin{theorem}[Conditional audit theorem with an estimated borrowing norm]
\label{l2:thm:same-sample-eval}
Suppose $u_{j,t}$ is bounded away from zero with probability tending to one,
$N_{j,t}^{\mathrm{eval}}\to\infty$, the conditional $2+\delta$ moment of
\eqref{l2:eq:same-sample-eval-influence} is uniformly bounded after
standardization, and $\omega_{\mathcal R,E,j,t}^{2}$ is bounded away from
zero.  Then
\begin{equation}
\widehat{\mathcal R}^{\mathrm{same}}_{j,t}
-
\mathcal R_{j,t}^{\mathrm{fit}}
=
\frac1{N_{j,t}^{\mathrm{eval}}}
\sum_{r=1}^{N_{j,t}^{\mathrm{eval}}}
\xi^{\mathcal R,E}_{j,t}(Z_{j,t,r}^{\mathrm{eval}})
+o_{\bm{P}}\{(N_{j,t}^{\mathrm{eval}})^{-1/2}\}.
\label{l2:eq:same-sample-linearization}
\end{equation}
Consequently, the corresponding regular conditional laws converge in
Sweeting's probability mode to $N(0,1)$ after normalization by
$\omega_{\mathcal R,E,j,t}/\sqrt{N_{j,t}^{\mathrm{eval}}}$.
If
\[
\widehat\xi^{\mathcal R,E}_{j,t,r}
=
(W_{j,t,r}-\widehat G_{j,t})
-
\frac{\tau_{j,t}}{\widehat u_{j,t}}
\left[
\{d^B_{j,t}(X_{j,t,r}^{\mathrm{eval}})\}^{2}
-
\widehat u_{j,t}^{2}
\right],
\]
then the empirical second moment of these influences is ratio consistent for
$\omega_{\mathcal R,E,j,t}^{2}$.
\end{theorem}

\begin{proof}
The exact identity
\[
\widehat u_{j,t}-u_{j,t}
=
\frac{\widehat u_{j,t}^{2}-u_{j,t}^{2}}{2u_{j,t}}
-
\frac{(\widehat u_{j,t}-u_{j,t})^{2}}{2u_{j,t}}
\]
gives
\[
\widehat{\mathcal R}^{\mathrm{same}}_{j,t}
-
\mathcal R_{j,t}^{\mathrm{fit}}
=
\frac1{N_{j,t}^{\mathrm{eval}}}
\sum_r\xi^{\mathcal R,E}_{j,t}(Z_{j,t,r}^{\mathrm{eval}})
+
\frac{\tau_{j,t}}{u_{j,t}}
(\widehat u_{j,t}-u_{j,t})^{2}.
\]
The last term is $o_{\bm{P}}\{(N_{j,t}^{\mathrm{eval}})^{-1/2}\}$ under the
norm floor and fourth-moment bound.  Conditional Lyapunov and the same
conditional von Bahr--Esseen argument used for the nominal gain give the
limit and variance consistency.  The variance includes both the norm term
and its covariance with the paired gain.
\end{proof}

Assume there is a deterministic $\overline\omega_{\mathcal R,E,j,t}^{2}>0$
such that
\[
\frac{\omega_{\mathcal R,E,j,t}^{2}}
{\overline\omega_{\mathcal R,E,j,t}^{2}}
\xrightarrow{\bm{P}}1,
\qquad
s_{\mathcal R,\mathrm{same},j,t}^{2}
=
s_{\mathcal R,\mathrm{fit},j,t}^{2}
+
\frac{\overline\omega_{\mathcal R,E,j,t}^{2}}
{N_{j,t}^{\mathrm{eval}}}.
\]
Then Theorems~\ref{l2:thm:one-client-fit} and
\ref{l2:thm:same-sample-eval}, together with the two-stage assembly theorem,
give
\begin{equation}
\frac{
\widehat{\mathcal R}^{\mathrm{same}}_{j,t}-\mathfrak R_{j,t}
}{s_{\mathcal R,\mathrm{same},j,t}}
\Longrightarrow N(0,1).
\label{l2:eq:same-sample-combined-clt}
\end{equation}
Adding the fitting-variance estimator to the empirical audit influence
variance in Theorem~\ref{l2:thm:same-sample-eval} yields a ratio-consistent
studentizing scale.  This is the regular implementation when the borrowing
norm and gain are estimated from the same audit sample.

\subsection{One-client combined theorem}

Put
\begin{equation}
s_{\mathcal R,j,t}^{2}
=
s_{\mathcal R,\mathrm{fit},j,t}^{2}
+
s_{\mathcal R,\mathrm{eval},j,t}^{2}.
\label{l2:eq:total-variance}
\end{equation}

\begin{theorem}[Combined $L_2$-robust gain CLT]
\label{l2:thm:one-client-combined}
Under Theorem~\ref{l2:thm:one-client-fit}, the conditional evaluation conditions
above, and \eqref{l2:eq:eval-stabilization},
\begin{equation}
\frac{
\widetilde{\mathcal R}_{j,t}-
\mathfrak R_{j,t}
}{
s_{\mathcal R,j,t}
}
\Longrightarrow N(0,1).
\label{l2:eq:one-client-combined-clt}
\end{equation}
The fitting and evaluation limits are asymptotically independent.
\end{theorem}

\begin{proof}
The exact decomposition is
\[
\widetilde{\mathcal R}_{j,t}-\mathfrak R_{j,t}
=
\{\mathcal R_{j,t}^{\mathrm{fit}}-\mathfrak R_{j,t}\}
+
\{\widetilde{\mathcal R}_{j,t}-
  \mathcal R_{j,t}^{\mathrm{fit}}\}.
\]
The first term is $\mathcal{F}_t^{\mathrm{fit}}$-measurable and satisfies
Theorem~\ref{l2:thm:one-client-fit}.  The second has the conditional limit
\eqref{l2:eq:conditional-eval-limit}.  For $u,v\in\mathbb{R}$, condition on
$\mathcal{F}_t^{\mathrm{fit}}$ in the joint characteristic function.  The inner
conditional characteristic function converges in probability to
$e^{-v^{2}/2}$ and is bounded by one, hence also converges in first mean.
The outer factor converges to $e^{-u^{2}/2}$.  This proves asymptotic
independence.  Combining the two variances and using the standard subsequence
argument for the deterministic variance fractions gives
\eqref{l2:eq:one-client-combined-clt}.
\end{proof}

Define
\begin{equation}
\widehat s_{\mathcal R,j,t}^{2}
=
\widehat s_{\mathcal R,\mathrm{fit},j,t}^{2}
+
\frac{\widehat\sigma_{E,j,t}^{2}}{N_{j,t}^{\mathrm{eval}}}.
\label{l2:eq:total-variance-estimator}
\end{equation}

\begin{corollary}[Studentization and robust-positive probability]
\label{l2:cor:one-client-studentized}
Under Theorem~\ref{l2:thm:one-client-combined},
Proposition~\ref{l2:prop:fit-variance-consistency}, and the evaluation
variance consistency,
\begin{equation}
\frac{
\widetilde{\mathcal R}_{j,t}-\mathfrak R_{j,t}
}{
\widehat s_{\mathcal R,j,t}
}
\Longrightarrow N(0,1).
\label{l2:eq:studentized-one-client}
\end{equation}
For every deterministic threshold $\varepsilon_{j,t}$,
\begin{equation}
\bm{P}\{\widetilde{\mathcal R}_{j,t}>\varepsilon_{j,t}\}
=
\Phi\left(
\frac{\mathfrak R_{j,t}-\varepsilon_{j,t}}
{s_{\mathcal R,j,t}}
\right)
+o(1).
\label{l2:eq:robust-positive-probability}
\end{equation}
\end{corollary}

\subsection{Grid-based norm and least-favorable function}

The manuscript implementation estimates the norm from a common query grid.
Let
\[
\widehat\nu_{j,t}^{2}
=
\sum_{\ell=1}^{M_t}
\widehat\omega_{j,t,\ell}
\,d^{B}_{j,t}(x_{t,\ell})^{2},
\]
where $\widehat\omega_{j,t,\ell}$ are nonnegative normalized estimates of the
client-$j$ evaluation-covariate weights.  Define the feasible robust-value estimator
\begin{equation}
\widehat{\mathcal R}_{j,t}
=
\widehat G_{j,t}
-
2\tau_{j,t}\widehat\nu_{j,t}.
\label{l2:eq:feasible-robust-value}
\end{equation}

\begin{assumption}[Grid norm accuracy]
\label{l2:ass:grid-norm}
For every client sequence covered by the theorem,
\begin{equation}
|\widehat\nu_{j,t}-\norm{d^{B}_{j,t}}_{j,t}|\xrightarrow{\bm{P}}0,
\label{l2:eq:grid-norm-consistency}
\end{equation}
and, more importantly, the norm error is negligible on the robust-value
standard-error scale:
\begin{equation}
\frac{
\tau_{j,t}
|\widehat\nu_{j,t}-\norm{d^{B}_{j,t}}_{j,t}|
}{s_{\mathcal R,j,t}}
\xrightarrow{\bm{P}}0.
\label{l2:eq:grid-norm-rate}
\end{equation}
Under Assumption~\ref{l2:ass:persistent-direction}, the sufficient squared-norm
condition
\begin{equation}
\frac{
\tau_{j,t}
|\widehat\nu_{j,t}^{2}-\norm{d^{B}_{j,t}}_{j,t}^{2}|
}{s_{\mathcal R,j,t}}
\xrightarrow{\bm{P}}0
\label{l2:eq:grid-squared-norm-rate}
\end{equation}
implies \eqref{l2:eq:grid-norm-rate}.  Ordinary norm consistency alone does
not.
\end{assumption}

\begin{corollary}[Feasible grid-based robust value]
\label{l2:cor:grid-robust-value}
Under Assumption~\ref{l2:ass:persistent-direction},
Assumption~\ref{l2:ass:grid-norm}, and the conditions of
Corollary~\ref{l2:cor:one-client-studentized},
\begin{equation}
\frac{
\widehat{\mathcal R}_{j,t}-\mathfrak R_{j,t}
}{
\widehat s_{\mathcal R,j,t}
}
\Longrightarrow N(0,1).
\label{l2:eq:grid-studentized}
\end{equation}
\end{corollary}

\begin{proof}
Because $\norm{d^{B}_{j,t}}_{j,t}\xrightarrow{\bm{P}}\norm{\overline d^{B}_{j,t}}_{j,t}$ and the
latter is bounded below by $c_v$, with probability tending to one
$\widehat\nu_{j,t}+\norm{d^{B}_{j,t}}_{j,t}$ is bounded below by $c_v/2$.
Therefore,
\[
|\widehat\nu_{j,t}-\norm{d^{B}_{j,t}}_{j,t}|
\le
\frac{2}{c_v}
|\widehat\nu_{j,t}^{2}-\norm{d^{B}_{j,t}}_{j,t}^{2}|.
\]
Assumption~\ref{l2:ass:grid-norm} makes the difference between
$\widehat{\mathcal R}_{j,t}$ and
$\widetilde{\mathcal R}_{j,t}$ negligible on the total standard-error scale.
Apply Corollary~\ref{l2:cor:one-client-studentized} and Slutsky's theorem.
\end{proof}

For the least-favorable function, let $\widehat m_{j,t}^{0}$ estimate the
center $m_{j,t}^{0}$ and define
\begin{equation}
\widehat m_{j,t}^{\circ}
=
\widehat m_{j,t}^{0}
-
\tau_{j,t}
\frac{d^{B}_{j,t}}{\widehat\nu_{j,t}}.
\label{l2:eq:estimated-lf-function}
\end{equation}
Let
\[
\overline m_{j,t}^{\circ}
=
m_{j,t}^{0}
-
\tau_{j,t}
\frac{\overline d^{B}_{j,t}}
{\norm{\overline d^{B}_{j,t}}_{j,t}}.
\]

\begin{proposition}[Consistency of the least-favorable function]
\label{l2:prop:lf-function-consistency}
Suppose
$\norm{\widehat m_{j,t}^{0}-m_{j,t}^{0}}_{j,t}\xrightarrow{\bm{P}}0$,
$\norm{e_{B,j,t}}_{j,t}\xrightarrow{\bm{P}}0$, and
$|\widehat\nu_{j,t}-\norm{d^{B}_{j,t}}_{j,t}|\xrightarrow{\bm{P}}0$.
Under Assumption~\ref{l2:ass:persistent-direction},
\begin{equation}
\norm{
\widehat m_{j,t}^{\circ}-
\overline m_{j,t}^{\circ}
}_{j,t}
\xrightarrow{\bm{P}}0.
\label{l2:eq:lf-function-consistency}
\end{equation}
\end{proposition}

\begin{proof}
On the event where both $\norm{d^{B}_{j,t}}_{j,t}$ and
$\widehat\nu_{j,t}$ are bounded below,
\begin{align*}
\norm{
\frac{d^{B}_{j,t}}{\widehat\nu_{j,t}}
-
\frac{\overline d^{B}_{j,t}}
{\norm{\overline d^{B}_{j,t}}_{j,t}}
}_{j,t}
&\le
\frac{\norm{d^{B}_{j,t}-\overline d^{B}_{j,t}}_{j,t}
}{\widehat\nu_{j,t}}
+
\norm{\overline d^{B}_{j,t}}_{j,t}
\left|
\frac1{\widehat\nu_{j,t}}
-
\frac1{\norm{\overline d^{B}_{j,t}}_{j,t}}
\right|.
\end{align*}
Every term tends to zero in probability.  Add the center-estimation error and
multiply by the bounded radius.
\end{proof}

\subsection{Finite-dimensional client process}

Fix $r$.  For sufficiently large $t$, clients $1,\ldots,r$ participate.
Define
\[
\widehat{\bm{\mathcal R}}_{r,t}
=
(\widehat{\mathcal R}_{1,t},\ldots,
 \widehat{\mathcal R}_{r,t})^{\mathsf{T}},
\qquad
\bm{\mathfrak R}_{r,t}
=
(\mathfrak R_{1,t},\ldots,
 \mathfrak R_{r,t})^{\mathsf{T}}.
\]
For fitting observation $a$, put
\[
\bm\lambda^{\mathcal R,(r)}_{a,t}
=
(\lambda^{\mathcal R}_{1,a,t},\ldots,
 \lambda^{\mathcal R}_{r,a,t})^{\mathsf{T}}.
\]
Define
\begin{equation}
\Sigma^{\mathcal R,\mathrm{fit}}_{r,t}
=
\sum_{a\in\mathcal{I}_t}
\bm{E}_t^{\mathrm{fit}}
\left[
\bm\lambda^{\mathcal R,(r)}_{a,t}
\bm\lambda^{\mathcal R,(r)\mathsf{T}}_{a,t}
\right].
\label{l2:eq:fit-cross-covariance}
\end{equation}
The off-diagonal entries retain all common-server dependence.

Let
\begin{equation}
D^{\mathcal R}_{r,t}
=
\operatorname{diag}(s_{\mathcal R,1,t},\ldots,s_{\mathcal R,r,t}),
\label{l2:eq:robust-diagonal-scale}
\end{equation}
and define the normalized fitting covariance
\[
A^{\mathcal R}_{r,t}
=
(D^{\mathcal R}_{r,t})^{-1}
\Sigma^{\mathcal R,\mathrm{fit}}_{r,t}
(D^{\mathcal R}_{r,t})^{-1}.
\]
The normalized evaluation covariance is diagonal:
\[
B^{\mathcal R}_{r,t}
=
\operatorname{diag}\left(
\frac{s_{\mathcal R,\mathrm{eval},1,t}^{2}}
{s_{\mathcal R,1,t}^{2}},\ldots,
\frac{s_{\mathcal R,\mathrm{eval},r,t}^{2}}
{s_{\mathcal R,r,t}^{2}}
\right).
\]

\begin{assumption}[Finite-block covariance and Lyapunov conditions]
\label{l2:ass:block-clt}
For every fixed $r$,
\[
A^{\mathcal R}_{r,t}\longrightarrow A^{\mathcal R}_{r},
\qquad
B^{\mathcal R}_{r,t}\longrightarrow B^{\mathcal R}_{r},
\]
and, for some $\delta>0$,
\begin{equation}
\sum_{a\in\mathcal{I}_t}
\bm{E}_t^{\mathrm{fit}}
\left[
\left\|
(D^{\mathcal R}_{r,t})^{-1}
\bm\lambda^{\mathcal R,(r)}_{a,t}
\right\|^{2+\delta}
\right]
\longrightarrow0.
\label{l2:eq:block-lyapunov}
\end{equation}
The vector remainders are negligible after multiplication by
$(D^{\mathcal R}_{r,t})^{-1}$.
\end{assumption}

\begin{theorem}[Fixed-block $L_2$-robust gain process]
\label{l2:thm:fixed-block}
Under Assumptions~\ref{l2:ass:persistent-direction}, \ref{l2:ass:nominal-input}, and \ref{l2:ass:fit-clt},
\ref{l2:ass:variance-estimation}, \ref{l2:ass:grid-norm}, and
\ref{l2:ass:block-clt}, together with the coordinatewise conditional
evaluation and variance-stabilization conditions,
\begin{equation}
(D^{\mathcal R}_{r,t})^{-1}
\left(
\widehat{\bm{\mathcal R}}_{r,t}
-
\bm{\mathfrak R}_{r,t}
\right)
\Longrightarrow
N_r(\bm0,R^{\mathcal R}_{r}),
\label{l2:eq:fixed-block-limit}
\end{equation}
where
\begin{equation}
R^{\mathcal R}_{r}
=
A^{\mathcal R}_{r}
+
B^{\mathcal R}_{r}.
\label{l2:eq:robust-covariance-decomposition}
\end{equation}
The limiting fitting and evaluation Gaussian vectors are independent.
\end{theorem}

\begin{proof}
The block Lyapunov and remainder conditions give the fitting CLT with
covariance $A^{\mathcal R}_{r}$.  The coordinatewise conditional audit theorem
and independent client audit streams give the diagonal conditional limit with
covariance $B^{\mathcal R}_{r}$.  Apply the two-stage client-array assembly
theorem, Theorem~\ref{gen:thm:two-stage-assembly}.
\end{proof}

The finite-block fitting covariance is estimated by
\begin{equation}
\widehat\Sigma^{\mathcal R,\mathrm{fit}}_{j\ell,t}
=
\sum_{a\in\mathcal{I}_t}
\widehat\lambda^{\mathcal R}_{j,a,t}
\widehat\lambda^{\mathcal R}_{\ell,a,t},
\label{l2:eq:fit-covariance-estimator}
\end{equation}
while the evaluation contribution is estimated diagonally by
$\widehat\sigma_{E,j,t}^{2}/N_{j,t}^{\mathrm{eval}}$.  Under the aggregate plug-in conditions
for every fixed block, the resulting covariance and correlation estimators
are consistent entrywise.

\subsection{Sequence-space limit and positivity}

Embed the standardized coordinates in
\[
\mathbb{R}^{\infty}=\prod_{j=1}^{\infty}\mathbb{R}
\]
with its product topology.  Put
\begin{equation}
T^{\mathcal R}_{j,t}
=
\begin{cases}
\dfrac{
\widehat{\mathcal R}_{j,t}-\mathfrak R_{j,t}
}{s_{\mathcal R,j,t}},&j\le K_t,\\[1.3ex]
0,&j>K_t.
\end{cases}
\label{l2:eq:embedded-process}
\end{equation}

\begin{assumption}[Compatible covariance limits]
\label{l2:ass:compatible}
The matrices $R^{\mathcal R}_{r}$ are compatible principal submatrices, and
for every fixed $j$ the sequence $T^{\mathcal R}_{j,t}$ is tight.
\end{assumption}

\begin{theorem}[Client-indexed $L_2$-robust gain process]
\label{l2:thm:process}
Suppose the conditions of Theorem~\ref{l2:thm:fixed-block} hold for every
fixed $r$, and Assumption~\ref{l2:ass:compatible} holds.  Then there is a
centered Gaussian random
sequence
\[
\bm Z^{\mathcal R}
=(Z^{\mathcal R}_{1},Z^{\mathcal R}_{2},\ldots)
\]
with the compatible finite-dimensional covariance matrices
$R^{\mathcal R}_{r}$, and
\begin{equation}
\bm T^{\mathcal R}_{t}
\Longrightarrow
\bm Z^{\mathcal R}
\qquad
\text{in }\mathbb{R}^{\infty}
\text{ with the product topology}.
\label{l2:eq:process-limit}
\end{equation}
\end{theorem}

\begin{proof}
Theorem~\ref{l2:thm:fixed-block} gives convergence of every fixed finite block.
Coordinatewise tightness implies tightness in the countable product topology:
for any $\eta>0$, choose compact intervals $K_j$ with
$\sup_t\bm{P}(T^{\mathcal R}_{j,t}\notin K_j)<\eta2^{-j}$, and use compactness
of $\prod_jK_j$.  Every subsequential limit has the same compatible Gaussian
finite-dimensional distributions, so the limit is unique.
\end{proof}

For deterministic thresholds $\varepsilon_{j,t}$, define the robust margins
\begin{equation}
\gamma^{\mathcal R}_{j,t}
=
\frac{\mathfrak R_{j,t}-\varepsilon_{j,t}}
{s_{\mathcal R,j,t}}.
\label{l2:eq:robust-margin}
\end{equation}

\begin{corollary}[Fixed-block robust-positive probability]
\label{l2:cor:orthant}
If, for fixed $r$,
$\gamma^{\mathcal R}_{j,t}\to\gamma^{\mathcal R}_{j}$ for
$j=1,\ldots,r$, then
\begin{align}
&\bm{P}\left
\{
\widehat{\mathcal R}_{j,t}>\varepsilon_{j,t},
\ j=1,\ldots,r
\right\}
\nonumber\\
&\qquad\longrightarrow
\bm{P}\left
\{
Z^{\mathcal R}_{j}>-\gamma^{\mathcal R}_{j},
\ j=1,\ldots,r
\right\}.
\label{l2:eq:orthant-limit}
\end{align}
\end{corollary}

For the moving event involving every client, define
\[
\delta^{\mathcal R}_{j,t}
=
\sup_{x\in\mathbb{R}}
\left|
\bm{P}(T^{\mathcal R}_{j,t}\le x)-\Phi(x)
\right|.
\]

\begin{corollary}[All-client robust positivity by summability]
\label{l2:cor:all-client}
For every $t$,
\begin{align}
&\bm{P}\left\{
\widehat{\mathcal R}_{j,t}>\varepsilon_{j,t}
\text{ for every }j\le K_t
\right\}
\nonumber\\
&\quad\ge
1-
\sum_{j=1}^{K_t}
\left
\{
\Phi(-\gamma^{\mathcal R}_{j,t})
+
\delta^{\mathcal R}_{j,t}
\right
\}.
\label{l2:eq:all-client-bound}
\end{align}
Consequently, if
\begin{equation}
\sum_{j=1}^{K_t}\Phi(-\gamma^{\mathcal R}_{j,t})
+
\sum_{j=1}^{K_t}\delta^{\mathcal R}_{j,t}
\longrightarrow0,
\label{l2:eq:all-client-summability}
\end{equation}
then every empirical $L_2$-robust value exceeds its threshold with probability
tending to one.
\end{corollary}

\begin{proof}
Apply the union bound and the definition of
$\delta^{\mathcal R}_{j,t}$.
\end{proof}

\subsection{Exact-zero mean borrowing direction}
\label{l2:subsec:vanishing-direction}

We now treat the nondifferentiable boundary
\begin{equation}
\overline d^{B}_{j,t}=0.
\label{l2:eq:vanishing-mean-direction}
\end{equation}
Put
\begin{equation}
S^B_{j,t}=\norm{d^{B}_{j,t}}_{j,t}^{2},
\qquad
\nu_{j,t}^{2}=\bm{E}_t^{\mathrm{fit}}S^B_{j,t}.
\label{l2:eq:direction-squared-norm}
\end{equation}
Although the norm is not differentiable at zero, the squared norm is a
quadratic functional.  The exact identity
\begin{align}
2\tau_{j,t}\sqrt{S^B_{j,t}}
&=
2\tau_{j,t}\nu_{j,t}
+
\kappa^B_{j,t}(S^B_{j,t}-\nu_{j,t}^{2})
-
R^{\sqrt{\ }}_{j,t},
\label{l2:eq:square-root-boundary-expansion}\\
\kappa^B_{j,t}
&=
\frac{\tau_{j,t}}{\nu_{j,t}},
\qquad
R^{\sqrt{\ }}_{j,t}
=
\frac{
\tau_{j,t}(S^B_{j,t}-\nu_{j,t}^{2})^{2}
}{
\nu_{j,t}(\sqrt{S^B_{j,t}}+\nu_{j,t})^{2}
}
\ge0
\label{l2:eq:square-root-boundary-remainder}
\end{align}
follows from the exact square-root expansion.

Let
\[
\bm\psi^B_{j,a,t}
=
\vartheta(\phi_{a,F,t}-\phi_{j,a,L,t}).
\]
Define the squared-direction diagonal and off-diagonal terms
\begin{align}
d^B_{j,a,t}
&=
\norm{\bm\psi^B_{j,a,t}}_{j,t}^{2}
-
\bm{E}_t^{\mathrm{fit}}
\norm{\bm\psi^B_{j,a,t}}_{j,t}^{2},
\label{l2:eq:direction-diagonal}\\
u^B_{j,ab,t}
&=
2\inner{\bm\psi^B_{j,a,t}}{\bm\psi^B_{j,b,t}}_{j,t},
\qquad a<b.
\label{l2:eq:direction-off-diagonal}
\end{align}
For the nominal gain, use the linear influence $\lambda^G_{j,a,t}$ from
\eqref{l2:eq:nominal-gain-representation} and define the gain diagonal and
off-diagonal kernels through the bilinear form
\eqref{sg:eq:boundary-bilinear-form}:
\begin{align}
d^G_{j,a,t}
&=
\mathcal B_{\alpha,j,t}(\bm\phi_{j,a,t},\bm\phi_{j,a,t})
-
\bm{E}_t^{\mathrm{fit}}
\mathcal B_{\alpha,j,t}(\bm\phi_{j,a,t},\bm\phi_{j,a,t}),\\
u^G_{j,ab,t}
&=
2\mathcal B_{\alpha,j,t}(\bm\phi_{j,a,t},\bm\phi_{j,b,t}).
\label{l2:eq:gain-boundary-kernels}
\end{align}
Here
$\bm\phi_{j,a,t}=(\phi_{j,a,L,t},\phi_{a,F,t})$, and
$\mathcal B_{\alpha,j,t}$ is \eqref{sg:eq:boundary-bilinear-form} with the
client-$j$ evaluation inner product.

After fixing an ordering of the independent fitting observations, define the
combined martingale increment
\begin{equation}
M^{\mathcal R,0}_{j,a,t}
=
\lambda^G_{j,a,t}
+d^G_{j,a,t}
-\kappa^B_{j,t}d^B_{j,a,t}
+
\sum_{b<a}
\{u^G_{j,ba,t}-\kappa^B_{j,t}u^B_{j,ba,t}\}.
\label{l2:eq:vanishing-direction-increment}
\end{equation}
Let
\begin{equation}
s_{\mathcal R,0,\mathrm{fit},j,t}^{2}
=
\sum_a\bm{E}_t^{\mathrm{fit}}
(M^{\mathcal R,0}_{j,a,t})^{2}.
\label{l2:eq:vanishing-direction-variance}
\end{equation}

\begin{assumption}[Exact-zero-direction martingale regime]
\label{l2:ass:vanishing-direction}
For every client sequence covered by the theorem, $\nu_{j,t}>0$ and
\begin{equation}
\frac{S^B_{j,t}}{\nu_{j,t}^{2}}\xrightarrow{\bm{P}}1.
\label{l2:eq:norm-concentration}
\end{equation}
The nominal gain and squared-direction functionals admit the linear--quadratic
expansions
\begin{align}
G_{j,t}^{\mathrm{fit}}-\Gamma_{j,t}
&=
\sum_a\lambda^G_{j,a,t}
+\sum_a d^G_{j,a,t}
+\sum_{a<b}u^G_{j,ab,t}
+r^{G,0}_{j,t},
\label{l2:eq:vanishing-gain-expansion}\\
S^B_{j,t}-\nu_{j,t}^{2}
&=
\sum_a d^B_{j,a,t}
+\sum_{a<b}u^B_{j,ab,t}
+r^{B,0}_{j,t},
\label{l2:eq:vanishing-norm-expansion}
\end{align}
where
$|r^{G,0}_{j,t}|+|\kappa^B_{j,t}r^{B,0}_{j,t}|=
o_{\bm{P}}(s_{\mathcal R,0,\mathrm{fit},j,t})$ and the same holds in
first mean.  The square-root remainder satisfies
\begin{equation}
\frac{R^{\sqrt{\ }}_{j,t}
+\bm{E}_t^{\mathrm{fit}}R^{\sqrt{\ }}_{j,t}}
{s_{\mathcal R,0,\mathrm{fit},j,t}}
\xrightarrow{\bm{P}}0.
\label{l2:eq:square-root-remainder-small}
\end{equation}
With $\mathcal H_{a,t}$ the fitting filtration, for every $\epsilon>0$,
\begin{align}
\frac1{s_{\mathcal R,0,\mathrm{fit},j,t}^{2}}
\sum_a
\bm{E}_t^{\mathrm{fit}}
\{(M^{\mathcal R,0}_{j,a,t})^{2}\mid\mathcal H_{a-1,t}\}
&\xrightarrow{\bm{P}}1,
\label{l2:eq:vanishing-pqv}\\
\frac1{s_{\mathcal R,0,\mathrm{fit},j,t}^{2}}
\sum_a
\bm{E}_t^{\mathrm{fit}}
\{(M^{\mathcal R,0}_{j,a,t})^{2}
\mathbf{1}(|M^{\mathcal R,0}_{j,a,t}|>
\epsilon s_{\mathcal R,0,\mathrm{fit},j,t})
\mid\mathcal H_{a-1,t}\}
&\xrightarrow{\bm{P}}0.
\label{l2:eq:vanishing-lindeberg}
\end{align}
\end{assumption}

\begin{theorem}[Fitting CLT at an exact-zero mean borrowing direction]
\label{l2:thm:vanishing-direction-fit}
Under Assumption~\ref{l2:ass:vanishing-direction},
\begin{equation}
\frac{
\mathcal R_{j,t}^{\mathrm{fit}}-\mathfrak R_{j,t}
}{s_{\mathcal R,0,\mathrm{fit},j,t}}
\Longrightarrow N(0,1).
\label{l2:eq:vanishing-direction-fit-clt}
\end{equation}
\end{theorem}

\begin{proof}
Using Proposition~\ref{l2:prop:exact-lf} and
\eqref{l2:eq:square-root-boundary-expansion}, then subtracting fitting
expectations, gives
\begin{align*}
\mathcal R_{j,t}^{\mathrm{fit}}-\mathfrak R_{j,t}
={}&
(G_{j,t}^{\mathrm{fit}}-\Gamma_{j,t})
-
\kappa^B_{j,t}(S^B_{j,t}-\nu_{j,t}^{2})\\
&+
R^{\sqrt{\ }}_{j,t}
-
\bm{E}_t^{\mathrm{fit}}R^{\sqrt{\ }}_{j,t}.
\end{align*}
The linear--quadratic expansions of the first two centered terms yield
$\sum_aM^{\mathcal R,0}_{j,a,t}$ plus a negligible remainder.  The final line
is negligible by \eqref{l2:eq:square-root-remainder-small}.  Conditions
\eqref{l2:eq:vanishing-pqv}--\eqref{l2:eq:vanishing-lindeberg} and the
martingale central limit theorem complete the proof.
\end{proof}

\begin{remark}[A usable sufficient square-root condition]
Since
\[
0\le R^{\sqrt{\ }}_{j,t}
\le
\frac{\tau_{j,t}}{\nu_{j,t}^{3}}
(S^B_{j,t}-\nu_{j,t}^{2})^{2},
\]
condition \eqref{l2:eq:square-root-remainder-small} follows from the
corresponding second-moment bound for the centered squared norm.  The basic
scale is the actual direction size
$\nu_{j,t}^{2}=\bm{E}_t^{\mathrm{fit}}S^B_{j,t}$ and the actual martingale
variance of $S^B_{j,t}$.  Comparisons with $\mathfrak r_{j,t}/h_t$ require a
separate directional noncancellation assumption and are not consequences of
the total local-plus-server leverage bound.
\end{remark}

\begin{corollary}[Variance-driven exact-model boundary]
\label{l2:cor:exact-vanishing-direction}
Suppose, in addition, that
$\overline L_{j,t}=\overline F_t=m^0_{j,t}$, so the nominal gain linear
influence vanishes.  Put
\begin{equation}
\kappa^B_{j,t}=\frac{\tau_{j,t}}{\nu_{j,t}},
\qquad
\mathcal B^{\mathcal R,0}_{j,t}
=
\mathcal B_{\alpha,j,t}
-
\kappa^B_{j,t}\mathcal B^B_{j,t},
\label{l2:eq:combined-boundary-bilinear-form}
\end{equation}
where
\[
\mathcal B^B_{j,t}(\bm u,\bm v)
=
\vartheta^2
\inner{u_F-u_L}{v_F-v_L}_{j,t}.
\]
Under the Hall--de Jong conditions for this stage-specific combined form,
Theorem~\ref{l2:thm:vanishing-direction-fit} holds.  If, in addition,
$\kappa^B_{j,t}\to\zeta_j\in[0,\infty)$ and
\begin{equation}
|\kappa^B_{j,t}-\zeta_j|\,
\{\Var_t^{\mathrm{fit}}(S^B_{j,t})\}^{1/2}
=o(s_{\mathcal R,0,\mathrm{fit},j,t}),
\label{l2:eq:limiting-kappa-stability}
\end{equation}
then the stage-specific form may be replaced, on the theorem scale, by
$\mathcal B_{\alpha,j,t}-\zeta_j\mathcal B^B_{j,t}$.  Moreover,
\begin{equation}
\mathfrak R_{j,t}
=
\Gamma_{j,t}-2\zeta_j\nu_{j,t}^{2}
+o(\nu_{j,t}^{2}).
\label{l2:eq:boundary-robust-center}
\end{equation}
The general local scaling is expressed by
$\kappa^B_{j,t}=\tau_{j,t}/\nu_{j,t}$; a nondegenerate local radius therefore
has order $\nu_{j,t}$.  If the additional directional noncancellation
condition
\[
0<c_B\le
\frac{\nu_{j,t}^{2}}{\mathfrak r_{j,t}/h_t}
\le C_B<\infty
\]
holds, this becomes
$\tau_{j,t}\asymp(\mathfrak r_{j,t}/h_t)^{1/2}$.  Only under a further
local-dominant sample-size-weight specialization does it reduce to
$(n_{j,t}h_{j,t})^{-1/2}$.
\end{corollary}

\begin{proof}
At the exact-model boundary, both the gain and squared-direction centered
terms are quadratic forms in the same first-order ratio influences.  The
exact coefficient multiplying the squared-direction form is
$\kappa^B_{j,t}$, which gives
\eqref{l2:eq:combined-boundary-bilinear-form} and the stage-specific
martingale CLT.  Replacing $\kappa^B_{j,t}$ by $\zeta_j$ changes the centered
quadratic statistic by
\[
(\kappa^B_{j,t}-\zeta_j)
(S^B_{j,t}-\nu_{j,t}^{2}),
\]
which is negligible under \eqref{l2:eq:limiting-kappa-stability}.  Put
$X_t=S^B_{j,t}/\nu_{j,t}^{2}$.  Then $X_t\xrightarrow{\bm{P}}1$ and
$\bm{E}_t^{\mathrm{fit}}X_t=1$.  Jensen gives
$\bm{E}_t^{\mathrm{fit}}\sqrt{X_t}\le1$, while the subsequence form of
Fatou gives the reverse lower limit.  Hence
$\bm{E}_t^{\mathrm{fit}}\sqrt{S^B_{j,t}}/\nu_{j,t}\to1$, and
$2\tau_{j,t}\nu_{j,t}=2\kappa^B_{j,t}\nu_{j,t}^{2}$ gives
\eqref{l2:eq:boundary-robust-center}.
\end{proof}

\begin{example}[Cancellation can reduce the borrowing-direction scale]
\label{l2:ex:cancellation}
Consider two homogeneous zero-regression clients with a common design and
bandwidth, sample sizes $n$ and $m$, sample-size weights, and
$m\to\infty$ with $m/n\to0$.  If $L_1,L_2$ are the local Nadaraya--Watson
estimators, then the pooled estimator satisfies
\[
F-L_1
=
\frac{m\widehat p_2}{n\widehat p_1+m\widehat p_2}
(L_2-L_1).
\]
Under regular overlap,
\[
\bm{E}\norm{F-L_1}^{2}
\asymp
\left(\frac mn\right)^2
\left(\frac1{mh}+\frac1{nh}\right)
\asymp
\frac{m}{n^2h},
\]
whereas the total leverage gives
$\mathfrak r_{1,t}/h\asymp1/(nh)$.  Hence
\[
\frac{\nu_{1,t}^{2}}{\mathfrak r_{1,t}/h}
\asymp\frac mn\longrightarrow0.
\]
The actual direction scale, rather than total estimator leverage, therefore
governs the $L_2$ boundary.
\end{example}

For feasible variance estimation, define cross-fitted increments
$\widehat M^{\mathcal R,0}_{j,a,t}$ by replacing all coefficient, ratio, and
evaluation-law nuisances in
\eqref{l2:eq:vanishing-direction-increment}, and set
\begin{equation}
\widehat s_{\mathcal R,0,\mathrm{fit},j,t}^{2}
=
\sum_a(\widehat M^{\mathcal R,0}_{j,a,t})^{2}.
\label{l2:eq:vanishing-direction-variance-estimator}
\end{equation}
\begin{assumption}[Exact-zero-direction variance estimation]
\label{l2:ass:vanishing-variance-estimation}
In addition to \eqref{l2:eq:vanishing-pqv},
\begin{equation}
\frac{
\sum_a\bm{E}_t^{\mathrm{fit}}
|M^{\mathcal R,0}_{j,a,t}|^4
}{s_{\mathcal R,0,\mathrm{fit},j,t}^{4}}
\longrightarrow0,
\label{l2:eq:vanishing-fourth-variation}
\end{equation}
and
\begin{equation}
\frac{
\sum_a(\widehat M^{\mathcal R,0}_{j,a,t}
-M^{\mathcal R,0}_{j,a,t})^{2}
}{s_{\mathcal R,0,\mathrm{fit},j,t}^{2}}
\xrightarrow{\bm{P}}0.
\label{l2:eq:vanishing-direction-plugin}
\end{equation}
\end{assumption}

When \eqref{l2:eq:vanishing-direction-plugin} is obtained from pairwise or
observationwise cross-fitting, each conditional bound is localized to the
good event measurable with respect to the nuisance estimate used in that
increment, and Lemma~\ref{sg:lem:localization} is applied to the simultaneous
good event.

\begin{proposition}[Exact-zero-direction fitting-variance consistency]
\label{l2:prop:vanishing-variance-consistency}
Under Assumptions~\ref{l2:ass:vanishing-direction} and
\ref{l2:ass:vanishing-variance-estimation},
\begin{equation}
\frac{\widehat s_{\mathcal R,0,\mathrm{fit},j,t}^{2}}
{s_{\mathcal R,0,\mathrm{fit},j,t}^{2}}
\xrightarrow{\bm{P}}1.
\label{l2:eq:vanishing-variance-consistency}
\end{equation}
\end{proposition}

\begin{proof}
The process
\[
\sum_a\left
\{(M^{\mathcal R,0}_{j,a,t})^2
-\bm{E}_t^{\mathrm{fit}}[(M^{\mathcal R,0}_{j,a,t})^2
\mid\mathcal H_{a-1,t}]
\right\}
\]
is a martingale.  Its variance is bounded by
$\sum_a\bm{E}_t^{\mathrm{fit}}|M^{\mathcal R,0}_{j,a,t}|^4$, so
\eqref{l2:eq:vanishing-fourth-variation}, Chebyshev's inequality, and
\eqref{l2:eq:vanishing-pqv} give
$\sum_a(M^{\mathcal R,0}_{j,a,t})^2/
 s_{\mathcal R,0,\mathrm{fit},j,t}^{2}\xrightarrow{\bm{P}}1$.
The aggregate plug-in condition and the usual $\ell^2$ Cauchy--Schwarz
inequality then give \eqref{l2:eq:vanishing-variance-consistency}.
\end{proof}

\begin{remark}[Communication scope]
The estimator in Proposition~\ref{l2:prop:vanishing-variance-consistency} is a
theoretical cross-fitted martingale estimator.  Its reconstruction from the
basic server summaries would require additional client-side calculations and
is not claimed here.
\end{remark}

\begin{assumption}[Exact-zero-direction evaluation stabilization]
\label{l2:ass:vanishing-evaluation}
Let $a^0_{j,t}=m^0_{j,t}-\overline L_{j,t}$ and define
\begin{equation}
v^{B}_{E,j,t}
=
4\bm{E}_t^{\mathrm{fit}}\left[
\Var_{Q_{j,t,X}}\{d^{B}_{j,t}(X)a^0_{j,t}(X)\}
+
\int\{d^{B}_{j,t}(x)\}^{2}
\sigma_{Q,j,t}^{2}(x)Q_{j,t,X}(\,d x)
\right].
\label{l2:eq:vanishing-evaluation-oracle}
\end{equation}
For every fixed client block, this quantity is positive, the conditional
evaluation variance satisfies
\begin{equation}
\frac{\sigma_{E,j,t}^{2}}{v^{B}_{E,j,t}}\xrightarrow{\bm{P}}1,
\label{l2:eq:vanishing-evaluation-stabilization}
\end{equation}
and the conditional standardized third moment divided by
$\sqrt{N_{j,t}^{\mathrm{eval}}}$ converges to zero in probability.
\end{assumption}

Put
\begin{equation}
s_{\mathcal R,0,\mathrm{eval},j,t}^{2}
=
\frac{v^{B}_{E,j,t}}{N_{j,t}^{\mathrm{eval}}},
\qquad
s_{\mathcal R,0,j,t}^{2}
=
s_{\mathcal R,0,\mathrm{fit},j,t}^{2}
+s_{\mathcal R,0,\mathrm{eval},j,t}^{2}.
\label{l2:eq:vanishing-total-scale}
\end{equation}

\begin{theorem}[Combined exact-zero-direction robust-gain inference]
\label{l2:thm:vanishing-direction-combined}
Under Theorem~\ref{l2:thm:vanishing-direction-fit},
Assumptions~\ref{l2:ass:vanishing-variance-estimation} and
\ref{l2:ass:vanishing-evaluation},
\begin{equation}
\frac{
\widetilde{\mathcal R}_{j,t}-\mathfrak R_{j,t}
}{s_{\mathcal R,0,j,t}}
\Longrightarrow N(0,1).
\label{l2:eq:vanishing-direction-combined}
\end{equation}
Moreover,
\begin{equation}
\widehat s_{\mathcal R,0,j,t}^{2}
=
\widehat s_{\mathcal R,0,\mathrm{fit},j,t}^{2}
+
\frac{\widehat\sigma_{E,j,t}^{2}}
{N_{j,t}^{\mathrm{eval}}}
\end{equation}
is ratio consistent for $s_{\mathcal R,0,j,t}^{2}$, and the corresponding
studentized statistic is asymptotically standard normal.
\end{theorem}

\begin{proof}
The evaluation fluctuation of $\widetilde{\mathcal R}_{j,t}$ equals
$\widehat G_{j,t}-G_{j,t}^{\mathrm{fit}}$ exactly, because the fitted
borrowing norm is held fixed conditionally on the fitting data.  The pathwise
Berry--Esseen theorem and
\eqref{l2:eq:vanishing-evaluation-stabilization} give the conditional Gaussian
limit on the scale $s_{\mathcal R,0,\mathrm{eval},j,t}$.  Conditioning in the
joint characteristic function gives asymptotic independence from the fitting
martingale in Theorem~\ref{l2:thm:vanishing-direction-fit}.  The usual
subsequence argument combines the two deterministic variance fractions.  The
conditional von Bahr--Esseen sample-variance argument gives
$\widehat\sigma_{E,j,t}^{2}/v^{B}_{E,j,t}\xrightarrow{\bm{P}}1$, and the fitting-variance
ratio consistency gives studentization.
\end{proof}

\begin{corollary}[Grid transfer at an exact-zero mean direction]
\label{l2:cor:boundary-grid-transfer}
Suppose the assumptions of
Theorem~\ref{l2:thm:vanishing-direction-combined} hold, and define
\[
\widehat{\mathcal R}_{j,t}
=
\widehat G_{j,t}-2\tau_{j,t}\widehat\nu_{j,t}.
\]
If
\begin{equation}
\frac{
\tau_{j,t}
|\widehat\nu_{j,t}-\sqrt{S^B_{j,t}}|
}{s_{\mathcal R,0,j,t}}
\xrightarrow{\bm{P}}0,
\label{l2:eq:boundary-grid-transfer}
\end{equation}
then the oracle-normalized and studentized conclusions of
Theorem~\ref{l2:thm:vanishing-direction-combined} remain valid with
$\widetilde{\mathcal R}_{j,t}$ replaced by
$\widehat{\mathcal R}_{j,t}$.
If, in addition,
$\widehat\nu_{j,t}+\sqrt{S^B_{j,t}}\asymp_{\bm{P}}\nu_{j,t}$, it suffices that
\begin{equation}
\frac{
\tau_{j,t}
|\widehat\nu_{j,t}^{2}-S^B_{j,t}|
}{\nu_{j,t}s_{\mathcal R,0,j,t}}
\xrightarrow{\bm{P}}0.
\label{l2:eq:boundary-grid-transfer-squared}
\end{equation}
\end{corollary}

\begin{proof}
The exact difference is
\[
\widehat{\mathcal R}_{j,t}-\widetilde{\mathcal R}_{j,t}
=
-2\tau_{j,t}
\frac{\widehat\nu_{j,t}^{2}-S^B_{j,t}}
{\widehat\nu_{j,t}+\sqrt{S^B_{j,t}}}.
\]
Condition \eqref{l2:eq:boundary-grid-transfer} and Slutsky prove the first
claim; the comparability condition converts it to
\eqref{l2:eq:boundary-grid-transfer-squared}.
\end{proof}

For a fixed client block $1,\ldots,r$, put
\[
\widetilde{\boldsymbol{\mathcal R}}_{r,t}
=(\widetilde{\mathcal R}_{1,t},\ldots,
  \widetilde{\mathcal R}_{r,t})^{\mathsf{T}},
\qquad
\boldsymbol{\mathfrak R}_{r,t}
=(\mathfrak R_{1,t},\ldots,\mathfrak R_{r,t})^{\mathsf{T}},
\]
and
\[
\bm M^{\mathcal R,0,(r)}_{a,t}
=
(M^{\mathcal R,0}_{1,a,t},\ldots,
 M^{\mathcal R,0}_{r,a,t})^{\mathsf{T}},
\]
and define
\begin{equation}
\Sigma^{\mathcal R,0,\mathrm{fit}}_{r,t}
=
\sum_a\bm{E}_t^{\mathrm{fit}}
\left[
\bm M^{\mathcal R,0,(r)}_{a,t}
\bm M^{\mathcal R,0,(r)\mathsf{T}}_{a,t}
\right].
\label{l2:eq:vanishing-block-fit-covariance}
\end{equation}
Let
\[
D^{\mathcal R,0}_{r,t}
=\operatorname{diag}(s_{\mathcal R,0,1,t},\ldots,s_{\mathcal R,0,r,t}),
\]
\[
A^{\mathcal R,0}_{r,t}
=(D^{\mathcal R,0}_{r,t})^{-1}
\Sigma^{\mathcal R,0,\mathrm{fit}}_{r,t}
(D^{\mathcal R,0}_{r,t})^{-1},
\]
and let $B^{\mathcal R,0}_{r,t}$ be diagonal with entries
$s_{\mathcal R,0,\mathrm{eval},j,t}^{2}/s_{\mathcal R,0,j,t}^{2}$.

\begin{assumption}[Fixed-block exact-zero-direction process]
\label{l2:ass:vanishing-block}
For every fixed $r$:
\begin{enumerate}[label=(\roman*)]
\item the vector of clientwise linear--quadratic remainders is negligible after
multiplication by $(D^{\mathcal R,0}_{r,t})^{-1}$;
\item $A^{\mathcal R,0}_{r,t}\to A^{\mathcal R,0}_r$ and
$B^{\mathcal R,0}_{r,t}\to B^{\mathcal R,0}_r$;
\item for every $\bm q\in\mathbb{R}^r$, the scalar martingale increments
$\bm q^{\mathsf{T}}(D^{\mathcal R,0}_{r,t})^{-1}
\bm M^{\mathcal R,0,(r)}_{a,t}$ satisfy the predictable quadratic-variation
and Lindeberg conditions with limiting variance
$\bm q^{\mathsf{T}}A^{\mathcal R,0}_r\bm q$;
\item the client evaluation samples are independent across clients and
Assumption~\ref{l2:ass:vanishing-evaluation} holds coordinatewise;
\item with
\[
\widehat\Sigma^{\mathcal R,0,\mathrm{fit}}_{r,t}
=
\sum_a
\widehat{\bm M}^{\mathcal R,0,(r)}_{a,t}
\widehat{\bm M}^{\mathcal R,0,(r)\mathsf{T}}_{a,t},
\]
\begin{equation}
\left\|
(D^{\mathcal R,0}_{r,t})^{-1}
(\widehat\Sigma^{\mathcal R,0,\mathrm{fit}}_{r,t}
-\Sigma^{\mathcal R,0,\mathrm{fit}}_{r,t})
(D^{\mathcal R,0}_{r,t})^{-1}
\right\|\xrightarrow{\bm{P}}0,
\label{l2:eq:vanishing-block-covariance-consistency}
\end{equation}
and the coordinatewise total variance estimators are ratio consistent.
\end{enumerate}
\end{assumption}

\begin{theorem}[Fixed-block exact-zero-direction process]
\label{l2:thm:vanishing-fixed-block}
Under Assumption~\ref{l2:ass:vanishing-block},
\begin{equation}
(D^{\mathcal R,0}_{r,t})^{-1}
(\widetilde{\boldsymbol{\mathcal R}}_{r,t}
-\boldsymbol{\mathfrak R}_{r,t})
\Longrightarrow
N_r\{0,R^{\mathcal R,0}_r\},
\qquad
R^{\mathcal R,0}_r
=A^{\mathcal R,0}_r+B^{\mathcal R,0}_r.
\label{l2:eq:vanishing-fixed-block-clt}
\end{equation}
The conclusion remains valid under coordinatewise studentization, and the
estimated fixed-block correlation matrix is consistent.  Compatible
finite-dimensional limits yield a Gaussian process in the product topology.
Fixed-block robust-positive orthant limits follow.  For thresholds
$\varepsilon_{j,t}$, define
\[
\gamma^{\mathcal R,0}_{j,t}
=
\frac{\mathfrak R_{j,t}-\varepsilon_{j,t}}
{s_{\mathcal R,0,j,t}}.
\]
For the moving all-client event, if
$\delta^{\mathcal R,0}_{j,t}$ denotes the coordinatewise normal
approximation error and
\begin{equation}
\sum_{j=1}^{K_t}\Phi(-\gamma^{\mathcal R,0}_{j,t})
+
\sum_{j=1}^{K_t}\delta^{\mathcal R,0}_{j,t}
\longrightarrow0,
\label{l2:eq:vanishing-all-client-summability}
\end{equation}
then every empirical exact-zero-direction robust value exceeds its threshold
with probability tending to one.
\end{theorem}

\begin{proof}
Apply the martingale central limit theorem to every fixed scalar contrast and
use Cram\'er--Wold for the fitting block.  The independent conditional
evaluation limits give the diagonal evaluation block, and conditioning in the
joint characteristic function gives asymptotic independence.  Assumption
\ref{l2:ass:vanishing-block}(v) yields studentization and covariance
consistency.  Product-topology tightness, orthant convergence, and the
union-bound summability argument are the same as in the persistent-direction
process theorem.
\end{proof}

\subsection{Scope}
The robust-value theorem is exact at the fitted level: positivity of
$\mathcal R_{j,t}^{\mathrm{fit}}$ is equivalent to positive gain throughout
the $L_2(Q_{j,t,X})$ ball.  Conditional audit inference targets this realized
fitted robust value; fitting-plus-audit inference targets its repeated-fitting
mean $\mathfrak R_{j,t}$.  The regular same-audit-sample implementation must
include the norm-estimation influence in
Theorem~\ref{l2:thm:same-sample-eval}; an external grid or auxiliary sample
may instead be used only under the stated standard-error-scale transfer
condition.  The exact-zero-direction theorem is for
$\overline d^B_{j,t}=0$ at every stage.  A local-to-zero sequence with
$\norm{\overline d^B_{j,t}}_{j,t}\to0$ may place the linear and quadratic
terms on comparable scales and is outside the present results.  Consistency
of a least-favorable function toward a deterministic direction is claimed
only in the persistent-direction regime.

\section{A fully worked regular regime}
\label{app:worked-regime}

This section verifies in one bounded two-client model the assumptions needed
for the regular pointwise, nominal-gain, KL-certificate, and $L_2$ procedures.
The purpose is not to replace the general triangular-array theorems, but to
show that their fitting, audit, nuisance, communication, and variance
requirements can hold simultaneously.

\subsection{Model and exact gain geometry}

Let $b(x)=b_0+b_1x$, where $b_0>|b_1|/2>0$.  There are two fitting clients,
with equal weights and equal sample sizes $n$.  Under both fitting laws,
$X$ is uniform on $[-1,1]$.  Let $\varepsilon$ be independent Rademacher
noise taking values $\pm\sigma$.  The fitting laws are
\[
P_1:\quad Y=\varepsilon,
\qquad
P_2:\quad Y=2b(X)+\varepsilon.
\]
For client 1, the audit law is
\[
Q_1:\quad X\sim\operatorname{Unif}[-1/2,1/2],
\qquad
Y=b(X)+\varepsilon.
\]
Use a bounded, nonnegative, symmetric, compactly supported second-order
kernel and common bandwidth
\begin{equation}
h_n=n^{-1/3}.
\label{work:eq:bandwidth}
\end{equation}
All predictions are stabilized and clipped at a deterministic level larger
than $2\norm{b}_\infty+\sigma$.

For all sufficiently large $n$, the audit region is separated from the design
boundary by more than the kernel bandwidth.  Symmetry of the kernel and
linearity of $b$ then give the exact finite-bandwidth targets
\[
m^h_{1,L}=0,
\qquad
m^h_F=b,
\qquad
m_1^Q=b.
\]
Thus the procedure-target biases are
$b^P_L=-b$ and $b^P_F=0$, and
\begin{equation}
G_1^P(\alpha)
=(1-\alpha^2)\norm{b}_{Q_{1,X}}^2>0.
\label{work:eq:positive-gain}
\end{equation}
Evaluating the same fitting configuration under the unchanged client law
$P_1$ instead gives
\begin{equation}
G_1^{P,P_1}(\alpha)
=-(1-\alpha)^2\norm{b}_{P_{1,X}}^2<0.
\label{work:eq:negative-gain}
\end{equation}
Hence positive and negative transfer occur within one supported model family,
with only the evaluation regression changed.

\subsection{Pointwise and fitting rates}

For the federated pointwise estimator,
\[
q_n
=
\sum_{j=1}^{2}\frac{w_j^2}{nh_n}
=
\frac{1}{2nh_n}.
\]
The choice \eqref{work:eq:bandwidth} gives
\[
nh_n=n^{2/3}\to\infty,
\qquad
\sqrt n\,h_n=n^{1/6}\to\infty,
\qquad
\sqrt{nh_n}\,h_n^2=n^{-1/3}\to0.
\]
Thus the pointwise consistency, oracle central limit theorem, and distributed
variance estimator apply on the interior audit region.

For the scalar gain theorem, the leverage quantities satisfy
\[
\mathfrak r_n\asymp n^{-1},
\qquad
\gamma_{\max,n}\asymp n^{-1},
\]
so
\[
\frac{\mathfrak r_n}{h_n}=n^{-2/3}\to0,
\qquad
\frac{\sqrt{\mathfrak r_n}}{h_n}=n^{-1/6}\to0,
\qquad
\frac{\gamma_{\max,n}}{\sqrt{\mathfrak r_n}}
=n^{-1/2}\to0.
\]
At the limiting targets, $\overline b_L=-b$ and $\overline b_F=0$; the local
derivative coefficient is nonzero because $b$ is bounded away from zero.
Consequently,
\[
s_{\mathrm{fit},n}^2\asymp n^{-1},
\qquad
\Gamma_n\longrightarrow
(1-\alpha^2)\norm{b}_{Q_{1,X}}^2>0.
\]
This verifies first-order nondegeneracy and positive gain separately.

At the population targets the paired audit gain is
\[
W
=
(1-\alpha^2)b(X)^2
+2(1-\alpha)b(X)\varepsilon.
\]
Its audit variance is positive because $b$ is bounded away from zero:
\begin{equation}
\Var_{Q_1}(W)
=
\Var\{(1-\alpha^2)b(X)^2\}
+4(1-\alpha)^2\sigma^2\bm{E} b(X)^2.
\label{work:eq:audit-variance}
\end{equation}
All standardized moments are uniformly bounded.  An independent audit sample
of size $N_{E,n}\asymp n$ therefore satisfies the conditional Berry--Esseen,
variance-consistency, and studentization requirements.

\subsection{Feasible nuisance and variance estimation}

Use an independent pilot sample from $Q_1$, of size $N_{\eta,n}\asymp n$, to
estimate the linear evaluation regression, the conditional noise variance,
and the audit-covariate integrals.  Estimate the finite-bandwidth local and
federated ratio targets on fitting folds separate from the observations whose
influence contributions are evaluated.  Bounded responses, denominator
floors, the compact interior region, and standard kernel concentration give,
for these nuisance estimates,
\[
a_n
=
O_{\bm{P}}\left
\{
 h_n^2+
 \sqrt{\frac{\log n}{nh_n}}+
 \sqrt{\frac{\log n}{N_{\eta,n}h_n}}
\right\}
=o_{\bm{P}}(1).
\]
The nuisance-specific denominator and error bounds define the good events in
Assumption~\ref{sg:ass:nuisance}; the localization lemma then gives
\[
\sum_a(\widehat\lambda_{a,n}-\lambda_{a,n})^2
=
O_{\bm{P}}(\mathfrak r_na_n^2).
\]
Together with the no-dominant-observation condition, this proves ratio
consistency of the fitting-variance estimator.  The same bounded pilot
construction verifies the regular KL certificate nuisance condition.

The stronger fourth-order leverage comparison used in a quadratic boundary
calculation is also checkable here but is not implied by ordinary
no-dominance.  Indeed,
\[
\sum_a\gamma_{a,n}^4\asymp n^{-3},
\qquad
\mathfrak r_n^2h_n\asymp n^{-7/3},
\]
so
\begin{equation}
\frac{\sum_a\gamma_{a,n}^4}{\mathfrak r_n^2h_n}
\asymp\frac1{nh_n}\longrightarrow0.
\label{work:eq:boundary-leverage}
\end{equation}

\subsection{Regular KL and $L_2$ robustness}

Bounded responses and clipped predictions imply a deterministic conditional
range bound $|W-\mu|\le B_W$.  Bernstein's conditional mgf inequality \citep{BoucheronLugosiMassart2013} gives
\[
\log\bm{E}\{
 e^{\lambda(\mu-W)}\mid\mathcal{F}_n^{\mathrm{fit}}
\}
\le
\frac{\lambda^2v_n}{2\{1-(B_W/3)\lambda\}},
\qquad
0\le\lambda<3/B_W.
\]
Thus the sub-gamma assumption holds with $c_n=B_W/3$.  By
\eqref{work:eq:audit-variance}, the gain variance converges to a positive
constant, so the model lies in the regular positive-variance KL regime.  Its
admissible radii are determined by the fitted critical-radius formulas in
Proposition~\ref{kl:prop:certified-radius}; no sample-size radius is imposed.

The deterministic borrowing direction converges to
$(1-\alpha)b$, and hence
\[
\norm{\overline d^B_n}_{Q_{1,X}}
\longrightarrow
(1-\alpha)\norm b_{Q_{1,X}}>0.
\]
The $L_2$ persistent-direction theorem therefore applies.  The limiting
fitted robust value is
\[
(1-\alpha^2)\norm b^2
-2\tau(1-\alpha)\norm b,
\]
which is positive exactly when
\begin{equation}
\tau<\frac{1+\alpha}{2}\norm b_{Q_{1,X}}.
\label{work:eq:l2-radius}
\end{equation}
If the borrowing norm is estimated from the audit covariates, the audit
variance uses the full influence in
\eqref{l2:eq:same-sample-eval-influence}; an external grid can replace it only
under Assumption~\ref{l2:ass:grid-norm}.

\subsection{Communication and confidence bounds}

At each prediction query, clients transmit the usual numerator and denominator
summaries.  For pointwise variance, each client sends the three residual
second-moment summaries from Appendix~\ref{app:fnw-pointwise}.  For fitting
gain variance, the server broadcasts the estimated coefficient functions on
the pilot covariates; each client computes its observation-level influences
locally.  Client 1 combines its local and server contributions before
squaring, retaining the shared-observation covariance, and transmits only
scalar sums of influence products.  Audit responses remain local; client 1
returns the empirical paired-gain mean, variance, and, when needed, scalar KL
or $L_2$ influence summaries.

The realized-fit and procedure-level lower confidence bounds are,
respectively,
\[
\operatorname{LCB}^{\mathrm{fit}}_n(\delta)
=
\widehat G_n
-z_{1-\delta}
\frac{\widehat\sigma_{E,n}}{\sqrt{N_{E,n}}},
\]
and
\[
\operatorname{LCB}^{\mathrm{proc}}_n(\delta)
=
\widehat G_n
-z_{1-\delta}
\left\{
\widehat s_{\mathrm{fit},n}^2
+
\frac{\widehat\sigma_{E,n}^2}{N_{E,n}}
\right\}^{1/2}.
\]
The first concerns the realized fitted predictor pair; the second concerns the
learning procedure under repeated fitting.

\bibliography{references}

\end{document}